\documentclass[10pt]{article}
\usepackage[preprint]{tmlr}
\usepackage[hidelinks]{hyperref}
\hypersetup{pdftitle={What Does a Harness Repair? Four Preregistered Studies Around a Longevity-Tuned Language Model},pdfsubject={Preprint}}
\usepackage{url}
\usepackage{xurl}
\usepackage{booktabs}
\usepackage{float}
\usepackage{placeins}
\usepackage{needspace}
\usepackage{amsmath}
\usepackage{amssymb}
\usepackage{array}
\usepackage{graphicx}
\graphicspath{{figures/}}
\makeatletter
\newcommand{\V}[1]{\@ifundefined{v@#1}{\PackageError{numbers}{Undefined value #1}{}}{\@nameuse{v@#1}}}
\makeatother
\expandafter\def\csname v@ab.h2.re.0\endcsname{30}
\expandafter\def\csname v@ab.h2.re.1\endcsname{32}
\expandafter\def\csname v@ab.h2.te.0\endcsname{34}
\expandafter\def\csname v@add.budget.h\endcsname{77}
\expandafter\def\csname v@add.gate.clean\endcsname{11{,}265}
\expandafter\def\csname v@add.gate.clean0\endcsname{3{,}849}
\expandafter\def\csname v@add.gate.n\endcsname{324}
\expandafter\def\csname v@add.gate.of\endcsname{1{,}204}
\expandafter\def\csname v@add.gepa.h\endcsname{6.63}
\expandafter\def\csname v@add.gpu.h\endcsname{9.33}
\expandafter\def\csname v@add.idle.h\endcsname{6.46}
\expandafter\def\csname v@add.last\endcsname{12}
\expandafter\def\csname v@add.numpy.local\endcsname{2.2.6}
\expandafter\def\csname v@add.rep.h\endcsname{22.07}
\expandafter\def\csname v@add.rep.runs\endcsname{17}
\expandafter\def\csname v@add.repeat.n\endcsname{3{,}700}
\expandafter\def\csname v@add.sel.n\endcsname{24}
\expandafter\def\csname v@add.test.calls\endcsname{267{,}056}
\expandafter\def\csname v@add.test.h\endcsname{25.83}
\expandafter\def\csname v@add.test.runs\endcsname{14}
\expandafter\def\csname v@add.total.h\endcsname{54.53}
\expandafter\def\csname v@ag.lab\endcsname{9}
\expandafter\def\csname v@ag.maxdiff\endcsname{0.010}
\expandafter\def\csname v@ag.n\endcsname{9}
\expandafter\def\csname v@ag.ncells\endcsname{54}
\expandafter\def\csname v@ag.nexcl\endcsname{2}
\expandafter\def\csname v@ag.q.gepa\endcsname{4}
\expandafter\def\csname v@ag.q.h4\endcsname{6}
\expandafter\def\csname v@ag.q.total\endcsname{19}
\expandafter\def\csname v@ag.rep\endcsname{9}
\expandafter\def\csname v@ag.test\endcsname{9}
\expandafter\def\csname v@an.te.age\endcsname{50.8}
\expandafter\def\csname v@an.te.dec\endcsname{0.346}
\expandafter\def\csname v@ar.band\endcsname{0.05}
\expandafter\def\csname v@ar.cnt.above\endcsname{6}
\expandafter\def\csname v@ar.cnt.below\endcsname{3}
\expandafter\def\csname v@ar.cnt.cells\endcsname{12}
\expandafter\def\csname v@ar.cnt.equal\endcsname{3}
\expandafter\def\csname v@ar.cnt.unres\endcsname{0}
\expandafter\def\csname v@ar.te.gs.b\endcsname{0.700}
\expandafter\def\csname v@ar.te.gs.d\endcsname{\ensuremath{-0.634}}
\expandafter\def\csname v@ar.te.gs.d.hi\endcsname{\ensuremath{-0.590}}
\expandafter\def\csname v@ar.te.gs.d.lo\endcsname{\ensuremath{-0.677}}
\expandafter\def\csname v@ar.te.gs.t\endcsname{0.066}
\expandafter\def\csname v@ar.te.lb.b\endcsname{0.217}
\expandafter\def\csname v@ar.te.lb.d\endcsname{\ensuremath{+0.414}}
\expandafter\def\csname v@ar.te.lb.d.hi\endcsname{\ensuremath{+0.433}}
\expandafter\def\csname v@ar.te.lb.d.lo\endcsname{\ensuremath{+0.395}}
\expandafter\def\csname v@ar.te.lb.t\endcsname{0.631}
\expandafter\def\csname v@ar.te.mm.b\endcsname{0.384}
\expandafter\def\csname v@ar.te.mm.d\endcsname{\ensuremath{-0.164}}
\expandafter\def\csname v@ar.te.mm.d.hi\endcsname{\ensuremath{-0.125}}
\expandafter\def\csname v@ar.te.mm.d.lo\endcsname{\ensuremath{-0.203}}
\expandafter\def\csname v@ar.te.mm.t\endcsname{0.220}
\expandafter\def\csname v@au.re.tot.nc.share\endcsname{23.7\%}
\expandafter\def\csname v@au.re.tot.ref.share\endcsname{22.3\%}
\expandafter\def\csname v@au.te.aut.l38.nc\endcsname{100.0\%}
\expandafter\def\csname v@au.te.aut.lb.nc\endcsname{43.6\%}
\expandafter\def\csname v@au.te.aut.lb.ref\endcsname{36.0\%}
\expandafter\def\csname v@au.te.bud.l38.nc\endcsname{99.9\%}
\expandafter\def\csname v@au.te.t1536.l38.nc\endcsname{99.6\%}
\expandafter\def\csname v@au.te.t512.lb.nc\endcsname{0.0\%}
\expandafter\def\csname v@au.te.t64.lb.nc\endcsname{0.0\%}
\expandafter\def\csname v@au.te.tot.nc.share\endcsname{23.8\%}
\expandafter\def\csname v@au.te.tot.ref.share\endcsname{22.3\%}
\expandafter\def\csname v@bg.rows\endcsname{556}
\expandafter\def\csname v@bl.map30\endcsname{120}
\expandafter\def\csname v@bl.map42\endcsname{2}
\expandafter\def\csname v@bl.te.l42.cbo\endcsname{0.840}
\expandafter\def\csname v@bl.te.lb.chance\endcsname{0.380}
\expandafter\def\csname v@bl.te.lb.clo\endcsname{0.389}
\expandafter\def\csname v@bl.te.lb.so\endcsname{0.539}
\expandafter\def\csname v@bl.te.lb.so.hi\endcsname{0.551}
\expandafter\def\csname v@bl.te.lb.so.lo\endcsname{0.527}
\expandafter\def\csname v@blm.te.l34.i.aut.do\endcsname{\ensuremath{+0.060}}
\expandafter\def\csname v@blm.te.l34.i.aut.do.hi\endcsname{\ensuremath{+0.117}}
\expandafter\def\csname v@blm.te.l34.i.aut.do.lo\endcsname{\ensuremath{+0.003}}
\expandafter\def\csname v@blm.te.l42.q.gc1.do\endcsname{\ensuremath{-0.010}}
\expandafter\def\csname v@blm.te.l42.q.gc1.do.hi\endcsname{\ensuremath{+0.015}}
\expandafter\def\csname v@blm.te.l42.q.gc1.do.lo\endcsname{\ensuremath{-0.034}}
\expandafter\def\csname v@blm.te.l42.q.t512.do\endcsname{\ensuremath{+0.027}}
\expandafter\def\csname v@blm.te.l42.q.t512.do.hi\endcsname{\ensuremath{+0.048}}
\expandafter\def\csname v@blm.te.l42.q.t512.do.lo\endcsname{\ensuremath{+0.007}}
\expandafter\def\csname v@blm.te.lb.i.aut.do\endcsname{\ensuremath{-0.225}}
\expandafter\def\csname v@blm.te.lb.q.aut.acc\endcsname{0.632}
\expandafter\def\csname v@blm.te.lb.q.aut.do\endcsname{\ensuremath{+0.093}}
\expandafter\def\csname v@blm.te.lb.q.aut.do.hi\endcsname{\ensuremath{+0.112}}
\expandafter\def\csname v@blm.te.lb.q.aut.do.lo\endcsname{\ensuremath{+0.074}}
\expandafter\def\csname v@blm.te.lb.q.t512.do\endcsname{\ensuremath{+0.096}}
\expandafter\def\csname v@blm.te.lb.q.t512.do.hi\endcsname{\ensuremath{+0.115}}
\expandafter\def\csname v@blm.te.lb.q.t512.do.lo\endcsname{\ensuremath{+0.077}}
\expandafter\def\csname v@blm.te.lb.s.t1536.do\endcsname{\ensuremath{-0.382}}
\expandafter\def\csname v@c.b\endcsname{50{,}000}
\expandafter\def\csname v@c.ci\endcsname{95\%}
\expandafter\def\csname v@c.fam1\endcsname{9}
\expandafter\def\csname v@c.fam2\endcsname{13}
\expandafter\def\csname v@c.fam3\endcsname{9}
\expandafter\def\csname v@c.fam4\endcsname{16}
\expandafter\def\csname v@c.lvl1\endcsname{99.44}
\expandafter\def\csname v@c.lvl2\endcsname{99.62}
\expandafter\def\csname v@c.lvl4\endcsname{99.69}
\expandafter\def\csname v@c.margin\endcsname{0.05}
\expandafter\def\csname v@c.minm\endcsname{0.05}
\expandafter\def\csname v@c.q\endcsname{0.025}
\expandafter\def\csname v@c.seedbase\endcsname{4242}
\expandafter\def\csname v@c.share\endcsname{0.8}
\expandafter\def\csname v@cb.all\endcsname{65}
\expandafter\def\csname v@cb.fam1\endcsname{12}
\expandafter\def\csname v@cb.fam3\endcsname{12}
\expandafter\def\csname v@cb.h1.visibility\endcsname{7}
\expandafter\def\csname v@cb.h3.holds\endcsname{9}
\expandafter\def\csname v@cb.h4.byc\endcsname{7}
\expandafter\def\csname v@cb.h4.evaluable\endcsname{20}
\expandafter\def\csname v@cb.h4.holds\endcsname{9}
\expandafter\def\csname v@cb.sens.max1\endcsname{3}
\expandafter\def\csname v@cell.re.q.gs.rows\endcsname{264}
\expandafter\def\csname v@cell.re.q.lb.rows\endcsname{1{,}990}
\expandafter\def\csname v@cell.re.q.mm.rows\endcsname{399}
\expandafter\def\csname v@cell.te.i.lb.rows\endcsname{3{,}942}
\expandafter\def\csname v@cell.te.i.mm.rows\endcsname{799}
\expandafter\def\csname v@cell.te.q.gs.rows\endcsname{527}
\expandafter\def\csname v@cell.te.q.lb.rows\endcsname{3{,}942}
\expandafter\def\csname v@cell.te.q.lb.units\endcsname{789}
\expandafter\def\csname v@cell.te.q.mm.rows\endcsname{799}
\expandafter\def\csname v@cfg.te.i.lb.bud.trunc\endcsname{0.021}
\expandafter\def\csname v@cfg.te.i.lb.t1536.trunc\endcsname{0.800}
\expandafter\def\csname v@cfg.te.q.gs.t512.parse\endcsname{0.992}
\expandafter\def\csname v@cfg.te.q.lb.aut.acc\endcsname{0.632}
\expandafter\def\csname v@cfg.te.q.lb.t512.parse\endcsname{1.000}
\expandafter\def\csname v@cfg.te.q.mm.t512.parse\endcsname{0.940}
\expandafter\def\csname v@cfg.te.s.gs.t1536.trunc\endcsname{0.066}
\expandafter\def\csname v@cfg.te.s.lb.bud.nc\endcsname{0.424}
\expandafter\def\csname v@cfg.te.s.mm.bud.trunc\endcsname{0.379}
\expandafter\def\csname v@cl.re.l30.const\endcsname{0.270}
\expandafter\def\csname v@cl.re.l30.h0.acc\endcsname{0.190}
\expandafter\def\csname v@cl.re.l30.h0.share\endcsname{0.5\%}
\expandafter\def\csname v@cl.re.l30.h1.acc\endcsname{0.229}
\expandafter\def\csname v@cl.re.l30.h1.share\endcsname{15.4\%}
\expandafter\def\csname v@cl.re.l34.const\endcsname{0.510}
\expandafter\def\csname v@cl.re.l34.h0.acc\endcsname{0.565}
\expandafter\def\csname v@cl.re.l34.h0.share\endcsname{83.6\%}
\expandafter\def\csname v@cl.re.l34.h1.acc\endcsname{0.580}
\expandafter\def\csname v@cl.re.l34.h1.share\endcsname{47.1\%}
\expandafter\def\csname v@cl.re.l42.const\endcsname{0.823}
\expandafter\def\csname v@cl.re.l42.h0.acc\endcsname{0.830}
\expandafter\def\csname v@cl.re.l42.h0.share\endcsname{94.9\%}
\expandafter\def\csname v@cl.re.l42.h1.acc\endcsname{0.828}
\expandafter\def\csname v@cl.re.l42.h1.share\endcsname{98.6\%}
\expandafter\def\csname v@cl.te.l30.const\endcsname{0.250}
\expandafter\def\csname v@cl.te.l30.h0.acc\endcsname{0.202}
\expandafter\def\csname v@cl.te.l30.h0.share\endcsname{33.4\%}
\expandafter\def\csname v@cl.te.l30.h1.acc\endcsname{0.232}
\expandafter\def\csname v@cl.te.l30.h1.share\endcsname{30.0\%}
\expandafter\def\csname v@cl.te.l30.weight\endcsname{40\%}
\expandafter\def\csname v@cl.te.l34.const\endcsname{0.515}
\expandafter\def\csname v@cl.te.l34.h0.acc\endcsname{0.602}
\expandafter\def\csname v@cl.te.l34.h0.share\endcsname{19.7\%}
\expandafter\def\csname v@cl.te.l34.h1.acc\endcsname{0.586}
\expandafter\def\csname v@cl.te.l34.h1.share\endcsname{58.3\%}
\expandafter\def\csname v@cl.te.l34.weight\endcsname{20\%}
\expandafter\def\csname v@cl.te.l42.aut.share\endcsname{16.2\%}
\expandafter\def\csname v@cl.te.l42.const\endcsname{0.840}
\expandafter\def\csname v@cl.te.l42.h0.acc\endcsname{0.847}
\expandafter\def\csname v@cl.te.l42.h0.share\endcsname{94.6\%}
\expandafter\def\csname v@cl.te.l42.h1.acc\endcsname{0.837}
\expandafter\def\csname v@cl.te.l42.h1.share\endcsname{98.2\%}
\expandafter\def\csname v@cl.te.l42.weight\endcsname{40\%}
\expandafter\def\csname v@claude.pt\endcsname{03:42}
\expandafter\def\csname v@claude.utc\endcsname{10:42}
\expandafter\def\csname v@cnt.all\endcsname{47}
\expandafter\def\csname v@cnt.h123\endcsname{31}
\expandafter\def\csname v@cnt.h1s.visibility\endcsname{8}
\expandafter\def\csname v@cnt.h2.res.holds\endcsname{7}
\expandafter\def\csname v@cnt.h4.fails\endcsname{4}
\expandafter\def\csname v@cnt.h4.holds\endcsname{6}
\expandafter\def\csname v@cnt.h4.inconclusive\endcsname{5}
\expandafter\def\csname v@cnt.lab.inconclusive\endcsname{1}
\expandafter\def\csname v@cnt.lab.notrepl\endcsname{0}
\expandafter\def\csname v@cnt.lab.replicated\endcsname{30}
\expandafter\def\csname v@codex.n\endcsname{16}
\expandafter\def\csname v@codex.pt\endcsname{04:23}
\expandafter\def\csname v@codex.utc\endcsname{11:23}
\expandafter\def\csname v@d.a.comp\endcsname{81}
\expandafter\def\csname v@d.a.rows\endcsname{1{,}204}
\expandafter\def\csname v@d.conc\endcsname{32}
\expandafter\def\csname v@d.e.expected\endcsname{14}
\expandafter\def\csname v@d.lb.comp\endcsname{2{,}104}
\expandafter\def\csname v@d.lb.rows\endcsname{14{,}720}
\expandafter\def\csname v@d.mm.sub\endcsname{2{,}000}
\expandafter\def\csname v@d.pilot.lb\endcsname{214}
\expandafter\def\csname v@d.pilot.n\endcsname{700}
\expandafter\def\csname v@d.resc.n\endcsname{300}
\expandafter\def\csname v@d.split\endcsname{40/20/40}
\expandafter\def\csname v@d.vllm\endcsname{0.30.0}
\expandafter\def\csname v@det.n\endcsname{600}
\expandafter\def\csname v@e3.a13.def\endcsname{no accepted proposal}
\expandafter\def\csname v@e3.a13.nochange\endcsname{No verdict, interval or label changes}
\expandafter\def\csname v@e3.a13.say\endcsname{the selected harness is the seed}
\expandafter\def\csname v@e3.a5.full\endcsname{33}
\expandafter\def\csname v@e3.add.last\endcsname{13}
\expandafter\def\csname v@e3.ag.re.bs.au.agree.pp\endcsname{92.7\%}
\expandafter\def\csname v@e3.ag.re.bs.au.nacc.pp\endcsname{38.6\%}
\expandafter\def\csname v@e3.ag.re.bs.au.oacc.pp\endcsname{36.3\%}
\expandafter\def\csname v@e3.ag.re.bs.au.same.pp\endcsname{80.2\%}
\expandafter\def\csname v@e3.ag.re.bs.th.agree.pp\endcsname{98.2\%}
\expandafter\def\csname v@e3.ag.re.sm.au.nag\endcsname{399}
\expandafter\def\csname v@e3.ag.re.sm.au.rows\endcsname{399}
\expandafter\def\csname v@e3.ag.te.bs.au.agree.pp\endcsname{92.2\%}
\expandafter\def\csname v@e3.ag.te.bs.au.nacc.pp\endcsname{39.2\%}
\expandafter\def\csname v@e3.ag.te.bs.au.nc\endcsname{34}
\expandafter\def\csname v@e3.ag.te.bs.au.oacc.pp\endcsname{38.4\%}
\expandafter\def\csname v@e3.ag.te.bs.au.oc\endcsname{28}
\expandafter\def\csname v@e3.ag.te.bs.au.same.pp\endcsname{79.7\%}
\expandafter\def\csname v@e3.ag.te.bs.th.agree.pp\endcsname{98.9\%}
\expandafter\def\csname v@e3.ag.te.sm.au.nag\endcsname{799}
\expandafter\def\csname v@e3.ag.te.sm.au.rows\endcsname{799}
\expandafter\def\csname v@e3.agr.sha8\endcsname{84691b34}
\expandafter\def\csname v@e3.ana.b\endcsname{50{,}000}
\expandafter\def\csname v@e3.ana.sha\endcsname{b091f544dc76ad0de59406695c7cdb7bc47e274be162a6c4d6eacec7f80e5cf0}
\expandafter\def\csname v@e3.cell.re.sm.au.rows\endcsname{384}
\expandafter\def\csname v@e3.cell.re.sm.bq.rows\endcsname{744}
\expandafter\def\csname v@e3.cell.re.sm.fp.rows\endcsname{982}
\expandafter\def\csname v@e3.cell.re.sm.mm.rows\endcsname{399}
\expandafter\def\csname v@e3.cell.te.sm.au.oracle.pp\endcsname{71.6\%}
\expandafter\def\csname v@e3.cell.te.sm.au.rows\endcsname{977}
\expandafter\def\csname v@e3.cell.te.sm.bq.rows\endcsname{1{,}572}
\expandafter\def\csname v@e3.cell.te.sm.fp.rows\endcsname{2{,}374}
\expandafter\def\csname v@e3.cell.te.sm.mm.const\endcsname{A}
\expandafter\def\csname v@e3.cell.te.sm.mm.oracle.label\endcsname{B}
\expandafter\def\csname v@e3.cell.te.sm.mm.rows\endcsname{799}
\expandafter\def\csname v@e3.cfg.re.bs.mm.t512.trunc.pp\endcsname{98.0\%}
\expandafter\def\csname v@e3.cfg.re.sm.mm.t512.trunc.pp\endcsname{90.7\%}
\expandafter\def\csname v@e3.cfg.te.bs.au.aut.acc.pp\endcsname{64.9\%}
\expandafter\def\csname v@e3.cfg.te.bs.au.t512.trunc.pp\endcsname{52.9\%}
\expandafter\def\csname v@e3.cfg.te.bs.fp.t512.parse.pp\endcsname{90.8\%}
\expandafter\def\csname v@e3.cfg.te.bs.mm.t512.trunc.pp\endcsname{97.4\%}
\expandafter\def\csname v@e3.cfg.te.sm.au.aut.acc.pp\endcsname{53.2\%}
\expandafter\def\csname v@e3.cfg.te.sm.au.t512.trunc.pp\endcsname{82.7\%}
\expandafter\def\csname v@e3.cfg.te.sm.fp.t512.parse.pp\endcsname{98.4\%}
\expandafter\def\csname v@e3.cfg.te.sm.mm.t512.trunc.pp\endcsname{90.5\%}
\expandafter\def\csname v@e3.cnt.coll.au.n\endcsname{8}
\expandafter\def\csname v@e3.cnt.coll.au.of\endcsname{8}
\expandafter\def\csname v@e3.cnt.coll.oth.n\endcsname{0}
\expandafter\def\csname v@e3.cnt.coll.oth.of\endcsname{24}
\expandafter\def\csname v@e3.cnt.fz.no\endcsname{16}
\expandafter\def\csname v@e3.cnt.h1.lab.inc\endcsname{1}
\expandafter\def\csname v@e3.cnt.h1.lab.rep\endcsname{3}
\expandafter\def\csname v@e3.cnt.h1.te.nog\endcsname{2}
\expandafter\def\csname v@e3.cnt.h1.te.vis\endcsname{2}
\expandafter\def\csname v@e3.cnt.h1.total\endcsname{4}
\expandafter\def\csname v@e3.cnt.h2.nontriv\endcsname{3}
\expandafter\def\csname v@e3.cnt.h2.te.fails\endcsname{4}
\expandafter\def\csname v@e3.cnt.h2.te.holds\endcsname{7}
\expandafter\def\csname v@e3.cnt.h2.te.inc\endcsname{1}
\expandafter\def\csname v@e3.cnt.h2.total\endcsname{12}
\expandafter\def\csname v@e3.cnt.h2.triv\endcsname{4}
\expandafter\def\csname v@e3.cnt.h5.all.both.above\endcsname{24}
\expandafter\def\csname v@e3.cnt.h5.all.n\endcsname{12}
\expandafter\def\csname v@e3.cnt.h5.total\endcsname{4}
\expandafter\def\csname v@e3.cnt.mm.runs\endcsname{4}
\expandafter\def\csname v@e3.cnt.n.all\endcsname{40}
\expandafter\def\csname v@e3.cnt.n.conf\endcsname{20}
\expandafter\def\csname v@e3.cnt.n.desc\endcsname{10}
\expandafter\def\csname v@e3.cnt.n.expl\endcsname{10}
\expandafter\def\csname v@e3.cnt.noacc.code\endcsname{4}
\expandafter\def\csname v@e3.cnt.pool.max\endcsname{37}
\expandafter\def\csname v@e3.cnt.pool.min\endcsname{31}
\expandafter\def\csname v@e3.cnt.pool.seed.max\endcsname{36}
\expandafter\def\csname v@e3.cnt.pool.seed.min\endcsname{31}
\expandafter\def\csname v@e3.cnt.runs\endcsname{16}
\expandafter\def\csname v@e3.cnt.seed.mm\endcsname{3}
\expandafter\def\csname v@e3.cnt.seed.n\endcsname{4}
\expandafter\def\csname v@e3.collapse.thr\endcsname{90}
\expandafter\def\csname v@e3.det.calls\endcsname{800}
\expandafter\def\csname v@e3.det.rows\endcsname{400}
\expandafter\def\csname v@e3.det.sha8\endcsname{56544ad6}
\expandafter\def\csname v@e3.fam.h1\endcsname{4}
\expandafter\def\csname v@e3.fam.h2\endcsname{12}
\expandafter\def\csname v@e3.fam.h5\endcsname{8}
\expandafter\def\csname v@e3.fam.h5.report\endcsname{4}
\expandafter\def\csname v@e3.gb.cand1\endcsname{a709e3e8a8d7}
\expandafter\def\csname v@e3.gb.cand1.sel\endcsname{6d3130f93679}
\expandafter\def\csname v@e3.gb.cand2\endcsname{e843b28f81bd}
\expandafter\def\csname v@e3.gb.cand2.sel\endcsname{4f7bb630b2c8}
\expandafter\def\csname v@e3.gb.fail.calls\endcsname{13{,}155}
\expandafter\def\csname v@e3.gb.fail.trip\endcsname{3}
\expandafter\def\csname v@e3.gb.grid.calls\endcsname{17{,}636}
\expandafter\def\csname v@e3.gb.rep153\endcsname{153}
\expandafter\def\csname v@e3.gb.rep3\endcsname{3}
\expandafter\def\csname v@e3.gb.rep37\endcsname{37}
\expandafter\def\csname v@e3.gb.said.calls\endcsname{96{,}006}
\expandafter\def\csname v@e3.gb.said.trip\endcsname{7}
\expandafter\def\csname v@e3.gb.sel.re.calls\endcsname{10{,}036}
\expandafter\def\csname v@e3.gb.sel.te.calls\endcsname{22{,}888}
\expandafter\def\csname v@e3.gb.share\endcsname{0.007\%}
\expandafter\def\csname v@e3.gpu.new\endcsname{RTX PRO 6000}
\expandafter\def\csname v@e3.gpu.old\endcsname{RTX 5090}
\expandafter\def\csname v@e3.h1.re.sm.bq.t.hi.pp\endcsname{\ensuremath{+2.4}}
\expandafter\def\csname v@e3.h1.re.sm.bq.t.lo.pp\endcsname{\ensuremath{-2.7}}
\expandafter\def\csname v@e3.h1.te.bs.au.delta.pp\endcsname{\ensuremath{+34.3}}
\expandafter\def\csname v@e3.h1.te.bs.au.delta.uhi.pp\endcsname{\ensuremath{+38.3}}
\expandafter\def\csname v@e3.h1.te.bs.au.delta.ulo.pp\endcsname{\ensuremath{+30.2}}
\expandafter\def\csname v@e3.h1.te.bs.bq.db.pp\endcsname{\ensuremath{-3.9}}
\expandafter\def\csname v@e3.h1.te.bs.bq.delta.hi.pp\endcsname{\ensuremath{+10.0}}
\expandafter\def\csname v@e3.h1.te.bs.bq.delta.lo.pp\endcsname{\ensuremath{+3.7}}
\expandafter\def\csname v@e3.h1.te.bs.bq.delta.pp\endcsname{\ensuremath{+6.8}}
\expandafter\def\csname v@e3.h1.te.bs.bq.dv.pp\endcsname{\ensuremath{+10.7}}
\expandafter\def\csname v@e3.h1.te.bs.fp.delta.hi.pp\endcsname{\ensuremath{+5.1}}
\expandafter\def\csname v@e3.h1.te.bs.fp.delta.lo.pp\endcsname{\ensuremath{0.0}}
\expandafter\def\csname v@e3.h1.te.bs.fp.delta.pp\endcsname{\ensuremath{+2.6}}
\expandafter\def\csname v@e3.h1.te.bs.mm.delta.pp\endcsname{\ensuremath{+36.7}}
\expandafter\def\csname v@e3.h1.te.sm.au.delta.pp\endcsname{\ensuremath{+40.7}}
\expandafter\def\csname v@e3.h1.te.sm.au.delta.uhi.pp\endcsname{\ensuremath{+44.4}}
\expandafter\def\csname v@e3.h1.te.sm.au.delta.ulo.pp\endcsname{\ensuremath{+37.1}}
\expandafter\def\csname v@e3.h1.te.sm.bq.db.pp\endcsname{\ensuremath{-0.6}}
\expandafter\def\csname v@e3.h1.te.sm.bq.delta.hi.pp\endcsname{\ensuremath{+10.0}}
\expandafter\def\csname v@e3.h1.te.sm.bq.delta.lo.pp\endcsname{\ensuremath{+4.7}}
\expandafter\def\csname v@e3.h1.te.sm.bq.delta.pp\endcsname{\ensuremath{+7.4}}
\expandafter\def\csname v@e3.h1.te.sm.bq.dv.pp\endcsname{\ensuremath{+8.0}}
\expandafter\def\csname v@e3.h1.te.sm.bq.t.lo.pp\endcsname{\ensuremath{+0.5}}
\expandafter\def\csname v@e3.h1.te.sm.fp.delta.hi.pp\endcsname{\ensuremath{+1.3}}
\expandafter\def\csname v@e3.h1.te.sm.fp.delta.lo.pp\endcsname{\ensuremath{-3.5}}
\expandafter\def\csname v@e3.h1.te.sm.fp.delta.pp\endcsname{\ensuremath{-1.1}}
\expandafter\def\csname v@e3.h1.te.sm.mm.delta.pp\endcsname{\ensuremath{+25.5}}
\expandafter\def\csname v@e3.h2.re.bs.mm.s0.theta.pp\endcsname{\ensuremath{+3.5}}
\expandafter\def\csname v@e3.h2.re.bs.mm.s0.verdict\endcsname{holds}
\expandafter\def\csname v@e3.h2.te.bs.au.s0.uhi.pp\endcsname{\ensuremath{-4.1}}
\expandafter\def\csname v@e3.h2.te.bs.au.s1.uhi.pp\endcsname{\ensuremath{-4.5}}
\expandafter\def\csname v@e3.h2.te.bs.mm.s0.hi.pp\endcsname{\ensuremath{+3.5}}
\expandafter\def\csname v@e3.h2.te.bs.mm.s0.lo.pp\endcsname{\ensuremath{-5.1}}
\expandafter\def\csname v@e3.h2.te.bs.mm.s0.theta.pp\endcsname{\ensuremath{-0.8}}
\expandafter\def\csname v@e3.h5.re.bs.au.s0.uhi.pp\endcsname{\ensuremath{+5.8}}
\expandafter\def\csname v@e3.h5.re.bs.au.s0.ulo.pp\endcsname{\ensuremath{+1.0}}
\expandafter\def\csname v@e3.hn.te.sm.au.s0.parse.pp\endcsname{98.7\%}
\expandafter\def\csname v@e3.hn.te.sm.au.s1.parse.pp\endcsname{99.6\%}
\expandafter\def\csname v@e3.items.sha8\endcsname{c946c23e}
\expandafter\def\csname v@e3.lab.re.au.lc.no\endcsname{267}
\expandafter\def\csname v@e3.lab.re.au.lc.yes\endcsname{117}
\expandafter\def\csname v@e3.lab.te.au.lc.no\endcsname{700}
\expandafter\def\csname v@e3.lab.te.au.lc.yes\endcsname{277}
\expandafter\def\csname v@e3.level.h1\endcsname{98.75\%}
\expandafter\def\csname v@e3.level.h2\endcsname{99.58\%}
\expandafter\def\csname v@e3.level.h5\endcsname{99.375\%}
\expandafter\def\csname v@e3.lic.bs\endcsname{Apache-2.0}
\expandafter\def\csname v@e3.lic.sm\endcsname{Apache-2.0}
\expandafter\def\csname v@e3.mm.opt.te.k3\endcsname{4}
\expandafter\def\csname v@e3.ov.chk\endcsname{ff4918215ea7}
\expandafter\def\csname v@e3.ov.dec.total\endcsname{10}
\expandafter\def\csname v@e3.ov.spec\endcsname{9a6cd1a6c9f4}
\expandafter\def\csname v@e3.ova.fp.rno.lo\endcsname{12.6}
\expandafter\def\csname v@e3.ova.fp.rno.max\endcsname{19.4}
\expandafter\def\csname v@e3.ova.fp.rno.min\endcsname{16.4}
\expandafter\def\csname v@e3.ova.fp.tall.max\endcsname{14.2}
\expandafter\def\csname v@e3.ova.fp.tall.min\endcsname{11.9}
\expandafter\def\csname v@e3.ova.fp.tno.lo\endcsname{10.3}
\expandafter\def\csname v@e3.ova.fp.tno.max\endcsname{16.0}
\expandafter\def\csname v@e3.ova.fp.tno.min\endcsname{12.8}
\expandafter\def\csname v@e3.ova.script\endcsname{cb0c23c926e4}
\expandafter\def\csname v@e3.ova.spec\endcsname{df9dc5a70ff7}
\expandafter\def\csname v@e3.ovr.all.g5max.n\endcsname{3}
\expandafter\def\csname v@e3.ovr.au.dno.max\endcsname{0.4}
\expandafter\def\csname v@e3.ovr.au.dno.min\endcsname{-3.4}
\expandafter\def\csname v@e3.ovr.au.dov.max\endcsname{3.2}
\expandafter\def\csname v@e3.ovr.au.dov.min\endcsname{-0.9}
\expandafter\def\csname v@e3.ovr.au.wo.n\endcsname{8}
\expandafter\def\csname v@e3.ovr.au.wo.negv\endcsname{-1.3}
\expandafter\def\csname v@e3.ovr.au.wo.npos\endcsname{7}
\expandafter\def\csname v@e3.ovr.au.wo.pmax\endcsname{6.6}
\expandafter\def\csname v@e3.ovr.au.wo.pmin\endcsname{1.1}
\expandafter\def\csname v@e3.ovr.fp.accgap.max\endcsname{2.7}
\expandafter\def\csname v@e3.ovr.fp.cno.max\endcsname{64.3}
\expandafter\def\csname v@e3.ovr.fp.cno.min\endcsname{59.2}
\expandafter\def\csname v@e3.ovr.fp.cov.max\endcsname{51.9}
\expandafter\def\csname v@e3.ovr.fp.cov.min\endcsname{34.5}
\expandafter\def\csname v@e3.ovr.fp.dall.max\endcsname{19.8}
\expandafter\def\csname v@e3.ovr.fp.dall.min\endcsname{15.9}
\expandafter\def\csname v@e3.ovr.fp.dno.hi\endcsname{17.4}
\expandafter\def\csname v@e3.ovr.fp.dno.lo\endcsname{9.1}
\expandafter\def\csname v@e3.ovr.fp.dno.max\endcsname{14.7}
\expandafter\def\csname v@e3.ovr.fp.dno.min\endcsname{11.7}
\expandafter\def\csname v@e3.ovr.fp.dov.max\endcsname{41.9}
\expandafter\def\csname v@e3.ovr.fp.dov.min\endcsname{22.1}
\expandafter\def\csname v@e3.ovr.fp.rno.lo\endcsname{5.3}
\expandafter\def\csname v@e3.ovr.fp.rno.max\endcsname{12.8}
\expandafter\def\csname v@e3.ovr.fp.rno.min\endcsname{9.1}
\expandafter\def\csname v@e3.ovr.fp.sno.max\endcsname{83.6}
\expandafter\def\csname v@e3.ovr.fp.sno.min\endcsname{73.7}
\expandafter\def\csname v@e3.ovs.y0\endcsname{1900}
\expandafter\def\csname v@e3.ovs.y1\endcsname{2099}
\expandafter\def\csname v@e3.prereg.sec.prompts\endcsname{2.4}
\expandafter\def\csname v@e3.rng.au.dc.re.max\endcsname{3.4}
\expandafter\def\csname v@e3.rng.au.dc.re.min\endcsname{0.3}
\expandafter\def\csname v@e3.rng.au.dc.te.max\endcsname{0.7}
\expandafter\def\csname v@e3.rng.au.dc.te.min\endcsname{0.1}
\expandafter\def\csname v@e3.rng.au.mj.re.max\endcsname{97.1}
\expandafter\def\csname v@e3.rng.au.mj.re.min\endcsname{92.1}
\expandafter\def\csname v@e3.rng.au.mj.te.max\endcsname{96.8}
\expandafter\def\csname v@e3.rng.au.mj.te.min\endcsname{93.1}
\expandafter\def\csname v@e3.rng.au.th.te.max\endcsname{-6.9}
\expandafter\def\csname v@e3.rng.au.th.te.min\endcsname{-19.1}
\expandafter\def\csname v@e3.rng.au.thn.te.max\endcsname{19.1}
\expandafter\def\csname v@e3.rng.au.thn.te.min\endcsname{6.9}
\expandafter\def\csname v@e3.rng.bq.dc.bo.max\endcsname{22.5}
\expandafter\def\csname v@e3.rng.bq.dc.bo.min\endcsname{14.5}
\expandafter\def\csname v@e3.rng.bq.th.bo.amax\endcsname{2.0}
\expandafter\def\csname v@e3.rng.fp.dc.bo.max\endcsname{19.8}
\expandafter\def\csname v@e3.rng.fp.dc.bo.min\endcsname{14.2}
\expandafter\def\csname v@e3.rng.fp.thn.bo.max\endcsname{17.0}
\expandafter\def\csname v@e3.rng.fp.thn.bo.min\endcsname{11.9}
\expandafter\def\csname v@e3.rng.mm.dc.bo.max\endcsname{28.5}
\expandafter\def\csname v@e3.rng.mm.dc.bo.min\endcsname{22.1}
\expandafter\def\csname v@e3.rng.oth.mj.re.max\endcsname{70.5}
\expandafter\def\csname v@e3.rng.oth.mj.re.min\endcsname{12.3}
\expandafter\def\csname v@e3.rng.oth.mj.te.max\endcsname{68.6}
\expandafter\def\csname v@e3.rng.oth.mj.te.min\endcsname{11.8}
\expandafter\def\csname v@e3.run.bs.fp.s1.evald\endcsname{70}
\expandafter\def\csname v@e3.run.bs.fp.s1.pre\endcsname{0}
\expandafter\def\csname v@e3.run.sm.bq.s0.pool\endcsname{33}
\expandafter\def\csname v@e3.run.sm.mm.s1.pre\endcsname{1{,}485}
\expandafter\def\csname v@e3.s0.au.R.rows\endcsname{384}
\expandafter\def\csname v@e3.s0.au.S.rows\endcsname{642}
\expandafter\def\csname v@e3.s0.au.T.largest\endcsname{2}
\expandafter\def\csname v@e3.s0.au.T.rows\endcsname{977}
\expandafter\def\csname v@e3.s0.au.considered\endcsname{2{,}003}
\expandafter\def\csname v@e3.s0.au.k\endcsname{142}
\expandafter\def\csname v@e3.s0.au.q.mc\endcsname{1{,}082}
\expandafter\def\csname v@e3.s0.au.q.num\endcsname{2{,}225}
\expandafter\def\csname v@e3.s0.au.q.tf\endcsname{3{,}225}
\expandafter\def\csname v@e3.s0.au.records\endcsname{6{,}532}
\expandafter\def\csname v@e3.s0.au.share\endcsname{68.22}
\expandafter\def\csname v@e3.s0.au.short\endcsname{23}
\expandafter\def\csname v@e3.s0.au.st.resolved\endcsname{3{,}748}
\expandafter\def\csname v@e3.s0.bq.limit\endcsname{4{,}074}
\expandafter\def\csname v@e3.s0.bq.moved\endcsname{954}
\expandafter\def\csname v@e3.s0.bq.share\endcsname{62.41}
\expandafter\def\csname v@e3.s0.fp.share\endcsname{60.20}
\expandafter\def\csname v@e3.s0.min.rep\endcsname{100}
\expandafter\def\csname v@e3.s0.min.search\endcsname{500}
\expandafter\def\csname v@e3.s0.min.test\endcsname{1{,}000}
\expandafter\def\csname v@e3.s0.mm.share\endcsname{12.59}
\expandafter\def\csname v@e3.s0.share.hi\endcsname{85}
\expandafter\def\csname v@e3.s0.share.lo\endcsname{55}
\expandafter\def\csname v@e3.s0.terms.au\endcsname{research only (README)}
\expandafter\def\csname v@e3.s0.terms.bq\endcsname{CC BY-SA 3.0}
\expandafter\def\csname v@e3.s0.terms.fp\endcsname{CC BY-NC-SA 3.0}
\expandafter\def\csname v@e3.s3.commit\endcsname{7010323}
\expandafter\def\csname v@e3.s3.date\endcsname{26 October 2026}
\expandafter\def\csname v@e3.s3.hash\endcsname{297a6b28}
\expandafter\def\csname v@e3.s3.hash.full\endcsname{297a6b288cc1c8f6f05c5720ea6c959a469b3150d43825920b189e2f3e73ad1e}
\expandafter\def\csname v@e3.s3.time\endcsname{09:16:03}
\expandafter\def\csname v@e3.sc.au.bs.correct\endcsname{133}
\expandafter\def\csname v@e3.sc.au.ntest\endcsname{977}
\expandafter\def\csname v@e3.sc.au.sm.const\endcsname{142}
\expandafter\def\csname v@e3.sc.au.sm.correct\endcsname{104}
\expandafter\def\csname v@e3.sc.au.sm.n\endcsname{200}
\expandafter\def\csname v@e3.sc.au.width\endcsname{0.032}
\expandafter\def\csname v@e3.sc.bq.bs.correct\endcsname{151}
\expandafter\def\csname v@e3.sc.bq.sm.const\endcsname{123}
\expandafter\def\csname v@e3.sc.bq.sm.correct\endcsname{160}
\expandafter\def\csname v@e3.sc.bq.sm.n\endcsname{200}
\expandafter\def\csname v@e3.sc.bq.width\endcsname{0.030}
\expandafter\def\csname v@e3.sc.fp.bs.correct\endcsname{127}
\expandafter\def\csname v@e3.sc.fp.sm.const\endcsname{132}
\expandafter\def\csname v@e3.sc.fp.sm.correct\endcsname{111}
\expandafter\def\csname v@e3.sc.fp.width\endcsname{0.033}
\expandafter\def\csname v@e3.sc.sm.g2b.c1.th.n\endcsname{120}
\expandafter\def\csname v@e3.sc.sm.g2b.c1.th.same\endcsname{120}
\expandafter\def\csname v@e3.sc.sm.shared.n\endcsname{600}
\expandafter\def\csname v@e3.sc.sm.shared.same\endcsname{600}
\expandafter\def\csname v@e3.scorer.commit\endcsname{cc583ed}
\expandafter\def\csname v@e3.scorer.sha\endcsname{0b8c86984419}
\expandafter\def\csname v@e3.scorer.tests\endcsname{35}
\expandafter\def\csname v@e4.a11.cr.bs\endcsname{6}
\expandafter\def\csname v@e4.a11.cr.bs.d\endcsname{477}
\expandafter\def\csname v@e4.a11.cr.sm\endcsname{4}
\expandafter\def\csname v@e4.a11.cr.sm.d\endcsname{446}
\expandafter\def\csname v@e4.a11.d\endcsname{2{,}168}
\expandafter\def\csname v@e4.a11.mf.bs\endcsname{7}
\expandafter\def\csname v@e4.a11.mf.bs.d\endcsname{642}
\expandafter\def\csname v@e4.a11.mf.sm\endcsname{12}
\expandafter\def\csname v@e4.a11.mf.sm.d\endcsname{603}
\expandafter\def\csname v@e4.a11.n\endcsname{29}
\expandafter\def\csname v@e4.a11.p\endcsname{1.34}
\expandafter\def\csname v@e4.a11.runs\endcsname{20}
\expandafter\def\csname v@e4.a11.thr\endcsname{1}
\expandafter\def\csname v@e4.boot\endcsname{50{,}000}
\expandafter\def\csname v@e4.cell.page\endcsname{e5}
\expandafter\def\csname v@e4.ci95\endcsname{95}
\expandafter\def\csname v@e4.cmp.n\endcsname{8{,}426}
\expandafter\def\csname v@e4.cx.bs\endcsname{\ensuremath{-1.0}}
\expandafter\def\csname v@e4.cx.bs.hi\endcsname{\ensuremath{+14.2}}
\expandafter\def\csname v@e4.cx.bs.lo\endcsname{\ensuremath{-16.2}}
\expandafter\def\csname v@e4.cx.sm\endcsname{\ensuremath{+3.1}}
\expandafter\def\csname v@e4.cx.sm.hi\endcsname{\ensuremath{+12.0}}
\expandafter\def\csname v@e4.cx.sm.lo\endcsname{\ensuremath{-5.7}}
\expandafter\def\csname v@e4.d.calls\endcsname{6{,}000}
\expandafter\def\csname v@e4.d.card.p\endcsname{0.95}
\expandafter\def\csname v@e4.d.card.t\endcsname{0.6}
\expandafter\def\csname v@e4.d.conc\endcsname{32}
\expandafter\def\csname v@e4.d.ctx.prop\endcsname{16{,}384}
\expandafter\def\csname v@e4.d.ctx.task\endcsname{8{,}192}
\expandafter\def\csname v@e4.d.deff.cr\endcsname{27.51}
\expandafter\def\csname v@e4.d.deff.mf\endcsname{3.46}
\expandafter\def\csname v@e4.d.evals.cr\endcsname{8{,}330}
\expandafter\def\csname v@e4.d.evals.mf\endcsname{2{,}696}
\expandafter\def\csname v@e4.d.gepa.ver\endcsname{0.1.4}
\expandafter\def\csname v@e4.d.half\endcsname{150}
\expandafter\def\csname v@e4.d.ins.chars\endcsname{3{,}000}
\expandafter\def\csname v@e4.d.lvl1\endcsname{98.75}
\expandafter\def\csname v@e4.d.lvl2\endcsname{99.375}
\expandafter\def\csname v@e4.d.maj80\endcsname{240}
\expandafter\def\csname v@e4.d.margin\endcsname{0.05}
\expandafter\def\csname v@e4.d.margin.pts\endcsname{5}
\expandafter\def\csname v@e4.d.maxtok\endcsname{1{,}500}
\expandafter\def\csname v@e4.d.mb\endcsname{3}
\expandafter\def\csname v@e4.d.min80\endcsname{60}
\expandafter\def\csname v@e4.d.reps\endcsname{5}
\expandafter\def\csname v@e4.d.rows\endcsname{300}
\expandafter\def\csname v@e4.d.runs\endcsname{60}
\expandafter\def\csname v@e4.d.sys.chars\endcsname{3{,}800}
\expandafter\def\csname v@e4.d.tl.cap\endcsname{2{,}048}
\expandafter\def\csname v@e4.d.val.maj\endcsname{120}
\expandafter\def\csname v@e4.d.val.min\endcsname{30}
\expandafter\def\csname v@e4.d.weights\endcsname{1:4}
\expandafter\def\csname v@e4.d.weights.min\endcsname{4}
\expandafter\def\csname v@e4.d001.date\endcsname{2026-11-07}
\expandafter\def\csname v@e4.d001.time\endcsname{23:57:39}
\expandafter\def\csname v@e4.floor\endcsname{0.02}
\expandafter\def\csname v@e4.hw.driver\endcsname{595.71.05}
\expandafter\def\csname v@e4.hw.gpu\endcsname{NVIDIA GeForce RTX 5090}
\expandafter\def\csname v@e4.hw.vllm\endcsname{0.30.0}
\expandafter\def\csname v@e4.lic.0.date\endcsname{2026-11-07}
\expandafter\def\csname v@e4.manifold.date\endcsname{2026-10-29}
\expandafter\def\csname v@e4.orig.all\endcsname{12}
\expandafter\def\csname v@e4.orig.sha8\endcsname{83d3dbf8}
\expandafter\def\csname v@e4.orig.sup\endcsname{5}
\expandafter\def\csname v@e4.osf.commit\endcsname{15b7ea6}
\expandafter\def\csname v@e4.osf.embargo\endcsname{2027-07-01}
\expandafter\def\csname v@e4.osf.reg.date\endcsname{2026-11-03}
\expandafter\def\csname v@e4.osf.reg.time\endcsname{07:23}
\expandafter\def\csname v@e4.pil.S\endcsname{5}
\expandafter\def\csname v@e4.pil.dev.date\endcsname{2026-11-01}
\expandafter\def\csname v@e4.pil.r3.d\endcsname{723}
\expandafter\def\csname v@e4.pil.r3.n\endcsname{4}
\expandafter\def\csname v@e4.pil.r3.p\endcsname{0.55}
\expandafter\def\csname v@e4.pil.reqs\endcsname{1{,}487}
\expandafter\def\csname v@e4.pil.runs\endcsname{12}
\expandafter\def\csname v@e4.pil.sd.bs\endcsname{0.047}
\expandafter\def\csname v@e4.pil.sd.sm\endcsname{0.098}
\expandafter\def\csname v@e4.pil.seed0\endcsname{0 of 12}
\expandafter\def\csname v@e4.pre.bytes\endcsname{77{,}516}
\expandafter\def\csname v@e4.pre.sha8\endcsname{df6ffb5d}
\expandafter\def\csname v@e4.ratio.bal\endcsname{50:50}
\expandafter\def\csname v@e4.ratio.mf\endcsname{43:57}
\expandafter\def\csname v@e4.ratio.skew\endcsname{80:20}
\expandafter\def\csname v@e4.rc.n\endcsname{12}
\expandafter\def\csname v@e4.rc.rows\endcsname{400}
\expandafter\def\csname v@e4.re.cr.sm.rows\endcsname{1{,}531}
\expandafter\def\csname v@e4.re.mf.sm.rows\endcsname{595}
\expandafter\def\csname v@e4.read.entry\endcsname{31}
\expandafter\def\csname v@e4.read.n\endcsname{76}
\expandafter\def\csname v@e4.read.sha8\endcsname{f08b483b}
\expandafter\def\csname v@e4.rep.sha8\endcsname{f3fed57f}
\expandafter\def\csname v@e4.rng.cr.hc.max\endcsname{\ensuremath{+34.5}}
\expandafter\def\csname v@e4.rng.cr.hc.min\endcsname{\ensuremath{+25.3}}
\expandafter\def\csname v@e4.rng.mf.ba.max\endcsname{52.7}
\expandafter\def\csname v@e4.rng.mf.ba.min\endcsname{50.1}
\expandafter\def\csname v@e4.rng.mf.hc.max\endcsname{\ensuremath{0.0}}
\expandafter\def\csname v@e4.rng.mf.hc.min\endcsname{\ensuremath{-3.6}}
\expandafter\def\csname v@e4.run.a5.stop\endcsname{22:53}
\expandafter\def\csname v@e4.run.card.date\endcsname{2026-11-04}
\expandafter\def\csname v@e4.run.ceiling\endcsname{700}
\expandafter\def\csname v@e4.run.ckpt.sha8\endcsname{7688850b}
\expandafter\def\csname v@e4.run.end.date\endcsname{2026-11-07}
\expandafter\def\csname v@e4.run.end.time\endcsname{20:04}
\expandafter\def\csname v@e4.run.gepa.h\endcsname{53.2}
\expandafter\def\csname v@e4.run.harn\endcsname{80}
\expandafter\def\csname v@e4.run.hours\endcsname{101.171}
\expandafter\def\csname v@e4.run.k\endcsname{1}
\expandafter\def\csname v@e4.run.list.sha8\endcsname{bf524187}
\expandafter\def\csname v@e4.run.rep\endcsname{99}
\expandafter\def\csname v@e4.run.rep.h\endcsname{3.2}
\expandafter\def\csname v@e4.run.spend\endcsname{291.4}
\expandafter\def\csname v@e4.run.sta.date\endcsname{2026-11-04}
\expandafter\def\csname v@e4.run.test\endcsname{95}
\expandafter\def\csname v@e4.rw.chen.n\endcsname{114}
\expandafter\def\csname v@e4.rw.chen.p\endcsname{22}
\expandafter\def\csname v@e4.rw.chen.rsi\endcsname{90}
\expandafter\def\csname v@e4.rw.ding.a\endcsname{54}
\expandafter\def\csname v@e4.rw.ding.b\endcsname{14}
\expandafter\def\csname v@e4.rw.ding.n\endcsname{21}
\expandafter\def\csname v@e4.rw.hu.hi\endcsname{20}
\expandafter\def\csname v@e4.rw.hu.lo\endcsname{13}
\expandafter\def\csname v@e4.rw.hu.n16\endcsname{16}
\expandafter\def\csname v@e4.rw.hu.n256\endcsname{256}
\expandafter\def\csname v@e4.rw.tomar90\endcsname{90}
\expandafter\def\csname v@e4.sec8\endcsname{8}
\expandafter\def\csname v@e4.sw.numpy\endcsname{2.2.6}
\expandafter\def\csname v@e4.sw.scipy\endcsname{1.15.3}
\expandafter\def\csname v@e4.te.cr.bs.cs.rmin\endcsname{\ensuremath{+11.0}}
\expandafter\def\csname v@e4.te.cr.bs.cs.rmin.hi\endcsname{\ensuremath{+18.8}}
\expandafter\def\csname v@e4.te.cr.bs.cs.rmin.lo\endcsname{\ensuremath{+2.8}}
\expandafter\def\csname v@e4.te.cr.bs.sm.c55.rmin\endcsname{74.9}
\expandafter\def\csname v@e4.te.cr.bs.sm.c82a.rmaj\endcsname{93.4}
\expandafter\def\csname v@e4.te.cr.bs.sm.c82a.rmin\endcsname{57.6}
\expandafter\def\csname v@e4.te.cr.sm.bl.aut.ba\endcsname{82.1}
\expandafter\def\csname v@e4.te.cr.sm.cs.rmin\endcsname{\ensuremath{+0.3}}
\expandafter\def\csname v@e4.te.cr.sm.cs.rmin.hi\endcsname{\ensuremath{+3.0}}
\expandafter\def\csname v@e4.te.cr.sm.cs.rmin.lo\endcsname{\ensuremath{-2.3}}
\expandafter\def\csname v@e4.te.cr.sm.sm.c55.rmin\endcsname{83.4}
\expandafter\def\csname v@e4.te.cr.sm.sm.c82a.ba\endcsname{80.9}
\expandafter\def\csname v@e4.te.cr.sm.sm.c82a.rmaj\endcsname{93.4}
\expandafter\def\csname v@e4.te.cr.sm.sm.c82a.rmin\endcsname{68.5}
\expandafter\def\csname v@e4.te.mf.bs.b2\endcsname{\ensuremath{+42.8}}
\expandafter\def\csname v@e4.te.mf.bs.bl.aut.ba\endcsname{52.9}
\expandafter\def\csname v@e4.te.mf.bs.bm\endcsname{\ensuremath{-37.8}}
\expandafter\def\csname v@e4.te.mf.bs.d2\endcsname{\ensuremath{+37.6}}
\expandafter\def\csname v@e4.te.mf.bs.dm\endcsname{\ensuremath{-32.5}}
\expandafter\def\csname v@e4.te.mf.bs.k.acc\endcsname{57.5}
\expandafter\def\csname v@e4.te.mf.bs.sm.c55.rmin\endcsname{19.2}
\expandafter\def\csname v@e4.te.mf.bs.sm.c82a.rmaj\endcsname{99.3}
\expandafter\def\csname v@e4.te.mf.bs.sm.c82a.rmin\endcsname{0.9}
\expandafter\def\csname v@e4.te.mf.bs.sm.c82b.ba\endcsname{52.6}
\expandafter\def\csname v@e4.te.mf.sm.bl.aut.ba\endcsname{40.7}
\expandafter\def\csname v@e4.te.mf.sm.sm.c55.rmin\endcsname{13.6}
\expandafter\def\csname v@e4.te.mf.sm.sm.c82a.rmaj\endcsname{98.5}
\expandafter\def\csname v@e4.te.mf.sm.sm.c82a.rmin\endcsname{1.9}
\expandafter\def\csname v@ex.b1.i\endcsname{0.028}
\expandafter\def\csname v@ex.b1.q\endcsname{0.201}
\expandafter\def\csname v@ex.b1.s\endcsname{0.098}
\expandafter\def\csname v@ex.b2.i\endcsname{0.276}
\expandafter\def\csname v@ex.b2.q\endcsname{0.000}
\expandafter\def\csname v@ex.b2.s\endcsname{0.196}
\expandafter\def\csname v@ex.d.half\endcsname{0.50}
\expandafter\def\csname v@ex.d.i\endcsname{0.150}
\expandafter\def\csname v@ex.d.q\endcsname{1.000}
\expandafter\def\csname v@ex.d.s\endcsname{0.327}
\expandafter\def\csname v@ex.d.within\endcsname{0.01}
\expandafter\def\csname v@ex.e.q\endcsname{0.159}
\expandafter\def\csname v@ex.h1.i.gs.s1\endcsname{0.45}
\expandafter\def\csname v@ex.h1.i.gs.s2\endcsname{0.06}
\expandafter\def\csname v@ex.h1.s.gs.s1\endcsname{0.52}
\expandafter\def\csname v@ex.h1.s.gs.s2\endcsname{0.00}
\expandafter\def\csname v@excl.re.reading_exposed\endcsname{7}
\expandafter\def\csname v@excl.re.v21_holdout\endcsname{147}
\expandafter\def\csname v@excl.te.v21_holdout\endcsname{371}
\expandafter\def\csname v@fa.ir4.n\endcsname{7}
\expandafter\def\csname v@fa.ir4.of\endcsname{7}
\expandafter\def\csname v@fa.prgain.n\endcsname{1}
\expandafter\def\csname v@fa.scf1.n\endcsname{4}
\expandafter\def\csname v@fa.scf2.n\endcsname{4}
\expandafter\def\csname v@fa.scf3.n\endcsname{2}
\expandafter\def\csname v@fa.tb2.n\endcsname{3}
\expandafter\def\csname v@fr.date\endcsname{30 September 2026}
\expandafter\def\csname v@fr.sha\endcsname{dec5dc25fdd0074a973926767a7a7f378e3dbc06eaccbf976d6df6f84dd2f515}
\expandafter\def\csname v@fr.time\endcsname{06:53:11}
\expandafter\def\csname v@ft.ctx\endcsname{2{,}048}
\expandafter\def\csname v@fu.gs.gb0.bad\endcsname{142}
\expandafter\def\csname v@fu.gs.gb0.iter\endcsname{975}
\expandafter\def\csname v@fu.gs.gb0.prop\endcsname{975}
\expandafter\def\csname v@fu.gs.gb0.seedval\endcsname{0.000}
\expandafter\def\csname v@fu.gs.gb0.soft\endcsname{0}
\expandafter\def\csname v@fu.gs.gb1.bad\endcsname{137}
\expandafter\def\csname v@fu.gs.gb1.soft\endcsname{0}
\expandafter\def\csname v@fu.gs.gc0.acc\endcsname{30}
\expandafter\def\csname v@fu.gs.gc0.bestval\endcsname{0.587}
\expandafter\def\csname v@fu.gs.gc0.iter\endcsname{241}
\expandafter\def\csname v@fu.gs.gc0.nonprop\endcsname{32}
\expandafter\def\csname v@fu.gs.gc0.prop\endcsname{209}
\expandafter\def\csname v@fu.gs.gc0.seedval\endcsname{0.180}
\expandafter\def\csname v@fu.gs.gc0.soft\endcsname{0}
\expandafter\def\csname v@fu.gs.gc1.acc\endcsname{29}
\expandafter\def\csname v@fu.gs.gc1.iter\endcsname{271}
\expandafter\def\csname v@fu.gs.gc1.nonprop\endcsname{32}
\expandafter\def\csname v@fu.gs.gc1.prop\endcsname{239}
\expandafter\def\csname v@fu.gs.gc1.soft\endcsname{0}
\expandafter\def\csname v@fu.re.gc0.minus.strong\endcsname{\ensuremath{-0.018}}
\expandafter\def\csname v@fu.re.gc1.minus.strong\endcsname{\ensuremath{-0.001}}
\expandafter\def\csname v@fu.te.aut.acc\endcsname{0.203}
\expandafter\def\csname v@fu.te.aut.minus.strong\endcsname{\ensuremath{-0.336}}
\expandafter\def\csname v@fu.te.base.strong\endcsname{0.539}
\expandafter\def\csname v@fu.te.bud.minus.strong\endcsname{\ensuremath{-0.135}}
\expandafter\def\csname v@fu.te.gc0.acc\endcsname{0.540}
\expandafter\def\csname v@fu.te.gc0.minus.strong\endcsname{\ensuremath{+0.001}}
\expandafter\def\csname v@fu.te.gc0.minus.strong.hi\endcsname{\ensuremath{+0.017}}
\expandafter\def\csname v@fu.te.gc1.acc\endcsname{0.545}
\expandafter\def\csname v@fu.te.gc1.minus.strong\endcsname{\ensuremath{+0.006}}
\expandafter\def\csname v@fu.te.gc1.minus.strong.hi\endcsname{\ensuremath{+0.020}}
\expandafter\def\csname v@fu.tok.po.aut.gs\endcsname{23.7}
\expandafter\def\csname v@fu.txt.arm\endcsname{300}
\expandafter\def\csname v@fu.txt.dec.minrows\endcsname{6}
\expandafter\def\csname v@fu.txt.dec.none\endcsname{2}
\expandafter\def\csname v@fu.txt.eval\endcsname{6{,}307}
\expandafter\def\csname v@fu.txt.gc0.ndec\endcsname{4}
\expandafter\def\csname v@fu.txt.gc0.windows\endcsname{4{,}681}
\expandafter\def\csname v@fu.txt.gc1.ndec\endcsname{8}
\expandafter\def\csname v@fu.txt.gc1.windows\endcsname{5{,}759}
\expandafter\def\csname v@gb.te.gs.co\endcsname{0.034}
\expandafter\def\csname v@gb.te.mm.co\endcsname{0.119}
\expandafter\def\csname v@gb.te.mm.items\endcsname{799}
\expandafter\def\csname v@gb.te.mm.k10\endcsname{670}
\expandafter\def\csname v@gbm.re.gs.q.t512.do\endcsname{\ensuremath{+0.011}}
\expandafter\def\csname v@gbm.re.gs.q.t512.do.hi\endcsname{\ensuremath{+0.049}}
\expandafter\def\csname v@gbm.re.gs.q.t512.do.lo\endcsname{\ensuremath{-0.027}}
\expandafter\def\csname v@gbm.te.gs.i.bud.acc\endcsname{0.922}
\expandafter\def\csname v@gbm.te.gs.q.t512.acc\endcsname{0.070}
\expandafter\def\csname v@gbm.te.gs.q.t512.do\endcsname{\ensuremath{+0.036}}
\expandafter\def\csname v@gbm.te.gs.q.t512.do.hi\endcsname{\ensuremath{+0.063}}
\expandafter\def\csname v@gbm.te.gs.q.t512.do.lo\endcsname{\ensuremath{+0.011}}
\expandafter\def\csname v@gbm.te.gs.q.t64.acc\endcsname{0.068}
\expandafter\def\csname v@gbm.te.gs.s.t1536.acc\endcsname{0.880}
\expandafter\def\csname v@gbm.te.mm.i.bud.acc\endcsname{0.542}
\expandafter\def\csname v@gbm.te.mm.q.t1536.acc\endcsname{0.234}
\expandafter\def\csname v@gbm.te.mm.q.t1536.do\endcsname{\ensuremath{+0.115}}
\expandafter\def\csname v@gbm.te.mm.q.t512.acc\endcsname{0.217}
\expandafter\def\csname v@gbm.te.mm.q.t512.do\endcsname{\ensuremath{+0.098}}
\expandafter\def\csname v@gbm.te.mm.s.res.acc\endcsname{0.378}
\expandafter\def\csname v@gepa.gb0.calls\endcsname{6{,}147}
\expandafter\def\csname v@gepa.gb0.of\endcsname{24}
\expandafter\def\csname v@gepa.gb0.seen\endcsname{377}
\expandafter\def\csname v@gepa.gb0.sel\endcsname{1}
\expandafter\def\csname v@gepa.gb0.val\endcsname{0.660}
\expandafter\def\csname v@gepa.gb1.of\endcsname{29}
\expandafter\def\csname v@gepa.gb1.sel\endcsname{1}
\expandafter\def\csname v@gepa.gb1.tacc\endcsname{0.620}
\expandafter\def\csname v@gepa.gb1.val\endcsname{0.667}
\expandafter\def\csname v@gepa.gc0.kept\endcsname{31}
\expandafter\def\csname v@gepa.gc0.of\endcsname{31}
\expandafter\def\csname v@gepa.gc0.sel\endcsname{0}
\expandafter\def\csname v@gepa.gc0.val\endcsname{0.680}
\expandafter\def\csname v@gepa.gc1.calls\endcsname{6{,}003}
\expandafter\def\csname v@gepa.gc1.of\endcsname{31}
\expandafter\def\csname v@gepa.gc1.seen\endcsname{192}
\expandafter\def\csname v@gepa.gc1.sel\endcsname{12}
\expandafter\def\csname v@gepa.gc1.tacc\endcsname{0.631}
\expandafter\def\csname v@gepa.gc1.val\endcsname{0.700}
\expandafter\def\csname v@gp.copy.rows\endcsname{10}
\expandafter\def\csname v@gp.copy.run\endcsname{40}
\expandafter\def\csname v@gp.ctx\endcsname{4{,}096}
\expandafter\def\csname v@gp.limit.sys\endcsname{3{,}800}
\expandafter\def\csname v@gp.limit.usr\endcsname{3{,}000}
\expandafter\def\csname v@gp.mb\endcsname{3}
\expandafter\def\csname v@gp.n.cand\endcsname{20}
\expandafter\def\csname v@gp.prop.tok\endcsname{8{,}192}
\expandafter\def\csname v@gp.rows.lb30\endcsname{120}
\expandafter\def\csname v@gp.rows.lb34\endcsname{60}
\expandafter\def\csname v@gp.rows.lb42\endcsname{120}
\expandafter\def\csname v@gp.seed\endcsname{20261011}
\expandafter\def\csname v@gp.six\endcsname{6}
\expandafter\def\csname v@gp.sp.a\endcsname{60}
\expandafter\def\csname v@gp.sp.b\endcsname{30}
\expandafter\def\csname v@gp.split\endcsname{150}
\expandafter\def\csname v@gp.ver\endcsname{0.1.4}
\expandafter\def\csname v@h1.te.i.gs.db\endcsname{\ensuremath{-0.343}}
\expandafter\def\csname v@h1.te.i.gs.delta\endcsname{\ensuremath{-0.287}}
\expandafter\def\csname v@h1.te.i.gs.delta.hi\endcsname{\ensuremath{-0.218}}
\expandafter\def\csname v@h1.te.i.gs.delta.lo\endcsname{\ensuremath{-0.355}}
\expandafter\def\csname v@h1.te.i.gs.t\endcsname{\ensuremath{+0.286}}
\expandafter\def\csname v@h1.te.i.lb.nb\endcsname{1}
\expandafter\def\csname v@h1.te.i.lb.share\endcsname{1.00}
\expandafter\def\csname v@h1.te.i.lb.t\endcsname{\ensuremath{+0.063}}
\expandafter\def\csname v@h1.te.i.mm.delta\endcsname{\ensuremath{+0.403}}
\expandafter\def\csname v@h1.te.i.mm.delta.hi\endcsname{\ensuremath{+0.456}}
\expandafter\def\csname v@h1.te.i.mm.delta.lo\endcsname{\ensuremath{+0.352}}
\expandafter\def\csname v@h1.te.i.mm.dv\endcsname{\ensuremath{+0.419}}
\expandafter\def\csname v@h1.te.i.mm.nb\endcsname{69}
\expandafter\def\csname v@h1.te.q.lb.delta\endcsname{\ensuremath{-0.003}}
\expandafter\def\csname v@h1.te.q.lb.delta.hi\endcsname{\ensuremath{+0.012}}
\expandafter\def\csname v@h1.te.q.lb.delta.lo\endcsname{\ensuremath{-0.018}}
\expandafter\def\csname v@h1.te.s.gs.db\endcsname{\ensuremath{-0.036}}
\expandafter\def\csname v@h1.te.s.gs.dl\endcsname{\ensuremath{-0.053}}
\expandafter\def\csname v@h1.te.s.gs.dv\endcsname{\ensuremath{+0.197}}
\expandafter\def\csname v@h1.te.s.gs.nb\endcsname{340}
\expandafter\def\csname v@h1.te.s.gs.share\endcsname{1.82}
\expandafter\def\csname v@h1.te.s.gs.t\endcsname{\ensuremath{+0.068}}
\expandafter\def\csname v@h1.te.s.gs.t.hi\endcsname{\ensuremath{+0.090}}
\expandafter\def\csname v@h1.te.s.gs.t.lo\endcsname{\ensuremath{+0.047}}
\expandafter\def\csname v@h1.te.s.lb.db\endcsname{\ensuremath{+0.001}}
\expandafter\def\csname v@h1.te.s.lb.delta\endcsname{\ensuremath{+0.060}}
\expandafter\def\csname v@h1.te.s.lb.delta.hi\endcsname{\ensuremath{+0.074}}
\expandafter\def\csname v@h1.te.s.lb.delta.lo\endcsname{\ensuremath{+0.046}}
\expandafter\def\csname v@h1.te.s.lb.nb\endcsname{448}
\expandafter\def\csname v@h1.te.s.lb.pnc\endcsname{543}
\expandafter\def\csname v@h1.te.s.lb.punp\endcsname{3{,}394}
\expandafter\def\csname v@h1.te.s.mm.nb\endcsname{81}
\expandafter\def\csname v@h2.te.i.gs.res.hi\endcsname{\ensuremath{-0.281}}
\expandafter\def\csname v@h2.te.i.gs.res.lo\endcsname{\ensuremath{-0.402}}
\expandafter\def\csname v@h2.te.i.gs.res.theta\endcsname{\ensuremath{-0.342}}
\expandafter\def\csname v@h2.te.i.lb.res.theta\endcsname{\ensuremath{0.000}}
\expandafter\def\csname v@h2.te.q.lb.gb1.theta\endcsname{\ensuremath{+0.012}}
\expandafter\def\csname v@h2.te.q.lb.gc0.theta\endcsname{\ensuremath{0.000}}
\expandafter\def\csname v@h2.te.q.lb.gc1.theta\endcsname{\ensuremath{+0.002}}
\expandafter\def\csname v@h2.te.s.gs.res.hi\endcsname{\ensuremath{-0.053}}
\expandafter\def\csname v@h2.te.s.gs.res.lo\endcsname{\ensuremath{-0.129}}
\expandafter\def\csname v@h2.te.s.gs.res.theta\endcsname{\ensuremath{-0.089}}
\expandafter\def\csname v@h3.te.i.lb.dt\endcsname{\ensuremath{+0.979}}
\expandafter\def\csname v@h3.te.i.mm.dp\endcsname{\ensuremath{+0.662}}
\expandafter\def\csname v@h3.te.q.gs.tr.hi\endcsname{\ensuremath{0.000}}
\expandafter\def\csname v@h3.te.q.lb.tr.hi\endcsname{\ensuremath{+0.001}}
\expandafter\def\csname v@h3.te.q.mm.dt\endcsname{\ensuremath{+0.060}}
\expandafter\def\csname v@h3.te.q.mm.dt.hi\endcsname{\ensuremath{+0.085}}
\expandafter\def\csname v@h3.te.q.mm.dt.lo\endcsname{\ensuremath{+0.038}}
\expandafter\def\csname v@h3.te.s.gs.dt\endcsname{\ensuremath{+0.260}}
\expandafter\def\csname v@h3.te.s.lb.dp\endcsname{\ensuremath{+0.151}}
\expandafter\def\csname v@h4.a.hcomp\endcsname{86}
\expandafter\def\csname v@h4.a.hrows\endcsname{602}
\expandafter\def\csname v@h4.a.i.dr\endcsname{\ensuremath{+0.164}}
\expandafter\def\csname v@h4.a.i.ds\endcsname{\ensuremath{+0.142}}
\expandafter\def\csname v@h4.a.i.g.hi\endcsname{\ensuremath{+0.108}}
\expandafter\def\csname v@h4.a.i.g.lo\endcsname{\ensuremath{-0.065}}
\expandafter\def\csname v@h4.a.rows.r\endcsname{565}
\expandafter\def\csname v@h4.a.rows.s\endcsname{562}
\expandafter\def\csname v@h4.a.seed\endcsname{20261012}
\expandafter\def\csname v@h4.a.t30\endcsname{344}
\expandafter\def\csname v@h4.a.t34\endcsname{172}
\expandafter\def\csname v@h4.a.t38\endcsname{344}
\expandafter\def\csname v@h4.a.t42\endcsname{344}
\expandafter\def\csname v@h4.b1.i.bound\endcsname{0.057}
\expandafter\def\csname v@h4.b1.i.ds\endcsname{\ensuremath{-0.012}}
\expandafter\def\csname v@h4.b1.q.bound\endcsname{0.205}
\expandafter\def\csname v@h4.b1.q.ds\endcsname{\ensuremath{-0.145}}
\expandafter\def\csname v@h4.b1.r.n\endcsname{794}
\expandafter\def\csname v@h4.b1.r.of\endcsname{1{,}577}
\expandafter\def\csname v@h4.b1.s.bound\endcsname{0.083}
\expandafter\def\csname v@h4.b1.s.ds\endcsname{\ensuremath{+0.001}}
\expandafter\def\csname v@h4.b1.s.n\endcsname{408}
\expandafter\def\csname v@h4.b1.s.of\endcsname{796}
\expandafter\def\csname v@h4.b1.share\endcsname{0.2050}
\expandafter\def\csname v@h4.b2.i.bound\endcsname{0.289}
\expandafter\def\csname v@h4.b2.i.ds\endcsname{\ensuremath{+0.289}}
\expandafter\def\csname v@h4.b2.q.bound\endcsname{0.000}
\expandafter\def\csname v@h4.b2.s.bound\endcsname{0.208}
\expandafter\def\csname v@h4.b2.s.ds\endcsname{\ensuremath{+0.208}}
\expandafter\def\csname v@h4.c.all\endcsname{8{,}830}
\expandafter\def\csname v@h4.c.pool\endcsname{8{,}305}
\expandafter\def\csname v@h4.c.s.dr.hi\endcsname{\ensuremath{+0.063}}
\expandafter\def\csname v@h4.c.s.dr.lo\endcsname{\ensuremath{-0.048}}
\expandafter\def\csname v@h4.c.s.ds\endcsname{\ensuremath{-0.001}}
\expandafter\def\csname v@h4.c.seed\endcsname{20261016}
\expandafter\def\csname v@h4.c.t1\endcsname{0.50}
\expandafter\def\csname v@h4.c.t2\endcsname{0.30}
\expandafter\def\csname v@h4.c.t3\endcsname{0.20}
\expandafter\def\csname v@h4.d.i.bound\endcsname{0.189}
\expandafter\def\csname v@h4.d.i.ds\endcsname{\ensuremath{+0.124}}
\expandafter\def\csname v@h4.d.i.g\endcsname{\ensuremath{+0.001}}
\expandafter\def\csname v@h4.d.q.bound\endcsname{1.000}
\expandafter\def\csname v@h4.d.q.ds\endcsname{\ensuremath{+0.638}}
\expandafter\def\csname v@h4.d.q.g\endcsname{\ensuremath{-0.003}}
\expandafter\def\csname v@h4.d.s.bound\endcsname{0.352}
\expandafter\def\csname v@h4.d.s.ds\endcsname{\ensuremath{+0.164}}
\expandafter\def\csname v@h4.d.s.g\endcsname{\ensuremath{-0.007}}
\expandafter\def\csname v@h4.e.q.bound\endcsname{0.139}
\expandafter\def\csname v@h4.e.q.dr\endcsname{\ensuremath{+0.014}}
\expandafter\def\csname v@h4.e.q.ds\endcsname{\ensuremath{+0.005}}
\expandafter\def\csname v@h4.evaluable\endcsname{15}
\expandafter\def\csname v@h4.thr.ir4.pts\endcsname{2}
\expandafter\def\csname v@h4.thr.ir4.smd\endcsname{0.05}
\expandafter\def\csname v@h4.thr.ir5\endcsname{10\%}
\expandafter\def\csname v@h4.thr.scf1.p\endcsname{0.90}
\expandafter\def\csname v@h4.thr.scf1.t\endcsname{0.10}
\expandafter\def\csname v@h4.thr.scf2\endcsname{0.05}
\expandafter\def\csname v@h4.thr.scf3\endcsname{80\%}
\expandafter\def\csname v@h4.thr.smi\endcsname{1\%}
\expandafter\def\csname v@h4.thr.tb1\endcsname{0.90}
\expandafter\def\csname v@h4.thr.tb2\endcsname{0.05}
\expandafter\def\csname v@h4.thr.tb4\endcsname{50}
\expandafter\def\csname v@k.ans\endcsname{1{,}500}
\expandafter\def\csname v@k.bud\endcsname{512}
\expandafter\def\csname v@k.bud.max\endcsname{2{,}012}
\expandafter\def\csname v@k.gepa.rows\endcsname{300}
\expandafter\def\csname v@k.k4.h\endcsname{275}
\expandafter\def\csname v@k.metric\endcsname{6{,}000}
\expandafter\def\csname v@k.t1536\endcsname{1{,}536}
\expandafter\def\csname v@k.t512\endcsname{512}
\expandafter\def\csname v@k.t64\endcsname{64}
\expandafter\def\csname v@lit.h4.fails\endcsname{3}
\expandafter\def\csname v@lit.h4.holds\endcsname{6}
\expandafter\def\csname v@lit.h4.inconclusive\endcsname{6}
\expandafter\def\csname v@mx.expected.ok\endcsname{12}
\expandafter\def\csname v@mx.n\endcsname{16}
\expandafter\def\csname v@nc.both.letter\endcsname{310}
\expandafter\def\csname v@nc.both.rows\endcsname{543}
\expandafter\def\csname v@nc.te.age\endcsname{37.3}
\expandafter\def\csname v@nc.te.dec\endcsname{0.022}
\expandafter\def\csname v@numpy\endcsname{2.3.5}
\expandafter\def\csname v@pt.gpu\endcsname{58\%}
\expandafter\def\csname v@pw.h1.i.b\endcsname{0.450}
\expandafter\def\csname v@pw.h1.i.db\endcsname{0.756}
\expandafter\def\csname v@pw.h1.i.l\endcsname{0.000}
\expandafter\def\csname v@pw.h1.i.v\endcsname{0.550}
\expandafter\def\csname v@pw.h1.s.b\endcsname{0.725}
\expandafter\def\csname v@pw.h1.s.db\endcsname{0.055}
\expandafter\def\csname v@pw.h1.s.l\endcsname{0.040}
\expandafter\def\csname v@pw.h1.s.v\endcsname{0.210}
\expandafter\def\csname v@pw.h2.d\endcsname{0.325}
\expandafter\def\csname v@pw.h2.n\endcsname{527}
\expandafter\def\csname v@pw.h2.unadj\endcsname{0.52}
\expandafter\def\csname v@pw.h2.z\endcsname{2.89}
\expandafter\def\csname v@pw.s2.acc\endcsname{0.3}
\expandafter\def\csname v@pw.s2.gap\endcsname{0.02}
\expandafter\def\csname v@pw.z.unadj\endcsname{1.96}
\expandafter\def\csname v@pw.z9\endcsname{2.77}
\expandafter\def\csname v@resc.n\endcsname{300}
\expandafter\def\csname v@rp.rg.share\endcsname{58.2\%}
\expandafter\def\csname v@rp.rs.share\endcsname{67.4\%}
\expandafter\def\csname v@rp.tg.common\endcsname{36{,}080}
\expandafter\def\csname v@rp.tg.finish\endcsname{72}
\expandafter\def\csname v@rp.tg.share\endcsname{57.6\%}
\expandafter\def\csname v@rp.ts.share\endcsname{67.2\%}
\expandafter\def\csname v@rw.calls\endcsname{1{,}006}
\expandafter\def\csname v@rw.cap\endcsname{200}
\expandafter\def\csname v@rw.date\endcsname{2026-10-03}
\expandafter\def\csname v@rw.liu.from\endcsname{23.81}
\expandafter\def\csname v@rw.liu.to\endcsname{13.50}
\expandafter\def\csname v@rw.pr41674\endcsname{41674}
\expandafter\def\csname v@rw.pr41674.date\endcsname{2026-05-15}
\expandafter\def\csname v@rw.win\endcsname{2024-01 to 2026-10-03}
\expandafter\def\csname v@s2.add.last\endcsname{9}
\expandafter\def\csname v@s2.date\endcsname{7 October 2026}
\expandafter\def\csname v@s2.gpu\endcsname{RTX PRO 6000}
\expandafter\def\csname v@s2.recheck\endcsname{900}
\expandafter\def\csname v@s2.shut\endcsname{8 October 2026}
\expandafter\def\csname v@s2.time\endcsname{00:28:49}
\expandafter\def\csname v@seed.h3.cells\endcsname{12}
\expandafter\def\csname v@seed.re.i.gs.t512.dtrunc\endcsname{\ensuremath{+0.098}}
\expandafter\def\csname v@seed.rest.max\endcsname{0.023}
\expandafter\def\csname v@seed.te.i.gs.h3.dt\endcsname{\ensuremath{+0.581}}
\expandafter\def\csname v@seed.te.i.gs.t512.dtrunc\endcsname{\ensuremath{+0.097}}
\expandafter\def\csname v@seed.te.i.gs.t512.trunc\endcsname{0.619}
\expandafter\def\csname v@seed.te.i.lb.h3.dt\endcsname{\ensuremath{+0.963}}
\expandafter\def\csname v@seed.te.i.lb.t512.trunc\endcsname{0.997}
\expandafter\def\csname v@seed.te.q.lb.bud.dacc\endcsname{\ensuremath{-0.018}}
\expandafter\def\csname v@seed.te.q.lb.t512.dacc\endcsname{\ensuremath{-0.013}}
\expandafter\def\csname v@seed.te.q.mm.h3.dt\endcsname{\ensuremath{+0.068}}
\expandafter\def\csname v@sens.flagged.h4.n\endcsname{2}
\expandafter\def\csname v@sens.l30.n\endcsname{2}
\expandafter\def\csname v@sens.never.n\endcsname{3}
\expandafter\def\csname v@sens.strict.n\endcsname{2}
\expandafter\def\csname v@sens.strict.rows\endcsname{31}
\expandafter\def\csname v@sens.v05.n\endcsname{1}
\expandafter\def\csname v@smp.i.k\endcsname{50}
\expandafter\def\csname v@smp.i.p\endcsname{1.0}
\expandafter\def\csname v@smp.i.temp\endcsname{0.8}
\expandafter\def\csname v@smp.q.k\endcsname{20}
\expandafter\def\csname v@smp.q.p\endcsname{0.95}
\expandafter\def\csname v@smp.q.temp\endcsname{0.6}
\expandafter\def\csname v@srv.te.s.bud.calls\endcsname{7{,}216}
\expandafter\def\csname v@srv.te.s.bud.rb\endcsname{4{,}384}
\expandafter\def\csname v@tied.maxdiff\endcsname{0.009}
\expandafter\def\csname v@tied.rows\endcsname{8{,}305}
\expandafter\def\csname v@tied.same.max\endcsname{0.741}
\expandafter\def\csname v@tied.same.min\endcsname{0.656}
\expandafter\def\csname v@tied.units\endcsname{1{,}188}
\expandafter\def\csname v@v05.band\endcsname{30--45}
\expandafter\def\csname v@v05.band.acc\endcsname{0.493}
\expandafter\def\csname v@v05.cfg.max\endcsname{0.493}
\expandafter\def\csname v@v05.cfg.min\endcsname{0.471}
\expandafter\def\csname v@v05.e2.est\endcsname{\ensuremath{+0.012}}
\expandafter\def\csname v@v05.e2.hi\endcsname{\ensuremath{+0.052}}
\expandafter\def\csname v@v05.e2.lo\endcsname{\ensuremath{-0.028}}
\expandafter\def\csname v@v05.freeze\endcsname{26 September 2026, 23:52:06~UTC}
\expandafter\def\csname v@v05.margin\endcsname{0.10}
\expandafter\def\csname v@v05.prior.rows\endcsname{216}
\expandafter\def\csname v@v05.tokens\endcsname{47\%}
\expandafter\def\csname v@ver.one\endcsname{1}
\expandafter\def\csname v@ver.two\endcsname{2}
\expandafter\def\csname v@x.power\endcsname{0.19}
\expandafter\def\csname v@z.cell.te.lb.rows\endcsname{3{,}942}
\expandafter\def\csname v@z.cfg.te.lb.t512.parse\endcsname{0.001}
\expandafter\def\csname v@z.cfg.te.lb.t512.trunc\endcsname{1.000}
\expandafter\def\csname v@z.cnt.all\endcsname{18}
\expandafter\def\csname v@z.cnt.h1.total\endcsname{3}
\expandafter\def\csname v@z.cnt.h123\endcsname{13}
\expandafter\def\csname v@z.cnt.h2.total\endcsname{7}
\expandafter\def\csname v@z.cnt.h3.total\endcsname{3}
\expandafter\def\csname v@z.cnt.lab.replicated\endcsname{13}
\expandafter\def\csname v@z.h1.te.gs.db\endcsname{\ensuremath{-0.342}}
\expandafter\def\csname v@z.h1.te.gs.delta\endcsname{\ensuremath{-0.241}}
\expandafter\def\csname v@z.h1.te.gs.delta.hi\endcsname{\ensuremath{-0.167}}
\expandafter\def\csname v@z.h1.te.gs.delta.lo\endcsname{\ensuremath{-0.315}}
\expandafter\def\csname v@z.h1.te.gs.dv\endcsname{\ensuremath{+0.101}}
\expandafter\def\csname v@z.h1.te.gs.t\endcsname{\ensuremath{+0.293}}
\expandafter\def\csname v@z.h1.te.lb.db\endcsname{\ensuremath{0.000}}
\expandafter\def\csname v@z.h1.te.lb.delta\endcsname{\ensuremath{+0.203}}
\expandafter\def\csname v@z.h1.te.lb.delta.hi\endcsname{\ensuremath{+0.219}}
\expandafter\def\csname v@z.h1.te.lb.delta.lo\endcsname{\ensuremath{+0.186}}
\expandafter\def\csname v@z.h1.te.lb.nb\endcsname{2}
\expandafter\def\csname v@z.h1.te.mm.db\endcsname{\ensuremath{-0.018}}
\expandafter\def\csname v@z.h1.te.mm.delta\endcsname{\ensuremath{+0.402}}
\expandafter\def\csname v@z.h1.te.mm.delta.hi\endcsname{\ensuremath{+0.453}}
\expandafter\def\csname v@z.h1.te.mm.delta.lo\endcsname{\ensuremath{+0.350}}
\expandafter\def\csname v@z.h1.te.mm.dv\endcsname{\ensuremath{+0.419}}
\expandafter\def\csname v@z.h1.te.mm.nb\endcsname{107}
\expandafter\def\csname v@z.h2.te.gs.res.hi\endcsname{\ensuremath{-0.281}}
\expandafter\def\csname v@z.h2.te.gs.res.lo\endcsname{\ensuremath{-0.400}}
\expandafter\def\csname v@z.h2.te.gs.res.theta\endcsname{\ensuremath{-0.340}}
\expandafter\def\csname v@z.h2.te.lb.gc0.hi\endcsname{\ensuremath{-0.311}}
\expandafter\def\csname v@z.h2.te.lb.gc0.lo\endcsname{\ensuremath{-0.362}}
\expandafter\def\csname v@z.h2.te.lb.gc0.theta\endcsname{\ensuremath{-0.337}}
\expandafter\def\csname v@z.h2.te.lb.gc1.hi\endcsname{\ensuremath{-0.314}}
\expandafter\def\csname v@z.h2.te.lb.gc1.lo\endcsname{\ensuremath{-0.369}}
\expandafter\def\csname v@z.h2.te.lb.gc1.theta\endcsname{\ensuremath{-0.342}}
\expandafter\def\csname v@z.h2.te.lb.res.theta\endcsname{\ensuremath{0.000}}
\expandafter\def\csname v@z.h3.te.gs.dp\endcsname{\ensuremath{+0.531}}
\expandafter\def\csname v@z.h3.te.gs.dt\endcsname{\ensuremath{+0.531}}
\expandafter\def\csname v@z.h3.te.lb.dt\endcsname{\ensuremath{+0.974}}
\expandafter\def\csname v@z.h3.te.mm.dp\endcsname{\ensuremath{+0.782}}
\expandafter\def\csname v@z.h4.a.dr\endcsname{\ensuremath{+0.404}}
\expandafter\def\csname v@z.h4.a.ds\endcsname{\ensuremath{+0.430}}
\expandafter\def\csname v@z.h4.a.g.hi\endcsname{\ensuremath{+0.096}}
\expandafter\def\csname v@z.h4.a.g.lo\endcsname{\ensuremath{-0.151}}
\expandafter\def\csname v@z.h4.b1.changed\endcsname{0.205}
\expandafter\def\csname v@z.h4.b1.dr\endcsname{\ensuremath{-0.090}}
\expandafter\def\csname v@z.h4.b1.ds\endcsname{\ensuremath{-0.097}}
\expandafter\def\csname v@z.h4.b1.startparse\endcsname{0.345}
\expandafter\def\csname v@z.h4.b2.bound\endcsname{0.237}
\expandafter\def\csname v@z.h4.b2.ds\endcsname{\ensuremath{+0.237}}
\expandafter\def\csname v@z.h4.c.dr\endcsname{\ensuremath{+0.082}}
\expandafter\def\csname v@z.h4.c.ds\endcsname{\ensuremath{+0.034}}
\expandafter\def\csname v@z.h4.d.ds\endcsname{\ensuremath{+0.239}}
\expandafter\def\csname v@z.h4.evaluable\endcsname{5}
\expandafter\def\csname v@z.level.h1\endcsname{99.44}
\expandafter\def\csname v@z.level.h2\endcsname{99.62}
\expandafter\def\csname v@z.level.h3\endcsname{99.44}
\expandafter\def\csname v@z.level.h4\endcsname{99.69}
\expandafter\def\csname v@z.sens.flagged.h4.n\endcsname{0}
\expandafter\def\csname v@z.sens.other.max\endcsname{0}
\expandafter\def\csname v@z.sens.strict.n\endcsname{1}
\expandafter\def\csname v@z.sens.strict.rows\endcsname{13}

\newcommand{\qwen}{\mbox{Qwen3-1.7B-Longevity}}
\newcommand{\smol}{\mbox{SmolLM3-3B}}
\newcommand{\intern}{\mbox{Intern-S1-mini}}
\newcommand{\basem}{\mbox{base Qwen3-1.7B}}
\newcommand{\cfg}[1]{\mbox{\textsf{\small #1}}}
\newcommand{\CI}[2]{[#1,\,#2]}

\newcommand{\verdict}[1]{\textit{#1}}
\newcommand{\harnessneed}{\needspace{12\baselineskip}}

\title{What Does a Harness Repair?\\ Four Preregistered Studies Around a Longevity-Tuned Language Model}
\author{\name Bowen Xu$^{*}$\textsuperscript{\dag} \email bxu666@stanford.edu \\
\addr Department of Biology, Stanford University, Stanford, CA, USA
\AND
\name Boyu Chen$^{*}$ \\
\addr Institute of Health Informatics, University College London, London, UK \\
$^{*}$Equal contribution. \textsuperscript{\dag}Corresponding author.}

\begin{document}

\maketitle

\begin{abstract}
A harness search around a frozen longevity-tuned 1.7B model reported a rescue.
An audit traced its gain to rows the start left unparsed and found that the caller sent reference answers as assistant turns; replication through the same pipeline missed this defect.
We preregistered four studies and report every endpoint.
In the first two, thinking off raised test accuracy over capped thinking in \V{cb.h1.visibility} of \V{cb.fam1} cells, mainly by parsing rows the cap left unparsed, and was not meaningfully worse than a rescue in \V{cnt.h2.res.holds} of \V{c.fam1} small-model test cells.
Replication reproduced an injected defect's distortion in \V{cb.h4.holds} of \V{cb.h4.evaluable} evaluable cases.
At 14B, two GEPA harnesses beat thinking off by \V{ab.h2.te.0} test points but were not detectably above the strongest constant label.
In the third study, GEPA beat the constant on FPB (confirmatory), thinking off was not meaningfully worse than any BoolQ harness, and \V{e3.cnt.seed.mm} of \V{e3.cnt.mm.runs} MMLU-Pro searches kept their seed; the fourth comparison was inconclusive on test.
On exploratory Autocast, harnesses were within \V{e3.rng.au.dc.te.max} points of the constant and gave one label in \V{e3.rng.au.mj.te.min}\% to \V{e3.rng.au.mj.te.max}\% of parsed answers on test.
On the prediction tasks (14B mortality, Autocast) the harnesses ended near the constant label.
The fourth study varied search-set skew and objective on CREAK and resolved Manifold questions.
All its results are descriptive because render hashes were not committed before the first replication call (D001).
Minority-recall estimates were lower at \V{e4.ratio.skew} than at \V{e4.ratio.bal} in all four test cells.
The balanced objective and re-selection from each search's pool had higher minority- and lower majority-recall estimates in all four cells.
Five of their eight minority-recall intervals included zero, as did all eight balanced-accuracy contrast intervals.
This does not show that skew explains the earlier collapse.
\end{abstract}

\section{Introduction}
\label{sec:intro}

\paragraph{A rescue that was not one.}
A harness is everything around a frozen model that changes what it is scored on: the prompts, reasoning switches, token budgets, retry rules and parsers.
Harness search proposes a change and keeps it if the score rises \citep{khattab2023dspy,agrawal2025gepa,metaharness2026}.
Self-improving systems use the same step: propose a change to their own scaffolding, score it, and keep it if the score improves \citep{dgm2025,modularrsi2026}.
Our starting point is such a search around a frozen longevity-tuned language model, \qwen\ (Appendix~\ref{app:v05}).
It started from a configuration that thinks within a small token budget and kept a conditional rescue: when the first call had used nearly all its tokens, call again with a no-think suffix.
The rescue's gain replicated on a second shard.
Later audits found that the starting configuration parsed none of the \V{bg.rows} rows of the corrected re-evaluation, that a thinking-off setting the search had never tried was non-inferior to the rescue at less than half the token cost, and that the caller had sent each benchmark row's reference answer to the model as an earlier assistant turn.
Replication could not have shown the last, because every call carried it; trivial baselines, input audits and prompt-token reconciliation did.
The rescue's gain came from rows the starting configuration left unparsed.

\begin{figure}[t]
\centering
\includegraphics[width=\textwidth]{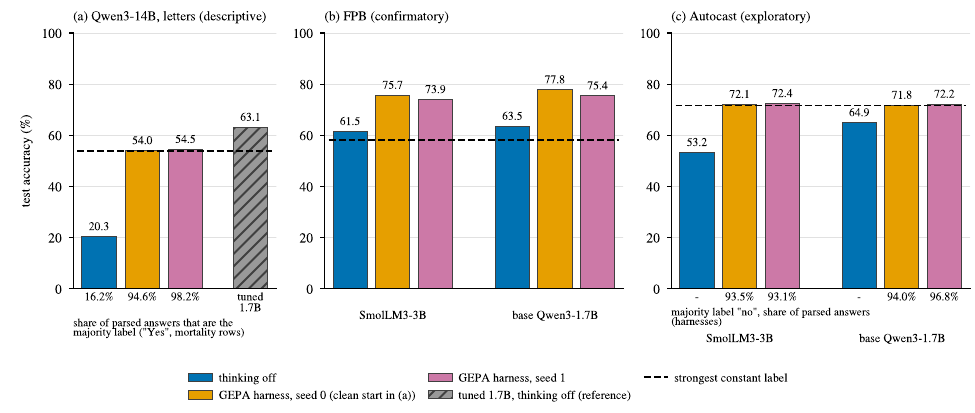}
\caption{Test accuracy of thinking off and of two GEPA harnesses against a constant label (dashed line); each panel carries its status. Panel (a) uses the strongest constant label on the test partition, as in the first study's analysis; panels (b) and (c) use the constant chosen on the search partition, which on test is also the most common label. (a)~Qwen3-14B on the LongevityBench letters (descriptive), with the tuned 1.7B model's thinking off for reference (the second study's rerun); the share under each bar is the share of parsed answers on the mortality rows that are ``Yes'', the majority label. (b)~FPB, \smol\ and \basem\ (confirmatory). (c)~Autocast, the same two models (exploratory); the share under each harness is its share of parsed answers that are ``no'', the majority label (not recorded for thinking off).}
\label{fig:open}
\end{figure}

\paragraph{The opposite case.}
On the same benchmark a general model behaves the other way (Figure~\ref{fig:open}a).
With thinking off, Qwen3-14B scores \V{fu.te.aut.acc} on the LongevityBench letters (test partition), far below the strongest constant label, which scores \V{fu.te.base.strong}; \qwen\ scores \V{ar.te.lb.t} in the same setting (the second study's rerun).
Two GEPA harnesses that start from an empty prompt gain \V{ab.h2.te.0} points on test over the 14B's thinking-off setting (\V{ab.h2.re.0} and \V{ab.h2.re.1} on replication), and neither is detectably above the constant: their test accuracy is \V{fu.te.gc0.acc} and \V{fu.te.gc1.acc}.
On the mortality task they answer ``Yes'' in \V{cl.te.l42.h0.share} and \V{cl.te.l42.h1.share} of their parsed answers on test (\V{cl.re.l42.h0.share} and \V{cl.re.l42.h1.share} on replication), the constant's answer, where the thinking-off setting answers ``Yes'' in \V{cl.te.l42.aut.share} on test.
In \qwen\ the thinking-off baseline already beat the constant and no harness was detectably better; in the 14B the search's large gain is a gain over a baseline that sits below the constant.
The comparison with the constant is descriptive in the second study, whose confirmatory endpoints compare harnesses with the baseline.

\paragraph{Why this matters beyond the domain.}
Domain-tuned small models are common parts of agent systems.
What a harness can repair depends on what the model already does without it, and tuning on a domain changes exactly that, so the same search result can mean opposite things for a tuned and for a general model.
The defects of the case study are generic ones of a pipeline (a caller, a scorer, a split, a served checkpoint), and every round of a self-improving loop is a selection of the kind that its search made.
Nothing here is evidence about aging biology.

\paragraph{What a replicated gain leaves open.}
A gain that replicates on untouched rows shows that the selected harness beat its parent by more than selection noise.
It does not show that the harness answers better than a constant answer would, and it leaves three further questions open.
The first is visibility: did the answers improve, or did rows that the parent left unparsed, because it ran out of tokens or never committed to an answer, become parsed?
The second is baseline adequacy: would a simple setting that the search never tried do as well, and does the model beat trivial baselines at all?
The third is evaluation validity: are the inputs, the gold labels, the split and the served model what they are said to be?
Each question has a literature (Section~\ref{sec:related}), and most of it works after the fact.
Harness and scaffold gains vanish against matched-budget test-time scaling and simple baselines \citep{wang2026rethinking,gideoni2026simple,liang2026auditing,kapoor2024agents}.
A gain can be completion rather than correctness \citep{seyoum2026hardware,lee2026repair,chen2026calibration}, and thinking off can beat a thinking cap \citep{nie2026coupling}.
Evaluation pipelines carry defects that move scores by tens of points \citep{zhu2025abc,berriche2026pipeline}.
We found no earlier study that fixes in advance, for a stated contrast, how much of a harness gain is visibility, none that tests the benchmark authors' own protocol against searched harnesses with a margin, and none that injects known defects one at a time to ask whether replication reproduces them.
The search behind these statements covered \V{rw.win} and was capped at \V{rw.cap} web searches, and three of its six parts had to be resumed after a rate limit, so these statements are bounded (Appendix~\ref{app:related}).

\paragraph{Four studies.}
We ask what a harness repairs in four preregistered studies, each frozen before any call on its replication and test partitions.
The first, frozen on \V{fr.date} at \V{fr.time}~UTC, uses three models of 1.7B to 8B parameters, three benchmark pools, an adaptive arm in which GEPA \citep{agrawal2025gepa} searches for harnesses around the 1.7B model, and six evaluation defects injected one at a time.
The second, frozen on \V{s2.date} at \V{s2.time}~UTC after the first study's results were known, puts the same hypotheses to Qwen3-14B with a GEPA arm of its own, and reruns the base and the tuned version of Qwen3-1.7B on a second GPU as a descriptive comparison.
The third, frozen on \V{e3.s3.date} at \V{e3.s3.time}~UTC, tests whether what we found around the longevity model holds on other benchmarks and models: \smol\ and \basem\ on BoolQ, FPB and MMLU-Pro, a new hypothesis that compares each harness with the constant label, and Autocast as an exploratory arm (Figure~\ref{fig:open}b and c).
The fourth, registered on the Open Science Framework and frozen at \V{e4.osf.reg.time}~UTC on \V{e4.osf.reg.date}, returns to the collapse itself: it varies the class ratio of the search set and the objective of GEPA on CREAK and Manifold with \smol\ and \basem\ (Section~\ref{sec:fourth}).

The paper makes five contributions.
It splits the gain of a thinking-off protocol over a capped thinking configuration exactly into rows only the protocol parses, rows both parse, and rows only the capped configuration parses, the split that \citet{chen2026calibration} use for self-revision, and fixes a dominance statistic and a gain gate in advance.
It tests that protocol for non-inferiority against a derived rescue and against GEPA-selected harnesses, with trivial baselines next to every accuracy and, in the third study, each harness against the constant label, and it tests whether the serving stack's thinking budget, which allows a longer answer as well, lowers truncation.
For each defect and model it fixes one claim in advance, that replication through the same pipeline confirms the defect's distortion, and it measures the false alarms of the audit checks.
It also compares the base and the tuned version of one model, and the same procedure on two GPUs, descriptively.
Its fourth study varies the class ratio and the objective of the search and reports the preregistered intervals as description, because a deviation was recorded after the results were known (Section~\ref{sec:fourth-results}).
All primary endpoints of the four studies are reported with every deviation and addendum, and everything computed after the outputs existed is labelled.

\paragraph{Limits.}
The results cover five models, one of which is the tuned model that the case study searched around, eight benchmarks and six defects, and nothing beyond them; the pattern across the studies rests on few tasks (Section~\ref{sec:pattern}).
We studied one round of selection and no recursive loop.
We do not claim that harness gains can be illusory, that thinking off can beat a thinking cap, or that agent and harness evaluations can be preregistered, since each has been shown or done before, and we do not claim that harness search finds genuine quality gains in general.

\section{Related work}
\label{sec:related}

We organise related work by the questions of Section~\ref{sec:intro}; Appendix~\ref{app:related} states how it was searched and what the search could not cover.
Work posted before the first version of this paper is prior work, and later work is cited as concurrent.

\paragraph{Harness search, and whether its gains are real.}
Prompt optimisers \citep{zhou2023ape,yang2024opro,khattab2023dspy,agrawal2025gepa} and harness search \citep{metaharness2026,modularrsi2026,dgm2025} treat the text and control flow around fixed weights as the object of search, and GEPA is the adaptive arm of our study.
Earlier work has shown, after the fact, that such gains can be illusory: evolved harnesses and scaffolds do not beat matched-budget test-time scaling or simple baselines \citep{wang2026rethinking,gideoni2026simple,liang2026auditing,kapoor2024agents}, a hardware-verification harness gains mostly completion \citep{seyoum2026hardware}, a search started from a defective seed fails \citep{liu2026reflection}, and self-improving loops tamper with or hack their evaluators \citep{wang2026tampering,yang2026hackprobe}.
Concurrent work reaches related conclusions in other settings.
In \V{e4.rw.ding.n} settings with significant search-set gains, \citet{ding2026hda} attribute an average of \V{e4.rw.ding.a}\% to overfitting and \V{e4.rw.ding.b}\% to test-time scaling.
With token budgets matched and scorer exploits removed, \citet{oved2026tokens} find that evolution's advantage over independent sampling largely vanishes.
\citet{tian2026ceiling} distinguish process failures, such as blocked calls and exhausted step budgets, from content failures, and argue that the latter call for weight training.
\citet{tomar2026textbook} reports three preregistered rounds with coding agents and a frontier model.
Raising an output-token limit, together with three other changes, recovers about \V{e4.rw.tomar90}\% of the best loop agent's gain.
This resembles the capped-output mechanism in our first study.
Tomar cites the first two versions of this paper but describes their result more broadly, as search mostly repairing a capped-thinking mismatch; our 14B results do not support extending that explanation to every harness gain.
Recursive harness self-improvement is now studied directly.
RRSI regularises which edits are proposed and kept, because recursive evolution can overfit its training tasks, and reports gains on its evolution split and on out-of-distribution benchmarks \citep{xia2026rrsi}.
Held-out and out-of-distribution checks test whether a gain generalizes, not whether the evaluation pipeline that produced it is sound, which is the question of our H4; we did not examine RRSI's pipeline and do not say that it has a defect.
\citet{lee2026rhsi} score harness revisions with an LLM judge on generated tasks and report no audit of the judge or of the validity of its tasks.
Concurrent work evaluates candidate harnesses with tests that an LLM generates, without ground-truth solutions (its abstract does not mention simple baselines) \citep{pan2026harnesspro}, and asks with pass@k analyses whether harness gains elicit what a model could already reach or expand what it can solve \citep{kim2026elicitation}; the abstract of the second does not mention a visibility split, constant baselines, preregistration or defect injection.
Our question is narrower: where the gain of one protocol comes from, and whether a known defect would be seen, with the questions and decision rules fixed in advance.

\paragraph{Selection on a reused set, and the class ratio of the search set (the fourth study).}
Selecting candidates using noisy scores can make their selection-set performance look too high.
\citet{hu2026winners} treat the keep-if-better step of a self-improvement loop as selection under measurement noise and find that the final selection-set score of greedy loops exceeded held-out accuracy by \V{e4.rw.hu.lo} to \V{e4.rw.hu.hi} points with \V{e4.rw.hu.n16} selection items and by far less with \V{e4.rw.hu.n256}, and that the validation scores of GEPA and MIPROv2 overstate held-out gains in the same way.
Their lever is the size of the selection set.
Our fourth study fixes the size and varies the class ratio of the search set and the objective.
The papers in our reading notes do not report a class-ratio intervention or a comparison of accuracy and balanced-accuracy objectives, so it asks a different question.
We evaluate the selected candidates on held-out partitions, and the re-selection compares two candidates of one search by balanced validation accuracy.
Audits of repeated full improvement loops remain uncommon: \citet{chen2026product} find that \V{e4.rw.chen.p}\% of the \V{e4.rw.chen.rsi} studies that describe their systems as recursive self-improvement audit two or more reruns of the full loop (of \V{e4.rw.chen.n} studies coded).
Our fourth study runs \V{e4.d.reps} searches per condition and reports their spread.

\paragraph{Visibility and quality (H1).}
\citet{chen2026calibration} split the accuracy change of a self-revision step into a content margin, a format-recovery margin, a format-loss margin and a dead cell, which is our split under other names but without a predefined dominance threshold.
\citet{nie2026coupling} and \citet{kaiser2026beyond} give the identity for a single configuration or between models, completion rate times conditional accuracy; under a shared budget thinking off beats capped thinking \citep{nie2026coupling,ma2025nothinking}, and capped single-call evaluation of hybrid models has been called an artifact of protocol \citep{lee2026dart}.
Other work finds that unparsed answers hide correct ones \citep{garware2026ruler}, separates traces cut by a cap from traces that finished without an answer \citep{goyal2026cap}, reads a harness-evolution gain informally as completion rather than correctness on the twelve tasks that had been searched \citep{seyoum2026hardware}, and counts repaired abstentions for constrained decoding \citep{lee2026repair}.
We add a pre-specified dominance statistic and gain gate, applied prospectively to the contrast between a thinking-off protocol and a capped thinking configuration.

\paragraph{Baseline adequacy and constant labels (H2, H5).}
Several studies find, after the fact, that simple baselines match searched or evolved harnesses and complex agents \citep{kapoor2024agents,wang2026rethinking,gideoni2026simple,liang2026auditing,changchen2026npo}, and a concurrent study finds the same for reported gains on text-attributed graphs, together with evaluation risks in released code (test-set selection, test labels in training, a loader misalignment) \citep{podstawski2026graphllm}.
A preregistered non-inferiority test of a simple alternative exists for agent memory \citep{sharma2026selection}, and equivalence margins have been used for scaffolds \citep{gringras2026scaffolding} and for compression \citep{singh2026certifying}.
Majority-label bias is known in few-shot prompting, where the bias is the model's and calibration removes it \citep{zhao2021calibrate}; here a search converges on the constant label while its gain over the start looks real, so the question is what the gain is measured against.
Our tests fix the margin in advance for the benchmark authors' own protocol, against a derived rescue and GEPA-selected harnesses, and compare each harness in the third study with the constant label.

\paragraph{Budget forcing (H3).}
Budget forcing and the Qwen3 thinking budget \citep{muennighoff2025s1,yang2025qwen3} serve as evaluation controls, with a forced answer or a forced end of thinking \citep{nie2026coupling,carlon2026earlier,lee2026dart,changchen2026npo}, and one study forces the answer delimiter at a cap in vLLM \citep{goyal2026cap}.
A concurrent study reads truncated reasoning traces with a separate decision model instead of forcing an answer \citep{wang2026truncated}.
We test the engine's thinking-token budget \citep{kwon2023vllm} as a harness setting against a capped configuration; the technique is not ours.

\paragraph{Defects and prospective replication (H4).}
Other work injects faults with known ground truth and measures false alarms: tool bugs in agent benchmarks \citep{bellibatlu2026contract}, tampering edits \citep{wang2026tampering}, reward-hacking channels \citep{yang2026hackprobe} and planted gains \citep{shawn2026pace}.
Defects like ours have been catalogued as they occurred, for answer-key access, label defaults, parser fallbacks, stop sequences, denominators and gold errors \citep{wahi2026judge,berriche2026pipeline,zhu2025abc}, and the reference scorer of MMLU-Pro credits a random option for an empty prediction.\footnote{\texttt{evaluate\_from\_local.py}, function \texttt{save\_res}, in the repository TIGER-AI-Lab/MMLU-Pro, read on \V{rw.date}.}
We found no study that injects pipeline defects one at a time into a harness-search pipeline and tests, for each defect and model, whether replication through the same pipeline reproduces the distortion.

\paragraph{Served-model identity and determinism.}
Serving stacks change outputs at fixed weights \citep{pape2026silent,wang2026interface,yuan2025fp32,he2025nondeterminism}, answers can flip while accuracy stays flat \citep{dutta2024flips,kubler2026degradation}, and a served route is not its model identifier \citep{stenstrom2026ib2,gao2025equality}.
vLLM used to decide tying from the configuration alone and silently discarded a stored head; its fix \citep{vllm51665} is in the version we used (vLLM~\V{d.vllm}), which loads the stored head.
We add a measured instance of the mismatch, since the mechanism is known.

\paragraph{GEPA from a broken start.}
\citet{liu2026reflection} report that GEPA, started from its official GSM8K seed (which puts the answer before the reasoning), falls from \V{rw.liu.from}\% to \V{rw.liu.to}\% and never repairs the order, and \citet{ursekar2026harnessopt} start optimisers from deliberately untuned seeds, one of them a non-functional stub.
Ours is a controlled comparison of clean and broken starts; why a search repairs one defect and not another is not tested here.

\paragraph{Preregistration and prospective design.}
Preregistered or frozen designs with deviation logs, holdouts or equivalence margins appear in \citet{gringras2026scaffolding,weinberger2026waste,sharma2026selection,lee2026repair,goyal2026cap,liang2026auditing,bellibatlu2026contract,singh2026certifying,tomar2026textbook}.
\citet{vaccaro2026preregister} argues in a position paper that experiments with AI agents should be preregistered and gives a template; our studies are an empirical case of preregistered rules frozen before the run with every deviation logged, and we did not follow that template.
\citet{thomas2026next} confirm on the first eligible model released after the plan, a stronger standard than ours, since our freeze guards the rules and not model-specific tuning.
We found no earlier study that combines the freeze with its hash and time in the commit, a pilot, replication then test, an adaptive arm before replication, every primary endpoint of the preregistrations reported, addenda that state what had been seen, and an exposure log.

\paragraph{Hardware and token-level agreement.}
Numerical differences between hardware change the tokens a model generates, even at zero temperature \citep{yuan2025fp32,he2025nondeterminism}.
\citet{cooper2026nondeterminism} report, in their abstract, fixed-reduction-order kernels whose linear-layer outputs are bitwise identical across three NVIDIA architectures; our stack has no such kernels, and on a second GPU \V{pt.gpu} of the tuned model's token sequences changed while the verdicts did not, whereas \smol's correctness readings and parsed labels on MMLU-Pro agreed with the earlier study's on \V{e3.ag.te.sm.au.nag} of \V{e3.ag.te.sm.au.rows} test rows on the same kind of GPU (Section~\ref{sec:third-repro}).
Our comparison of the same procedure on two GPUs asks whether the conclusions of a study depend on that agreement, which is a different question from whether the outputs match.

\section{Design}
\label{sec:design}

The design is that of two frozen preregistrations, the first study's and the second study's (hashes and times in Appendix~\ref{app:repro}); the second reuses the first's rules and changes the model and the families (Section~\ref{sec:second}), and this section states only what the results need.

\subsection{Setting}
\paragraph{Models and stack.}
The first study uses \qwen\ \citep{zhavoronkov2026cell}, a Qwen3 \citep{yang2025qwen3} model tuned on aging-biology data, \smol\ \citep{smollm3} and \intern\ \citep{bai2025interns1}.
The second study uses Qwen3-14B and, descriptively, the base and the tuned version of Qwen3-1.7B (Section~\ref{sec:second}).
The models of both studies ran on vLLM~\V{d.vllm} \citep{kwon2023vllm} in batch-invariant mode \citep{he2025nondeterminism} at zero temperature and client concurrency \V{d.conc}, the first study on one RTX~5090 and the second on an \V{s2.gpu}, with three exceptions: the GEPA proposer and the seeded-sampling arms sample at the model cards' values, and the FreeToken \citep{yang2026freetoken} continuity runs of the tied-output-layer defect (E) ran on FreeToken 0.1.3 instead of vLLM, at temperature 0 and client concurrency 1, in a context of \V{ft.ctx} tokens.
The \qwen\ checkpoint declares tied embeddings but stores a different output matrix, and the study model uses the stored one (untied).
Two models of the first study and the two 1.7B arms of the second also have a descriptive seeded-sampling arm.

\paragraph{Pools, cells and partitions.}
The LongevityBench letter pool combines, by row, LB-0030 (age band, five options), LB-0034 (which of two participants is older) and LB-0042 (ten-year survival), and LB-0038, an age regression, is reported apart.
MMLU-Pro \citep{wang2024mmlupro} (a stratified subset of \V{d.mm.sub} items) and GSM8K \citep{cobbe2021gsm8k} are the other two pools, and a \emph{cell} is a model with a pool, so there are nine cells in the first study and three in the second, which uses the same items, partitions and analysis rows.
The items are split into search, replication and test partitions (\V{d.split}); LongevityBench's \V{d.lb.rows} rows, in \V{d.lb.comp} participant components, are split by component so that no participant spans partitions, and GEPA searches on the search partition only.
The \emph{analysis rows} leave out \V{excl.re.v21_holdout} replication and \V{excl.te.v21_holdout} test rows that the case study had held out and later called, and \V{excl.re.reading_exposed} replication rows whose reference answers could be recovered from an upload made by mistake; putting them back is a sensitivity analysis.
The test cells then have \V{cell.te.q.lb.rows} letter rows (\V{cell.te.q.lb.units} participant components, the bootstrap unit), \V{cell.te.q.mm.rows} MMLU-Pro items and \V{cell.te.q.gs.rows} GSM8K items, and the replication cells have \V{cell.re.q.lb.rows}, \V{cell.re.q.mm.rows} and \V{cell.re.q.gs.rows}.

\subsection{Configurations and scoring}
Table~\ref{tab:configs} lists the six configurations of the grid and the GEPA harnesses.
A call is \emph{unparsed} if the scorer cannot read an answer from it, and it then scores as wrong, so an unparsed answer may hide a correct answer, a wrong one or none (Appendix~\ref{app:scorer} states the scorer's rules).
A call is \emph{truncated} if it ended at the token limit (finish reason \texttt{length}), whatever the scorer makes of it, and \emph{non-committed} if it ended by itself (finish reason \texttt{stop}) and is unparsed; for \cfg{RESCUE-1500} both properties are those of the call that gives the answer.

For H1, let P be the baseline configuration and R the comparison, paired on the $n$ rows of a cell, with $u$ the parse indicator and $c$ the correct indicator of a call and a bar negating an indicator.
Row $i$ is in V if only R parses it ($\bar u_{P,i}u_{R,i}$), in B if both parse it ($u_{P,i}u_{R,i}$) and in L if only P parses it ($u_{P,i}\bar u_{R,i}$).
With $\Delta=\sum_i(c_{R,i}-c_{P,i})/n$ and the contributions $\Delta_V=\sum_iV_ic_{R,i}/n$ (P scores zero on V), $\Delta_B=\sum_iB_i(c_{R,i}-c_{P,i})/n$ and $\Delta_L=-\sum_iL_ic_{P,i}/n$, the identity $\Delta=\Delta_V+\Delta_B+\Delta_L$ is exact, because a row that neither configuration parses adds nothing.
Each contribution is divided by the cell's $n$ and not by the size of its row set, as in the tables, so $\Delta_V$ is R's number of correct answers on V per row of the cell.
\citet{chen2026calibration} use this additive split for self-revision (content, format-recovery and format-loss margins); the dominance statistic, its fixed share and the gain gate are ours.
The dominance statistic $T=\Delta_V-\V{c.share}\,(\Delta_V+\Delta_B)$ is positive when visibility supplies more than \V{c.share} of what the rows R parses contribute.
The split locates the gain; it does not show that a correct answer was already there and only hidden, since the two configurations generate differently.

\begin{table}[t]
\caption{Configurations (``tokens'' is \texttt{max\_tokens}; thinking on or off is set in every request). RESCUE-1500 is derived from the two grid runs and verified against the real two-call sequence on \V{d.resc.n} items per model.}
\label{tab:configs}
\centering
\footnotesize
\setlength{\tabcolsep}{3pt}
\begin{tabular}{@{}>{\raggedright\arraybackslash}p{0.37\columnwidth}>{\raggedright\arraybackslash}p{0.57\columnwidth}@{}}
\toprule
Configuration & Request \\
\midrule
\cfg{THINK-64}, \cfg{THINK-512}, \cfg{THINK-1536} & thinking on; \V{k.t64}, \V{k.t512} or \V{k.t1536} tokens for thinking and answer together (\cfg{THINK-64} is the case study's parent) \\
\cfg{AUTHORS-1500} & thinking off; \V{k.ans} tokens (the benchmark authors' serving protocol) \\
\cfg{BUDGET-512-1500} & thinking on; vLLM ends thinking after \V{k.bud} tokens \citep{muennighoff2025s1}; up to \V{k.ans} tokens of answer follow, so \V{k.bud.max} tokens in all against \V{k.t512} for \cfg{THINK-512} \\
\cfg{RESCUE-1500} & \cfg{THINK-512}; where its call ended at the token limit, the \cfg{AUTHORS-1500} call on the same row (the call that gives the answer decides the properties of the row) \\
GEPA harnesses & in each study four runs (clean or broken start, two seeds), around \qwen\ in the first and Qwen3-14B in the second, \V{k.metric} metric calls on \V{k.gepa.rows} search rows, \intern\ as proposer; the selected harness edits the system prompt and appends an instruction, so the arm searches prompt text only (Appendix~\ref{app:gepa}) \\
\bottomrule
\end{tabular}
\end{table}

\subsection{Hypotheses and endpoints}
Table~\ref{tab:hyp} states the four hypotheses.
The first study has \V{cnt.all} primary endpoints and the second \V{z.cnt.all}, \V{cb.all} in all; they are the only confirmatory results, and their verdicts come from percentile-bootstrap intervals adjusted within each family.

\begin{table*}[t]
\caption{Hypotheses, estimands and verdict rules. $m$ is the minimal difference and $\delta$ the non-inferiority margin, both \V{c.margin}. Every bound in a verdict is a bound of the family-adjusted interval, and H1's gate uses the same resamples and the same adjusted level as $T$. The last column gives the first study's family and the second study's number of endpoints: the second study has \V{z.cnt.h1.total}, \V{z.cnt.h2.total}, \V{z.cnt.h3.total} and \V{z.h4.evaluable} endpoints in the four families and is tested at the first study's family sizes and levels (Section~\ref{sec:second}).}
\label{tab:hyp}
\centering
\footnotesize
\setlength{\tabcolsep}{4pt}
\begin{tabular}{@{}l>{\raggedright\arraybackslash}p{0.32\textwidth}>{\raggedright\arraybackslash}p{0.34\textwidth}>{\raggedright\arraybackslash}p{0.22\textwidth}@{}}
\toprule
 & Claim and estimand & Verdict & Endpoints (family) \\
\midrule
H1 & Most of the accuracy that \cfg{AUTHORS-1500} gains over \cfg{THINK-512} comes from rows \cfg{THINK-512} leaves unparsed. Gate: $\Delta$; statistic: $T$. & \verdict{no gain: not tested} unless the lower bound of $\Delta$ is above zero; if it is, \verdict{visibility dominates} if the lower bound of $T$ is above zero, \verdict{quality matters} if the upper bound of $T$ is below zero, \verdict{inconclusive} otherwise. & \V{c.fam1}: one per cell; second study \V{z.cnt.h1.total} \\
H2 & \cfg{AUTHORS-1500} is not meaningfully worse than a comparator: $\theta=\mathrm{acc}(\text{\cfg{AUTHORS-1500}})-\mathrm{acc}(\text{comparator})$. & \verdict{holds} if the lower bound of $\theta>-\delta$; \verdict{fails} if the upper bound $<-\delta$; else \verdict{inconclusive}. & \V{c.fam2}: \cfg{RESCUE-1500} in nine cells; four GEPA harnesses (\qwen, letters); second study \V{z.cnt.h2.total}: \cfg{RESCUE-1500} in three cells, four harnesses (Qwen3-14B, letters) \\
H3 & Against \cfg{THINK-512}, \cfg{BUDGET-512-1500} (a forced end of thinking and a larger allowance) lowers truncation ($D_T$) and raises parsing ($D_P$). & \verdict{holds} if both lower bounds $>m$; \verdict{fails} if either upper bound $<m$; \verdict{no room} if the upper bound of \cfg{THINK-512}'s truncation is $<m$ (applied first); else \verdict{inconclusive}. & \V{c.fam3}: one per cell; second study \V{z.cnt.h3.total} \\
H4 & For each defect and model, prospective replication confirms the defect's distortion (Section~\ref{sec:h4design}). & \verdict{holds}, \verdict{fails} or \verdict{inconclusive} from three intervals; \verdict{not evaluable} if the defect cannot move the start by more than $m$. & \V{c.fam4} pairs, fixed; second study \V{z.h4.evaluable} pairs \\
\bottomrule
\end{tabular}
\end{table*}

\subsection{Defects, claims and checks (H4)}
\label{sec:h4design}
Six defects arose in the case study, and we inject each into a harness-search pipeline, one at a time, and compare the result with the clean pipeline on the same rows.
The pipeline's start $s$ is \cfg{THINK-1536}, and the reported quantity is the start's accuracy on the letter pool, measured on a selection sample $S$ (the replication partition) and on a check sample $R$ (the test partition); for the broken start it is the gain over the start.
Let $D_S$ and $D_R$ be a defect's distortion of the reported quantity on $S$ and on $R$, and $\Gamma=D_R-D_S$.
The defects are the reference answer sent as an assistant turn (A), a gold fallback (B1) and an unparsed-as-A scoring (B2) for LB-0042, the case study's skewed split (C), the broken start \cfg{THINK-64} (D) and the tied output layer served in place of the untied one (E, \qwen\ only, as an engine that decides tying from the configuration alone does \citep{vllm51665}); the second study applies the other five, since Qwen3-14B has one served variant.
For the broken start the selection does not depend on the start, so it cancels, and the distortion is $D_S=\mathrm{acc}_S(\text{\cfg{THINK-1536}})-\mathrm{acc}_S(\text{\cfg{THINK-64}})$, with the same difference on $R$ for $D_R$.
Appendix~\ref{app:h4} gives each defect's construction, clean counterpart, samples and evaluability bound, and each check's predicate and threshold.

The claim for a defect and a model holds if $D_S$ and $D_R$ lie wholly outside $\pm m$ with the same sign and $\Gamma$ lies wholly inside $\pm m$; it fails if any of the three lies wholly on the other side of its boundary, and it is inconclusive otherwise.
A pair is evaluable only if the largest distortion its mechanism can cause on $S$ exceeds $m$, a bound that we compute from parse, gold-change and answer-difference indicators without accuracy; the skewed split changes the sample and has no row-level bound, so it is evaluable by rule.
The family stays at \V{c.fam4} whatever is evaluable.
Five checks read the defects' signatures: per-call prompt-token reconciliation (TR), trivial baselines (TB), the simple configuration family (SCF), input reading (IR) and the gain and level rules of prospective replication (PR), plus served-model identity (SMI) for the tied output layer.
Their detection matrix is largely true by construction and descriptive, and what is empirical is their false alarms on clean pipelines.

\subsection{Statistics and what was fixed in advance}
Every contrast is paired on rows.
In each of $B=\V{c.b}$ bootstrap resamples we draw the cell's units with replacement (participant components for LongevityBench, items for MMLU-Pro and GSM8K), so that a unit drawn $k$ times counts its rows $k$ times; every statistic of the cell is recomputed on the resample, and all contrasts of a cell share it (Appendix~\ref{app:repro} gives the seeds and the index rule).
Every estimate carries an unadjusted \V{c.ci} interval ($q=\V{c.q}$), and a verdict uses the Bonferroni-adjusted interval of its family, with $q=\V{c.q}/m$ for $m$ endpoints (levels \V{c.lvl1}\%, \V{c.lvl2}\%, \V{c.lvl1}\% and \V{c.lvl4}\% for H1--H4).
These levels are nominal: a percentile bootstrap covers only approximately, so the family-wise error rate is controlled approximately and is not guaranteed.
For H4, the resample of an instance's units also serves its clean counterpart where the two share their rows (all defects except the skewed split, which is resampled independently), so that $D_S$ and $D_R$ are paired.
A test verdict is \emph{replicated} if the replication partition gives the same verdict, \emph{not replicated} if it gives the opposite one, and \emph{inconclusive replication} otherwise.
The error control of H4 covers ``holds'' only, and ``fails'' is a descriptive label.
The first preregistration was frozen at \V{fr.time}~UTC on \V{fr.date}, before any replication or test item was called, and the pilot called \V{d.pilot.n} pilot items only; the second was frozen at \V{s2.time}~UTC on \V{s2.date} on the same terms.
The directions of H1 and H3 were visible in the pilot's parse and truncation rates, and the primary set of H1 changed twice before the freeze after those rates were seen (Section~\ref{sec:deviations}).

\subsection{The second study}
\label{sec:second}
The second preregistration was written after the first study's results were known.
It reuses the first study's rules by reference, keeps its hypotheses and verdict rules, and changes the model and the families: Qwen3-14B with \V{z.cnt.h1.total} H1 cells, \V{z.cnt.h2.total} H2 endpoints (the rescue in three cells and four GEPA harnesses on the letters), \V{z.cnt.h3.total} H3 cells and \V{z.h4.evaluable} H4 pairs, \V{z.cnt.all} primary endpoints in all, tested at the first study's family sizes and so at the levels \V{z.level.h1}\%, \V{z.level.h2}\%, \V{z.level.h3}\% and \V{z.level.h4}\%.
Its GEPA arm has the first study's design around Qwen3-14B, and its defect-A pass runs after the test partition, so the checkpoint locked no selection for it and that pair is evaluable by rule.
The checkpoint, which records the four selected harnesses and the replication results before the first test call, is committed and its hash logged before that call.
The two 1.7B arms run the base model and the tuned model on the same items, configurations and schedule as a secondary, descriptive comparison with fixed wording and a fixed limit on its interpretation (Section~\ref{sec:finetune}); they carry no verdict and no adjusted interval.
The second study ran on an \V{s2.gpu} against the first study's RTX~5090, the same chip architecture.

A platform shutdown on \V{s2.shut}, when the account balance ran out, ended the instance, and its data disk was cloned to another host with another \V{s2.gpu}; the 14B's test grid had finished on the first GPU and every other test run ran on the second.
The preregistered rule for a change of the physical GPU was applied: the batch-invariance check was repeated for the three models and passed, and \V{s2.recheck} replication calls of each model, in three configurations, were rerun and were identical token for token, so the confirmatory analysis stands.
The GEPA harness evaluation that the shutdown interrupted was excluded and rerun from scratch; Appendix~\ref{app:deviations} records every deviation of the second study.

\section{What a harness can repair depends on the model and the task}
\label{sec:results}

This section reports the first two studies.
Five models enter it: the tuned \qwen, whose thinking-off setting beats the strongest constant label on the LongevityBench letters; its base model, which scores below that label (a descriptive comparison); Qwen3-14B, which scores below it with thinking off and reaches it with GEPA; and \smol\ and \intern, which leave most capped calls unparsed.
What a harness can repair differs among them.
The third study's models and benchmarks are in Section~\ref{sec:third-results}, and what the three studies together show and do not show is in Section~\ref{sec:pattern}; the fourth study, on the class ratio of the search set, is in Section~\ref{sec:fourth-results}.

Results are on the test partition, the confirmatory one, unless stated, and the replication partition enters through the replication label.
Primary endpoints carry the family-adjusted interval and everything else the unadjusted \V{c.ci} interval; Appendix~\ref{app:tables} gives every endpoint of the first study and Appendix~\ref{app:z} every endpoint of the second.
A star marks an endpoint whose quantity is fixed by an identity of the design or by the observed parse pattern, so that its verdict says little beyond the mechanism; the verdict still depends on the intervals, and the text says why where the endpoint appears.

\subsection{The \V{cb.all} primary endpoints at a glance}
Table~\ref{tab:glance} (Appendix~\ref{app:tables}) gives the \V{cnt.h123} endpoints of H1--H3 of the first study and the \V{z.cnt.h123} of the second, and Table~\ref{tab:h4} (Appendix~\ref{app:h4}) the pairs of H4: \V{c.fam4} in the first study and \V{z.h4.evaluable} in the second.
Of the \V{cnt.h123} endpoints of H1--H3 in the first study, \V{cnt.lab.replicated} are \emph{replicated} and \V{cnt.lab.inconclusive} is an \emph{inconclusive replication}, and the number labelled \emph{not replicated} is \V{cnt.lab.notrepl}; all \V{z.cnt.lab.replicated} of the \V{z.cnt.h123} in the second study are replicated.
Figure~\ref{fig:h123} (Appendix~\ref{app:tables}) plots the adjusted intervals, and the sections below take the endpoints by question.
In the first study no sensitivity analysis changes more than \V{cb.sens.max1} of the \V{cnt.h123} H1--H3 verdicts (Appendix~\ref{app:sens}); in the second the strict scorer changes \V{z.sens.strict.n} of the \V{z.sens.strict.rows}, the rescue on MMLU-Pro (\verdict{holds} becomes \verdict{fails}), and no other analysis changes any (Appendix~\ref{app:z}).

\subsection{Where the gains come from (H1)}
\label{sec:h1}
In the first study \verdict{visibility dominates} in \smol\ on all three pools and in \intern\ on the letters and MMLU-Pro.
In these five cells \cfg{AUTHORS-1500} gains over \cfg{THINK-512} between $\Delta=\V{h1.te.s.lb.delta}$ \CI{\V{h1.te.s.lb.delta.lo}}{\V{h1.te.s.lb.delta.hi}} (\smol, letters) and $\V{h1.te.i.mm.delta}$ \CI{\V{h1.te.i.mm.delta.lo}}{\V{h1.te.i.mm.delta.hi}} (\intern, MMLU-Pro), and the gain is visibility: $\Delta_V$ is at least $\Delta$ in all five (shares from \V{h1.te.i.lb.share} to \V{h1.te.s.gs.share}), because the rows both configurations parse contribute at most $\Delta_B\le\V{h1.te.s.lb.db}$ and the rows only \cfg{THINK-512} parses cost $\Delta_L$ (for \smol\ on GSM8K, $\Delta_V=\V{h1.te.s.gs.dv}$, $\Delta_B=\V{h1.te.s.gs.db}$ and $\Delta_L=\V{h1.te.s.gs.dl}$).
Two of the five are nearly fixed by the parse pattern, so their verdicts say little beyond the mechanism (the stars of Table~\ref{tab:glance}, explained in Appendix~\ref{app:tables}), and the verdict is informative in \smol's three cells.

The other four cells fail the gain gate.
\qwen\ needs no visibility, since \cfg{THINK-512} parses \V{cfg.te.q.lb.t512.parse} of the letter rows, \V{cfg.te.q.mm.t512.parse} of MMLU-Pro and \V{cfg.te.q.gs.t512.parse} of GSM8K, and $\Delta=\V{h1.te.q.lb.delta}$ \CI{\V{h1.te.q.lb.delta.lo}}{\V{h1.te.q.lb.delta.hi}} on the letters.
In \intern\ on GSM8K the gate fails the other way, because \cfg{AUTHORS-1500} loses accuracy ($\Delta=\V{h1.te.i.gs.delta}$ \CI{\V{h1.te.i.gs.delta.lo}}{\V{h1.te.i.gs.delta.hi}}); there $T=\V{h1.te.i.gs.t}$ is positive only because $\Delta_B=\V{h1.te.i.gs.db}$ is large and negative, the reading the gate exists to stop, and \cfg{THINK-512} is the more accurate configuration on the rows both parse.

\paragraph{Qwen3-14B.}
In the second study visibility dominates again where the capped setting leaves rows unparsed.
\verdict{Visibility dominates} on the letters ($\Delta=\V{z.h1.te.lb.delta}$ \CI{\V{z.h1.te.lb.delta.lo}}{\V{z.h1.te.lb.delta.hi}}) and on MMLU-Pro ($\Delta=\V{z.h1.te.mm.delta}$ \CI{\V{z.h1.te.mm.delta.lo}}{\V{z.h1.te.mm.delta.hi}}), and the gate fails on GSM8K ($\Delta=\V{z.h1.te.gs.delta}$ \CI{\V{z.h1.te.gs.delta.lo}}{\V{z.h1.te.gs.delta.hi}}).
In the letter cell the parse pattern fixes the quantity: \cfg{THINK-512} parses \V{z.cfg.te.lb.t512.parse} of the letter rows and every unparsed call ended at the token limit, so B holds \V{z.h1.te.lb.nb} of \V{z.cell.te.lb.rows} rows, $\Delta_B$ is \V{z.h1.te.lb.db} and the whole gain is visibility; the verdict still rests on the gate and on the interval of $T$.
On GSM8K \cfg{AUTHORS-1500} is less accurate than \cfg{THINK-512} on the rows both configurations parse, the pattern of \intern: $T=\V{z.h1.te.gs.t}$ is positive only because $\Delta_B$ is large and negative, the reading the gate exists to stop (Appendix~\ref{app:z}).
Over both studies visibility dominates in \V{cb.h1.visibility} of \V{cb.fam1} cells.

\subsection{Whether simple baselines are adequate (H2)}
\label{sec:h2}
In the first study \cfg{AUTHORS-1500} is non-inferior to \cfg{RESCUE-1500} in \V{cnt.h2.res.holds} of nine cells and to all four GEPA harnesses.
It fails on GSM8K for \smol\ ($\theta=\V{h2.te.s.gs.res.theta}$ \CI{\V{h2.te.s.gs.res.lo}}{\V{h2.te.s.gs.res.hi}}) and for \intern\ ($\theta=\V{h2.te.i.gs.res.theta}$ \CI{\V{h2.te.i.gs.res.lo}}{\V{h2.te.i.gs.res.hi}}), where the rescue is more accurate than the thinking-off configuration by more than the margin.
The study does not isolate which part of the rescue, the thinking within \V{k.t512} tokens or the second call, accounts for the difference.
For \smol\ on GSM8K the replication partition is inconclusive, so the label is an \emph{inconclusive replication}.
In \intern's letter cell the rescue equals \cfg{AUTHORS-1500} ($\theta=\V{h2.te.i.lb.res.theta}$, since every \cfg{THINK-512} call is cut off), so \verdict{holds} is trivial there.

In the first study GEPA searched prompt text only around \qwen\ (Appendix~\ref{app:gepa}), repaired its broken starts, and found no harness whose accuracy differs detectably from the thinking-off baseline's.
In the clean-start run with seed zero no proposal beat the seed on the validation rows, so its selected harness is the seed, \cfg{AUTHORS-1500} itself ($\theta=\V{h2.te.q.lb.gc0.theta}$ because the harness is the same text and request).
The other three selected harnesses parse every test row and reach accuracy \V{gepa.gb1.tacc} to \V{gepa.gc1.tacc}, against \V{cfg.te.q.lb.aut.acc} for \cfg{AUTHORS-1500}; $\theta$ runs from \V{h2.te.q.lb.gc1.theta} to \V{h2.te.q.lb.gb1.theta}, and all three adjusted intervals contain zero.
\paragraph{Qwen3-14B.}
Against the rescue, \cfg{AUTHORS-1500} holds on MMLU-Pro (under the strict scorer, a sensitivity analysis, it fails; Appendix~\ref{app:z}) and fails on GSM8K ($\theta=\V{z.h2.te.gs.res.theta}$ \CI{\V{z.h2.te.gs.res.lo}}{\V{z.h2.te.gs.res.hi}}), as it did for \smol\ and \intern.
On the letters the rescue equals \cfg{AUTHORS-1500} ($\theta=\V{z.h2.te.lb.res.theta}$), because every \cfg{THINK-512} call is cut off, so \verdict{holds} is trivial there, as it is for \intern.
Qwen3-14B often ends a call without a readable answer, which the frozen scorer counts as wrong: \V{au.te.tot.nc.share} of its analysed test calls across the five grid configurations are non-committed (descriptive; Appendix~\ref{app:z}).
Against GEPA the result is the opposite of the first study's.
Both clean-start harnesses are more accurate than \cfg{AUTHORS-1500} by more than the margin ($\theta=\V{z.h2.te.lb.gc0.theta}$ \CI{\V{z.h2.te.lb.gc0.lo}}{\V{z.h2.te.lb.gc0.hi}} and $\V{z.h2.te.lb.gc1.theta}$ \CI{\V{z.h2.te.lb.gc1.lo}}{\V{z.h2.te.lb.gc1.hi}}), with test accuracy \V{fu.te.gc0.acc} and \V{fu.te.gc1.acc} against \V{fu.te.aut.acc}, and the verdict is \verdict{fails} for both, on both partitions.
The two broken-start searches each kept their seed, which is one text, so their two endpoints are one result: the harness is the start that answers nothing, and the verdict \verdict{holds} rests on \cfg{AUTHORS-1500} being far above it.

\paragraph{Why the broken start could not be repaired.}
The broken start is the request of \cfg{THINK-64}, thinking on with a cap of \V{k.t64} tokens, and it is truncated in every call of every cell, so it answers nothing.
In each of the two runs GEPA made \V{fu.gs.gb0.prop} proposals, and \V{fu.gs.gb0.bad} and \V{fu.gs.gb1.bad} of them failed the interface check and were scored zero without a call; every proposal scored zero against its parent's zero on a minibatch of three rows, so the strict-improvement rule accepted none, no candidate held a soft switch (\texttt{/think} or \texttt{/no\_think}), and the selected harness is the seed.
In these two searches, under their budget and acceptance rule, no proposal scored above zero, whereas the \qwen\ model of the first study answers within the cap and there GEPA repaired its broken starts, and the 14B's clean-start searches made \V{fu.gs.gc0.prop} and \V{fu.gs.gc1.prop} proposals and accepted \V{fu.gs.gc0.acc} and \V{fu.gs.gc1.acc}.
The arm searched prompt text only, so this says nothing about a search that may also change the request.

\paragraph{Qwen3-14B against the strongest constant.}
Table~\ref{tab:acc14}, computed after the outputs existed, gives the letter accuracy of the 14B's configurations and harnesses next to the strongest oracle constant, the constant label of the first study's analysis (below).
The two clean-start harnesses are not detectably above the constant: accuracy minus the constant is $\V{fu.te.gc0.minus.strong}$ and $\V{fu.te.gc1.minus.strong}$ on test and $\V{fu.re.gc0.minus.strong}$ and $\V{fu.re.gc1.minus.strong}$ on replication, and all four unadjusted intervals contain zero.
The intervals do not show equality either: on test their upper bounds are $\V{fu.te.gc0.minus.strong.hi}$ and $\V{fu.te.gc1.minus.strong.hi}$, so an advantage of up to about two points is not excluded.
\cfg{AUTHORS-1500} and \cfg{BUDGET-512-1500} are below it by $\V{fu.te.aut.minus.strong}$ and $\V{fu.te.bud.minus.strong}$ on test.

\begin{table}[t]
\caption{Qwen3-14B on the LongevityBench letters: accuracy with its unadjusted interval, parse rate, and the paired difference between accuracy and the strongest oracle constant (the constant label). The two broken-start harnesses are the same text and answer nothing. Computed after the outputs existed.}
\label{tab:acc14}
\centering
\footnotesize
\setlength{\tabcolsep}{3pt}
\resizebox{\ifdim\width>\linewidth\linewidth\else\width\fi}{!}{%
\begin{tabular}{@{}lccc@{}}
\toprule
Configuration & Accuracy & Parse rate & Accuracy minus the strongest constant \\
\midrule
\multicolumn{4}{@{}l}{\textbf{Replication} (1{,}990 rows, 399 components)} \\
\cfg{THINK-1536} & 0.239 [0.223, 0.255] & 0.345 & \ensuremath{-0.300} [\ensuremath{-0.324}, \ensuremath{-0.276}] \\
\cfg{AUTHORS-1500} & 0.222 [0.205, 0.239] & 0.567 & \ensuremath{-0.318} [\ensuremath{-0.346}, \ensuremath{-0.289}] \\
\cfg{BUDGET-512-1500} & 0.416 [0.397, 0.434] & 0.728 & \ensuremath{-0.124} [\ensuremath{-0.149}, \ensuremath{-0.098}] \\
\cfg{GEPA-CLEAN-s0} & 0.521 [0.504, 0.539] & 0.992 & \ensuremath{-0.018} [\ensuremath{-0.038}, \ensuremath{+0.003}] \\
\cfg{GEPA-CLEAN-s1} & 0.539 [0.518, 0.559] & 0.994 & \ensuremath{-0.001} [\ensuremath{-0.023}, \ensuremath{+0.022}] \\
\cfg{GEPA-BROKEN-s0} & 0.000 [0.000, 0.000] & 0.000 & \ensuremath{-0.539} [\ensuremath{-0.559}, \ensuremath{-0.519}] \\
\bottomrule
\end{tabular}}
\end{table}

\paragraph{How the harnesses reach the constant.}
The pooled accuracy adds up three different behaviours, one for each task, and they differ by partition (Appendix~\ref{app:s0} gives both partitions, computed after the outputs existed); each partition has its own constant label, chosen on it.
On the mortality task the strongest constant is ``Yes'' (accuracy \V{cl.te.l42.const} on test, \V{cl.re.l42.const} on replication), both clean-start harnesses answer ``Yes'' in \V{cl.te.l42.h0.share} and \V{cl.te.l42.h1.share} of their parsed answers on test and in \V{cl.re.l42.h0.share} and \V{cl.re.l42.h1.share} on replication, and their accuracy, \V{cl.te.l42.h0.acc} and \V{cl.te.l42.h1.acc} on test and \V{cl.re.l42.h0.acc} and \V{cl.re.l42.h1.acc} on replication, is the constant's.
On the participant-comparison task they exceed the constant in both partitions but lean on one label to different degrees, and on the age-band task they fall below it in both partitions (Appendix~\ref{app:s0} gives the accuracies and shares).
The pooled accuracy at the constant is therefore a net result: equal to it on the mortality task, which carries \V{cl.te.l42.weight} of the rows and which they answer ``Yes'' almost throughout, above it on the comparison task and below it on the age-band task.
\cfg{AUTHORS-1500} of the same model answers ``Yes'' in only \V{cl.te.l42.aut.share} of its parsed answers on the mortality task on test, and the tuned 1.7B model's \cfg{AUTHORS-1500} is above the constant on all three tasks in both partitions.
The harness texts tell the model not to infer survival beyond the snapshot data, or to conclude that survival cannot be determined when follow-up data are missing (Appendix~\ref{app:gepa14}), and the harnesses still answer ``Yes'' almost throughout on the mortality task; we did not read their outputs, so why they do is not examined (Appendix~\ref{app:scorer}).

\paragraph{The first study against the trivial baselines.}
On the letters only \qwen\ beat the strongest constant, the oracle constant label, in every configuration that parses (for \cfg{AUTHORS-1500} $\V{blm.te.lb.q.aut.do}$ \CI{\V{blm.te.lb.q.aut.do.lo}}{\V{blm.te.lb.q.aut.do.hi}}); \smol\ and \intern\ were below it in every configuration, because most of their calls end unparsed (Appendix~\ref{app:baselines-first}, computed after the outputs existed).

\subsection{The tuned model against its base (descriptive)}
\label{sec:finetune}
The second study ran the base and the tuned version of Qwen3-1.7B on the same items, configurations and schedule, as a descriptive comparison with unadjusted intervals, no verdict, and a limit on its reading fixed in advance.
The two models differ in every weight, in the data the tuned one was fine-tuned on and in the handling of the output layer, so a difference shows the effect of the whole fine-tuning on a benchmark.
It is consistent with, but does not isolate, the effect of supervised training on the same task formats from other NHANES survey waves; it does not show general domain knowledge, and it is not instance-level leakage if the wave split of the model's authors holds, which we take from their audit and do not verify.
A difference on MMLU-Pro or GSM8K is not evidence about aging biology.

On the headline cell, \cfg{AUTHORS-1500}, the base model scored below the tuned model by more than five points on the letters (Table~\ref{tab:arms}, Appendix~\ref{app:z}; tuned minus base $\V{ar.te.lb.d}$ \CI{\V{ar.te.lb.d.lo}}{\V{ar.te.lb.d.hi}} on test, \V{ar.te.lb.t} against \V{ar.te.lb.b}), and above it on MMLU-Pro ($\V{ar.te.mm.d}$ \CI{\V{ar.te.mm.d.lo}}{\V{ar.te.mm.d.hi}}; \V{ar.te.mm.t} against \V{ar.te.mm.b}) and on GSM8K ($\V{ar.te.gs.d}$ \CI{\V{ar.te.gs.d.lo}}{\V{ar.te.gs.d.hi}}; \V{ar.te.gs.t} against \V{ar.te.gs.b}); on LB-0038 the exact-match rates are within five points of each other.
The fine-tuning raised letter accuracy and lowered accuracy on both general pools, most on GSM8K.
Tuning and size are not separated by the studies as a whole: GEPA ran around the tuned 1.7B model and around the 14B on these letters, which differ in both, and the base model was searched only in the third study, on other benchmarks.

\subsection{Budget forcing (H3)}
\label{sec:h3}
In the first study \cfg{BUDGET-512-1500} lowers truncation and raises parsing in the six cells where \cfg{THINK-512} truncates a large share, with $D_T$ from \V{h3.te.s.gs.dt} (\smol, GSM8K) to \V{h3.te.i.lb.dt} (\intern, letters), $D_P$ from \V{h3.te.s.lb.dp} to \V{h3.te.i.mm.dp}, and both adjusted lower bounds above the minimal difference.
In the \qwen\ letter and GSM8K cells \cfg{THINK-512} does not truncate (adjusted upper bounds \V{h3.te.q.lb.tr.hi} and \V{h3.te.q.gs.tr.hi}), so there is \emph{no room} for a drop of \V{c.minm}.
In \qwen\ on MMLU-Pro $D_T=D_P=\V{h3.te.q.mm.dt}$ with adjusted interval \CI{\V{h3.te.q.mm.dt.lo}}{\V{h3.te.q.mm.dt.hi}}, which contains \V{c.minm}, so the verdict is \verdict{inconclusive}.
For Qwen3-14B \cfg{BUDGET-512-1500} lowers truncation and raises parsing in all three cells, with $D_T$ from \V{z.h3.te.gs.dt} (GSM8K) to \V{z.h3.te.lb.dt} (letters), $D_P$ from \V{z.h3.te.gs.dp} to \V{z.h3.te.mm.dp} (MMLU-Pro) and both adjusted lower bounds above the minimal difference; \cfg{THINK-512} truncates \V{z.cfg.te.lb.t512.trunc} of its letter calls.
Over both studies the verdict is \verdict{holds} in \V{cb.h3.holds} of \V{cb.fam3} cells.

\cfg{BUDGET-512-1500} differs from \cfg{THINK-512} in two ways at once: vLLM ends thinking after \V{k.bud} tokens, and the total allowance is \V{k.bud.max} tokens against \V{k.t512}.
H3 therefore supports the combined change and not forced termination alone.
The partial comparison with \cfg{THINK-1536}, which has a larger allowance and no forcing, is in Appendix~\ref{app:tables}.
In the two models of the first study with a seeded arm, substantial truncation persists under sampling at the model cards' values (descriptive; Appendix~\ref{app:seeded}).

\subsection{Whether replication reveals known defects (H4)}
\label{sec:h4}
In the first study the claim holds in six pairs: the broken start in all three models, the unparsed-as-A scoring in \smol\ and \intern, and the gold fallback in \qwen.
It fails in four and is inconclusive in five, and one pair, the unparsed-as-A scoring in \qwen, is not evaluable because its start leaves no row unparsed (Table~\ref{tab:h4} and the pairs one by one in Appendix~\ref{app:h4}).
Five of the six that hold carry a star, because the quantity is fixed by an identity of the design: the distortion of the unparsed-as-A scoring equals the share the start leaves unparsed with gold~A, which is also its evaluability bound, and the distortion under the broken start is the difference between the accuracies of two starts, \cfg{THINK-1536} and \cfg{THINK-64}.
The verdict still depends on the intervals: for the broken start the preregistration expected the claim to hold only about \V{ex.d.half} of the time with no true difference, since it rests on $\Gamma$ being within about \V{ex.d.within} of zero (Appendix~\ref{app:expected}).
The gold fallback in \qwen\ is the informative pair, with $D_S=\V{h4.b1.q.ds}$.
In all six the distortion is large and reappears on $R$, so a replication would have reproduced the distorted report instead of revealing it.
These counts follow the general rule of the preregistration for $D_R$, which the pinned code implements (Section~\ref{sec:deviations}); under the narrower reading that grades only the two sub-cases listed after the general rule, the skewed split in \smol\ is inconclusive and the counts are \V{lit.h4.holds} holds, \V{lit.h4.fails} fails and \V{lit.h4.inconclusive} inconclusive (Appendix~\ref{app:deviations}).

In the second study the claim holds for B1, B2 and D and is inconclusive for A and C.
Two of the three carry a star: B2's distortion, $D_S=\V{z.h4.b2.ds}$, equals its evaluability bound \V{z.h4.b2.bound}, and D's is the difference of two starts' accuracies, $D_S=\V{z.h4.d.ds}$; their verdicts still depend on the intervals.
B1 is the informative pair: the start parses \V{z.h4.b1.startparse} of the letter rows, the gold fallback changes the gold of a share \V{z.h4.b1.changed} of the rows, and the distortion, $D_S=\V{z.h4.b1.ds}$ on the selection sample and $D_R=\V{z.h4.b1.dr}$ on the check sample, is the same on both.
Over both studies the claim holds in \V{cb.h4.holds} of the \V{cb.h4.evaluable} evaluable pairs, \V{cb.h4.byc} of them pairs whose quantity is fixed by an identity of the design (the stars).
Appendix~\ref{app:checks} summarises the audit checks of Section~\ref{sec:h4design}: their detections are descriptive and largely true by construction, and what is empirical is their false alarms on the \V{fa.ir4.of} clean instances: IR4 fires on \V{fa.ir4.n}, SCF1 on \V{fa.scf1.n}, SCF2 on \V{fa.scf2.n}, TB2 on \V{fa.tb2.n}, SCF3 on \V{fa.scf3.n} and the gain rule of PR on \V{fa.prgain.n}.

\subsection{Whether the conclusions change across GPUs}
\label{sec:gpu}
The tuned 1.7B model ran in both studies, on an RTX~5090 in the first and on an \V{s2.gpu} in the second, with the same environment, the same batch-invariant server, the same request bodies and the same items.
Its outputs are not the same: token IDs differ in \V{rp.rg.share} of the replication grid's calls and \V{rp.tg.share} of the test grid's, and in \V{rp.rs.share} and \V{rp.ts.share} of the two seeded arms' calls (Table~\ref{tab:gpu}, Appendix~\ref{app:z}); the request body never differs, and the finish reason differs in \V{rp.tg.finish} of the \V{rp.tg.common} test-grid calls.
The conclusions are the same.
Recomputed on the rerun's outputs with the first study's rules, seeds and family sizes, \V{ag.n} of the first study's \V{ag.q.total} primary endpoints for this model, the endpoints of the grid's H1--H3, give the same verdict in \V{ag.test} cases on the test partition and in \V{ag.rep} on the replication partition and the same replication label in \V{ag.lab}, and none flips.
The \V{ag.q.gepa} GEPA endpoints and the \V{ag.q.h4} H4 pairs of this model were not rerun.
The accuracy of every configuration in the three pools differs by at most \V{ag.maxdiff} between the two runs, and \V{ag.nexcl} of the \V{ag.ncells} unadjusted intervals of the difference exclude zero (Appendix~\ref{app:z}).
The comparison is descriptive and covers one model, \V{ag.n} grid endpoints and two GPUs of one architecture; it says that these conclusions did not depend on the token-level agreement, which numerical differences between hardware are known to break \citep{yuan2025fp32,he2025nondeterminism}.

\section{Deviations and addenda}
\label{sec:deviations}

The preregistration allows a change only if it alters no request bytes, parsing, scoring or call order, and then only with a written note and a dry run that reproduces the recorded render hashes; any other change needs a new registration.
After the first freeze, the requests, the scoring and the primary criteria stayed as preregistered, with two exceptions that were not mechanical.
We waived one stopping rule, once and for one stage: the check on \intern's thinking switch had stopped the run that sends the reference answer, and we let that check record its failures and the run continue.
And we settled one descriptive label after seeing the outputs: while auditing this paper we found that the two listed cases of the claim rule for the check sample are incomplete, so its general rule governs, and under it the claim for the skewed split in \smol\ fails where the narrower reading leaves it inconclusive (Section~\ref{sec:h4} gives both totals for H4).

Four items of the second study's record bear on a result.
The physical GPU changed during the study and the preregistered recheck found no difference, and the interrupted harness evaluation was rerun from scratch (Section~\ref{sec:second}).
A pinned file that lists the replication rows left out of every analysis because their reference answers could be recovered was missing on the instance; the checkpoint stopped when it first ran, before any run was read, the file was copied by hand and its hash checked, and the checkpoint was repeated.
And the non-commitment of Qwen3-14B was decided, before the study's outputs were read, to be reported descriptively with the frozen scorer primary (Appendix~\ref{app:z}).
Table~\ref{tab:dev} in Appendix~\ref{app:deviations} lists the deviations that bear on a result, and the appendix lists every exception, known deviation, pre-freeze decision and addendum with its date, reason, rule and the outputs seen at the time; the platform shutdown, the clone of the instance and the copy of the outputs change no result.
No call touched a replication or test item before a freeze, and in the first two studies we saw only counts, and no accuracy, score or verdict of the main study, until every stage had run and the checkpoint's record had been committed.
Everything computed after the outputs existed is labelled where it appears.
This paper's tables and figures only format aggregates committed in our repository, and its audit recomputes every verdict from its interval and agrees with the report on all of them (Appendix~\ref{app:deviations}); the third study's record is in Section~\ref{sec:dev3}.

\section{The third study: design}
\label{sec:third}

The second study left a pattern that it was not built to test.
For Qwen3-14B on the LongevityBench letters, both clean-start GEPA harnesses were far more accurate than the thinking-off baseline and not detectably above the strongest constant label, and on the mortality task they answered ``Yes'' almost throughout (Section~\ref{sec:results}).
The third study asks whether that pattern recurs on other tasks and models, and what separates the cases where a search finds a real gain from those where it reaches the constant.
It reuses the first study's rules and the second study's analysis, and it was written after both studies' results were known, so it is not blind to the pattern that its third hypothesis (H5) tests.
Its preregistration was frozen at \V{e3.s3.time}~UTC on \V{e3.s3.date} (commit \V{e3.s3.commit}, SHA256 beginning \V{e3.s3.hash}), before any call touched a replication or test item; the only earlier calls, in a screening step, used search-partition rows (Appendix~\ref{app:third-prereg}).

\subsection{Models, benchmarks and partitions}
The study runs \smol\ and the base model of the tuned Qwen3-1.7B (the secondary arm of the second study, called \basem\ here), on one \V{e3.gpu.old} with the earlier studies' stack: vLLM \V{d.vllm} in batch-invariant mode, temperature zero and client concurrency \V{d.conc}.
\intern\ proposes harnesses and samples at the model card's values.
The first study ran \smol\ on the same kind of GPU, and the second study's base-model arm ran on \V{e3.gpu.new} GPUs; Section~\ref{sec:third-repro} uses this.

A fixed list of candidate benchmarks was screened before the freeze for a strong majority label, a test partition of at least \V{e3.s0.min.test} rows and licence terms that allow research use.
BoolQ \citep{clark2019boolq} asks a yes-or-no question about a passage, and Financial PhraseBank (FPB) \citep{malo2014phrasebank} asks for the sentiment of a sentence from financial news (negative, neutral or positive).
Both passed.
MMLU-Pro \citep{wang2024mmlupro} has up to ten options (ten for \V{gb.te.mm.k10} of the \V{gb.te.mm.items} test questions and three, the fewest, for \V{e3.mm.opt.te.k3}) and no majority label; it is the control, with the first two studies' items and partitions.
Autocast \citep{zou2022autocast} asks yes-or-no forecasting questions whose outcomes were later resolved; it is the exploratory arm described below.
Items are split by group (the passage, the sentence or the question, so that no group spans two partitions) into search, replication and test partitions, and the group is the bootstrap unit (the item for MMLU-Pro).
The test partitions have \V{e3.cell.te.sm.bq.rows} (BoolQ), \V{e3.cell.te.sm.fp.rows} (FPB), \V{e3.cell.te.sm.mm.rows} (MMLU-Pro) and \V{e3.cell.te.sm.au.rows} (Autocast) rows, and the replication partitions \V{e3.cell.re.sm.bq.rows}, \V{e3.cell.re.sm.fp.rows}, \V{e3.cell.re.sm.mm.rows} and \V{e3.cell.re.sm.au.rows}.
MMLU-Pro was called again, for both models and both configurations, so that each harness and its seed are paired on one stack; the earlier outputs are kept for an agreement check.
Here \cfg{AUTHORS-1500} is the first study's thinking-off configuration applied unchanged to the new tasks, not a protocol that the authors of these benchmarks supplied.
The prompts, and the readers for yes-or-no and sentiment answers with their rules for conflicts, refusals and labels that occur only in reasoning, are fixed in Section~\V{e3.prereg.sec.prompts} of the third preregistration, an ancillary file of this arXiv version; Appendix~\ref{app:third-scoring} prints them with synthetic examples.

\subsection{Screening and the premise of H5}
H5 compares a harness with the constant label chosen on the search partition, which says something only where the model's own thinking-off accuracy does not already exceed that constant.
A screening step therefore decided, for each model and benchmark, whether H5 is confirmatory.
Eligibility was checked from counts before any model call: a search partition of at least \V{e3.s0.min.search} rows, a replication partition of at least \V{e3.s0.min.rep} and a test partition of at least \V{e3.s0.min.test}, rows that fit the context, a majority share between \V{e3.s0.share.lo}\% and \V{e3.s0.share.hi}\%, and terms that allow research use.
For each eligible benchmark we then ran \cfg{AUTHORS-1500} on \V{e3.sc.bq.sm.n} search rows with both models.
A pair \emph{qualifies} if the model's number of correct answers is at most the number of rows that carry the constant label; equality qualifies.
The two eligible benchmarks with the most qualifying pairs were kept, and the rule was applied once to the first complete attempt, with no rescreening.
The H5 endpoints of a pair that qualified are confirmatory, and those of a pair that did not are descriptive (``premise not met'') and keep their place in the family.
Autocast failed the eligibility rule by \V{e3.s0.au.short} rows: its test partition has \V{e3.sc.au.ntest} rows against the minimum of \V{e3.s0.min.test}, and the rule stands.

We decided, from the counts alone and before any model call, to run Autocast as a declared exploratory arm.
It follows the same procedure and carries unadjusted \V{c.ci} intervals and no verdict label of any kind; it enters no family, and its calls in each step come after every confirmatory call.
Its prompt is the question text only, because the benchmark's background field can state the outcome of a question.
Its questions resolved before these models were trained, so a result there may reflect memory of the answers rather than forecasting.

\subsection{The GEPA arm}
GEPA runs as in the earlier studies, with \intern\ as the proposer, on a clean start only and on prompt text only: a system prompt and an appended instruction.
Each model and benchmark has one seed harness, an empty system prompt and no instruction, so four per model; its requests equal those of \cfg{AUTHORS-1500} on every row of its sample (text and token ids, checked before each run).
Each pair has two search seeds, so there are \V{e3.cnt.runs} runs: twelve on BoolQ, FPB and MMLU-Pro, and four on Autocast.
A run spends \V{k.metric} metric calls on a \V{k.gepa.rows}-row sample of the search partition, half for training and half for validation, and the harness with the highest mean validation score is selected (ties to the lowest candidate index, and the seed if no proposal was accepted).
Each selected harness is then evaluated on the replication partition and, after a checkpoint that locks it and is committed before the first test call, on the test partition.
A check that a thinking-off call does not begin with the model's own thinking token (gate~(b)) stops the grid and, as decided after the freeze, records in the searches and in the evaluations of selected harnesses (Section~\ref{sec:dev3}).

\subsection{Hypotheses, families and statuses}
The contrast and the verdict rules of H1 and H2 are the first study's (Table~\ref{tab:hyp}), at its margin $\delta=m=\V{c.margin}$.
H5 is new: for a harness, $\Delta_c$ is its accuracy minus the accuracy of the constant label, paired on rows.
Its verdict reads the whole interval of $\Delta_c$ against $\pm m$: \verdict{at the constant} if the interval lies within $\pm m$, \verdict{above the constant} if its lower bound exceeds $m$, \verdict{below the constant} if its upper bound is below $-m$, and \verdict{inconclusive} otherwise.
The replication label follows the same rule on the replication partition.
The stated reading of the 14B pattern is H2 \verdict{fails} together with H5 \verdict{at the constant} for the same harness, and it counts as confirmatory only where both endpoints are.

The families are fixed at \V{e3.fam.h1}, \V{e3.fam.h2} and \V{e3.fam.h5} endpoints (levels \V{e3.level.h1}, \V{e3.level.h2} and \V{e3.level.h5}) whatever the screening shows.
H1 has one endpoint per model and kept benchmark, H2 one per model, benchmark (the two kept ones and MMLU-Pro) and seed, and H5 one per model, kept benchmark and seed.
The analysis report's own summary field gives the H5 family as \V{e3.fam.h5.report}, the number of confirmatory H5 endpoints; the preregistration fixes it at \V{e3.fam.h5}, and the report gives the confirmatory H5 endpoints the adjusted interval of that family of \V{e3.fam.h5}, which is what we use; its descriptive and exploratory H5 estimates carry unadjusted \V{c.ci} intervals.
Of the \V{e3.cnt.n.all} endpoints of the study, \V{e3.cnt.n.conf} are confirmatory, \V{e3.cnt.n.desc} descriptive and \V{e3.cnt.n.expl} exploratory.
MMLU-Pro is descriptive for H1 and H5, because the gain of \cfg{AUTHORS-1500} over \cfg{THINK-512} is known from the first two studies; its H2 is confirmatory, because the harnesses are new and its search items were never called.
Descriptive endpoints carry the unadjusted interval, and a statement that combines endpoints is confirmatory only if each of them is.
Fixed in advance and reported next to every H5 verdict, without changing any verdict, are the \emph{label collapse} (at least \V{e3.collapse.thr}\% of the harness's parsed test answers are the majority label) and four flags: a literal constant predictor, no accepted proposal in the run, identical selected harnesses for the two seeds, and pervasive parse failure.

\section{The third study: results}
\label{sec:third-results}

Results are on the test partition unless stated, and every share and range names its partition.
Confirmatory endpoints carry the family-adjusted interval, descriptive ones the unadjusted \V{c.ci} interval, and the exploratory Autocast arm carries unadjusted intervals and no verdict; a range that spans both partitions says so.
Table~\ref{tab:third-glance} gives the verdicts, Table~\ref{tab:third-acc} the accuracies, and Appendix~\ref{app:third-endpoints} every endpoint with both partitions and the replication label.

\subsection{Screening and the premise}
BoolQ qualified for neither model: \smol\ answered \V{e3.sc.bq.sm.correct} of the \V{e3.sc.bq.sm.n} screening rows correctly and \basem\ \V{e3.sc.bq.bs.correct}, against \V{e3.sc.bq.sm.const} rows that carry the constant label ``yes''.
Its four H5 endpoints are therefore descriptive, with the premise not met.
FPB qualified for both models (\V{e3.sc.fp.sm.correct} and \V{e3.sc.fp.bs.correct} correct against \V{e3.sc.fp.sm.const} for the constant ``neutral''), so its four H5 endpoints are confirmatory.
Both benchmarks were kept.
Autocast, outside the ranking, would have qualified for both models (\V{e3.sc.au.sm.correct} and \V{e3.sc.au.bs.correct} against \V{e3.sc.au.sm.const}).

\begin{table}[t]
\caption{Verdicts of the third study on the test partition. H1 compares \cfg{AUTHORS-1500} with \cfg{THINK-512}, H2 compares \cfg{AUTHORS-1500} with each GEPA harness (seed zero and seed one of the search), and H5 compares each harness with the constant label of the search partition. H1 and H5 on MMLU-Pro and H5 on BoolQ are descriptive (\textsuperscript{d}); every other cell is confirmatory, and Autocast is exploratory with no verdict. A star marks an H2 endpoint whose selected harness is the seed, so that $\theta$ is exactly zero and \verdict{holds} is trivial. A dagger marks an endpoint whose replication label is \emph{inconclusive replication}.}
\label{tab:third-glance}
\centering
\footnotesize
\setlength{\tabcolsep}{3pt}
\resizebox{\ifdim\width>\linewidth\linewidth\else\width\fi}{!}{%
\begin{tabular}{@{}llccccc@{}}
\toprule
Model & Benchmark & \shortstack{H1\\ \cfg{AUTHORS-1500} vs\\ \cfg{THINK-512}} & \shortstack{H2, seed 0\\ \cfg{AUTHORS-1500} vs\\ harness} & \shortstack{H2, seed 1\\ \cfg{AUTHORS-1500} vs\\ harness} & \shortstack{H5, seed 0\\ harness vs\\ constant} & \shortstack{H5, seed 1\\ harness vs\\ constant} \\
\midrule
SmolLM3-3B & BoolQ & visibility dominates\textsuperscript{\dag} & holds & holds\textsuperscript{$\star$} & above the constant\textsuperscript{d} & above the constant\textsuperscript{d} \\
 & FPB & no gain: not tested & fails & fails & above the constant & above the constant \\
 & MMLU-Pro & visibility dominates\textsuperscript{d} & holds\textsuperscript{$\star$} & holds\textsuperscript{$\star$} & above the constant\textsuperscript{d} & above the constant\textsuperscript{d} \\
 & Autocast & \multicolumn{5}{l}{exploratory, no verdict} \\
\addlinespace
base Qwen3-1.7B & BoolQ & visibility dominates & holds & holds & above the constant\textsuperscript{d} & above the constant\textsuperscript{d} \\
 & FPB & no gain: not tested & fails & fails & above the constant & above the constant \\
 & MMLU-Pro & visibility dominates\textsuperscript{d} & inconclusive\textsuperscript{\dag} & holds\textsuperscript{$\star$} & above the constant\textsuperscript{d} & above the constant\textsuperscript{d} \\
 & Autocast & \multicolumn{5}{l}{exploratory, no verdict} \\
\bottomrule
\end{tabular}}
\end{table}

\subsection{Where the gains come from (H1)}
\label{sec:third-h1}
On the BoolQ test rows, \cfg{AUTHORS-1500} gains over \cfg{THINK-512} by $\Delta=\V{e3.h1.te.sm.bq.delta.pp}$ points \CI{\V{e3.h1.te.sm.bq.delta.lo.pp}}{\V{e3.h1.te.sm.bq.delta.hi.pp}} for \smol\ and by $\V{e3.h1.te.bs.bq.delta.pp}$ \CI{\V{e3.h1.te.bs.bq.delta.lo.pp}}{\V{e3.h1.te.bs.bq.delta.hi.pp}} for \basem, and the verdict is \verdict{visibility dominates} for both.
The gain comes from rows that \cfg{THINK-512} leaves unparsed, since $\Delta_V$ is $\V{e3.h1.te.sm.bq.dv.pp}$ and $\V{e3.h1.te.bs.bq.dv.pp}$ points while $\Delta_B$ is $\V{e3.h1.te.sm.bq.db.pp}$ and $\V{e3.h1.te.bs.bq.db.pp}$.
The base model's verdict replicates.
\smol's is an \emph{inconclusive replication}: on test the lower bound of $T$ is only \V{e3.h1.te.sm.bq.t.lo.pp} points, and on replication its interval \CI{\V{e3.h1.re.sm.bq.t.lo.pp}}{\V{e3.h1.re.sm.bq.t.hi.pp}} contains zero.
On FPB the gate fails for both models on both partitions (\verdict{no gain: not tested}); on test, $\Delta=\V{e3.h1.te.sm.fp.delta.pp}$ \CI{\V{e3.h1.te.sm.fp.delta.lo.pp}}{\V{e3.h1.te.sm.fp.delta.hi.pp}} for \smol\ and $\V{e3.h1.te.bs.fp.delta.pp}$ \CI{\V{e3.h1.te.bs.fp.delta.lo.pp}}{\V{e3.h1.te.bs.fp.delta.hi.pp}} for \basem; \cfg{THINK-512} already parses \V{e3.cfg.te.sm.fp.t512.parse.pp} and \V{e3.cfg.te.bs.fp.t512.parse.pp} of the FPB test rows, so there is little for visibility to supply.
Of the \V{e3.cnt.h1.total} confirmatory endpoints, \V{e3.cnt.h1.te.vis} are \verdict{visibility dominates} and \V{e3.cnt.h1.te.nog} \verdict{no gain: not tested} on test; the labels are \emph{replicated} for \V{e3.cnt.h1.lab.rep} and \emph{inconclusive replication} for \V{e3.cnt.h1.lab.inc}.

On MMLU-Pro (descriptive), \cfg{THINK-512} is truncated on \V{e3.cfg.te.sm.mm.t512.trunc.pp} (\smol) and \V{e3.cfg.te.bs.mm.t512.trunc.pp} (\basem) of the test calls and on \V{e3.cfg.re.sm.mm.t512.trunc.pp} and \V{e3.cfg.re.bs.mm.t512.trunc.pp} of the replication calls, so the gains are large ($\Delta=\V{e3.h1.te.sm.mm.delta.pp}$ and $\V{e3.h1.te.bs.mm.delta.pp}$ points on test) and \verdict{visibility dominates} in both models on both partitions.
On Autocast (exploratory, no verdict) the test gains are $\V{e3.h1.te.sm.au.delta.pp}$ \CI{\V{e3.h1.te.sm.au.delta.ulo.pp}}{\V{e3.h1.te.sm.au.delta.uhi.pp}} and $\V{e3.h1.te.bs.au.delta.pp}$ \CI{\V{e3.h1.te.bs.au.delta.ulo.pp}}{\V{e3.h1.te.bs.au.delta.uhi.pp}} points, where \cfg{THINK-512} is truncated on \V{e3.cfg.te.sm.au.t512.trunc.pp} and \V{e3.cfg.te.bs.au.t512.trunc.pp} of the test calls.

\subsection{Whether the thinking-off baseline is adequate (H2)}
\label{sec:third-h2}
Of the \V{e3.cnt.h2.total} confirmatory H2 endpoints, \V{e3.cnt.h2.te.fails} fail on test, all on FPB, \V{e3.cnt.h2.te.holds} hold and \V{e3.cnt.h2.te.inc} is inconclusive.
On FPB the harnesses are more accurate than \cfg{AUTHORS-1500} by \V{e3.rng.fp.thn.bo.min} to \V{e3.rng.fp.thn.bo.max} points over both partitions, so the baseline is not adequate there.
On BoolQ the baseline is not meaningfully worse than any harness: no $\theta$ is further from zero than \V{e3.rng.bq.th.bo.amax} points on either partition.

Four of the seven that hold are trivial.
In \V{e3.cnt.seed.n} of the \V{e3.cnt.runs} runs the selected harness is the seed: \smol\ on BoolQ with seed one, \smol\ on MMLU-Pro with seed zero and with seed one, and \basem\ on MMLU-Pro with seed one.
Their pools held \V{e3.cnt.pool.seed.min} to \V{e3.cnt.pool.seed.max} candidates, and the seed was strictly the highest on validation in each (Table~\ref{tab:third-a13} in Appendix~\ref{app:third-runs} gives the pool, the proposals and the validation scores of each run).
The seed equals \cfg{AUTHORS-1500}, so $\theta$ is exactly zero on both partitions and \verdict{holds} says nothing about the search.
The two \smol\ harnesses on MMLU-Pro are the same text, so their endpoints are one measurement counted twice.
These runs also expose a mismatch in a descriptive flag.
The preregistration defines outcome flag (ii) as \V{e3.a13.def}: the run accepted no candidate.
The frozen analysis code sets it when \V{e3.a13.say}.
The report's flag is therefore yes for these \V{e3.cnt.noacc.code} runs, and under the preregistered definition it is no for all \V{e3.cnt.fz.no} (every pool held \V{e3.cnt.pool.min} to \V{e3.cnt.pool.max} candidates).
We report both and did not rename the flag. \V{e3.a13.nochange}, because no test uses the flag (Appendix~\ref{app:third-deviations}).

The other \V{e3.cnt.h2.nontriv} endpoints that hold are BoolQ's, for \smol\ with seed zero and for \basem\ with both seeds.
The inconclusive endpoint is \basem\ on MMLU-Pro with seed zero, where $\theta=\V{e3.h2.te.bs.mm.s0.theta.pp}$ points \CI{\V{e3.h2.te.bs.mm.s0.lo.pp}}{\V{e3.h2.te.bs.mm.s0.hi.pp}} straddles the margin on test and the replication partition gives \verdict{\V{e3.h2.re.bs.mm.s0.verdict}} ($\theta=\V{e3.h2.re.bs.mm.s0.theta.pp}$); its label is an \emph{inconclusive replication}.

\subsection{Whether a harness reaches the constant (H5)}
\label{sec:third-h5}
All \V{e3.cnt.h5.total} confirmatory endpoints, which are the FPB harnesses, are \verdict{above the constant} on both partitions and replicated: $\Delta_c$ lies between \V{e3.rng.fp.dc.bo.min} and \V{e3.rng.fp.dc.bo.max} points over both partitions, against the constant ``neutral''.
The descriptive endpoints agree: BoolQ's harnesses are \V{e3.rng.bq.dc.bo.min} to \V{e3.rng.bq.dc.bo.max} points above the constant ``yes'' and MMLU-Pro's \V{e3.rng.mm.dc.bo.min} to \V{e3.rng.mm.dc.bo.max} points above the search partition's constant (``\V{e3.cell.te.sm.mm.const}''; the oracle constant on test is \V{e3.cell.te.sm.mm.oracle.label}), over both partitions.
Three of the four MMLU-Pro harnesses are the seed, so for them H5 says that \cfg{AUTHORS-1500} itself is above the constant.
All \V{e3.cnt.h5.all.n} H5 endpoints that carry a verdict are \verdict{above the constant} on both partitions (\V{e3.cnt.h5.all.both.above} verdicts), and none is \verdict{at the constant}, \verdict{below} or \verdict{inconclusive}.
The 14B pattern, H2 \verdict{fails} with H5 \verdict{at the constant}, therefore occurs in no confirmatory or descriptive cell.
The four FPB harnesses show the opposite: H2 \verdict{fails} and H5 is \verdict{above the constant}, both confirmatory, so a search found a harness that beats the baseline and the constant label.
No harness of BoolQ, FPB or MMLU-Pro collapses onto the majority label: \V{e3.cnt.coll.oth.n} of \V{e3.cnt.coll.oth.of} cells (twelve harnesses on two partitions) reach the collapse threshold, with majority shares of \V{e3.rng.oth.mj.te.min}\% to \V{e3.rng.oth.mj.te.max}\% of parsed answers on test and \V{e3.rng.oth.mj.re.min}\% to \V{e3.rng.oth.mj.re.max}\% on replication.

\begin{table}[t]
\caption{Accuracy on the test partition of the constant label chosen on the search partition (in brackets, the label), of \cfg{THINK-512}, \cfg{AUTHORS-1500}, \cfg{RESCUE-1500} and of the two selected harnesses. A star marks a harness that is the seed, equal to \cfg{AUTHORS-1500}. Autocast is exploratory. Intervals are in Appendix~\ref{app:third-endpoints}.}
\label{tab:third-acc}
\centering
\footnotesize
\setlength{\tabcolsep}{3pt}
\resizebox{\ifdim\width>\linewidth\linewidth\else\width\fi}{!}{%
\begin{tabular}{@{}llcccccc@{}}
\toprule
Model & Benchmark & Constant (label) & \cfg{THINK-512} & \cfg{AUTHORS-1500} & \cfg{RESCUE-1500} & Harness, seed 0 & Harness, seed 1 \\
\midrule
SmolLM3-3B & BoolQ & 0.616 (yes) & 0.767 & 0.841 & 0.847 & 0.838 & 0.841\textsuperscript{$\star$} \\
 & FPB & 0.580 (neutral) & 0.626 & 0.615 & 0.633 & 0.757 & 0.739 \\
 & MMLU-Pro & 0.114 (A) & 0.098 & 0.353 & 0.378 & 0.353\textsuperscript{$\star$} & 0.353\textsuperscript{$\star$} \\
 & Autocast (exploratory) & 0.716 (no) & 0.125 & 0.532 & 0.542 & 0.721 & 0.724 \\
\addlinespace
base Qwen3-1.7B & BoolQ & 0.616 (yes) & 0.725 & 0.793 & 0.832 & 0.799 & 0.780 \\
 & FPB & 0.580 (neutral) & 0.610 & 0.635 & 0.658 & 0.778 & 0.754 \\
 & MMLU-Pro & 0.114 (A) & 0.025 & 0.392 & 0.393 & 0.399 & 0.392\textsuperscript{$\star$} \\
 & Autocast (exploratory) & 0.716 (no) & 0.306 & 0.649 & 0.653 & 0.718 & 0.722 \\
\bottomrule
\end{tabular}}
\end{table}

\subsection{Autocast, an exploratory arm}
\label{sec:third-autocast}
Autocast has no verdict, no label and no family, because it fell \V{e3.s0.au.short} rows short of the size rule; everything below is descriptive with unadjusted intervals.
The constant ``no'' scores \V{e3.cell.te.sm.au.oracle.pp} on the test partition, and \cfg{AUTHORS-1500} is far below it, at \V{e3.cfg.te.sm.au.aut.acc.pp} for \smol\ and \V{e3.cfg.te.bs.au.aut.acc.pp} for \basem.
All four harnesses are more accurate than \cfg{AUTHORS-1500} ($\theta$ from $\V{e3.rng.au.th.te.min}$ to $\V{e3.rng.au.th.te.max}$ points on test) and within \V{e3.rng.au.dc.te.max} points of the constant on test (\V{e3.rng.au.dc.re.max} points on replication): $\Delta_c$ is \V{e3.rng.au.dc.te.min} to \V{e3.rng.au.dc.te.max} points on test and \V{e3.rng.au.dc.re.min} to \V{e3.rng.au.dc.re.max} on replication (the unadjusted intervals are in Appendix~\ref{app:third-autocast}).
All \V{e3.cnt.coll.au.n} of \V{e3.cnt.coll.au.of} harness cells (four harnesses on two partitions) answer ``no'' for \V{e3.rng.au.mj.te.min}\% to \V{e3.rng.au.mj.te.max}\% of their parsed answers on test and \V{e3.rng.au.mj.re.min}\% to \V{e3.rng.au.mj.re.max}\% on replication.
The point estimates and label shares resemble the 14B's pattern, but Autocast has no verdict and its unadjusted intervals do not establish the pattern (Tables~\ref{tab:third-h2} and~\ref{tab:third-h5}).
For \basem, the upper ends of the test intervals of $\theta$ against \cfg{AUTHORS-1500} are \V{e3.h2.te.bs.au.s0.uhi.pp} and \V{e3.h2.te.bs.au.s1.uhi.pp} points, so neither lies beyond the margin of five points, and the replication interval of $\Delta_c$ for seed zero, \CI{\V{e3.h5.re.bs.au.s0.ulo.pp}}{\V{e3.h5.re.bs.au.s0.uhi.pp}} points, is not within the margin of the constant.
The questions resolved before these models were trained, so the outcomes may be in their pretraining data, and these results do not show whether remembered outcomes helped: a small difference in the mean can hide gains and losses on different questions.
The four Autocast texts state events, people, organisations and dates as facts for the model to use, and some of these statements look wrong; two ask for reasoning after the answer, one of them with a third label (Appendix~\ref{app:third-gepa}).

\subsection{The pattern across the three studies, and what it does not show}
\label{sec:pattern}
Figure~\ref{fig:open} plots three of the cases. On questions about the input or about knowledge (FPB, BoolQ, MMLU-Pro), GEPA either found a gain over the constant or found no clear gain over thinking off.
On FPB the harnesses beat \cfg{AUTHORS-1500} and the constant label in all four cells, on both partitions (confirmatory).
Thinking off was non-inferior to all four BoolQ harnesses; \V{e3.cnt.seed.mm} of the four MMLU-Pro searches kept the seed, and the remaining comparison was inconclusive on test (non-inferiority is confirmatory, the comparison with the constant descriptive).
On questions that ask for a prediction, the harness ended near the constant and answered one label almost throughout while its gain over \cfg{AUTHORS-1500} looked large: the 14B on the mortality task in the second study (descriptive; ``Yes'' in \V{cl.te.l42.h0.share} and \V{cl.te.l42.h1.share} of parsed answers on test) and Autocast here (exploratory; \V{e3.rng.au.mj.te.min}\% to \V{e3.rng.au.mj.te.max}\% on test, gains of \V{e3.rng.au.thn.te.min} to \V{e3.rng.au.thn.te.max} points over \cfg{AUTHORS-1500}).
In each case where a harness beat \cfg{AUTHORS-1500} by a wide margin (the 14B, FPB, Autocast), the comparison with the constant showed whether there was a gain beyond a trivial answer, and the share of the majority label showed the collapse.

The pattern rests on few tasks, and it is a pattern in these cases and not a finding about their cause.
The tasks differ in label skew (the constant covers \V{e3.s0.bq.share}\% of BoolQ's search partition, \V{e3.s0.fp.share}\% of FPB's, \V{e3.s0.au.share}\% of Autocast's and \V{e3.s0.mm.share}\% of MMLU-Pro's, and \V{cl.te.l42.const} of the mortality task's test rows), in the number of labels (two, three, or up to ten) and in task type, so the studies do not show that the difference between the cases depends on the kind of question.
The collapse side is descriptive and exploratory only: one exploratory arm here, and in the second study a descriptive result for one model.
The two models of the third study are small, the 14B ran only in the second study, and the search covers prompt text only.

\subsection{Reproduction, determinism and gate (b)}
\label{sec:third-repro}
The determinism recheck was identical for each of the two models: for each, \V{e3.det.calls} of \V{e3.det.calls} calls on \V{e3.det.rows} test rows matched the test grid token for token.
MMLU-Pro lets us compare with the earlier studies' outputs, a comparison of parsed labels and correctness readings and not of tokens (computed after the outputs existed).
\smol's parsed labels and correctness readings agree on all \V{e3.ag.te.sm.au.rows} test and \V{e3.ag.re.sm.au.rows} replication rows, on both configurations, and the first study ran on the same kind of GPU (an \V{e3.gpu.old}).
For \basem, \cfg{AUTHORS-1500} agrees in correctness on \V{e3.ag.te.bs.au.agree.pp} of the test rows and \V{e3.ag.re.bs.au.agree.pp} of the replication rows, and in the parsed label on \V{e3.ag.te.bs.au.same.pp} and \V{e3.ag.re.bs.au.same.pp}; the second study's arm ran on \V{e3.gpu.new} GPUs, which fits a difference between GPUs, a reading we did not test.
Gate~(b) recorded \V{e3.gb.said.trip} trips in \V{e3.gb.said.calls} thinking-off calls of the searches (\V{e3.gb.share}), all in two runs of the base model and in candidates that were not selected, and none in the selected harnesses' \V{e3.gb.sel.re.calls} calls on replication and \V{e3.gb.sel.te.calls} on test or in the \V{e3.gb.grid.calls} calls of the grid, rescue and determinism runs (Appendix~\ref{app:third-runs}).

\section{After the freeze, in the third study}
\label{sec:dev3}

The third study's addenda run from A1 to A\V{e3.add.last} (Appendix~\ref{app:third-deviations}), and four of them bear on a result.
A5 narrowed the scope of an operational check.
The stop that looks for a request stating the row's gold answer had also read the harness texts, and it halted the first search on a format line (``Answer: Yes/No'') in a candidate's system prompt; it now reads each row's own prompt only, so a candidate that carries such a line can be evaluated, as it must be where a harness that names one label is the behaviour H5 measures.
A7 restarts both servers before every search run, after memory ran short next to a task server that had grown, and retries a run once after an out-of-memory failure.
A retried search run draws a new proposer sample, so its search path can differ from the stopped attempt's; one retry nevertheless reproduced the whole sequence of \V{e3.gb.rep37} candidates, including the one that tripped gate~(b).
A10 made gate~(b), the check that a thinking-off call does not begin with the model's own thinking token, record instead of stop in the searches and in the evaluations of the selected harnesses; it stays a stop in the start checks, the grid, the rescue verification and the determinism recheck, and \V{e3.gb.said.trip} of \V{e3.gb.said.calls} search calls tripped it.
A13 records that the descriptive flag of Section~\ref{sec:third-h2} is implemented differently from its frozen definition.
Three statements separate what these entries show.
First, no request, score, parsing or verdict rule changed: each entry states that request bytes, scoring, parsing and call order are unchanged and gives the evidence (the pinned render lists reproduce, the seed's requests equal those of \cfg{AUTHORS-1500}, no test reads the flag).
Second, A5, A7 and A10 changed when a search stops or restarts: the frozen code would have stopped those searches, at a format line in a candidate, after an out-of-memory failure and at gate~(b), so the searches that went on after those stops are not ones the frozen code would have completed, and what it would have selected is unknown.
A5, A7 and A10 were committed before the search ended, and so before any replication or test call; A13 was written after the analysis.
Third, A13 changes no verdict, interval or label, because it concerns the name of a descriptive flag that no test uses.

Before the freeze we computed accuracy only for \cfg{AUTHORS-1500} on the screening rows of the search partition, as the screening rule required.
Everything computed after the outputs existed is labelled where it appears: the agreement with the earlier studies' outputs on MMLU-Pro and the gate~(b) counts of each search run.

\section{The fourth study: design}
\label{sec:fourth}

\paragraph{Why a fourth study.}
Two of the earlier results ended in the same place.
The case study searched on a split that was skewed (the skewed split of the first study, defect C), and in the second and third studies several GEPA harnesses ended near the constant label and answered one label almost throughout: Qwen3-14B on the mortality task, and the Autocast arm (Section~\ref{sec:pattern}).
We had offered one reason, untested: on a binary task the correctness feedback tells the proposer the gold label and with it the base rate, so a search may settle on the majority label.
That reason points at the search set.
If the share of the majority label in the search set drives the collapse, we would expect minority recall to be higher after a search on a balanced set than after one on a skewed set, and higher under an objective that weights the classes than under one that does not.
The fourth study puts that question to two tasks with two small models, and it changes only the search set and the objective; the longevity model itself is not in it, because the question is about the search and the longevity case is where the question arose.

\paragraph{The question and the four comparisons.}
We call the label with the smaller share in a skewed search set the minority label and the other the majority label, and we measure recall of each label on the test partition, which keeps its natural class balance.
Four comparisons are fixed in advance, each per model and task (Table~\ref{tab:fourth-hyp}).
(a) Under the accuracy objective, is minority recall lower for a harness searched at \V{e4.ratio.skew} than for one searched at \V{e4.ratio.bal}?
(b) At \V{e4.ratio.skew}, do a balanced-accuracy objective that weights the classes \V{e4.d.weights} and the accuracy objective differ in balanced accuracy and in minority recall?
(c) Is the size of (a) different on the two tasks?
(d) Within the pool of candidates that one accuracy-objective search already evaluated, does choosing the candidate with the best balanced validation accuracy, instead of the best accuracy, change minority recall, with balanced accuracy not lower by more than \V{e4.d.margin}?
Hypothesis (d) was added after the pilot, which makes it the one comparison that the pilot's outputs shaped; the others were fixed before the pilot (below).
We report majority recall beside (d) and under (b), because a difference in minority recall can come with the opposite difference in majority recall.

\begin{table}[t]
\caption{The fourth study's comparisons, as the preregistration fixes them. Each is per model and task, on the test partition; recalls are of the minority and the majority label, and the estimates are differences of means over \V{e4.d.reps} searches. A comparison with a level is a family of the preregistration, with a Bonferroni level split within the family and no level across families; (c) has no family and no level.}
\label{tab:fourth-hyp}
\centering
\footnotesize
\setlength{\tabcolsep}{3pt}
\resizebox{\ifdim\width>\linewidth\linewidth\else\width\fi}{!}{%
\begin{tabular}{@{}lp{6.3cm}p{3.6cm}p{3.2cm}@{}}
\toprule
Id & Comparison & Estimate & Family and level \\
\midrule
(a) & minority recall, searched at 50:50 against 80:20 (accuracy objective) & difference of means over 5 searches & 4 claims, 98.75\% \\
(b) & balanced objective against accuracy objective at 80:20, in balanced accuracy and in minority recall & difference of means & 8 endpoints, 99.375\% \\
(c) & size of (a), CREAK against Manifold & difference of the two (a) & none, 95\% interval \\
(d) & selecting by balanced validation accuracy within the same pool, balanced accuracy not lower by more than 0.05 & mean of the paired differences per search & 8 endpoints, 99.375\% \\
\bottomrule
\end{tabular}}
\end{table}

\subsection{Tasks, models and partitions}
Arm~A is CREAK \citep{onoe2021creak}, claims about entities to be judged true or false, with the label ``true'' as the minority label in the skewed conditions and the natural balance at \V{e4.ratio.bal}.
Arm~B is a set of resolved binary markets from Manifold Markets, created after the models' knowledge cutoffs, with the question text only and the answer ``Yes'' or ``No''; its natural balance is \V{e4.ratio.mf}, so ``Yes'' is the minority label.
Arm~B is retrospective answering of resolved questions, not forecasting: the text was fetched after resolution without a revision history, so it may carry hints of the outcome.
CREAK's labels are mostly unchecked labels of its training split, and its claims may be in the models' pretraining data.
Both limits apply to every condition, but a shared exposure does not guarantee an equal effect on responses to different harnesses.
The two tasks also differ in information, class balance, difficulty and group structure, so (c) cannot say that knowledge and forecasting differ in a general way.
The models are \smol\ and Qwen3-1.7B (the base of the tuned model of the earlier studies, called \basem\ in the third study and in the report's keys), with thinking off, temperature zero and \V{e4.d.maxtok} tokens at most, on vLLM~\V{e4.hw.vllm} in batch-invariant mode.
The proposer is \intern.
Table~\ref{tab:fourth-rows} gives the sizes.
Items are grouped by text, entity and shared rare phrases so that no group spans two partitions, and the group is the bootstrap unit; CREAK's test partition has a design-effect bound of \V{e4.d.deff.cr} and Manifold's \V{e4.d.deff.mf}.
The licences of the data and our permission to use Manifold's are in the section on licences and use.

\begin{table}[t]
\caption{Sizes of the fourth study's sources. Pool, development and search rows are drawn from the pool before the partitions are fixed; the replication and test partitions are the rows that the studies call; the minority and majority columns count rows on the test partition.}
\label{tab:fourth-rows}
\centering
\footnotesize
\setlength{\tabcolsep}{3pt}
\resizebox{\ifdim\width>\linewidth\linewidth\else\width\fi}{!}{%
\begin{tabular}{@{}lrrrrrrr@{}}
\toprule
Task & Pool & Development & Search & Replication & Test & Test minority & Test majority \\
\midrule
CREAK & 11{,}543 & 554 & 2{,}659 & 1{,}531 & 6{,}799 & 3{,}408 & 3{,}391 \\
Manifold & 4{,}120 & 225 & 1{,}199 & 595 & 2{,}101 & 893 & 1{,}208 \\
\bottomrule
\end{tabular}}
\end{table}

\subsection{Conditions, search and selection}
Each model and task has three conditions.
In the first the search set is balanced at \V{e4.ratio.bal} (the two objectives are one function there).
In the second it is skewed at \V{e4.ratio.skew} with the accuracy objective, and in the third at \V{e4.ratio.skew} with a balanced objective that gives a right answer on a minority row \V{e4.d.weights.min} times the weight of one on a majority row.
Each condition has \V{e4.d.reps} searches, so the study has \V{e4.d.runs} GEPA searches.
A search set has \V{e4.d.rows} rows, \V{e4.d.half} for training and \V{e4.d.half} for validation, drawn from the search partition only; at the skewed ratio it has \V{e4.d.maj80} majority and \V{e4.d.min80} minority rows, and its validation half \V{e4.d.val.maj} and \V{e4.d.val.min}.
All searches of a condition share one fixed pair of halves, and the two skewed conditions share the same pair, so the searches differ in their random streams, which are fixed by a hash of the model, task, condition, replicate and stream name, and not in their rows.
The uncertainty that our intervals cover is therefore variation between searches and between evaluation rows; it does not cover the choice of search items, the direction of the skew or other tasks.
GEPA \citep{agrawal2025gepa} (version \V{e4.d.gepa.ver}) starts from the empty harness, proposes a new system prompt and instruction, tests it on a minibatch of \V{e4.d.mb} rows, and evaluates a candidate on the whole validation half when the minibatch score rises; a search stops after \V{e4.d.calls} metric calls.
The harness is a system prompt of at most \V{e4.d.sys.chars} characters and an instruction of at most \V{e4.d.ins.chars}.
The proposer sees the harness texts and the rows' inputs, outputs and correctness feedback, and no score or weight field; the feedback still lets it infer the gold label, in every condition alike.
The selected harness of a search is the candidate with the best validation score under the search's objective, with ties going to the seed.

The re-selection of (d) uses the \V{e4.d.reps} accuracy-objective searches at the skewed ratio of each model and task.
From each search's own pool it takes the candidate with the highest balanced validation accuracy, computed exactly from integer counts, with ties going to the lowest index, and it evaluates that candidate on the same rows.
If that candidate is the selected one, no call is made and the contrast is zero, which happened in \V{e4.run.k} of the pools.
It costs no search.

\subsection{Evaluation, baselines and statistics}
Every selected harness and every re-selected harness that differs from its search's selection is called once on every row of the replication and the test partition: \V{e4.d.evals.cr} calls per harness on CREAK and \V{e4.d.evals.mf} on Manifold.
This gives \V{e4.run.rep} evaluations on replication and \V{e4.run.test} on test.
Baselines are evaluated but not searched: \cfg{AUTHORS-1500} (the empty harness with the inference settings of the longevity model's authors \citep[STAR Methods, ``LLM inference'', page~\V{e4.cell.page}]{zhavoronkov2026cell}: thinking off, temperature zero and \V{e4.d.maxtok} tokens at most, here applied to models that are not theirs), \cfg{INSTR-1500} (the empty harness plus a short hand-written instruction), \cfg{THINK-512}, \cfg{THINK-LONG-2048} (thinking on, \V{e4.d.tl.cap} tokens) and, on replication only, \cfg{THINK-LONG-2048} with the model card's sampling (temperature \V{e4.d.card.t}, top-p \V{e4.d.card.p}).
The two constant labels are scored on every partition.
All comparisons with baselines and constants are descriptive.

The estimate for (a) is the difference of two conditions' means of minority recall over their \V{e4.d.reps} searches; the interval is the estimate plus and minus a $t$ quantile times the square root of $V=s_A^2/S_A+s_B^2/S_B+V_\text{rows}$, where each condition's own seed variance is used, without pooling, and $V_\text{rows}$ is the variance of the difference over resamples of the partition's groups with the harnesses fixed (\V{e4.boot} resamples).
The degrees of freedom are those of the seed part alone, and each condition's seed standard deviation is floored at \V{e4.floor} before it enters, with the wider of the floored and the unfloored interval reported.
The levels are \V{e4.d.lvl1}\% for the four endpoints of (a) and \V{e4.d.lvl2}\% for the eight of (b) and the eight of (d).
For (d) the seeds are paired and the unit is the difference per search; for (c) the interval is a \V{e4.ci95}\% one without a family.
The number of searches per condition is small, so every statement about the search is indicative: the pilot's seed variation was too large for any number of searches up to five to meet the margin that the preregistration wanted.

\subsection{The pilot and what it changed}
The pilot ran \V{e4.pil.runs} searches on CREAK at the skewed ratio, on development rows and search-partition rows only, from 2026-10-30 to \V{e4.pil.dev.date}.
Its seed standard deviation of minority recall was \V{e4.pil.sd.sm} for \smol\ and \V{e4.pil.sd.bs} for Qwen3-1.7B, which fixed the number of searches at \V{e4.pil.S} per condition and made the claims indicative.
The selected harness was the seed in \V{e4.pil.seed0}.
The cap for the long-thinking baseline was set at \V{e4.d.tl.cap} tokens.
The pilot's check of whether the weighted objective changes the minibatch decision, with a threshold of \V{e4.a11.thr}\% of decisions, did not meet that threshold as defined: \V{e4.pil.r3.n} of \V{e4.pil.r3.d} decisions differed (\V{e4.pil.r3.p}\%).
We did not redefine that check, and in the main study it is reported and is not a stop.
A diagnosis of the pilot's logs showed that the two objectives pick different candidates at the end of the search and that the runs of one seed under the two objectives diverge after a few proposals, through GEPA's ordering of programs, which depends on the aggregate validation score.
So seeds of the two objectives are replicates and not pairs, and the final selection, which the re-selection isolates, became a comparison of its own, (d).
The pilot's requests carried no score or weight field (all \V{e4.pil.reqs} checked).
These changes, and the change of (d)'s balanced-accuracy requirement from a rise to a non-inferiority guard, were made on 2026-11-01, before the preregistration, and were not allowed by the pilot's own rules, which said that no pilot output changes the hypotheses; each is dated and was authorised by us in the preregistration (Appendix~\ref{app:fourth-prereg}).
The pilot's harnesses and outputs do not enter the main study.

\subsection{Preregistration, the run and the change of card}
\label{sec:fourth-run}
The preregistration (SHA256 beginning \V{e4.pre.sha8}) was registered on the Open Science Framework at \V{e4.osf.reg.time}~UTC on \V{e4.osf.reg.date}, \url{https://osf.io/g5apw}, and frozen in commit \V{e4.osf.commit} four minutes later.
The registration is public from the posting of this version, and the ancillary files carry the preregistration, the addenda, the deviation journal and both analysis reports (see the statement on code and data availability).
No call touched a replication or test row before the freeze.
The preregistration names the instance, the card and the driver, and allows one change of card and driver before the first replication call.
The instance's host never freed a card, so on \V{e4.run.card.date} we cloned its disks to a new instance with one \V{e4.hw.gpu} (driver \V{e4.hw.driver}) and used the allowance.
Addenda A2 to A5 record the change, the clone's identity, the pins that we rewrote, and a first start of the chain that stopped before any call because the clone has no \texttt{python3} on the path of a non-login shell.
All \V{e4.d.runs} searches, the replication stage and the test stage ran on that one card and driver, which the chain checked before each stage.
The searches took \V{e4.run.gepa.h} hours, the replication stage \V{e4.run.rep.h} hours, and the study finished on \V{e4.run.end.date}; \V{e4.rc.n} determinism rechecks, which call a thinking-off baseline, \cfg{THINK-512} and the first selected harness of the skewed accuracy condition again for each model and task on \V{e4.rc.rows} rows each, were identical to the first calls in every comparison.
The logged estimate of the spend was \V{e4.run.spend}~CNY of a ceiling of \V{e4.run.ceiling}~CNY; the console's figure is ours to reconcile and is not used here.

\section{The fourth study: results}
\label{sec:fourth-results}

Results are on the test partition unless stated, and each estimate is the difference of means over \V{e4.d.reps} searches per condition, in points, with the preregistered interval at its family level (\V{e4.d.lvl1}\% for (a), \V{e4.d.lvl2}\% for (b) and (d)); the percentile intervals of the report are given in Appendix~\ref{app:fourth-endpoints} with both partitions.
Table~\ref{tab:fourth-ab} and Table~\ref{tab:fourth-d} give the comparisons, and Table~\ref{tab:fourth-means} the means that they are differences of.

\paragraph{Deviation D001: every result of this study is descriptive.}
Section~\V{e4.sec8} of the preregistration lists, among the values computed at the sealed stage and logged before the first replication call, the render-hash lists, which give the SHA256 of the rendered request text of every row.
We did not compute or commit them then.
The pinned code has no separate rendering pass at that stage: each evaluation rendered its own rows just before its calls and saved its list and hash on the instance, in the render record, the chain log and the request log, so every call's full request and the hash of its text were saved before it was sent, and none of the lists was committed before the first replication call.
We disclosed this difference in addendum A4, before the sealed stage, as a change that left every request, row, selection, endpoint and analysis as it was, and we did not enter it as a deviation.
After all results existed and had been read, an independent review of the analysis report (Codex, addendum A11) found that our rule treats a departure from a registered procedure as a deviation, and that on any recorded deviation every claim is descriptive only.
We accepted that, recorded the deviation as D001 on \V{e4.d001.date} at \V{e4.d001.time}~UTC, and reran the frozen analysis with it applied (addendum A13).
All \V{e4.cmp.n} numeric values of the report are equal to those of the original report; only the labels changed.
Our rule therefore makes every result of this study descriptive, and we use no verdict word for it.
The analysis had labelled \V{e4.orig.sup} of \V{e4.orig.all} claims as meeting their criteria before D001 was recorded; we do not use those labels.
The stricter reading was adopted after the results were known, which is a decision taken with the outcome in view, and the intervals below are the preregistered ones, reported without a conclusion drawn from where they fall.

\begin{table}[t]
\caption{Comparisons (a) and (b) on the test partition, in points (a signed difference of means over \V{e4.d.reps} searches per condition, with its interval). (a): minority recall at \V{e4.ratio.bal} minus minority recall at \V{e4.ratio.skew} under the accuracy objective (\V{e4.d.lvl1}\% interval). (b): the balanced objective minus the accuracy objective at \V{e4.ratio.skew}, in balanced accuracy (BA) and in minority recall ($R_{\min}$) (\V{e4.d.lvl2}\% intervals). The last column is the same difference in majority recall (\V{e4.ci95}\% percentile interval, descriptive). Every cell is descriptive (deviation D001).}
\label{tab:fourth-ab}
\centering
\footnotesize
\setlength{\tabcolsep}{3pt}
\resizebox{\ifdim\width>\linewidth\linewidth\else\width\fi}{!}{%
\begin{tabular}{@{}llcccc@{}}
\toprule
Task & Model & (a) $R_{\min}$ & (b) BA & (b) $R_{\min}$ & (b) $R_{\max}$ \\
\midrule
CREAK & Qwen3-1.7B & \ensuremath{+17.3} [\ensuremath{+1.9}, \ensuremath{+32.6}] & \ensuremath{+2.5} [\ensuremath{-2.2}, \ensuremath{+7.2}] & \ensuremath{+8.7} [\ensuremath{-5.9}, \ensuremath{+23.4}] & \ensuremath{-3.7} [\ensuremath{-8.0}, \ensuremath{+0.3}] \\
 & SmolLM3-3B & \ensuremath{+14.9} [\ensuremath{+3.2}, \ensuremath{+26.6}] & \ensuremath{+3.7} [\ensuremath{-1.2}, \ensuremath{+8.6}] & \ensuremath{+11.6} [\ensuremath{-2.7}, \ensuremath{+25.9}] & \ensuremath{-4.2} [\ensuremath{-7.2}, \ensuremath{-0.8}] \\
Manifold & Qwen3-1.7B & \ensuremath{+18.3} [\ensuremath{-4.4}, \ensuremath{+40.9}] & \ensuremath{+2.5} [\ensuremath{-3.2}, \ensuremath{+8.2}] & \ensuremath{+42.8} [\ensuremath{+22.4}, \ensuremath{+63.2}] & \ensuremath{-37.8} [\ensuremath{-44.3}, \ensuremath{-31.3}] \\
 & SmolLM3-3B & \ensuremath{+11.8} [\ensuremath{+4.4}, \ensuremath{+19.2}] & \ensuremath{+0.5} [\ensuremath{-4.3}, \ensuremath{+5.3}] & \ensuremath{+6.7} [\ensuremath{-9.4}, \ensuremath{+22.7}] & \ensuremath{-5.6} [\ensuremath{-11.1}, \ensuremath{-1.0}] \\
\bottomrule
\end{tabular}}
\end{table}

\begin{table}[t]
\caption{Comparison (d) on the test partition, in points: the harness that the same search's pool gives under selection by balanced validation accuracy minus the one selected by accuracy, per model and task (\V{e4.d.reps} paired searches; \V{e4.d.lvl2}\% intervals). The last column is the difference in majority recall (\V{e4.ci95}\% percentile interval, descriptive). Every cell is descriptive (deviation D001).}
\label{tab:fourth-d}
\centering
\footnotesize
\setlength{\tabcolsep}{3pt}
\resizebox{\ifdim\width>\linewidth\linewidth\else\width\fi}{!}{%
\begin{tabular}{@{}llccc@{}}
\toprule
Task & Model & BA & $R_{\min}$ & $R_{\max}$ beside \\
\midrule
CREAK & Qwen3-1.7B & \ensuremath{+3.9} [\ensuremath{-1.1}, \ensuremath{+8.8}] & \ensuremath{+19.7} [\ensuremath{+0.9}, \ensuremath{+38.5}] & \ensuremath{-12.0} [\ensuremath{-20.7}, \ensuremath{-3.3}] \\
 & SmolLM3-3B & \ensuremath{+3.6} [\ensuremath{-2.6}, \ensuremath{+9.8}] & \ensuremath{+11.9} [\ensuremath{-6.3}, \ensuremath{+30.1}] & \ensuremath{-4.7} [\ensuremath{-8.4}, \ensuremath{-1.1}] \\
Manifold & Qwen3-1.7B & \ensuremath{+2.6} [\ensuremath{-3.7}, \ensuremath{+8.9}] & \ensuremath{+37.6} [\ensuremath{+7.6}, \ensuremath{+67.7}] & \ensuremath{-32.5} [\ensuremath{-47.1}, \ensuremath{-17.8}] \\
 & SmolLM3-3B & \ensuremath{+0.6} [\ensuremath{-4.6}, \ensuremath{+5.8}] & \ensuremath{+10.8} [\ensuremath{-10.4}, \ensuremath{+32.0}] & \ensuremath{-9.5} [\ensuremath{-19.6}, \ensuremath{+0.5}] \\
\bottomrule
\end{tabular}}
\end{table}

\begin{table}[t]
\caption{Mean test recall of the minority and majority labels, in percent, over \V{e4.d.reps} searches per condition: the balanced search set (\V{e4.ratio.bal}), the skewed one (\V{e4.ratio.skew}) with the accuracy objective, with the balanced objective, and the accuracy-objective searches re-selected by balanced validation accuracy. $R_{\min}$ and $R_{\max}$ are the recalls of the minority and the majority label.}
\label{tab:fourth-means}
\centering
\footnotesize
\setlength{\tabcolsep}{3pt}
\resizebox{\ifdim\width>\linewidth\linewidth\else\width\fi}{!}{%
\begin{tabular}{@{}llcccccccc@{}}
\toprule
Task & Model & \multicolumn{2}{c}{balanced set} & \multicolumn{2}{c}{skewed, accuracy} & \multicolumn{2}{c}{skewed, re-selected} & \multicolumn{2}{c}{skewed, balanced} \\
& & $R_{\min}$ & $R_{\max}$ & $R_{\min}$ & $R_{\max}$ & $R_{\min}$ & $R_{\max}$ & $R_{\min}$ & $R_{\max}$ \\
\midrule
CREAK & Qwen3-1.7B & 74.9 & 81.1 & 57.6 & 93.4 & 77.3 & 81.4 & 66.3 & 89.7 \\
 & SmolLM3-3B & 83.4 & 85.8 & 68.5 & 93.4 & 80.4 & 88.6 & 80.1 & 89.2 \\
Manifold & Qwen3-1.7B & 19.2 & 83.8 & 0.9 & 99.3 & 38.6 & 66.9 & 43.7 & 61.5 \\
 & SmolLM3-3B & 13.6 & 89.1 & 1.9 & 98.5 & 12.7 & 89.0 & 8.5 & 92.9 \\
\bottomrule
\end{tabular}}
\end{table}

\subsection{(a) Minority recall at a skewed search ratio}
The estimate of (a) is positive in all four cells (Table~\ref{tab:fourth-ab}): the harnesses searched at \V{e4.ratio.skew} have lower mean minority recall than those searched at \V{e4.ratio.bal}.
The interval excludes zero in three cells and includes it for Qwen3-1.7B on Manifold, whose seed-to-seed spread is large.
On CREAK the means fall from \V{e4.te.cr.bs.sm.c55.rmin}\% to \V{e4.te.cr.bs.sm.c82a.rmin}\% (Qwen3-1.7B) and from \V{e4.te.cr.sm.sm.c55.rmin}\% to \V{e4.te.cr.sm.sm.c82a.rmin}\% (\smol) while mean majority recall rises, to \V{e4.te.cr.bs.sm.c82a.rmaj}\% and \V{e4.te.cr.sm.sm.c82a.rmaj}\%.
Under the skewed accuracy objective, mean minority recall on Manifold was \V{e4.te.mf.bs.sm.c82a.rmin}\% for Qwen3-1.7B and \V{e4.te.mf.sm.sm.c82a.rmin}\% for \smol, against \V{e4.te.mf.bs.sm.c55.rmin}\% and \V{e4.te.mf.sm.sm.c55.rmin}\% at \V{e4.ratio.bal}, and mean majority recall was \V{e4.te.mf.bs.sm.c82a.rmaj}\% and \V{e4.te.mf.sm.sm.c82a.rmaj}\%.
The replication partition gives estimates of the same sign in all four cells (Appendix~\ref{app:fourth-endpoints}).

\subsection{(b) The balanced objective}
Relative to the accuracy objective, the balanced objective has a positive minority-recall estimate in all four cells (Table~\ref{tab:fourth-ab}).
The interval excludes zero in one cell, Qwen3-1.7B on Manifold, and includes it in the other three.
The balanced-accuracy estimates are positive and small, and every interval includes zero, so in no cell are both of (b)'s lower bounds above zero.
The majority-recall estimates are negative in all four cells, and the interval excludes zero in three of them (it includes zero for Qwen3-1.7B on CREAK).
For Qwen3-1.7B on Manifold the balanced objective differs from the accuracy objective by \V{e4.te.mf.bs.b2} points in minority recall and by \V{e4.te.mf.bs.bm} points in majority recall (signed differences of means, Table~\ref{tab:fourth-ab}), and its mean balanced accuracy is near \V{e4.te.mf.bs.sm.c82b.ba}\%.

\subsection{(d) Selecting from the same pool by balanced validation accuracy}
Re-selection had a positive minority-recall estimate in every test cell (Table~\ref{tab:fourth-d}).
Its interval excluded zero for Qwen3-1.7B on both tasks and included zero for \smol\ on both.
The balanced-accuracy estimates are positive, every interval includes zero, and every lower bound is above $-\V{e4.d.margin.pts}$ points, the margin of the guard of (d).
The majority-recall estimates, shown beside (d), are negative in every cell; the interval excludes zero in three cells and includes it for \smol\ on Manifold.
The re-selected harness of \smol\ on CREAK differs from the harness of the balanced-objective searches by \V{e4.te.cr.sm.cs.rmin} points in minority recall (\V{e4.te.cr.sm.cs.rmin.lo} to \V{e4.te.cr.sm.cs.rmin.hi}), and that of Qwen3-1.7B on CREAK by \V{e4.te.cr.bs.cs.rmin} points \CI{\V{e4.te.cr.bs.cs.rmin.lo}}{\V{e4.te.cr.bs.cs.rmin.hi}} (descriptive, independent searches).

\subsection{(c) The two tasks}
The difference between the two tasks' (a), CREAK minus Manifold, is \V{e4.cx.bs} points \CI{\V{e4.cx.bs.lo}}{\V{e4.cx.bs.hi}} for Qwen3-1.7B and \V{e4.cx.sm} \CI{\V{e4.cx.sm.lo}}{\V{e4.cx.sm.hi}} for \smol\ (\V{e4.ci95}\% intervals).
Both intervals contain zero, which says that the data do not separate the two tasks' sizes of (a); it does not say that they are equal, and the study defines no equivalence margin.

\subsection{What the tasks' accuracy leaves for a collapse to mean}
On CREAK the harnesses answer far above the constant label: the means of accuracy minus the constant label's accuracy run from \V{e4.rng.cr.hc.min} to \V{e4.rng.cr.hc.max} points over the four conditions of the two models, so minority recall there is a statement about answers.
On Manifold, mean accuracy did not exceed that of the constant ``No'' (\V{e4.te.mf.bs.k.acc}\%): the differences run from \V{e4.rng.mf.hc.min} to \V{e4.rng.mf.hc.max} points, and mean balanced accuracy ranged from \V{e4.rng.mf.ba.min}\% to \V{e4.rng.mf.ba.max}\%.
There, higher minority-recall estimates coincided with lower majority-recall estimates, and the balanced-accuracy contrasts were positive but all their intervals included zero.
That is also why Qwen3-1.7B's differences in minority recall on Manifold, \V{e4.te.mf.bs.b2} points under (b) and \V{e4.te.mf.bs.d2} under (d), come with differences in majority recall of \V{e4.te.mf.bs.bm} and \V{e4.te.mf.bs.dm} points.

\subsection{Baselines, the replication partition and the search itself}
The baselines are descriptive (Appendix~\ref{app:fourth-endpoints}).
On CREAK, both \cfg{INSTR-1500} and \cfg{THINK-LONG-2048} had higher mean balanced accuracy than every searched condition for both models.
Most searched-condition means were above \cfg{AUTHORS-1500}; the exception was \smol\ under the skewed accuracy objective, at \V{e4.te.cr.sm.sm.c82a.ba}\% versus \V{e4.te.cr.sm.bl.aut.ba}\%.
On Manifold, \cfg{AUTHORS-1500} reaches \V{e4.te.mf.bs.bl.aut.ba}\% and \V{e4.te.mf.sm.bl.aut.ba}\%.
No selected harness had a parse rate below one half, and the thinking-switch check found no call with reasoning content in any search or evaluation.
In the \V{e4.run.rep} replication evaluations and the \V{e4.run.test} test evaluations the harnesses of each search differ in their ratio and objective only.
In the \V{e4.a11.runs} balanced-objective searches, weighted and unweighted decisions differed in \V{e4.a11.n} of \V{e4.a11.d} minibatch comparisons (\V{e4.a11.p}\%).
The counts were \V{e4.a11.cr.bs} of \V{e4.a11.cr.bs.d} and \V{e4.a11.cr.sm} of \V{e4.a11.cr.sm.d} on CREAK, and \V{e4.a11.mf.bs} of \V{e4.a11.mf.bs.d} and \V{e4.a11.mf.sm} of \V{e4.a11.mf.sm.d} on Manifold, for Qwen3-1.7B and \smol\ respectively.
The overall share exceeded the pilot's \V{e4.a11.thr}\% threshold, which the pilot had not met (addendum A11).
This diagnostic was descriptive and did not stop the main study.

\subsection{What the fourth study shows and does not show}
The estimates go in one direction in every cell: lower minority recall at the skewed ratio, and, for the balanced objective and for the re-selection, higher minority-recall and lower majority-recall estimates. Three of the four minority-recall intervals for (b), and both \smol\ intervals for (d), include zero. All eight balanced-accuracy contrast intervals include zero; their point estimates are positive.
The study cannot tell how much of this is the search set's ratio and how much is the proposer's use of the correctness feedback, because it varies the ratio and not what the proposer sees.
It covers one direction of skew, two tasks, one of which has near-chance mean balanced accuracy for these models, two models and \V{e4.d.reps} searches per condition, so its statements about the search are indicative, and under D001 none is more than a description.
On the question that started it, the pattern in the 14B and Autocast harnesses, the study adds a controlled variation on two other tasks and does not show that class ratio explains those harnesses.

\section{Discussion and limitations}
\label{sec:discussion}

\paragraph{What the results say.}
The gain of the thinking-off configuration over \cfg{THINK-512} comes mostly from rows that \cfg{THINK-512} left unparsed in \V{cb.h1.visibility} of \V{cb.fam1} cells of the first two studies (test verdicts), and again on BoolQ and MMLU-Pro in the third, so a search that starts from a configuration like this can find ``repairs'' that a thinking-off baseline already provides.
A thinking budget in the serving engine lowers truncation as well (H3), which is another way of reducing unparsed rows.

Baseline adequacy depends on the model and the task.
In the tuned 1.7B model the thinking-off setting was not meaningfully worse than any of four GEPA harnesses, and it beat the strongest constant label on the letters.
In Qwen3-14B it was below that constant, and the clean-start harnesses were more accurate than it by \V{ab.h2.te.0} points on test (\V{ab.h2.re.0} and \V{ab.h2.re.1} on replication).
In the third study thinking off was non-inferior to all four BoolQ harnesses; \V{e3.cnt.seed.mm} of the four MMLU-Pro searches kept it and the fourth comparison was inconclusive on test; and it fell short on FPB, where the harnesses beat it and the constant label.
On the LongevityBench letters, the baseline beat the strongest constant label only for the model tuned on that format, and GEPA beat the baseline only where it was below that constant; FPB shows that this is not a rule, because the baseline there was close to the constant (Table~\ref{tab:third-acc}) and the harnesses beat both.
On the letters GEPA ran around the tuned 1.7B model and the 14B only, which differ in tuning as well as in size, so the studies cannot separate the two.
A harness gain has to be read against trivial baselines: the 14B's harnesses gain \V{ab.h2.te.0} points on test over the thinking-off setting and are not detectably above the strongest constant label, so the same gain is large against one baseline and not detectable against another.

Where prospective replication confirmed the distortion (\V{cb.h4.holds} of \V{cb.h4.evaluable} evaluable pairs, \V{cb.h4.byc} of them pairs whose quantity is fixed by an identity of the design), the distortion was large and, within five points, the same on the check sample as on the selection sample.
Some systematic distortions therefore persist across samples, and a replication run through the same pipeline reproduces them instead of revealing them.
In the other pairs the claim fails (a descriptive label) or is inconclusive, and a defect whose claim fails need not be harmless.
Of the audit checks, the token reconciliation fired on no clean instance.
The others also fired on clean pipelines: the gain rule of prospective replication on \V{fa.prgain.n} of the \V{fa.ir4.of} clean instances, the trivial baselines (TB2) on \V{fa.tb2.n}, the simple-configuration-family rules on up to \V{fa.scf1.n} and the input-reading rule IR4 on all \V{fa.ir4.n}, so IR4 does not separate defective from clean pipelines.
The detections themselves are largely true by construction.
The grid verdicts of H1--H3 did not depend on the GPU for the one model that ran on both, although most of its token sequences did, and the longevity fine-tuning bought letter accuracy and cost general accuracy, with the limits stated in Section~\ref{sec:finetune}.

\paragraph{Lessons for harness optimization.}
Measure what the model already does before searching: a thinking-off baseline, the constant labels and, where one exists, the base model.
Read a gain against those and not against the start, because the start may answer nothing and the baseline may sit at the constant.
Report truncated and non-committed answers apart from wrong ones, since a gain that comes from unparsed rows is a gain in visibility, and reconcile the prompt tokens of each call with the request that was meant, which found the defect that replication could not.
Run these checks after every round of a self-improving loop, because each round is a selection of the kind studied here.
We studied one round of selection and no recursive loop; whether early visibility gains turn into quality later in a loop, and whether a loop can leave a start like the one that GEPA could not repair at 14B, this study does not answer.

\paragraph{What the fourth study adds, and what it does not.}
The fourth study moves the collapse from an observation to a variation we control.
On CREAK and Manifold, with \smol\ and \basem, the estimate of minority recall is lower for harnesses searched on a set skewed at \V{e4.ratio.skew} than for those searched on a balanced set in all four test cells (the interval excludes zero in three), and the estimates for a balanced objective and for a re-selection of the same search's pool by balanced validation accuracy are higher for minority recall and lower for majority recall in all four cells (Section~\ref{sec:fourth-results}).
Three of the four minority-recall intervals for (b), and both \smol\ intervals for (d), include zero.
These estimates go in the direction that the case study's skewed split, the 14B and Autocast suggested; they are differences between procedures on these two fixed search sets, not a measure of how a search treats a class in general.
The two tasks differ in what the models can do: on CREAK the harnesses answer far above the constant label, and on Manifold no condition's mean accuracy exceeds the constant ``No'', so differences in minority recall there may mainly reflect which label the harness favours.
The study cannot separate the search set's ratio from what the proposer infers from the correctness feedback, and it does not show that class ratio explains the 14B or Autocast harnesses.
Every result of the fourth study is descriptive, because the render-hash lists that its preregistration promised were produced per evaluation and not committed before the first replication call, which we recorded as deviation D001 only after the results were known, on an independent review.
The lesson for a search is the plain one: report recall of each label and the share of the majority label next to accuracy, and, where the classes are unequal, check what a balanced objective or a selection by balanced validation accuracy does to both, since in every cell the estimate of minority recall was higher where the estimate of majority recall was lower.

\paragraph{What the pattern does not show.}
The first three studies cannot separate label skew, label count and task type. The fourth varies search-set ratio and objective on two binary tasks, but does not show that skew explains the earlier collapse.
The pattern of Section~\ref{sec:pattern} rests on few tasks, and its collapse side is descriptive (the 14B on the mortality task) and exploratory (Autocast) only.
Two relations to other work are hypotheses, not findings.
\citet{braun2025acquiescence} finds a lean toward ``no'' in yes/no questions, for English only; the 14B's thinking-off setting answers ``Yes'' in \V{cl.te.l42.aut.share} of its parsed mortality answers on test, against \V{cl.te.l42.h0.share} and \V{cl.te.l42.h1.share} for the two harnesses, so the setting leans toward ``No'' where the harnesses lean toward ``Yes'', and whether the first is such a lean we did not test.
\citet{cho2026optimismbench} find that, for Qwen models, post-training moves the skew of forecasts negative in each of five pairs of base and chat models; the Autocast harnesses converge on the constant ``no'', which may be the same direction, and this exploratory arm cannot test it.

\paragraph{What the harness texts contain.}
We read the third study's selected harness texts after the outputs existed, and the checks in this paragraph were computed after the outputs existed, are descriptive and were not preregistered (Appendix~\ref{app:third-gepa}).
The proposer wrote statements that look derived from search rows into eleven of the twelve texts that are not the seed: figures and short quoted fragments in the BoolQ, FPB and MMLU-Pro texts, and events and dates stated as facts in the Autocast texts, a channel that the \V{gp.copy.run}-character copy rule does not catch.
The twelfth, shown in Appendix~\ref{app:third-gepa}, holds general rules and two short quoted legal phrases.
A lexical check flags the rows that share a capitalised word or an acronym, a date, a number or a run of five words with a text; it compares strings and is not named-entity recognition.
Each FPB harness stays above the constant and above \cfg{AUTHORS-1500} on the rows the check does not flag, which are \V{e3.ovr.fp.sno.min}\% to \V{e3.ovr.fp.sno.max}\% of the test rows.
On test, its gain over the constant on those rows is \V{e3.ovr.fp.dno.min} to \V{e3.ovr.fp.dno.max} points (unadjusted intervals from \V{e3.ovr.fp.dno.lo} to \V{e3.ovr.fp.dno.hi}), against \V{e3.ovr.fp.dall.min} to \V{e3.ovr.fp.dall.max} on all test rows, and its gain over \cfg{AUTHORS-1500} is \V{e3.ova.fp.tno.min} to \V{e3.ova.fp.tno.max} points (unadjusted lower bounds from \V{e3.ova.fp.tno.lo}), against \V{e3.ova.fp.tall.min} to \V{e3.ova.fp.tall.max} on all test rows, so the gain over thinking off is not concentrated on the flagged rows.
On replication the gain on the rows the check does not flag is \V{e3.ovr.fp.rno.min} to \V{e3.ovr.fp.rno.max} points over the constant, with lower bounds from \V{e3.ovr.fp.rno.lo}, and \V{e3.ova.fp.rno.min} to \V{e3.ova.fp.rno.max} points over \cfg{AUTHORS-1500}, with lower bounds from \V{e3.ova.fp.rno.lo}.
On test, a harness's accuracy differs by at most \V{e3.ovr.fp.accgap.max} points between the flagged and not flagged rows, while the constant is weaker on the flagged rows (\V{e3.ovr.fp.cov.min}\% to \V{e3.ovr.fp.cov.max}\%, against \V{e3.ovr.fp.cno.min}\% to \V{e3.ovr.fp.cno.max}\%), so the gain over the constant is larger there (\V{e3.ovr.fp.dov.min} to \V{e3.ovr.fp.dov.max} points).
Rows the check does not flag are not shown to be free of shared content, because the check misses paraphrase.
The rule is broad: its entity rule has no list of common words, so it over-counts shared content, and overlap of five words is nearly absent (at most \V{e3.ovr.all.g5max.n} rows of any partition), so the count says little about claims written in plain lowercase words.
On Autocast the difference from the constant on test lies between $\V{e3.ovr.au.dno.min}$ and $\V{e3.ovr.au.dno.max}$ points on the not flagged rows and between $\V{e3.ovr.au.dov.min}$ and $\V{e3.ovr.au.dov.max}$ on the flagged rows.
In \V{e3.ovr.au.wo.npos} of the \V{e3.ovr.au.wo.n} comparisons (four on test, four on replication), the difference is larger on the flagged rows, by \V{e3.ovr.au.wo.pmin} to \V{e3.ovr.au.wo.pmax} points ($\V{e3.ovr.au.wo.negv}$ in the eighth).
We did not test this.
It fits a small gain on rows that the check flags as sharing a string with a text, and the two groups of rows also differ in content.
As a hypothesis that we did not test, on a binary task the correctness feedback reveals the gold label and with it the base rate, which may help a search settle on the majority label, as happened with the 14B on the mortality task and with the Autocast harnesses.

\paragraph{Limitations.}
The studies cover five models, one of them the tuned model that the case study searched around and one a single larger model from its family, eight benchmarks and six defects chosen from one line of work, so H4 says how replication behaves for these defects and not for the space of defects.
``Prospective'' has limits: earlier work had called most LongevityBench rows with other configurations, callers and one model variant, and the rows the case study had held out are excluded.
Several endpoints carry a star in Table~\ref{tab:glance} and Table~\ref{tab:h4}: their quantity is fixed by an identity of the design or by the observed parse pattern, so their verdicts say little beyond the mechanism, although they still depend on the intervals.
For the skewed split (C) the frozen analysis resamples the defective and the clean instance independently although they reuse the same participant components, so their covariance is ignored and the intervals may be too wide or too narrow; the four claims for C (one fails, three are inconclusive) are the ones that could depend on it (Appendix~\ref{app:h4}).
Power is mixed.
The H2 endpoint of \intern\ on GSM8K was expected to be inconclusive and is not (an informal expectation, Appendix~\ref{app:power}), H4 has no planned power, and its error control covers ``holds'' only.
The GEPA arms searched prompt text only, with one proposer, four runs in each of the first two studies and \V{e3.cnt.runs} in the third, so their results are about that space; the broken start of the second study could not be repaired within it.
The two 1.7B arms are descriptive: the tuned and the base model differ in every weight and in training data, the intervals are unadjusted, and the classes are not corrected for the number of cells read.
The extended scorer and the role-break rule were fixed after the pilot's parse audit.
Unparsed counts as wrong by rule, so accuracy mixes parsing and correctness, and an unparsed answer may hide a correct answer, a wrong one or none; the study measures differences between configurations as served and not what the models know.
The batch-invariant server differs from the standard one call for call, and each study ran on one GPU type, the first and third on an RTX~5090 and the second on an \V{s2.gpu}; the second study's test grid for Qwen3-14B ran on one physical GPU and its other test runs on another of the same type, after a recheck that found no difference.
Our intervals resample participant components (items in the two general pools) at a frozen harness, zero temperature and fixed prompts, so they cover item sampling and not prompt or judge variance \citep{messing2026hidden}.
The freeze guards the rules and not model-specific tuning, which a preregistration for the next model would \citep{thomas2026next}, and the second and third preregistrations were written after the earlier studies' results were known, though before any call on their own partitions.
Zero temperature may turn some thinking into repetition; for the two models with a seeded arm in the first study substantial truncation persists under sampling (Appendix~\ref{app:seeded}), and for the other models it is untested.

The third study adds its own limits.
H5 is confirmatory for FPB only: BoolQ did not meet its premise, MMLU-Pro is descriptive for H1 and H5, and Autocast fell \V{e3.s0.au.short} rows short of the size rule and is exploratory, with no verdict and unadjusted intervals.
\V{e3.cnt.h2.triv} of the \V{e3.cnt.h2.te.holds} H2 endpoints that hold are trivial, because the selected harness is the seed.
Public benchmarks cannot be cleared of pretraining exposure, which may raise baseline accuracy and so let fewer pairs qualify, and its effect on the comparison with the constant is unknown; for Autocast the questions resolved before these models were trained, so the outcomes may be in their pretraining data.
The fourth study has a recorded deviation (D001, Section~\ref{sec:fourth-results}) and its power is low: its \V{e4.d.reps} searches per condition were too few for the margin the preregistration wanted, so its statements about the search are indicative, and its intervals cover search-to-search and row variation and not the choice of search items, the direction of the skew or other tasks.
Its second task is retrospective answering of resolved market questions whose text may carry hints of the outcome, and for these models its mean balanced accuracy is near chance, and its first task's labels are largely unchecked and may be in the models' pretraining data.
The fourth study did not use the tuned longevity model.
A diagnostic that the pilot had set as a gate, whether the weighted objective changes minibatch decisions, came out at \V{e4.a11.p}\% in the main study and was descriptive there.
The addenda of the third study run from A1 to A\V{e3.add.last} (Appendix~\ref{app:third-deviations}): four bear on a result.
None changed a request, a score, parsing or a verdict rule, but A5, A7 and A10 changed when a search stops or restarts, so the frozen code would have stopped those searches and what it would have selected is unknown; A13 changes no verdict, interval or label (Section~\ref{sec:dev3}).
One of them, A13, records that the descriptive flag for a run that accepted no candidate is implemented as the selected harness being the seed, which is not its frozen definition; \V{e3.cnt.noacc.code} runs are affected and no verdict.
The analysis report's own field gives the H5 family as \V{e3.fam.h5.report}, where the preregistration and the adjusted interval of every confirmatory H5 endpoint use \V{e3.fam.h5}; the descriptive and exploratory H5 estimates carry unadjusted \V{c.ci} intervals.
A retried search draws a new proposer sample, so its path can differ from the stopped attempt's.
The base model's agreement with the second study's outputs on MMLU-Pro is below \smol's, and we did not test whether the GPU explains it.
We read the selected harness texts of the third study only after the outputs existed, and the lexical check on them was computed afterwards, is descriptive and was not preregistered, and uses a broad rule that cannot see claims in plain lowercase words or paraphrase, so a row it does not flag is not shown to be free of shared content (Appendix~\ref{app:third-gepa}).
We used Claude (Claude Code and claude.ai) and Codex for code, drafting and review, and we verified all analyses and claims.

\paragraph{Broader impact.}
This paper contributes methodology for evaluating harness search and the pipelines that report it: a preregistered design, audit checks and a record of what each result rests on.
It builds no new capability, since the models and benchmarks are public and the benchmark behind the longevity tasks derives from public survey data, and the accuracies, those of small and mid-sized models on research tasks, do not support any use of these models for medical advice or clinical decisions.
One misreading to avoid is that a thinking-off baseline is always enough, and the studies show cells in which it is not.

\section{Conclusion}
\label{sec:conclusion}

A harness search around a longevity-tuned model reported a rescue whose gain came from rows the start left unparsed.
In three designs fixed in advance, most of the gain of a thinking-off configuration over a capped thinking configuration came from rows the capped configuration left unparsed, in \V{cb.h1.visibility} of \V{cb.fam1} cells of the first two studies (test verdicts), and that configuration was non-inferior to the rescue in most cells of three models of 1.7B to 8B parameters and to four GEPA-selected harnesses at 1.7B.
At 14B the clean-start harnesses beat it by \V{ab.h2.te.0} points on test (\V{ab.h2.re.0} and \V{ab.h2.re.1} on replication) but were not detectably above the strongest constant label, which they matched on the mortality task by answering ``Yes'' almost throughout.
In the third study GEPA beat the constant label on FPB (confirmatory); thinking off was non-inferior to all four BoolQ harnesses, \V{e3.cnt.seed.mm} of the four MMLU-Pro searches kept the seed and the remaining comparison was inconclusive on test; and on the exploratory Autocast arm the harnesses ended near the constant while answering one label almost throughout.
In the fourth study, minority-recall estimates were lower after \V{e4.ratio.skew} than after \V{e4.ratio.bal} searches in all four test cells.
Relative to accuracy-based search and selection, the balanced objective and re-selection had higher minority-recall estimates and lower majority-recall estimates.
Five of these eight minority-recall intervals, and all eight balanced-accuracy contrast intervals, included zero.
All results are descriptive under D001: render-hash lists were not computed at the sealed stage or committed before the first replication call, and the departure was classified as a deviation only after the results were known.
What a harness can repair therefore depends on the model and the task.
In the cases with large gains over thinking off, constant baselines showed whether accuracy exceeded a trivial answer; majority-label shares showed whether the harness collapsed to one label.
The pattern rests on few tasks that differ in label skew, label count and task type, and its collapse evidence is descriptive or exploratory.
In \V{cb.h4.holds} of the \V{cb.h4.evaluable} evaluable (defect, model) pairs, the defect's distortion of the reported number was larger than five points on the selection sample and reappeared, within five points, on the check sample; \V{cb.h4.byc} of these pairs are ones whose quantity is fixed by an identity of the design.
Reporting truncated and non-committed answers apart and running a thinking-off baseline and constant-label baselines cost little and changed what the gains meant.
Of the audit checks, the token reconciliation fired on no clean instance, and the others also fired on clean pipelines, from \V{fa.prgain.n} of the \V{fa.ir4.of} clean instances (the gain rule of prospective replication) to all \V{fa.ir4.n} (IR4, the rule that reads the rendered inputs).

\section*{Licences and use}
No model's weights were trained or modified in this work: every model was used as released, and the four studies evaluate harnesses around frozen models.
We checked each licence or term of use in its primary source on \V{e4.lic.0.date}: the model card or dataset card on the Hugging Face Hub, the licence file of the repository, or the owner's terms page.
Table~\ref{tab:licences} gives each item, what the source states and where, and the date.

\begin{table}[t]
\caption{Licences and terms of the datasets, models and main software, each read in its primary source on the date shown. The columns give the terms as the source states them; they are not legal advice.}
\label{tab:licences}
\centering
\scriptsize
\setlength{\tabcolsep}{3pt}
\begin{tabular}{@{}p{2.6cm}p{5.9cm}p{4.9cm}p{1.6cm}@{}}
\toprule
Item & Licence or terms as the source states them & Source & Checked \\
\midrule
LongevityBench & CC-BY-NC-4.0; the card adds that the underlying source datasets (NHANES, GTEx, GEO, Olink, OpenGenes, SynergyAge) retain their original licences and access conditions & \texttt{huggingface.\allowbreak{}co/\allowbreak{}datasets/\allowbreak{}insilicomedicine/\allowbreak{}longebench} & 2026-11-07 \\
MMLU-Pro & MIT & \texttt{huggingface.\allowbreak{}co/\allowbreak{}datasets/\allowbreak{}TIGER-\allowbreak{}Lab/\allowbreak{}MMLU-\allowbreak{}Pro} & 2026-11-07 \\
GSM8K & MIT & \texttt{huggingface.\allowbreak{}co/\allowbreak{}datasets/\allowbreak{}openai/\allowbreak{}gsm8k} & 2026-11-07 \\
BoolQ & CC-BY-SA-3.0 & \texttt{huggingface.\allowbreak{}co/\allowbreak{}datasets/\allowbreak{}google/\allowbreak{}boolq} & 2026-11-07 \\
Financial PhraseBank & CC-BY-NC-SA-3.0 & \texttt{huggingface.\allowbreak{}co/\allowbreak{}datasets/\allowbreak{}takala/\allowbreak{}financial\_phrasebank} & 2026-11-07 \\
Autocast & code MIT; the README says the data is hosted on GitHub with Metaculus's permission for research purposes only & \texttt{github.\allowbreak{}com/\allowbreak{}andyzoujm/\allowbreak{}autocast} & 2026-11-07 \\
CREAK & MIT & \texttt{github.\allowbreak{}com/\allowbreak{}yasumasaonoe/\allowbreak{}creak} & 2026-11-07 \\
Manifold Markets & the terms of service permit academic research, personal projects and non-commercial use, and data from outside the public API may not be collected; this study also has Manifold's written permission of 2026-10-29 for non-commercial research (no user identified, no personal information, no question text published) & \texttt{manifold.\allowbreak{}markets/\allowbreak{}terms} & 2026-11-07 \\
\addlinespace
Qwen3-1.7B-Longevity & CC-BY-ND-4.0 & \texttt{huggingface.\allowbreak{}co/\allowbreak{}insilicomedicine/\allowbreak{}Qwen3-\allowbreak{}1.\allowbreak{}7B-\allowbreak{}Longevity} & 2026-11-07 \\
Qwen3-1.7B & Apache-2.0 & \texttt{huggingface.\allowbreak{}co/\allowbreak{}Qwen/\allowbreak{}Qwen3-\allowbreak{}1.\allowbreak{}7B} & 2026-11-07 \\
Qwen3-14B & Apache-2.0 & \texttt{huggingface.\allowbreak{}co/\allowbreak{}Qwen/\allowbreak{}Qwen3-\allowbreak{}14B} & 2026-11-07 \\
SmolLM3-3B & Apache-2.0 & \texttt{huggingface.\allowbreak{}co/\allowbreak{}HuggingFaceTB/\allowbreak{}SmolLM3-\allowbreak{}3B} & 2026-11-07 \\
Intern-S1-mini & Apache-2.0 & \texttt{huggingface.\allowbreak{}co/\allowbreak{}internlm/\allowbreak{}Intern-\allowbreak{}S1-\allowbreak{}mini} & 2026-11-07 \\
\addlinespace
vLLM & Apache-2.0 & \texttt{github.\allowbreak{}com/\allowbreak{}vllm-\allowbreak{}project/\allowbreak{}vllm} & 2026-11-07 \\
GEPA & MIT & \texttt{github.\allowbreak{}com/\allowbreak{}gepa-\allowbreak{}ai/\allowbreak{}gepa} & 2026-11-07 \\
NumPy & BSD-3-Clause (the licence file also lists the licences of bundled code) & \texttt{github.\allowbreak{}com/\allowbreak{}numpy/\allowbreak{}numpy} & 2026-11-07 \\
SciPy & BSD-3-Clause & \texttt{github.\allowbreak{}com/\allowbreak{}scipy/\allowbreak{}scipy} & 2026-11-07 \\
\bottomrule
\end{tabular}
\end{table}

LongevityBench is licensed for non-commercial use (CC-BY-NC-4.0), and its card adds that the underlying source datasets keep their own licences and access conditions, which apply to our use as well; this work is non-commercial research.
The longevity model that the first study searched around is licensed CC-BY-ND-4.0; we used it unchanged, trained nothing on it and redistribute no weights.
Financial PhraseBank is CC-BY-NC-SA-3.0 and Autocast's data is hosted with the questions' source's permission for research only; we use Autocast's question text alone.
Manifold Markets' terms permit academic research and non-commercial use through its public API.
We also have Manifold's written permission of \V{e4.manifold.date} for this non-commercial research, on the conditions that we identify no user, publish no personal information, and do not redistribute question text; for Manifold, the paper and the ancillary files publish aggregate results and no question text or user information.
The first study's tuned model and its benchmark are the work of the model's authors \citep{zhavoronkov2026cell}, whose paper is open access under CC BY 4.0; the inference settings of our \cfg{AUTHORS-1500} configuration (thinking off, temperature zero, top-p one and a generation length of \V{e4.d.maxtok} tokens at most) are those that paper gives for its own evaluation of its fine-tuned models (STAR Methods, ``LLM inference'', page \V{e4.cell.page}; the passage is quoted in the licence check).
The software is vLLM~\V{e4.hw.vllm}, GEPA~\V{e4.d.gepa.ver}, and NumPy~\V{e4.sw.numpy} and SciPy~\V{e4.sw.scipy} for the analysis.

\section*{Code and data availability}
The repository of the studies is not public at present.
The paths, commits, tags and SHA256 values that this paper names belong to files in our repository, kept so that a later public export can be checked against them (Appendix~\ref{app:repro}); the run logs hold benchmark item text and model outputs and will not be released.

This arXiv version carries ancillary files and a short README that describes them: the frozen preregistration of each of the four studies, the addenda file of each, the aggregate analysis reports of all four (the fourth study's two: the report as first produced and the report rerun with deviation D001), the fourth study's deviation journal, and the specification of the third study's lexical check (Appendix~\ref{app:third-gepa}).
The fourth study's preregistration is also registered at \url{https://osf.io/g5apw}, which is public from the posting of this version; it was registered with an embargo until \V{e4.osf.embargo}, which we end then.
The preregistrations, the reports, the journal and the specification are attached byte for byte, and the addenda files as they stood at the posting with the row keys of any rows that they name withheld, because a row key identifies a benchmark row (in the fourth study's addenda we withheld three host-key fingerprints of a cloud instance in the same way).
The freeze times of the first three studies are recorded only in our private repository, in the hash file that each freeze commit adds, and they are not independently timestamped; the fourth study's registration time is recorded by the Open Science Framework, and its freeze commit follows it.
The SHA256 of each frozen file (Appendix~\ref{app:repro}) therefore lets a reader check that the ancillary copy is the file that was frozen, and, for the first three studies, not when it was frozen.
A filtered public export of the rest is planned and will be linked in a later version of this paper: the code and its tests and the hash manifests, with no item-level or row-level content derived from the held-out rows.

The licences and terms of every dataset, model and main software, each read in its primary source, are in the section on licences and use before this statement; no model's weights were trained or modified. We redistribute no model weights or Manifold question text. The frozen third-study preregistration contains a quoted Autocast screening question, as disclosed in the ancillary README; benchmark item text is otherwise withheld from the reported results.
The harness texts that a language model wrote in the first two studies are printed in Appendices~\ref{app:gepa} and~\ref{app:gepa14} with row-like passages withheld, under the rule stated there; version \V{ver.one} of this paper printed the first study's texts in full and version \V{ver.two} printed those of both studies.

\section*{Author Contributions}

\noindent\textbf{Bowen Xu:} Conceptualization, Methodology, Software, Investigation, Formal analysis,
Visualization, Writing -- original draft, Writing -- review \& editing, Project administration. Xu conceived and led the study.

\noindent\textbf{Boyu Chen:} Methodology, Supervision, Writing -- review \& editing.

\bibliographystyle{tmlr}
\bibliography{references}

\appendix

\section{Background: the case study this study follows}
\label{app:v05}

\paragraph{The case study.}
The earlier study (v0.5 in the project's records) ran a harness search around a frozen Qwen3-1.7B longevity model, served in its tied variant by FreeToken \citep{yang2026freetoken}.
The search started from a parent configuration, \cfg{CFG-THINK-64-ORIG} (thinking on, \V{k.t64} tokens for thinking and answer), and, driven by its failures on one shard, produced B1, a conditional rescue: when the first call used nearly all of its tokens, call again with a no-think suffix.
B1's gain in parse rate replicated on a second shard.
A later audit found a simpler comparator, thinking disabled, that was as accurate as B1; it then found that the project's caller had sent each benchmark row's reference answer to the model as a prior assistant turn in every historical call that could be checked.
Prompt-token reconciliation, the server's prompt-token count against the local count of the intended request, was the evidence: the counts differed by the length of the reference turn.

\paragraph{The corrected re-evaluation.}
After the caller was corrected, a preregistration was frozen (\V{v05.freeze}) and run once on \V{bg.rows} holdout rows that no earlier experiment had called, though the comparator configurations had been defined after another line of experiments had called other holdout rows, among them \V{v05.prior.rows} with a request equivalent to the no-think suffix.
The parent parsed none of the \V{bg.rows} rows, so the preregistered reliability endpoint was nearly guaranteed.
Thinking disabled was non-inferior to B1 at a margin of \V{v05.margin} (difference \V{v05.e2.est}, interval \CI{\V{v05.e2.lo}}{\V{v05.e2.hi}}) at \V{v05.tokens} of B1's token cost; since every B1 first call reached the rescue threshold, B1 reduced to always adding the suffix.
Post hoc, on LB-0030 no configuration clearly exceeded a constant guess of the age band most common in the two shards (band \V{v05.band}, accuracy \V{v05.band.acc}, against \V{v05.cfg.min} to \V{v05.cfg.max} for the configurations).
The case study's conclusion was that prospective replication controlled selection noise but detected neither the inadequate comparator nor the evaluation defect, and that trivial baselines, input audits and prompt-token reconciliation did.

\begin{table}[H]
\caption{Lessons of the case study and where this study tests them.}
\label{tab:lessons}
\centering
\small
\begin{tabular}{@{}p{0.6\linewidth}p{0.34\linewidth}@{}}
\toprule
Lesson & Tested by \\
\midrule
The parent parsed none of the prospective rows; the rescue made answers visible & H1, H3 \\
A thinking-off setting that the search never tried was non-inferior to the rescue & H2 \\
The caller sent the reference answer & H4, defect A \\
LB-0042's gold came from a fallback rule; the scorer could fail open & H4, defects B1, B2 \\
The search's split put a giant component into one partition & H4, defect C \\
The parent was a broken start & H4, defect D \\
The tied variant was served, not the checkpoint's stored output layer & H4, defect E \\
A constant guess matched the configurations on one task & trivial baselines (Section~\ref{sec:results}) \\
\bottomrule
\end{tabular}
\end{table}

\paragraph{Why this study is not a replication of it.}
The stack (vLLM in batch-invariant mode, not FreeToken), the model variant (untied output layer), two of the three models, two of the three pools and five of the six configurations are new, and the rules were fixed in advance for all of them.
It is not blind to direction: the pilot showed that the parent truncates where the budget is small, and the preregistration says so (Section~\ref{sec:design}).

\section{Full tables}
\label{app:tables}

Every number in this appendix is read from the analysis report committed in our repository (the preregistered analysis) or from the three files computed after the outputs existed; Appendix~\ref{app:repro} lists them.
Intervals of primary endpoints are the family-adjusted ones, all others unadjusted \V{c.ci}.
$\Delta$, $\theta$, $D_T$ and $D_P$ are accuracy or rate differences; $V$, $B$ and $L$ count the rows only \cfg{AUTHORS-1500} parses, both parse and only \cfg{THINK-512} parses.

\begin{table*}[t]
\caption{Verdicts of the \V{cnt.h123} endpoints of H1--H3 of the first study (the first nine rows) and the \V{z.cnt.h123} of the second (Qwen3-14B) on the test partition. A dagger marks the one endpoint whose replication label is not ``replicated'': \smol\ on GSM8K, H2, which \verdict{fails} on test and is \verdict{inconclusive} on replication (an \emph{inconclusive replication}). A star marks an endpoint whose quantity is fixed by an identity of the design or by the observed parse pattern; its verdict says little beyond that mechanism, although it still depends on the intervals (Sections~\ref{sec:h1} and~\ref{sec:h2}). The last two rows are H2 against the GEPA harnesses of each study.}
\label{tab:glance}
\centering
\footnotesize
\setlength{\tabcolsep}{3pt}
\resizebox{\ifdim\width>\linewidth\linewidth\else\width\fi}{!}{%
\begin{tabular}{@{}llccc@{}}
\toprule
Model & Pool & \shortstack{H1\\ \cfg{AUTHORS-1500} vs\\ \cfg{THINK-512}} & \shortstack{H2\\ \cfg{AUTHORS-1500} vs\\ \cfg{RESCUE-1500}} & \shortstack{H3\\ \cfg{BUDGET-512-1500} vs\\ \cfg{THINK-512}} \\
\midrule
Qwen3-1.7B-Longevity & LB letters & no gain: not tested & holds & no room \\
 & MMLU-Pro & no gain: not tested & holds & inconclusive \\
 & GSM8K & no gain: not tested & holds & no room \\
\addlinespace
SmolLM3-3B & LB letters & visibility dominates & holds & holds \\
 & MMLU-Pro & visibility dominates & holds & holds \\
 & GSM8K & visibility dominates & fails\textsuperscript{\dag} & holds \\
\addlinespace
Intern-S1-mini & LB letters & visibility dominates\textsuperscript{$\star$} & holds\textsuperscript{$\star$} & holds \\
 & MMLU-Pro & visibility dominates\textsuperscript{$\star$} & holds & holds \\
 & GSM8K & no gain: not tested & fails & holds \\
\addlinespace
Qwen3-14B & LB letters & visibility dominates\textsuperscript{$\star$} & holds\textsuperscript{$\star$} & holds \\
 & MMLU-Pro & visibility dominates & holds & holds \\
 & GSM8K & no gain: not tested & fails & holds \\
\addlinespace
\midrule
Qwen3-1.7B-Longevity & LB letters & \multicolumn{3}{l}{H2 against the four GEPA harnesses (clean 0, broken 0, clean 1, broken 1): holds\textsuperscript{$\star$}, holds, holds, holds} \\
Qwen3-14B & LB letters & \multicolumn{3}{l}{H2 against the four GEPA harnesses (clean 0, clean 1, broken 0, broken 1): fails, fails, holds\textsuperscript{$\star$}, holds\textsuperscript{$\star$}} \\
\bottomrule
\end{tabular}}
\end{table*}

\begin{figure*}[t]
\centering
\includegraphics[width=\textwidth]{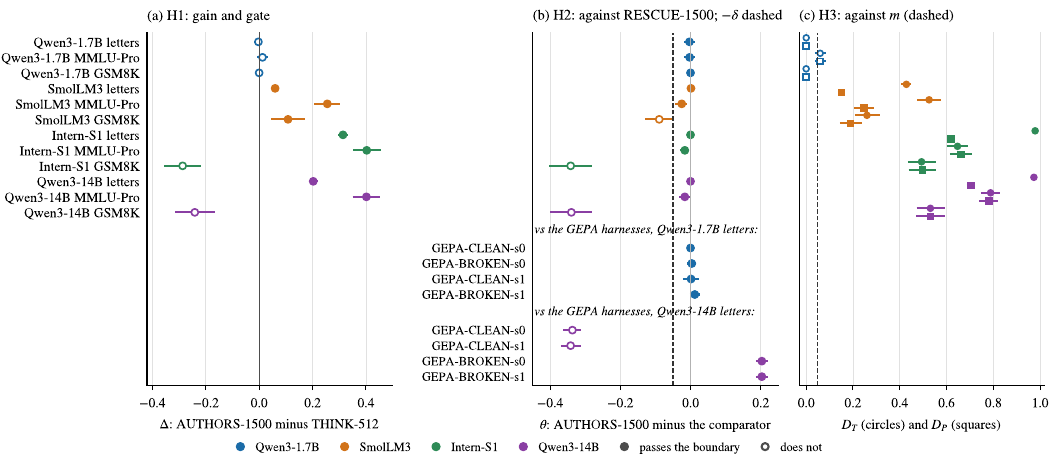}
\caption{Adjusted intervals of the primary endpoints of H1--H3 on the test partition, the first study's nine cells above the second study's three. (a)~H1: $\Delta$, the gain gate; (b)~H2: $\theta$ against the non-inferiority margin $-\delta$; (c)~H3: $D_T$ and $D_P$ against the minimal difference $m$. Cells are model$\times$pool; in (b) the harnesses of each study are below the cells; filled markers pass the verdict's boundary, open markers do not.}
\label{fig:h123}
\end{figure*}

\subsection{H1 to H3}

\paragraph{H1 of the first study in detail.}
Almost all of \cfg{THINK-512}'s unparsed calls ended at the token limit; the exception is \smol\ on the letters, where \V{h1.te.s.lb.pnc} of \V{h1.te.s.lb.punp} ended by themselves without an answer (non-committed).
In two of the five the parse pattern nearly fixes the quantity, as the preregistration expected for the first.
In \intern\ on the letters both configurations parse only \V{h1.te.i.lb.nb} row of \V{cell.te.i.lb.rows}, so $\Delta_B$ is nearly zero and $T$, nearly a fifth of $\Delta_V$, is positive (the preregistration listed the cell as ``$T>0$ nearly inevitable'').
In \intern\ on MMLU-Pro, B has \V{h1.te.i.mm.nb} of \V{cell.te.i.mm.rows} rows, so $\Delta_B$ cannot exceed that share, which is below a quarter of the observed $\Delta_V=\V{h1.te.i.mm.dv}$, and $T>0$ follows from B's row count and that observed $\Delta_V$.
The verdict is informative in \smol's three cells, where B has \V{h1.te.s.lb.nb}, \V{h1.te.s.mm.nb} and \V{h1.te.s.gs.nb} rows.

The sanity checks (\cfg{THINK-64} against \cfg{AUTHORS-1500}) give \verdict{visibility dominates} in \V{cnt.h1s.visibility} of nine cells, as expected, and fail the gate where \qwen\ already parses GSM8K.

We expected the \intern\ endpoint of H2 on GSM8K to be inconclusive, an informal expectation from a planned adjusted power of~\V{x.power} at a true $\theta$ of zero (Appendix~\ref{app:power}), and it is not close.

\paragraph{The first study's GEPA harnesses.}
The validation scores of the four selected harnesses (\V{gepa.gb0.val} to \V{gepa.gc1.val} over the four runs) exceed their test accuracy in every run, as selection makes them optimistic, and none of the candidates GEPA saw (\V{gepa.gc1.seen} to \V{gepa.gb0.seen} per run) contained a soft switch (\texttt{/think} or \texttt{/no\_think}).

\paragraph{H3 of the first study in detail.}
The parse gain is not the whole story for \smol: after the forced end of thinking it re-opens a turn in \V{srv.te.s.bud.rb} of its \V{srv.te.s.bud.calls} test calls, and \cfg{BUDGET-512-1500} still ends by itself without an answer on \V{cfg.te.s.lb.bud.nc} of its letter rows.

As a partial and descriptive comparison, \cfg{THINK-1536}, with an allowance of \V{k.t1536} tokens and no forcing, still ends at the token limit in \V{cfg.te.s.gs.t1536.trunc} (\smol, GSM8K) to \V{cfg.te.i.lb.t1536.trunc} (\intern, letters) of its calls in the six cells where \cfg{THINK-512} truncates a large share, against \V{cfg.te.i.lb.bud.trunc} to \V{cfg.te.s.mm.bud.trunc} for \cfg{BUDGET-512-1500}, which truncates less in all six; the two differ in allowance as well (\V{k.bud.max} against \V{k.t1536} tokens), so the comparison does not separate forcing from allowance.

\begin{table}[H]
\caption{H1, primary endpoints: \cfg{THINK-512} against \cfg{AUTHORS-1500}, test partition (family of \V{c.fam1}). Repl.: R replicated, I inconclusive replication, N not replicated. A star marks an endpoint whose quantity is fixed by an identity of the design or by the observed parse pattern; its verdict says little beyond that mechanism, although it still depends on the intervals.}
\label{tab:app-h1}
\centering
\scriptsize
\setlength{\tabcolsep}{2pt}
\resizebox{\ifdim\width>\linewidth\linewidth\else\width\fi}{!}{%
\begin{tabular}{@{}lcccccclc@{}}
\toprule
Cell & $\Delta$ & $\Delta_V$ & $\Delta_B$ & $\Delta_L$ & $T$ & $V$ / $B$ / $L$ & Verdict & Repl. \\
\midrule
Qwen3-1.7B LB letters & \ensuremath{-0.003} [\ensuremath{-0.018}, \ensuremath{+0.012}] & \ensuremath{0.000} & \ensuremath{-0.003} & \ensuremath{0.000} & \ensuremath{+0.002} [\ensuremath{-0.009}, \ensuremath{+0.014}] & 1 / 3{,}941 / 0 & no gain: not tested & R \\
Qwen3-1.7B MMLU-Pro & \ensuremath{+0.013} [\ensuremath{-0.006}, \ensuremath{+0.033}] & \ensuremath{+0.015} & \ensuremath{-0.003} & \ensuremath{0.000} & \ensuremath{+0.005} [\ensuremath{-0.008}, \ensuremath{+0.018}] & 48 / 751 / 0 & no gain: not tested & R \\
Qwen3-1.7B GSM8K & \ensuremath{0.000} [\ensuremath{0.000}, \ensuremath{0.000}] & \ensuremath{0.000} & \ensuremath{0.000} & \ensuremath{0.000} & \ensuremath{0.000} [\ensuremath{0.000}, \ensuremath{0.000}] & 0 / 523 / 0 & no gain: not tested & R \\
\addlinespace
SmolLM3 LB letters & \ensuremath{+0.060} [\ensuremath{+0.046}, \ensuremath{+0.074}] & \ensuremath{+0.071} & \ensuremath{+0.001} & \ensuremath{-0.012} & \ensuremath{+0.013} [\ensuremath{+0.009}, \ensuremath{+0.018}] & 749 / 448 / 100 & visibility dominates & R \\
SmolLM3 MMLU-Pro & \ensuremath{+0.255} [\ensuremath{+0.207}, \ensuremath{+0.304}] & \ensuremath{+0.282} & \ensuremath{-0.014} & \ensuremath{-0.013} & \ensuremath{+0.067} [\ensuremath{+0.054}, \ensuremath{+0.081}] & 529 / 81 / 13 & visibility dominates & R \\
SmolLM3 GSM8K & \ensuremath{+0.108} [\ensuremath{+0.046}, \ensuremath{+0.173}] & \ensuremath{+0.197} & \ensuremath{-0.036} & \ensuremath{-0.053} & \ensuremath{+0.068} [\ensuremath{+0.047}, \ensuremath{+0.090}] & 141 / 340 / 29 & visibility dominates & R \\
\addlinespace
Intern-S1 LB letters & \ensuremath{+0.314} [\ensuremath{+0.297}, \ensuremath{+0.331}] & \ensuremath{+0.314} & \ensuremath{0.000} & \ensuremath{0.000} & \ensuremath{+0.063} [\ensuremath{+0.059}, \ensuremath{+0.066}] & 2{,}407 / 1 / 0 & visibility dominates\textsuperscript{$\star$} & R \\
\bottomrule
\end{tabular}}
\end{table}

\begin{table}[H]
\caption{H1, sanity checks: \cfg{THINK-64} against \cfg{AUTHORS-1500}, test partition, at the family's adjusted level, outside the family and carrying no claim.}
\label{tab:app-h1s}
\centering
\scriptsize
\setlength{\tabcolsep}{2pt}
\resizebox{\ifdim\width>\linewidth\linewidth\else\width\fi}{!}{%
\begin{tabular}{@{}lcccccclc@{}}
\toprule
Cell & $\Delta$ & $\Delta_V$ & $\Delta_B$ & $\Delta_L$ & $T$ & $V$ / $B$ / $L$ & Verdict & Repl. \\
\midrule
Qwen3-1.7B LB letters & \ensuremath{+0.632} [\ensuremath{+0.613}, \ensuremath{+0.651}] & \ensuremath{+0.632} & \ensuremath{0.000} & \ensuremath{0.000} & \ensuremath{+0.126} [\ensuremath{+0.123}, \ensuremath{+0.130}] & 3{,}942 / 0 / 0 & visibility dominates &  \\
Qwen3-1.7B MMLU-Pro & \ensuremath{+0.089} [\ensuremath{+0.063}, \ensuremath{+0.119}] & \ensuremath{+0.089} & \ensuremath{0.000} & \ensuremath{0.000} & \ensuremath{+0.018} [\ensuremath{+0.013}, \ensuremath{+0.024}] & 274 / 525 / 0 & visibility dominates &  \\
Qwen3-1.7B GSM8K & \ensuremath{+0.002} [\ensuremath{0.000}, \ensuremath{+0.009}] & \ensuremath{+0.002} & \ensuremath{0.000} & \ensuremath{0.000} & \ensuremath{0.000} [\ensuremath{0.000}, \ensuremath{+0.002}] & 1 / 522 / 0 & no gain: not tested &  \\
\addlinespace
SmolLM3 LB letters & \ensuremath{+0.129} [\ensuremath{+0.116}, \ensuremath{+0.143}] & \ensuremath{+0.129} & \ensuremath{0.000} & \ensuremath{0.000} & \ensuremath{+0.026} [\ensuremath{+0.023}, \ensuremath{+0.029}] & 1{,}197 / 0 / 0 & visibility dominates &  \\
SmolLM3 MMLU-Pro & \ensuremath{+0.353} [\ensuremath{+0.307}, \ensuremath{+0.401}] & \ensuremath{+0.353} & \ensuremath{0.000} & \ensuremath{0.000} & \ensuremath{+0.071} [\ensuremath{+0.061}, \ensuremath{+0.080}] & 610 / 0 / 0 & visibility dominates &  \\
SmolLM3 GSM8K & \ensuremath{+0.789} [\ensuremath{+0.738}, \ensuremath{+0.837}] & \ensuremath{+0.789} & \ensuremath{0.000} & \ensuremath{0.000} & \ensuremath{+0.158} [\ensuremath{+0.148}, \ensuremath{+0.167}] & 481 / 0 / 0 & visibility dominates &  \\
\addlinespace
Intern-S1 LB letters & \ensuremath{+0.314} [\ensuremath{+0.297}, \ensuremath{+0.331}] & \ensuremath{+0.314} & \ensuremath{0.000} & \ensuremath{0.000} & \ensuremath{+0.063} [\ensuremath{+0.059}, \ensuremath{+0.066}] & 2{,}408 / 0 / 0 & visibility dominates &  \\
\bottomrule
\end{tabular}}
\end{table}

\begin{table}[H]
\caption{H2, primary endpoints: $\theta$ of \cfg{AUTHORS-1500} against \cfg{RESCUE-1500} in nine cells and against the four GEPA harnesses (family of \V{c.fam2}), with the replication partition's verdict and the label. A star marks an endpoint whose quantity is fixed by an identity of the design or by the observed parse pattern; its verdict says little beyond that mechanism, although it still depends on the intervals.}
\label{tab:app-h2}
\centering
\scriptsize
\setlength{\tabcolsep}{3pt}
\resizebox{\ifdim\width>\linewidth\linewidth\else\width\fi}{!}{%
\begin{tabular}{@{}lcccc@{}}
\toprule
Endpoint & $\theta$ (adjusted interval) & Verdict & Replication verdict & Label \\
\midrule
Qwen3-1.7B LB letters, RESCUE-1500 & \ensuremath{-0.003} [\ensuremath{-0.018}, \ensuremath{+0.012}] & holds & holds & replicated \\
Qwen3-1.7B MMLU-Pro, RESCUE-1500 & \ensuremath{-0.003} [\ensuremath{-0.019}, \ensuremath{+0.014}] & holds & holds & replicated \\
Qwen3-1.7B GSM8K, RESCUE-1500 & \ensuremath{0.000} [\ensuremath{0.000}, \ensuremath{0.000}] & holds & holds & replicated \\
\addlinespace
SmolLM3 LB letters, RESCUE-1500 & \ensuremath{+0.001} [\ensuremath{-0.007}, \ensuremath{+0.009}] & holds & holds & replicated \\
SmolLM3 MMLU-Pro, RESCUE-1500 & \ensuremath{-0.025} [\ensuremath{-0.044}, \ensuremath{-0.009}] & holds & holds & replicated \\
SmolLM3 GSM8K, RESCUE-1500 & \ensuremath{-0.089} [\ensuremath{-0.129}, \ensuremath{-0.053}] & fails & inconclusive & inconclusive replication \\
\addlinespace
Intern-S1 LB letters, RESCUE-1500 & \ensuremath{0.000} [\ensuremath{0.000}, \ensuremath{0.000}] & holds\textsuperscript{$\star$} & holds & replicated \\
Intern-S1 MMLU-Pro, RESCUE-1500 & \ensuremath{-0.016} [\ensuremath{-0.030}, \ensuremath{-0.005}] & holds & holds & replicated \\
Intern-S1 GSM8K, RESCUE-1500 & \ensuremath{-0.342} [\ensuremath{-0.402}, \ensuremath{-0.281}] & fails & fails & replicated \\
\midrule
Qwen3 LB letters, GEPA-CLEAN-s0 & \ensuremath{0.000} [\ensuremath{0.000}, \ensuremath{0.000}] & holds\textsuperscript{$\star$} & holds & replicated \\
Qwen3 LB letters, GEPA-BROKEN-s0 & \ensuremath{+0.004} [\ensuremath{-0.009}, \ensuremath{+0.016}] & holds & holds & replicated \\
Qwen3 LB letters, GEPA-CLEAN-s1 & \ensuremath{+0.002} [\ensuremath{-0.022}, \ensuremath{+0.025}] & holds & holds & replicated \\
Qwen3 LB letters, GEPA-BROKEN-s1 & \ensuremath{+0.012} [\ensuremath{-0.002}, \ensuremath{+0.026}] & holds & holds & replicated \\
\bottomrule
\end{tabular}}
\end{table}

\begin{table}[H]
\caption{H2, descriptive: $\theta$ of \cfg{AUTHORS-1500} against three other configurations, test partition, unadjusted intervals and the verdict rule applied to them.}
\label{tab:app-h2d}
\centering
\scriptsize
\setlength{\tabcolsep}{3pt}
\resizebox{\ifdim\width>\linewidth\linewidth\else\width\fi}{!}{%
\begin{tabular}{@{}lccc@{}}
\toprule
Cell & vs THINK-512 & vs THINK-1536 & vs BUDGET-512-1500 \\
\midrule
Qwen3-1.7B LB letters & \ensuremath{-0.003} [\ensuremath{-0.013}, \ensuremath{+0.008}] holds & \ensuremath{-0.003} [\ensuremath{-0.013}, \ensuremath{+0.007}] holds & \ensuremath{-0.003} [\ensuremath{-0.013}, \ensuremath{+0.007}] holds \\
Qwen3-1.7B MMLU-Pro & \ensuremath{+0.013} [\ensuremath{-0.001}, \ensuremath{+0.026}] holds & \ensuremath{-0.005} [\ensuremath{-0.018}, \ensuremath{+0.008}] holds & \ensuremath{-0.003} [\ensuremath{-0.015}, \ensuremath{+0.010}] holds \\
Qwen3-1.7B GSM8K & \ensuremath{0.000} [\ensuremath{0.000}, \ensuremath{0.000}] holds & \ensuremath{0.000} [\ensuremath{0.000}, \ensuremath{0.000}] holds & \ensuremath{0.000} [\ensuremath{0.000}, \ensuremath{0.000}] holds \\
\addlinespace
SmolLM3 LB letters & \ensuremath{+0.060} [\ensuremath{+0.050}, \ensuremath{+0.070}] holds & \ensuremath{-0.028} [\ensuremath{-0.040}, \ensuremath{-0.015}] holds & \ensuremath{+0.009} [\ensuremath{-0.003}, \ensuremath{+0.020}] holds \\
SmolLM3 MMLU-Pro & \ensuremath{+0.255} [\ensuremath{+0.220}, \ensuremath{+0.290}] holds & \ensuremath{+0.039} [\ensuremath{+0.003}, \ensuremath{+0.075}] holds & \ensuremath{+0.095} [\ensuremath{+0.059}, \ensuremath{+0.131}] holds \\
SmolLM3 GSM8K & \ensuremath{+0.108} [\ensuremath{+0.063}, \ensuremath{+0.154}] holds & \ensuremath{-0.091} [\ensuremath{-0.127}, \ensuremath{-0.055}] fails & \ensuremath{-0.046} [\ensuremath{-0.083}, \ensuremath{-0.008}] inconclusive \\
\addlinespace
Intern-S1 LB letters & \ensuremath{+0.314} [\ensuremath{+0.302}, \ensuremath{+0.326}] holds & \ensuremath{+0.189} [\ensuremath{+0.175}, \ensuremath{+0.204}] holds & \ensuremath{+0.061} [\ensuremath{+0.046}, \ensuremath{+0.076}] holds \\
\bottomrule
\end{tabular}}
\end{table}

\begin{table}[H]
\caption{H3, primary endpoints: \cfg{THINK-512} against \cfg{BUDGET-512-1500}, test partition (family of \V{c.fam3}); the truncation rate of \cfg{THINK-512} decides \verdict{no room}.}
\label{tab:app-h3}
\centering
\scriptsize
\setlength{\tabcolsep}{2pt}
\resizebox{\ifdim\width>\linewidth\linewidth\else\width\fi}{!}{%
\begin{tabular}{@{}lcccccc@{}}
\toprule
Cell & $D_T$ & $D_P$ & THINK-512 truncation & Verdict & Replication verdict & Label \\
\midrule
Qwen3-1.7B LB letters & \ensuremath{0.000} [\ensuremath{0.000}, \ensuremath{+0.001}] & \ensuremath{0.000} [\ensuremath{0.000}, \ensuremath{+0.001}] & \ensuremath{0.000} [\ensuremath{0.000}, \ensuremath{+0.001}] & no room & no room & replicated \\
Qwen3-1.7B MMLU-Pro & \ensuremath{+0.060} [\ensuremath{+0.038}, \ensuremath{+0.085}] & \ensuremath{+0.060} [\ensuremath{+0.038}, \ensuremath{+0.085}] & \ensuremath{+0.060} [\ensuremath{+0.038}, \ensuremath{+0.085}] & inconclusive & inconclusive & replicated \\
Qwen3-1.7B GSM8K & \ensuremath{0.000} [\ensuremath{0.000}, \ensuremath{0.000}] & \ensuremath{0.000} [\ensuremath{0.000}, \ensuremath{0.000}] & \ensuremath{0.000} [\ensuremath{0.000}, \ensuremath{0.000}] & no room & no room & replicated \\
\addlinespace
SmolLM3 LB letters & \ensuremath{+0.428} [\ensuremath{+0.407}, \ensuremath{+0.450}] & \ensuremath{+0.151} [\ensuremath{+0.136}, \ensuremath{+0.167}] & \ensuremath{+0.725} [\ensuremath{+0.708}, \ensuremath{+0.742}] & holds & holds & replicated \\
SmolLM3 MMLU-Pro & \ensuremath{+0.526} [\ensuremath{+0.476}, \ensuremath{+0.574}] & \ensuremath{+0.248} [\ensuremath{+0.205}, \ensuremath{+0.292}] & \ensuremath{+0.905} [\ensuremath{+0.875}, \ensuremath{+0.932}] & holds & holds & replicated \\
SmolLM3 GSM8K & \ensuremath{+0.260} [\ensuremath{+0.209}, \ensuremath{+0.315}] & \ensuremath{+0.190} [\ensuremath{+0.144}, \ensuremath{+0.239}] & \ensuremath{+0.300} [\ensuremath{+0.245}, \ensuremath{+0.357}] & holds & holds & replicated \\
\addlinespace
Intern-S1 LB letters & \ensuremath{+0.979} [\ensuremath{+0.972}, \ensuremath{+0.985}] & \ensuremath{+0.619} [\ensuremath{+0.603}, \ensuremath{+0.636}] & \ensuremath{+1.000} [\ensuremath{+0.999}, \ensuremath{+1.000}] & holds & holds & replicated \\
\bottomrule
\end{tabular}}
\end{table}

\subsection{H4}
\begin{table}[H]
\caption{H4, the \V{c.fam4} pairs with their adjusted intervals (family of \V{c.fam4}, level \V{c.lvl4}\%) and, in parentheses, which side of its boundary each interval lies on (\emph{holds}, \emph{fails} or \emph{open}). Stability of $\Gamma$ is a descriptive fact about the same interval. A star marks a claim whose quantity is fixed by an identity of the design; its verdict says little beyond that mechanism, although it still depends on the intervals.}
\label{tab:app-h4}
\centering
\scriptsize
\setlength{\tabcolsep}{2pt}
\resizebox{\ifdim\width>\linewidth\linewidth\else\width\fi}{!}{%
\begin{tabular}{@{}llcccll@{}}
\toprule
Pair & Bound & $D_S$ & $D_R$ & $\Gamma$ & Stability of $\Gamma$ & Claim \\
\midrule
A Qwen3-1.7B & 1.000 & \ensuremath{-0.052} [\ensuremath{-0.106}, \ensuremath{+0.002}] (open) & \ensuremath{-0.030} [\ensuremath{-0.087}, \ensuremath{+0.030}] (open) & \ensuremath{+0.022} [\ensuremath{-0.057}, \ensuremath{+0.102}] (open) & inconclusive & inconclusive \\
A SmolLM3 & 1.000 & \ensuremath{+0.097} [\ensuremath{+0.022}, \ensuremath{+0.173}] (open) & \ensuremath{+0.119} [\ensuremath{+0.042}, \ensuremath{+0.199}] (open) & \ensuremath{+0.022} [\ensuremath{-0.089}, \ensuremath{+0.133}] (open) & inconclusive & inconclusive \\
A Intern-S1 & 1.000 & \ensuremath{+0.142} [\ensuremath{+0.082}, \ensuremath{+0.201}] (holds) & \ensuremath{+0.164} [\ensuremath{+0.102}, \ensuremath{+0.226}] (holds) & \ensuremath{+0.022} [\ensuremath{-0.065}, \ensuremath{+0.108}] (open) & inconclusive & inconclusive \\
B1 Qwen3-1.7B & 0.205 & \ensuremath{-0.145} [\ensuremath{-0.170}, \ensuremath{-0.118}] (holds) & \ensuremath{-0.148} [\ensuremath{-0.166}, \ensuremath{-0.130}] (holds) & \ensuremath{-0.003} [\ensuremath{-0.035}, \ensuremath{+0.028}] (holds) & stable & holds \\
B1 SmolLM3 & 0.083 & \ensuremath{+0.001} [\ensuremath{-0.019}, \ensuremath{+0.021}] (fails) & \ensuremath{+0.007} [\ensuremath{-0.006}, \ensuremath{+0.021}] (fails) & \ensuremath{+0.006} [\ensuremath{-0.018}, \ensuremath{+0.031}] (holds) & stable & fails \\
B1 Intern-S1 & 0.057 & \ensuremath{-0.012} [\ensuremath{-0.028}, \ensuremath{+0.004}] (fails) & \ensuremath{-0.019} [\ensuremath{-0.030}, \ensuremath{-0.007}] (fails) & \ensuremath{-0.007} [\ensuremath{-0.027}, \ensuremath{+0.013}] (holds) & stable & fails \\
B2 Qwen3-1.7B & 0.000 & \multicolumn{4}{c}{not evaluable: the start leaves no row unparsed} & -- \\
\bottomrule
\end{tabular}}
\end{table}

\begin{table}[H]
\caption{The detection matrix (descriptive; largely true by construction). Checks: PR prospective replication, TR token reconciliation, TB trivial baselines, SCF simple configuration family, IR input reading, SMI served-model identity.}
\label{tab:app-matrix}
\centering
\scriptsize
\setlength{\tabcolsep}{2pt}
\resizebox{\ifdim\width>\linewidth\linewidth\else\width\fi}{!}{%
\begin{tabular}{@{}lccccccp{0.5\linewidth}@{}}
\toprule
Pair & PR & TR & TB & SCF & IR & SMI & Rules fired (expected ones in bold) \\
\midrule
A Qwen3-1.7B & U & D & U & U & U & -- & \textbf{IR1}, \textbf{IR2}, IR4, PR.gain, SCF2, TB2, \textbf{TB3}, \textbf{TR} \\
A SmolLM3 & D & D & U & U & U & -- & \textbf{IR1}, \textbf{IR2}, IR4, PR.gain, SCF1, SCF2, SCF3, TB2, \textbf{TR}; expected, not fired: TB3 \\
A Intern-S1 & M & D & U & U & U & -- & \textbf{IR1}, \textbf{IR2}, IR4, SCF1, SCF3, TB2, \textbf{TB3}, \textbf{TR} \\
B1 Qwen3-1.7B & M & M & M & M & U & -- & \textbf{IR3}, IR4 \\
B1 SmolLM3 & M & M & M & U & U & -- & \textbf{IR3}, IR4, SCF1 \\
B1 Intern-S1 & M & M & M & U & U & -- & \textbf{IR3}, IR4, SCF1, SCF2, SCF3 \\
B2 Qwen3-1.7B & D & M & D & D & U & -- & IR4, PR.gain, SCF2, \textbf{TB4} \\
B2 SmolLM3 & D & M & D & U & U & -- & IR4, PR.gain, SCF1, SCF2, \textbf{TB4} \\
B2 Intern-S1 & M & M & D & U & U & -- & IR4, SCF1, \textbf{TB4} \\
C Qwen3-1.7B & D & M & D & D & U & -- & \textbf{IR4}, \textbf{IR5}, PR.gain, SCF2, TB2; expected, not fired: TB1 \\
C SmolLM3 & M & M & D & U & U & -- & \textbf{IR4}, \textbf{IR5}, SCF1, TB2; expected, not fired: TB1 \\
C Intern-S1 & M & M & D & U & U & -- & \textbf{IR4}, \textbf{IR5}, SCF1, SCF2, SCF3, TB2; expected, not fired: TB1 \\
D Qwen3-1.7B & M & M & M & D & U & -- & IR4, \textbf{SCF1}, SCF2, \textbf{SCF3} \\
D SmolLM3 & M & M & M & U & U & -- & IR4, \textbf{SCF1}, SCF2, \textbf{SCF3} \\
D Intern-S1 & M & M & M & U & U & -- & IR4, \textbf{SCF1}, SCF2, \textbf{SCF3} \\
E Qwen3-1.7B & D & M & M & U & U & D & IR4, PR.level, SCF2, \textbf{SMI} \\
\bottomrule
\end{tabular}}

\par\vspace{2pt}{\scriptsize D detects: fires on the defective instance and not on the clean one (5, 3, 6, 3, 0, 1 instances for PR, TR, TB, SCF, IR, SMI); U uninformative: fires on both (1, 0, 3, 12, 16, 0); M misses.}
\end{table}

\begin{table}[H]
\caption{False alarms of each rule on the clean instances (the main clean instance and the clean counterparts of A and E, per model).}
\label{tab:app-fa}
\centering
\scriptsize
\setlength{\tabcolsep}{3pt}
\resizebox{\ifdim\width>\linewidth\linewidth\else\width\fi}{!}{%
\begin{tabular}{@{}llp{0.55\linewidth}r@{}}
\toprule
Check & Rule & Fired on & Of \\
\midrule
PR & PR.gain & 1: clean/A / Qwen3 & 7 \\
PR & PR.level & 0: none & 7 \\
TR & TR & 0: none & 7 \\
TB & TB1 & 0: none & 7 \\
TB & TB2 & 3: clean/A / Intern-S1, clean/A / Qwen3, clean/A / SmolLM3 & 7 \\
TB & TB3 & 0: none & 7 \\
TB & TB4 & 0: none & 7 \\
SCF & SCF1 & 4: clean / Intern-S1, clean / SmolLM3, clean/A / Intern-S1, clean/A / SmolLM3 & 7 \\
SCF & SCF2 & 4: clean / Intern-S1, clean/A / Intern-S1, clean/A / Qwen3, clean/E / Qwen3 & 7 \\
SCF & SCF3 & 2: clean / Intern-S1, clean/A / Intern-S1 & 7 \\
IR & IR1 & 0: none & 7 \\
IR & IR2 & 0: none & 7 \\
IR & IR3 & 0: none & 7 \\
IR & IR4 & 7: clean / Intern-S1, clean / Qwen3, clean / SmolLM3, clean/A / Intern-S1, clean/A / Qwen3, clean/A / SmolLM3, clean/E / Qwen3 & 7 \\
IR & IR5 & 0: none & 7 \\
\bottomrule
\end{tabular}}
\end{table}

\subsection{Configurations}
\begin{table}[H]
\caption{Accuracy, parse, truncation and non-commitment rates of the six configurations in the nine cells, test partition, with unadjusted \V{c.ci} intervals. Unparsed answers count as wrong. The trivial baselines of the three pools are in Table~\ref{tab:acc} and Appendix~\ref{app:baselines}.}
\label{tab:app-cfg}
\centering
\scriptsize
\setlength{\tabcolsep}{2pt}
\resizebox{\ifdim\width>\linewidth\linewidth\else\width\fi}{!}{%
\begin{tabular}{@{}llcccc@{}}
\toprule
Cell & Configuration & Accuracy & Parse & Truncated & Non-committed \\
\midrule
Qwen3-1.7B LB letters & THINK-64 & 0.000 [0.000, 0.000] & 0.000 [0.000, 0.000] & 1.000 [1.000, 1.000] & 0.000 [0.000, 0.000] \\
 & THINK-512 & 0.635 [0.621, 0.649] & 1.000 [0.999, 1.000] & 0.000 [0.000, 0.001] & 0.000 [0.000, 0.000] \\
 & THINK-1536 & 0.635 [0.621, 0.649] & 1.000 [1.000, 1.000] & 0.000 [0.000, 0.000] & 0.000 [0.000, 0.000] \\
 & AUTHORS-1500 & 0.632 [0.618, 0.646] & 1.000 [1.000, 1.000] & 0.000 [0.000, 0.000] & 0.000 [0.000, 0.000] \\
 & BUDGET-512-1500 & 0.635 [0.621, 0.649] & 1.000 [1.000, 1.000] & 0.000 [0.000, 0.000] & 0.000 [0.000, 0.000] \\
 & RESCUE-1500 & 0.635 [0.621, 0.649] & 1.000 [1.000, 1.000] & 0.000 [0.000, 0.000] & 0.000 [0.000, 0.000] \\
\addlinespace
Qwen3-1.7B MMLU-Pro & THINK-64 & 0.140 [0.116, 0.165] & 0.657 [0.625, 0.690] & 0.343 [0.310, 0.375] & 0.000 [0.000, 0.000] \\
\bottomrule
\end{tabular}}
\end{table}

\subsection{Trivial baselines, computed after the outputs existed}
\label{app:baselines}
Chance is the mean of $1/k$ over the rows ($k$ options).
A constant letter or a constant label is the same answer on every row; the non-oracle constant is the most common gold letter or label of the other partition, the oracle constant the most common one of the partition itself (ties to the smallest).
On LB-0030 and LB-0042 the order of the options is shuffled from row to row, so a constant letter cannot exploit an uneven label distribution and a constant label can.
The strongest baseline of a task is the larger of its two oracle constants; the pool's values are row-weighted means.
For MMLU-Pro and GSM8K (addendum A12) the constants are a letter and an answer (the most common gold answer of the other partition, and of the partition itself); the strongest baseline is the oracle constant, and a free-form number has no chance level.
Cells give accuracy and, in parentheses, the paired difference to the strongest oracle baseline; intervals for the pool are in Table~\ref{tab:acc}, the rest in the files of our repository (Appendix~\ref{app:repro}).

\begin{table}[H]
\caption{Accuracy and difference to the strongest baseline (oracle constant label), test partition.}
\label{tab:app-base-te}
\centering
\scriptsize
\setlength{\tabcolsep}{2pt}
\resizebox{\ifdim\width>\linewidth\linewidth\else\width\fi}{!}{%
\begin{tabular}{@{}lcccccc@{}}
\toprule
Model & THINK-64 & THINK-512 & THINK-1536 & AUTHORS-1500 & BUDGET-512-1500 & RESCUE-1500 \\
\midrule
\multicolumn{7}{@{}p{\linewidth}@{}}{\textbf{LB-0030} (1{,}577 rows). Chance 0.200; constant letter 0.216 (non-oracle), 0.216 (oracle); constant label 0.242 (non-oracle), 0.250 (oracle). Strongest, oracle: 0.250 [0.230, 0.269].} \\
Qwen3-1.7B & 0.000 (\ensuremath{-0.250}) & 0.369 (\ensuremath{+0.119}) & 0.369 (\ensuremath{+0.119}) & 0.344 (\ensuremath{+0.094}) & 0.369 (\ensuremath{+0.119}) & 0.369 (\ensuremath{+0.119}) \\
SmolLM3 & 0.000 (\ensuremath{-0.250}) & 0.000 (\ensuremath{-0.250}) & 0.021 (\ensuremath{-0.229}) & 0.053 (\ensuremath{-0.197}) & 0.032 (\ensuremath{-0.218}) & 0.053 (\ensuremath{-0.197}) \\
Intern-S1 & 0.000 (\ensuremath{-0.250}) & 0.000 (\ensuremath{-0.250}) & 0.004 (\ensuremath{-0.245}) & 0.023 (\ensuremath{-0.227}) & 0.055 (\ensuremath{-0.195}) & 0.023 (\ensuremath{-0.227}) \\
Qwen3, GEPA & \multicolumn{6}{p{0.8\linewidth}}{CLEAN-s0 0.344 (\ensuremath{+0.094}); BROKEN-s0 0.353 (\ensuremath{+0.103}); CLEAN-s1 0.388 (\ensuremath{+0.138}); BROKEN-s1 0.321 (\ensuremath{+0.072})} \\
\addlinespace
\multicolumn{7}{@{}p{\linewidth}@{}}{\textbf{LB-0034} (788 rows). Chance 0.500; constant letter 0.485 (non-oracle), 0.515 (oracle); constant label 0.485 (non-oracle), 0.515 (oracle). Strongest, oracle: 0.515 [0.480, 0.551].} \\
Qwen3-1.7B & 0.000 (\ensuremath{-0.515}) & 0.702 (\ensuremath{+0.187}) & 0.703 (\ensuremath{+0.188}) & 0.717 (\ensuremath{+0.202}) & 0.703 (\ensuremath{+0.188}) & 0.703 (\ensuremath{+0.188}) \\
\bottomrule
\end{tabular}}
\end{table}

\begin{table}[H]
\caption{The same for the replication partition.}
\label{tab:app-base-re}
\centering
\scriptsize
\setlength{\tabcolsep}{2pt}
\resizebox{\ifdim\width>\linewidth\linewidth\else\width\fi}{!}{%
\begin{tabular}{@{}lcccccc@{}}
\toprule
Model & THINK-64 & THINK-512 & THINK-1536 & AUTHORS-1500 & BUDGET-512-1500 & RESCUE-1500 \\
\midrule
\multicolumn{7}{@{}p{\linewidth}@{}}{\textbf{LB-0030} (796 rows). Chance 0.200; constant letter 0.212 (non-oracle), 0.212 (oracle); constant label 0.255 (non-oracle), 0.270 (oracle). Strongest, oracle: 0.270 [0.242, 0.299].} \\
Qwen3-1.7B & 0.000 (\ensuremath{-0.270}) & 0.358 (\ensuremath{+0.088}) & 0.358 (\ensuremath{+0.088}) & 0.335 (\ensuremath{+0.065}) & 0.358 (\ensuremath{+0.088}) & 0.358 (\ensuremath{+0.088}) \\
SmolLM3 & 0.000 (\ensuremath{-0.270}) & 0.000 (\ensuremath{-0.270}) & 0.025 (\ensuremath{-0.245}) & 0.041 (\ensuremath{-0.229}) & 0.031 (\ensuremath{-0.239}) & 0.041 (\ensuremath{-0.229}) \\
Intern-S1 & 0.000 (\ensuremath{-0.270}) & 0.000 (\ensuremath{-0.270}) & 0.005 (\ensuremath{-0.265}) & 0.011 (\ensuremath{-0.259}) & 0.046 (\ensuremath{-0.224}) & 0.011 (\ensuremath{-0.259}) \\
Qwen3, GEPA & \multicolumn{6}{p{0.8\linewidth}}{CLEAN-s0 0.335 (\ensuremath{+0.065}); BROKEN-s0 0.345 (\ensuremath{+0.075}); CLEAN-s1 0.389 (\ensuremath{+0.119}); BROKEN-s1 0.307 (\ensuremath{+0.036})} \\
\addlinespace
\multicolumn{7}{@{}p{\linewidth}@{}}{\textbf{LB-0034} (398 rows). Chance 0.500; constant letter 0.490 (non-oracle), 0.510 (oracle); constant label 0.490 (non-oracle), 0.510 (oracle). Strongest, oracle: 0.510 [0.462, 0.558].} \\
Qwen3-1.7B & 0.000 (\ensuremath{-0.510}) & 0.741 (\ensuremath{+0.231}) & 0.744 (\ensuremath{+0.234}) & 0.744 (\ensuremath{+0.234}) & 0.744 (\ensuremath{+0.234}) & 0.744 (\ensuremath{+0.234}) \\
\bottomrule
\end{tabular}}
\end{table}

\begin{table}[H]
\caption{MMLU-Pro and GSM8K, test partition: accuracy of the six configurations and, in parentheses, the paired difference to the strongest baseline (the oracle constant).}
\label{tab:app-base-gen-te}
\centering
\scriptsize
\setlength{\tabcolsep}{2pt}
\resizebox{\ifdim\width>\linewidth\linewidth\else\width\fi}{!}{%
\begin{tabular}{@{}lcccccc@{}}
\toprule
Model & THINK-64 & THINK-512 & THINK-1536 & AUTHORS-1500 & BUDGET-512-1500 & RESCUE-1500 \\
\midrule
\multicolumn{7}{@{}p{\linewidth}@{}}{\textbf{MMLU-Pro} (799 items). Chance 0.110; constant letter B, chosen on the other partition and oracle alike, 0.119 [0.096, 0.141].} \\
Qwen3-1.7B & 0.140 (\ensuremath{+0.021}) & 0.217 (\ensuremath{+0.098}) & 0.234 (\ensuremath{+0.115}) & 0.229 (\ensuremath{+0.110}) & 0.232 (\ensuremath{+0.113}) & 0.232 (\ensuremath{+0.113}) \\
SmolLM3 & 0.000 (\ensuremath{-0.119}) & 0.098 (\ensuremath{-0.021}) & 0.314 (\ensuremath{+0.195}) & 0.353 (\ensuremath{+0.234}) & 0.258 (\ensuremath{+0.139}) & 0.378 (\ensuremath{+0.259}) \\
Intern-S1 & 0.000 (\ensuremath{-0.119}) & 0.074 (\ensuremath{-0.045}) & 0.279 (\ensuremath{+0.160}) & 0.477 (\ensuremath{+0.358}) & 0.542 (\ensuremath{+0.423}) & 0.493 (\ensuremath{+0.374}) \\
\midrule
\multicolumn{7}{@{}p{\linewidth}@{}}{\textbf{GSM8K} (527 items). Constant answer 10 (chosen on the other partition) 0.019 [0.008, 0.032], 2 (oracle) 0.034 [0.019, 0.051]; no chance level.} \\
Qwen3-1.7B & 0.068 (\ensuremath{+0.034}) & 0.070 (\ensuremath{+0.036}) & 0.070 (\ensuremath{+0.036}) & 0.070 (\ensuremath{+0.036}) & 0.070 (\ensuremath{+0.036}) & 0.070 (\ensuremath{+0.036}) \\
SmolLM3 & 0.000 (\ensuremath{-0.034}) & 0.681 (\ensuremath{+0.647}) & 0.880 (\ensuremath{+0.846}) & 0.789 (\ensuremath{+0.755}) & 0.835 (\ensuremath{+0.801}) & 0.879 (\ensuremath{+0.844}) \\
Intern-S1 & 0.000 (\ensuremath{-0.034}) & 0.476 (\ensuremath{+0.442}) & 0.833 (\ensuremath{+0.799}) & 0.190 (\ensuremath{+0.156}) & 0.922 (\ensuremath{+0.888}) & 0.531 (\ensuremath{+0.497}) \\
\bottomrule
\end{tabular}}
\end{table}

\begin{table}[H]
\caption{MMLU-Pro and GSM8K, replication partition.}
\label{tab:app-base-gen-re}
\centering
\scriptsize
\setlength{\tabcolsep}{2pt}
\resizebox{\ifdim\width>\linewidth\linewidth\else\width\fi}{!}{%
\begin{tabular}{@{}lcccccc@{}}
\toprule
Model & THINK-64 & THINK-512 & THINK-1536 & AUTHORS-1500 & BUDGET-512-1500 & RESCUE-1500 \\
\midrule
\multicolumn{7}{@{}p{\linewidth}@{}}{\textbf{MMLU-Pro} (399 items). Chance 0.111; constant letter B, chosen on the other partition and oracle alike, 0.133 [0.100, 0.168].} \\
Qwen3-1.7B & 0.108 (\ensuremath{-0.025}) & 0.180 (\ensuremath{+0.048}) & 0.193 (\ensuremath{+0.060}) & 0.180 (\ensuremath{+0.048}) & 0.190 (\ensuremath{+0.058}) & 0.185 (\ensuremath{+0.053}) \\
SmolLM3 & 0.000 (\ensuremath{-0.133}) & 0.090 (\ensuremath{-0.043}) & 0.333 (\ensuremath{+0.201}) & 0.378 (\ensuremath{+0.246}) & 0.291 (\ensuremath{+0.158}) & 0.398 (\ensuremath{+0.266}) \\
Intern-S1 & 0.000 (\ensuremath{-0.133}) & 0.105 (\ensuremath{-0.028}) & 0.293 (\ensuremath{+0.160}) & 0.446 (\ensuremath{+0.313}) & 0.566 (\ensuremath{+0.434}) & 0.454 (\ensuremath{+0.321}) \\
\midrule
\multicolumn{7}{@{}p{\linewidth}@{}}{\textbf{GSM8K} (264 items). Constant answer 2 (chosen on the other partition) 0.030 [0.011, 0.053], 10 (oracle) 0.045 [0.023, 0.072]; no chance level.} \\
Qwen3-1.7B & 0.057 (\ensuremath{+0.011}) & 0.057 (\ensuremath{+0.011}) & 0.057 (\ensuremath{+0.011}) & 0.057 (\ensuremath{+0.011}) & 0.057 (\ensuremath{+0.011}) & 0.057 (\ensuremath{+0.011}) \\
SmolLM3 & 0.000 (\ensuremath{-0.045}) & 0.705 (\ensuremath{+0.659}) & 0.909 (\ensuremath{+0.864}) & 0.814 (\ensuremath{+0.769}) & 0.830 (\ensuremath{+0.784}) & 0.886 (\ensuremath{+0.841}) \\
Intern-S1 & 0.000 (\ensuremath{-0.045}) & 0.462 (\ensuremath{+0.417}) & 0.845 (\ensuremath{+0.799}) & 0.189 (\ensuremath{+0.144}) & 0.936 (\ensuremath{+0.890}) & 0.542 (\ensuremath{+0.496}) \\
\bottomrule
\end{tabular}}
\end{table}

\subsection{Other descriptive items}
\begin{table}[H]
\caption{Seeded sampling at the models' card values (test partition): parse and truncation rates, and the difference seeded minus zero temperature with unadjusted intervals.}
\label{tab:app-seeded}
\centering
\scriptsize
\setlength{\tabcolsep}{1pt}
\resizebox{\ifdim\width>\linewidth\linewidth\else\width\fi}{!}{%
\begin{tabular}{@{}lllccccc@{}}
\toprule
Model & Pool & Configuration & Parse (seeded) & Truncated (seeded) & $\Delta$ accuracy & $\Delta$ parse & $\Delta$ truncation \\
\midrule
Qwen3-1.7B & LB letters & THINK-64 & 0.000 & 1.000 & \ensuremath{0.000} [\ensuremath{0.000}, \ensuremath{0.000}] & \ensuremath{0.000} [\ensuremath{0.000}, \ensuremath{0.000}] & \ensuremath{0.000} [\ensuremath{0.000}, \ensuremath{0.000}] \\
 &  & THINK-512 & 0.999 & 0.001 & \ensuremath{-0.013} [\ensuremath{-0.026}, \ensuremath{-0.001}] & \ensuremath{0.000} [\ensuremath{-0.001}, \ensuremath{+0.001}] & \ensuremath{0.000} [\ensuremath{-0.001}, \ensuremath{+0.001}] \\
 &  & THINK-1536 & 1.000 & 0.000 & \ensuremath{-0.014} [\ensuremath{-0.027}, \ensuremath{-0.002}] & \ensuremath{0.000} [\ensuremath{0.000}, \ensuremath{0.000}] & \ensuremath{0.000} [\ensuremath{0.000}, \ensuremath{0.000}] \\
 &  & BUDGET-512-1500 & 1.000 & 0.000 & \ensuremath{-0.018} [\ensuremath{-0.030}, \ensuremath{-0.005}] & \ensuremath{0.000} [\ensuremath{0.000}, \ensuremath{0.000}] & \ensuremath{0.000} [\ensuremath{0.000}, \ensuremath{0.000}] \\
 & MMLU-Pro & THINK-64 & 0.641 & 0.359 & \ensuremath{-0.005} [\ensuremath{-0.026}, \ensuremath{+0.016}] & \ensuremath{-0.016} [\ensuremath{-0.039}, \ensuremath{+0.005}] & \ensuremath{+0.016} [\ensuremath{-0.005}, \ensuremath{+0.039}] \\
 &  & THINK-512 & 0.932 & 0.068 & \ensuremath{-0.004} [\ensuremath{-0.029}, \ensuremath{+0.021}] & \ensuremath{-0.008} [\ensuremath{-0.026}, \ensuremath{+0.011}] & \ensuremath{+0.008} [\ensuremath{-0.011}, \ensuremath{+0.026}] \\
 &  & THINK-1536 & 0.997 & 0.003 & \ensuremath{+0.004} [\ensuremath{-0.021}, \ensuremath{+0.029}] & \ensuremath{0.000} [\ensuremath{-0.004}, \ensuremath{+0.004}] & \ensuremath{0.000} [\ensuremath{-0.004}, \ensuremath{+0.004}] \\
\bottomrule
\end{tabular}}
\end{table}

\begin{table*}[t]
\caption{H3's statistics recomputed on the seeded outputs (the cards' sampling values, a seed per request) for the two models with a seeded arm; descriptive. $\tau_g$ and $\tau_s$ are the shares of \cfg{THINK-512} calls cut at the token limit with greedy decoding and seeded, and $\tau_s-\tau_g$ their paired difference. $D_T$ and $D_P$ are the seeded drop in truncation and gain in parsing of \cfg{BUDGET-512-1500} against \cfg{THINK-512}, with unadjusted intervals. The preregistered verdicts use family-adjusted intervals at another level and are not compared with these.}
\label{tab:app-seeded-h3}
\centering
\scriptsize
\setlength{\tabcolsep}{2pt}
\resizebox{\ifdim\width>\linewidth\linewidth\else\width\fi}{!}{%
\begin{tabular}{@{}lllccccc@{}}
\toprule
Partition & Model & Pool & $\tau_g$ & $\tau_s$ & $\tau_s-\tau_g$ & $D_T$ (seeded) & $D_P$ (seeded) \\
\midrule
Test & Qwen3-1.7B & LB letters & 0.000 & 0.001 & \ensuremath{0.000} [\ensuremath{-0.001}, \ensuremath{+0.001}] & \ensuremath{+0.001} [\ensuremath{0.000}, \ensuremath{+0.001}] & \ensuremath{+0.001} [\ensuremath{0.000}, \ensuremath{+0.001}] \\
 &  & MMLU-Pro & 0.060 & 0.068 & \ensuremath{+0.008} [\ensuremath{-0.011}, \ensuremath{+0.026}] & \ensuremath{+0.068} [\ensuremath{+0.051}, \ensuremath{+0.085}] & \ensuremath{+0.066} [\ensuremath{+0.050}, \ensuremath{+0.084}] \\
 &  & GSM8K & 0.000 & 0.000 & \ensuremath{0.000} [\ensuremath{0.000}, \ensuremath{0.000}] & \ensuremath{0.000} [\ensuremath{0.000}, \ensuremath{0.000}] & \ensuremath{-0.002} [\ensuremath{-0.006}, \ensuremath{0.000}] \\
 & Intern-S1 & LB letters & 1.000 & 0.997 & \ensuremath{-0.003} [\ensuremath{-0.005}, \ensuremath{-0.001}] & \ensuremath{+0.963} [\ensuremath{+0.957}, \ensuremath{+0.969}] & \ensuremath{+0.702} [\ensuremath{+0.689}, \ensuremath{+0.716}] \\
 &  & MMLU-Pro & 0.915 & 0.902 & \ensuremath{-0.013} [\ensuremath{-0.034}, \ensuremath{+0.009}] & \ensuremath{+0.618} [\ensuremath{+0.584}, \ensuremath{+0.652}] & \ensuremath{+0.589} [\ensuremath{+0.554}, \ensuremath{+0.625}] \\
 &  & GSM8K & 0.522 & 0.619 & \ensuremath{+0.097} [\ensuremath{+0.051}, \ensuremath{+0.142}] & \ensuremath{+0.581} [\ensuremath{+0.539}, \ensuremath{+0.622}] & \ensuremath{+0.567} [\ensuremath{+0.526}, \ensuremath{+0.609}] \\
\addlinespace
Replication & Qwen3-1.7B & LB letters & 0.001 & 0.000 & \ensuremath{-0.001} [\ensuremath{-0.002}, \ensuremath{0.000}] & \ensuremath{0.000} [\ensuremath{0.000}, \ensuremath{0.000}] & \ensuremath{0.000} [\ensuremath{0.000}, \ensuremath{0.000}] \\
\bottomrule
\end{tabular}}
\end{table*}

\begin{table}[H]
\caption{Tied against untied output layer of \qwen\ on FreeToken, the case study's six configurations, all four LongevityBench tasks pooled (\V{tied.rows} rows in \V{tied.units} participant components, unflagged rows of both partitions); computed after the outputs existed.}
\label{tab:app-tied}
\centering
\scriptsize
\setlength{\tabcolsep}{2pt}
\resizebox{\ifdim\width>\linewidth\linewidth\else\width\fi}{!}{%
\begin{tabular}{@{}lccccccc@{}}
\toprule
Configuration & Tied & Untied & Tied $-$ untied & Parse t / u & Same answers & Only t / u right & $p$ \\
\midrule
CFG-THINK-64-ORIG & 0.000 & 0.000 & \ensuremath{0.000} [\ensuremath{0.000}, \ensuremath{0.000}] & 0.000 / 0.000 & 1.000 & 0 / 0 & 1.000 \\
B1 & 0.452 & 0.449 & \ensuremath{+0.003} [\ensuremath{-0.001}, \ensuremath{+0.007}] & 1.000 / 1.000 & 0.741 & 182 / 157 & 0.192 \\
CFG-THINKING-FALSE-64-ORIG & 0.460 & 0.460 & \ensuremath{0.000} [\ensuremath{-0.005}, \ensuremath{+0.005}] & 1.000 / 1.000 & 0.731 & 194 / 194 & 1.000 \\
CFG-THINK-1536-ORIG & 0.469 & 0.463 & \ensuremath{+0.006} [\ensuremath{0.000}, \ensuremath{+0.013}] & 1.000 / 1.000 & 0.656 & 384 / 333 & 0.062 \\
CFG-NOTHINK-SUFFIX-64-ORIG & 0.452 & 0.449 & \ensuremath{+0.003} [\ensuremath{-0.001}, \ensuremath{+0.007}] & 1.000 / 1.000 & 0.741 & 182 / 157 & 0.192 \\
CFG-THINK-64-ANSWERONLY & 0.479 & 0.470 & \ensuremath{+0.009} [\ensuremath{+0.004}, \ensuremath{+0.014}] & 1.000 / 1.000 & 0.731 & 260 / 185 & 0.000 \\
\bottomrule
\end{tabular}}
\end{table}

\begin{table}[H]
\caption{The four GEPA runs (computed after the outputs existed): proposals and candidates, refused by the interface check, kept by GEPA, metric calls, the selected candidate, its prompt lengths, its validation score and its accuracy on the letter rows of the replication and test partitions.}
\label{tab:app-gepa}
\centering
\tiny
\setlength{\tabcolsep}{1pt}
\resizebox{\ifdim\width>\linewidth\linewidth\else\width\fi}{!}{%
\begin{tabular}{@{}lccccccccccc@{}}
\toprule
Harness & Proposals & Candidates seen & Refused & Kept & Metric calls & Selected / kept & Characters (system / instruction) & Valset & Replication acc. & Test acc. & Test parse \\
\midrule
GEPA-CLEAN-s0 & 207 & 204 & 28 & 31 & 6{,}039 & 0 / 31 & 103 / 0 & 0.680 & 0.633 & 0.632 [0.618, 0.646] & 1.000 \\
GEPA-BROKEN-s0 & 377 & 377 & 24 & 24 & 6{,}147 & 1 / 24 & 103 / 1{,}146 & 0.660 & 0.635 & 0.629 [0.615, 0.643] & 1.000 \\
GEPA-CLEAN-s1 & 195 & 192 & 36 & 31 & 6{,}003 & 12 / 31 & 2{,}469 / 922 & 0.700 & 0.641 & 0.631 [0.616, 0.645] & 1.000 \\
GEPA-BROKEN-s1 & 243 & 238 & 67 & 29 & 6{,}003 & 1 / 29 & 103 / 2{,}797 & 0.667 & 0.619 & 0.620 [0.606, 0.634] & 1.000 \\
\bottomrule
\end{tabular}}
\end{table}

\begin{table}[H]
\caption{LB-0038 (age regression), test partition, descriptive.}
\label{tab:app-lb38}
\centering
\scriptsize
\setlength{\tabcolsep}{2pt}
\resizebox{\ifdim\width>\linewidth\linewidth\else\width\fi}{!}{%
\begin{tabular}{@{}llcccccc@{}}
\toprule
Model & Measure & THINK-64 & THINK-512 & THINK-1536 & AUTHORS-1500 & BUDGET-512-1500 & RESCUE-1500 \\
\midrule
Qwen3-1.7B & parse rate & 0.000 & 1.000 & 1.000 & 1.000 & 1.000 & 1.000 \\
 & mean abs. error (years) & -- & 15.69 & 15.69 & 16.50 & 15.69 & 15.69 \\
 & exact match & 0.000 & 0.027 & 0.027 & 0.030 & 0.027 & 0.027 \\
\addlinespace
SmolLM3 & parse rate & 0.000 & 0.000 & 0.004 & 0.001 & 0.003 & 0.001 \\
 & mean abs. error (years) & -- & -- & 24.71 & 57.70 & 35.08 & 57.70 \\
 & exact match & 0.000 & 0.000 & 0.000 & 0.000 & 0.000 & 0.000 \\
\addlinespace
Intern-S1 & parse rate & 0.000 & 0.000 & 0.000 & 0.004 & 0.000 & 0.004 \\
 & mean abs. error (years) & -- & -- & -- & 16.57 & -- & 16.57 \\
 & exact match & 0.000 & 0.000 & 0.000 & 0.000 & 0.000 & 0.000 \\
\midrule
\multicolumn{8}{@{}p{\linewidth}@{}}{Mean absolute error is over parsed rows; exact match over all rows. Constant median 50.0 years (chosen on replication): mean absolute error 16.41 (oracle median 16.41), exact match 0.013.} \\
\bottomrule
\end{tabular}}
\end{table}

\subsection{Sensitivity analyses}
\begin{table}[H]
\caption{Sensitivity analyses: the number of the primary endpoints whose verdict changes, and which.}
\label{tab:app-sens}
\centering
\scriptsize
\setlength{\tabcolsep}{3pt}
\resizebox{\ifdim\width>\linewidth\linewidth\else\width\fi}{!}{%
\begin{tabular}{@{}lccp{0.55\linewidth}@{}}
\toprule
Analysis & Endpoints & Changed & Which (primary verdict $\to$ here) \\
\midrule
Strict scorer & 31 & 2 & H2 SmolLM3 MMLU-Pro, RESCUE: holds $\to$ inconclusive; H1 SmolLM3 GSM8K: visibility dominates $\to$ no gain: not tested \\
Never-called rows only & 13 & 3 & H2 Qwen3 letters, RESCUE: holds $\to$ inconclusive; H2 Qwen3 letters, GEPA-BROKEN-s1: holds $\to$ inconclusive; H1 SmolLM3 letters: visibility dominates $\to$ inconclusive \\
Flagged rows included (H1--H3) & 31 & 0 & none \\
Case study's pool (LB-0030 and LB-0034) & 13 & 1 & H2 Qwen3 letters, GEPA-CLEAN-s1: holds $\to$ inconclusive \\
LB-0030 alone & 13 & 2 & H2 Qwen3 letters, RESCUE: holds $\to$ inconclusive; H2 Qwen3 letters, GEPA-CLEAN-s1: holds $\to$ inconclusive \\
LB-0034 alone & 13 & 0 & none \\
LB-0042 alone & 13 & 0 & none \\
Role-break rule off & 31 & 0 & none \\
Role-break rule, first version & 31 & 0 & none \\
Replication and test pooled & 31 & 0 & none \\
Flagged rows included (H4 claims) & 16 & 2 & C SmolLM3: fails $\to$ inconclusive; C Intern-S1: inconclusive $\to$ fails \\
\bottomrule
\end{tabular}}
\end{table}

\begin{table}[H]
\caption{Composition of the never-called rows (computed after the outputs existed) next to the analysis rows. The share deceased at follow-up is the share of LB-0042 rows whose follow-up turn says that the participant was deceased, whenever the death occurred; it is not the ten-year label of the task, whose gold is ``Yes'' for a participant who was alive at ten years or who died later.}
\label{tab:app-never}
\centering
\scriptsize
\setlength{\tabcolsep}{3pt}
\resizebox{\ifdim\width>\linewidth\linewidth\else\width\fi}{!}{%
\begin{tabular}{@{}llcccc@{}}
\toprule
Rows & Partition & Rows & Letter rows & LB-0030 mean age & LB-0042 share deceased at follow-up \\
\midrule
Never called & replication & 187 & 103 & 36.2 & 0.022 \\
Analysis rows & replication & 1{,}990 (pool) & -- & 50.8 & 0.345 \\
Never called & test & 356 & 207 & 37.3 & 0.022 \\
Analysis rows & test & 3{,}942 (pool) & -- & 50.8 & 0.346 \\
\bottomrule
\end{tabular}}
\end{table}

\section{Deviations, exceptions and addenda in full}
\label{app:deviations}

In our repository (Appendix~\ref{app:repro}) the frozen preregistration (\texttt{docs/icml\_ext/PREREG\_ICML\_EXT.md}) records exceptions and known deviations before the freeze, together with its decisions of the days before it.
After the freeze, departures are appended to \texttt{docs/icml\_ext/PREREG\_ICML\_EXT\_ADDENDA.md}, one entry each, under its section on deviation rules.
This appendix lists them all: Table~\ref{tab:dev} gives the deviations that bear on a result, and the sections below give every exception, known deviation, pre-freeze decision and addendum.
``Seen'' states what had been seen of the study's outputs when an entry was made.

\begin{table}[H]
\caption{Deviations and addenda that bear on a result; the entries below give the rest. ``Seen'' is what had been seen of the study's outputs when the entry was made.}
\label{tab:dev}
\centering
\footnotesize
\setlength{\tabcolsep}{3pt}
\begin{tabular}{@{}>{\raggedright\arraybackslash}p{0.12\textwidth}>{\raggedright\arraybackslash}p{0.42\textwidth}>{\raggedright\arraybackslash}p{0.27\textwidth}>{\raggedright\arraybackslash}p{0.12\textwidth}@{}}
\toprule
Entry & What happened & Bearing on the results & Seen \\
\midrule
Before the freeze & The primary set of H1 changed twice (after the pilot's parse rates, and after a review in Codex); the scoring defect was split in two; the tied-output-layer defect was redesigned; one evaluability rule was fixed; the family of H4 was fixed at \V{c.fam4}; the margins were set to \V{c.margin}. & Defines the endpoints; no study accuracy was used. & pilot parse rates \\
Rows left out & \V{excl.re.v21_holdout} replication and \V{excl.te.v21_holdout} test rows that the case study had called, and \V{excl.re.reading_exposed} rows whose answers could be recovered from an upload made by mistake, were called but left out. & Putting them back changes no verdict of H1--H3 and \V{sens.flagged.h4.n} H4 claims. & no outputs \\
A change of GPU & After a restart the platform assigned another GPU of the same model; the determinism check rerun on the pilot rows passed, with outputs equal to the old GPU's. & None. \V{add.gpu.h} GPU-hours outside the budget. & pilot rows \\
Two script stops & The chain script stopped twice during the GEPA search, both times from a fault in the script. & None; no run was affected. & none \\
The thinking-switch check & A check on \intern's thinking switch stopped the run that sends the reference answer, and we let it record instead of stop. The repeat reproduced the first \V{add.repeat.n} calls token for token, and the check failed on \V{add.gate.n} of the run's \V{add.gate.of} thinking-off calls and on none of \V{add.gate.clean} in the clean stages. & The reference-answer defect in \intern\ only through the descriptive selection, since the claim reads the start. \V{add.idle.h} idle server-hours counted. & counts only \\
The checkpoint & \V{h4.evaluable} of the \V{c.fam4} pairs were evaluable; the pilot had suggested \V{d.e.expected}. & The unparsed-as-A scoring in \qwen\ is \verdict{not evaluable}; the gold fallback in \intern\ (bound \V{h4.b1.i.bound}) is evaluable. & counts and bounds \\
The determinism recheck & A recheck of \V{det.n} test calls per model was identical on all three models. & No verdict is relabelled descriptive. & counts \\
The analysis environment & The preregistered analysis ran on the instance with numpy~\V{numpy}, not the local \V{add.numpy.local}; the resampling stream was checked equal. & None. & all aggregates \\
The claim rule & The two listed cases of the claim rule for the check sample are incomplete, so the general rule governs, and the pinned code implements it. & None: the verdicts are the code's, and the skewed split in \smol\ fails, a descriptive label. & all aggregates \\
Computed after the outputs & The composition of the never-called rows, the trivial baselines of the three pools, the tied-versus-untied comparison, the GEPA candidates and the serving counts. & Labelled, and they decide nothing. & aggregates \\
\bottomrule
\end{tabular}
\end{table}

\subsection{Exceptions accepted before the freeze (X1--X9)}
\begin{itemize}
\item[X1] Three \texttt{RECORD} entries of the vLLM virtual environment do not match their files (two PyNvVideoCodec libraries and \texttt{build\_backend.py}); they date from the original installation, not from the clone. Decided 2026-09-28.
\item[X2] FreeToken 0.1.3 opens a process-group store on its serving port plus one that PyTorch binds off loopback; it carries no model traffic. The runner allows exactly this one listener, on FreeToken only, while no connection to it comes from off the machine. The case study's FreeToken runs very likely had it too.
\item[X3] \texttt{FREETOKEN\_ALLOW\_CUDA\_MISMATCH}, the case study's variable, is exempt from the credential-name check although its name contains ``token''.
\item[X4] \texttt{VLLM\_USE\_FLASHINFER\_SAMPLER=0} in every vLLM run, because FlashInfer's sampling kernel could not compile against the image's CUDA toolkit.
\item[X5] A change of instance on 2026-09-28: the source instance's GPU was no longer available, so the instance was cloned to one of the same configuration; the GPU identifier changed and the clone was verified against the recorded freeze, wheel, record and model hashes.
\item[X6] A restart of the container at 2026-09-29 00:57:25 UTC, which we did not initiate; no server was running, and the GPU identifier and the SSH settings were unchanged.
\item[X7] The first gate check (G1) ran on the checkpoint's own directory, whose configuration declares tied embeddings, before the output-layer decision; its results were operational only.
\item[X8] AI-assisted readings and uploads (item-exposure contacts 1 and 2); the reading of the main export (contact 3) and the copy of the item files to the project's own instance (contact 4) are in Appendix~\ref{app:exposure}.
\item[X9] A change of the physical GPU after a restart (addenda A1, A2).
\end{itemize}

\subsection{Known deviations from the plans (K1--K9)}
\begin{itemize}
\item[K1] Three models instead of four; the FreeToken continuity on \qwen\ only.
\item[K2] GEPA with a local proposer replaces Meta-Harness, so the proposer through an API and its budget fall away.
\item[K3] A single answer allowance of \V{k.ans} tokens replaces a calibrated rule that left out the calls the calibration allowance cut.
\item[K4] The batch-invariant server with batching replaces concurrency one on the standard server (about \V{k.k4.h} GPU-hours at concurrency one).
\item[K5] The pilot's LongevityBench rows come from the case study's evolution file or validation shards, not from never-called rows; calibration reported no accuracy.
\item[K6] The extended scorer rules, their narrowing and the role-break rule were decided after the pilot's parse audit and spot-checks on pilot items; the strict parser is a preregistered sensitivity analysis.
\item[K7] The GEPA interface changed between its two spikes, on the counts of refused proposals.
\item[K8] The environment that counts tokens locally differs from the instance's in the Python and numpy versions and other bundled tools; none takes part in tokenization, and the render lists were checked on both machines.
\item[K9] The pilot report was written after the first draft of the preregistration, from aggregates only.
\end{itemize}

\subsection{Decisions between the first draft and the freeze}
We revised the draft after our decisions on its first version, after a review in claude.ai, after two reviews and a final check in Codex (each given the file and not our repository), after our decisions on the summary of the drafting session, and after the reading of the rendered requests.
Table~\ref{tab:dev} calls these decisions ``Before the freeze''.
The decisions that shape the endpoints were these:
\begin{itemize}
\item The primary set of H1 changed twice, first after the pilot's parse rates and parsed answers were seen (no accuracy was used) and then after the first review in Codex, with \cfg{THINK-64} against \cfg{AUTHORS-1500} moved out to nine sanity checks.
\item The scoring defect B was split into B1 and B2.
\item The tied-output-layer defect E was redesigned so that the confirmatory instance serves the tied variant on both samples.
\item One evaluability rule applies to every defect, and the family of H4 is fixed at \V{c.fam4} whatever is evaluable.
\item The margin and the minimal difference are \V{c.margin}, with Bonferroni within each family.
\end{itemize}
After the reading, seven replication rows whose reference answers could be recovered from render-hash lists uploaded by mistake were flagged and left out of every analysis (the reading itself, item-exposure contact~3, concluded pass); Table~\ref{tab:dev} calls this ``Rows left out''.
The exclusion came after the last review in Codex and was not reviewed there.

\subsection{Addenda after the freeze (A1--A\V{add.last})}
\begin{description}
\item[A1, A2 (2026-09-30)] \emph{A change of GPU.} After we restarted the instance, the platform had assigned a different physical GPU of the same model. By the preregistration's rules the work stopped before any copy or call. Every check passed on the new GPU; the determinism check (G2~(b)) was rerun on the pilot rows and passed for all three models, and its outputs equal the old GPU's token for token. \V{add.gpu.h} GPU-hours, outside the budget; the new identifier is pinned. Seen: pilot rows only.
\item[A3 (2026-09-30)] \emph{Two script stops.} The chain script stopped twice during the GEPA search, both times from a fault in the script (a password check that failed under a shell option; a step that required an empty GPU while the task server stayed up). No run, call or check was affected. Seen: nothing of replication or test.
\item[A4, A5 (2026-10-01, 2026-10-02)] \emph{The thinking-switch check.} The defect-A pass of \intern\ stopped itself at gate~(b): five of its first thinking-off calls (\cfg{AUTHORS-1500}) were not clean (cut at the limit with a think block opened and not closed). The gate held on all \V{add.gate.clean0} thinking-off calls of the clean runs up to then. The preregistration had not foreseen that the injected defect could trigger the gate. We decided that in that stage only the gate records instead of stopping (commit \texttt{aae8b11}, runner \texttt{a9478cee}), that the idle server stops and its hours are counted (about \V{add.idle.h} of its server-hours were idle), that the partial folder is moved out of the run root, and that the run is repeated from scratch under a new name. The repeat reproduced all \V{add.repeat.n} calls of the failed attempt token for token and failed the gate on \V{add.gate.n} of the pass's \V{add.gate.of} thinking-off calls (the gate reads the completion, and all \V{add.gate.of} prompts carry think tags that the chat template writes around the reference turn). By the repair rules this is confirmatory: the failed attempt is kept and reported, and only the repeat enters the analysis. Seen: counts only.
\item[A6 (2026-10-02)] \emph{The checkpoint.} It ran on the instance, since the local machine lacked memory. All \V{add.rep.runs} replication runs were complete with matching manifests; the derived rescue equals the real two-call sequence on \V{resc.n} of \V{resc.n} verification items per model; \V{h4.evaluable} of \V{c.fam4} pairs are evaluable; and the selections of the pipelines whose selection sample lies in replication were locked in a record (\texttt{f033f23f}) whose hash was committed before the first test call. Seen: counts, bounds and the record's hash.
\item[A7 (2026-10-02, 2026-10-03)] \emph{The determinism recheck.} The test partition ran \V{add.test.runs} runs and \V{add.test.calls} calls, with no retry and no stop, and the recheck of \V{det.n} calls per model was identical on all three; the study used \V{add.total.h} of the \V{add.budget.h} GPU-hours of the cost bound. Seen: counts.
\item[A8 (2026-10-03)] \emph{The analysis environment.} The preregistered analysis ran once on the instance (numpy~\V{numpy}, not the local \V{add.numpy.local}; the resampling stream was checked equal for three seeds) after every pinned file matched its frozen hash, and the selections it reports equal the locked ones on all \V{add.sel.n} comparisons. The code that computes three descriptive items the pinned driver lacks (tied versus untied, GEPA candidates, serving counts) was written after the outputs existed and is labelled so. Seen: all aggregates, for the first time.
\item[A9 (2026-10-03)] \emph{Computed after the outputs.} We shut the instance down and do not release it; the report and its tables were committed as aggregates (\texttt{49e5f9c}); the never-called rows' composition (\V{nc.both.rows} rows, \V{nc.both.letter} on the letter tasks) and constant-letter baselines were computed after the outputs existed.
\item[A10 (2026-10-03)] \emph{Computed after the outputs.} Constant-label baselines, because a constant letter is a weak floor where the option order is shuffled: the strongest baseline of the pool is \V{bl.te.lb.so} on test against \V{bl.te.lb.clo} for the oracle constant letter. The strongest baseline is the one used in this paper.
\item[A11 (2026-10-03)] \emph{The claim rule.} The rule for $D_R$ in the claim of H4. The audit of this paper recomputes the claim from its three adjusted intervals; the two sub-cases that the preregistration lists after its general rule were incomplete, the general rule governs, and the pinned code implements it (our decision). No verdict changes. Seen: all aggregates. Details below.
\item[A12 (2026-10-03)] \emph{Computed after the outputs.} The trivial baselines of MMLU-Pro (chance and the constant letter) and GSM8K (the constant answer), chosen on the other partition and oracle, next to every model's accuracy in every grid configuration (holdout contact \#57). Seen: aggregates, no item key or text.
\end{description}

\subsection{The rule for $D_R$ (addendum A11)}
For $D_R$ the preregistration says that it fails if its interval lies ``wholly on the other side of its boundary'' and lists two sub-cases, wholly within the band and wholly outside it with the opposite sign.
The pinned code (\texttt{scripts/icml\_ext/analysis\_h4.py}, \texttt{claim\_verdict}) implements the general phrase: $D_R$ is graded in the direction of $D_S$'s sign, holds if it lies wholly beyond the band there, and fails if its interval lies wholly on the other side of that one boundary, which also covers an interval partly within the band and partly beyond its far edge.
The general rule governs and the two sub-cases were incomplete.
It matters for one pair, C in \smol: $D_S=\V{h4.c.s.ds}$ points down, so $D_R$ would have to lie below $-\V{c.minm}$, and its adjusted interval \CI{\V{h4.c.s.dr.lo}}{\V{h4.c.s.dr.hi}} lies wholly above it, so the claim cannot hold and is \verdict{fails}, a descriptive label.
Read as the two sub-cases only, the claim would be \verdict{inconclusive}, and H4 would count \V{lit.h4.holds} holds, \V{lit.h4.fails} fails and \V{lit.h4.inconclusive} inconclusive instead of \V{cnt.h4.holds}, \V{cnt.h4.fails} and \V{cnt.h4.inconclusive}; every other claim has the same verdict under both readings.

\subsection{What this paper adds}
This paper's text, tables and figures format aggregates committed in our repository and add no computation except the counts of its own tables (verdict counts by family, the largest and smallest of a column) and the independent recomputation of every verdict from its interval, which agrees with the report on every endpoint that has a verdict (Appendix~\ref{app:repro}).

\subsection{The second study: decisions before the freeze, and addenda A1--A\V{s2.add.last} after it}
\label{app:deviations2}
The decisions between the second study's first draft and its freeze, the exceptions it accepted and the differences from the first study are listed in \texttt{docs/icml\_ext2/DIFF\_FROM\_FROZEN.md}.
Its addenda after the freeze are in \texttt{docs/icml\_ext2/PREREG\_ICML\_EXT2\_ADDENDA.md}; each states what happened, the rule applied, the outputs seen at the time and the status.
\paragraph{A1: a pinned file was missing on the instance.}
The checkpoint stopped when it first ran, because the list of replication rows flagged as exposed, which the frozen code pins by hash, was not on the instance; the deployment script had omitted it.
No run had been read.
The file was copied by hand from our machine, its SHA256 was checked against the pin, the copy was logged first as a holdout contact, and the checkpoint was run again alone.
\paragraph{A2: the checkpoint record.}
The checkpoint record, which holds the selections, was not opened; its SHA256 (given in Appendix~\ref{app:repro}) was logged and committed before the first test call, and later checks compared hashes only.
\paragraph{A3: the shutdown of the instance.}
A platform shutdown on \V{s2.shut} ended the instance because the account balance ran out; it was not a platform fault.
The 14B's test grid had finished and verified, and the evaluation of one selected harness on the test rows had started.
No output had been read before or after.
\paragraph{A4: the clone, its verification and the change of GPU.}
We cloned the instance to another host, with another \V{s2.gpu} of the same model and driver.
The clone was verified file by file, by hashes and counts of the environment, the models, the item files, the frozen guard, every completed run and the checkpoint record, before anything else ran; nothing was missing.
The interrupted evaluation, which had no manifest and a few recorded responses, was moved out of the run folder as a failed attempt, kept and excluded from every analysis, and the evaluation was rerun from scratch with the same requests.
The preregistration's rule for a change of the physical GPU after the first replication call applied.
\paragraph{A5: the finished test grid.}
We decided that the 14B's test grid, which had completed and verified on the first GPU, is kept, and that the recheck of the replication subset decides: an identical result for Qwen3-14B lets it stand, and any difference would have stopped the sequence for our decision.
\paragraph{A6: the recheck.}
On the new GPU the batch-invariance check on the pilot rows was identical for the three models that were still to be called or rechecked, and \V{s2.recheck} replication calls of each model in three configurations were identical token for token to the outputs of the first GPU; part 2 resumed as confirmatory.
\paragraph{A7: the guard and the watchdog.}
While we were away the instance carried a guard that shut it down with the platform's own command when part 2 and the analysis had finished and the outputs were copied, or at a fixed time at the latest, and a watchdog on our machine checked it later; neither was triggered by a fault.
\paragraph{A8: the copy of the outputs.}
Part 2 finished with every check passed and no stop.
The external drive that holds the repository's run folders was not available, so the outputs were copied to a folder on our machine as compressed archives; every file was checked against the instance's list and every manifest against its files, the marker that the copy was verified was written, and the guard shut the instance down.
\paragraph{A9: the second copy, the extraction and the descriptive analyses.}
Windows had kept the external drive offline by policy and an administrator set it online; the archives were copied to it, extracted and verified again.
The descriptive analyses that the preregistration specifies (the two 1.7B arms with the reproduction record, the agreement with the first study, the non-commitment and refusal shares) were run, each after its holdout contact was logged, and the four aggregate reports were scanned for item keys, row fields and item text before they were committed.

\section{Expected in advance against observed}
\label{app:expected}

The preregistration states what it expected for several endpoints, from the pilot's parse and truncation rates and parsed answers (never from accuracy).
Table~\ref{tab:expected} sets each expectation next to what was observed.
The directions of H1 and H3 were visible in the pilot, and the preregistration says so; this table shows the places where the outcome differed from the expectation, which are as informative as the places where it did not.
``Inconclusive'' in the Expected column is an informal expectation read off planned power under assumed effects (Appendix~\ref{app:power}), not a prediction with a decision rule.

\begin{table}[H]
\caption{Expected in advance (preregistration, the sections on H1, H2 and H4) and observed.}
\label{tab:expected}
\centering
\small
\begin{tabular}{@{}p{0.2\linewidth}p{0.33\linewidth}p{0.4\linewidth}@{}}
\toprule
Item & Expected & Observed \\
\midrule
H1 sanity checks (\cfg{THINK-64}) & $T>0$ in eight cells; \qwen\ on GSM8K fails the gate & \verdict{visibility dominates} in \V{cnt.h1s.visibility} of nine cells; \qwen\ on GSM8K fails the gate. As expected. \\
H1, \qwen\ (three cells) & Fails the gate (on MMLU-Pro, or quality matters) & \verdict{no gain: not tested} in all three. As expected. \\
H1, \intern\ on the letters & $T>0$ nearly inevitable & \verdict{visibility dominates}, $T=\V{h1.te.i.lb.t}$. As expected. \\
H1, \smol\ and \intern\ on GSM8K & Inconclusive on $T$ (planned power \V{ex.h1.s.gs.s1} and \V{ex.h1.s.gs.s2} for \smol, \V{ex.h1.i.gs.s1} and \V{ex.h1.i.gs.s2} for \intern, in the two scenarios) & \smol: \verdict{visibility dominates}, $T=\V{h1.te.s.gs.t}$ \CI{\V{h1.te.s.gs.t.lo}}{\V{h1.te.s.gs.t.hi}}. \intern: the gate fails, $\Delta=\V{h1.te.i.gs.delta}$. Both differ. \\
H2, \intern\ on GSM8K against \cfg{RESCUE-1500} & Inconclusive (planned adjusted power \V{x.power}) & \verdict{fails}, $\theta=\V{h2.te.i.gs.res.theta}$ \CI{\V{h2.te.i.gs.res.lo}}{\V{h2.te.i.gs.res.hi}}. Differs. \\
H2, GEPA harnesses & Power unknown in advance; may well be inconclusive & \verdict{holds} for all four; all intervals contain zero. \\
H4, evaluable pairs & \V{d.e.expected} of \V{c.fam4}: B1 in \intern\ and B2 in \qwen\ not evaluable & \V{h4.evaluable} of \V{c.fam4}: only B2 in \qwen\ is not; B1 in \intern\ has bound \V{h4.b1.i.bound}. Differs. \\
H4, bounds on the pilot's \V{d.pilot.lb} letter rows against $S$ & B1 \V{ex.b1.q}, \V{ex.b1.s}, \V{ex.b1.i}; B2 \V{ex.b2.q}, \V{ex.b2.s}, \V{ex.b2.i}; D \V{ex.d.q}, \V{ex.d.s}, \V{ex.d.i}; E \V{ex.e.q} (\qwen, \smol, \intern) & B1 \V{h4.b1.q.bound}, \V{h4.b1.s.bound}, \V{h4.b1.i.bound}; B2 \V{h4.b2.q.bound}, \V{h4.b2.s.bound}, \V{h4.b2.i.bound}; D \V{h4.d.q.bound}, \V{h4.d.s.bound}, \V{h4.d.i.bound}; E \V{h4.e.q.bound}. \\
H4, D & The claim rests on $\Gamma$ and holds only if its point estimate is within about \V{ex.d.within} of zero (about \V{ex.d.half} of the time with no true difference) & Holds in all three models, $\Gamma=\V{h4.d.q.g}$, $\V{h4.d.s.g}$, $\V{h4.d.i.g}$. Consistent with the expectation, which allowed it. \\
H4, B2 & Can be large in \smol\ and \intern & $D_S=\V{h4.b2.s.ds}$ and $\V{h4.b2.i.ds}$. As expected. \\
H4, E & No expectation on its size & $D_S=\V{h4.e.q.ds}$, $D_R=\V{h4.e.q.dr}$; \verdict{fails}. \\
Detection matrix & Each defect detected by its signature rules (A: TR, IR1, IR2, TB3; B1: IR3; B2: TB4; C: IR4, IR5, TB1, not TB2 or PR; D: SCF1, SCF3; E: SMI) & All expected rules fired in \V{mx.expected.ok} of \V{mx.n} instances. TB1 did not fire in any of the three C instances, TB3 did not fire for A in \smol, and TB2 fired in all three C instances, with the gain rule of PR in \qwen. \\
\bottomrule
\end{tabular}
\end{table}

\section{Reproducibility}
\label{app:repro}

Every path, commit, tag and SHA256 value in this paper is that of a file in our repository, which is not public at present (see the statement on code and data availability after the Conclusion).
They are kept so that the planned public export can be checked against them; a reader cannot open these files until the export is linked.

\paragraph{The frozen record.}
In our repository the preregistration (\texttt{docs/icml\_ext/PREREG\_ICML\_EXT.md}) was frozen at \V{fr.time}~UTC on \V{fr.date}, in commit \texttt{7f645ae}, whose parent carries the tag \texttt{pre-icml-ext-freeze}; the tag \texttt{prereg-icml-ext-frozen} marks the freeze.
Its SHA256 is
\begin{center}\small\texttt{\V{fr.sha}}\end{center}
The file \texttt{docs/icml\_ext/PREREG\_ICML\_EXT.sha256} holds the hash and the time, and the addenda are in \texttt{docs/icml\_ext/PREREG\_ICML\_EXT\_ADDENDA.md}.
The statement on code and data availability says which files are attached to this arXiv version and why the freeze time cannot be checked independently.
The hash above lets a reader check the ancillary preregistration, and the report's hash below does the same for the report.

\paragraph{The outputs.}
The report of the preregistered analysis and its tables are committed as aggregates in our repository, in \texttt{docs/icml\_ext/ANALYSIS\_20261003/}: the report has SHA256 \texttt{1447c936d52523a1494169f30af239c7d36ce7e79e81655a6a91016570c22e0b}, the folder's manifest \texttt{37d3268a980974af6ab414af65b8b2b1a5b3d6071632f8f1867ffc3d871a8403}.
The items computed after the outputs are in three subfolders (the never-called rows and the constant-letter baselines; the constant-label baselines; the baselines of MMLU-Pro and GSM8K), each with its own manifest.
The study's \V{add.rep.runs} replication runs and \V{add.test.runs} test runs, the GEPA search and the failed attempt of A4 hold item text and model output, are not committed and will not be released; each has a manifest of SHA256 values that matched its files after the copy back.

\paragraph{The code.}
The files of our repository that fix the analysis are pinned in the preregistration by SHA256: \texttt{scripts/icml\_ext/analysis\_stats.py} (\texttt{e6a2c4dc}), \texttt{analysis\_records.py} (\texttt{7507fd6c}), \texttt{analysis\_h4.py} (\texttt{2f5a82e7}), \texttt{analysis\_run.py} (\texttt{d92ebfe2}), \texttt{defects.py} (\texttt{a2ad7e2e}), \texttt{scoring.py} (\texttt{3bce48ef}) and \texttt{runner\_checks.py} (\texttt{ec77476e}), and the caller \texttt{scripts/paper\_review/messages\_for\_inference.py} (\texttt{cc8debc7}).
The runner changed once after the freeze (addendum A4, commit \texttt{aae8b11}).
The environment is vLLM \V{d.vllm} with the packages listed in the setup log (\texttt{docs/icml\_ext/SERVER\_SETUP\_LOG.md}); the analysis ran with numpy~\V{numpy}.
vLLM pull request \V{rw.pr41674} (merged \V{rw.pr41674.date}) fixed a thinking-token budget that was silently ignored \citep{vllm41674}; the version used contains the fix, and H3 shows the budget acting.

\paragraph{This paper's numbers.}
Every number of the text is a value macro (\verb|\V| applied to a key) that \texttt{tools/values.py} registers with the file of our repository that it reads (at git \texttt{HEAD}, never the working tree) and the path inside it; \texttt{tools/gen.py} writes \texttt{common/numbers.tex} and the tables from that registry, and \texttt{--check} fails if a committed generated file would change.
\texttt{audit/number\_audit.py} then (i)~re-reads every registered value with its own path reader and formatter, (ii)~recomputes the verdict of every endpoint that has one (all but the unevaluable pair of the \V{cb.all} primary endpoints of the two studies) from its interval with the preregistered rules and compares it with the report's, (iii)~fails on any number in the hand-written text that is not a macro or an identifier, (iv)~checks every hash, path and tag named in the text against our repository, and (v)~checks the plotted values of the figures (\texttt{tools/make\_figures.py}).
\texttt{audit/CLAIM\_AUDIT.md} maps each claim of the text to its evidence.

\paragraph{The second study.}
Its preregistration (\texttt{docs/icml\_ext2/PREREG\_ICML\_EXT2.md}) has SHA256
\begin{center}\scriptsize\texttt{a2b8ec439d5ba941e8de2bda74fb74600f8fe9e7f0650a74e673cedf4ac85b61}\end{center}
it was frozen by the commit that adds its hash file, with the tags \texttt{pre-icml-ext2-freeze} and \texttt{prereg-icml-ext2-frozen}.
The checkpoint record, which locks the four selected harnesses and was committed by hash before the first test call, has SHA256
\begin{center}\scriptsize\texttt{4aa7da7484e5f117cf9a17180ce672c1b666e20d3ec3bd878fbc8afd88656a28}\end{center}
The environment of the instance equals the first study's by the SHA256 of its package list,
\begin{center}\scriptsize\texttt{fee9f31063305f6754e1020f9ad8cd063bc24f2b6ca59fca72924a938bc2f16d}\end{center}
on an \V{s2.gpu}; the clone of the instance after the shutdown was verified against the same hashes (Appendix~\ref{app:deviations2}).
The aggregate reports of its analyses are in \texttt{docs/icml\_ext2/ANALYSIS\_20261023/} and the numbers computed after the outputs existed in \texttt{docs/icml\_ext2/FOLLOWUP\_JSON\_20261023/}.

\paragraph{The bootstrap in full.}
The seed of a cell derives from a hash of its partition, model and pool (base \V{c.seedbase}), and a unit drawn $k$ times weights each of its rows by $k$.
The $B=\V{c.b}$ statistics are sorted ascending and indexed from zero, and for a two-sided tail probability $q$ the lower bound of the interval is the value at index $\lfloor Bq\rfloor-1$ and the upper bound the value at index $\lceil B(1-q)\rceil-1$, clamped to the range of indices.
For H4, $\Gamma$ pairs the $b$-th resample of $S$ with the $b$-th of $R$, which are disjoint samples drawn with their own seeds.

\paragraph{The fourth study.}
Its preregistration (\texttt{docs/icml\_ext4/PREREG\_ICML\_EXT4.md}) has SHA256
\begin{center}\scriptsize\texttt{df6ffb5d8555d3eed92b1fa83abea2476e732280d5fc8d9cf54d41458981a912}\end{center}
it was registered on the Open Science Framework (\url{https://osf.io/g5apw}) and frozen by commit \texttt{15b7ea6}, with the tags \texttt{pre-icml-ext4-freeze} and \texttt{prereg-icml-ext4-frozen}; \texttt{docs/icml\_ext4/PREREG\_ICML\_EXT4.sha256} holds the hash and the registration time, and the addenda are in \texttt{docs/icml\_ext4/PREREG\_ICML\_EXT4\_ADDENDA.md}.
The harness list, committed before the first replication call, has SHA256
\begin{center}\scriptsize\texttt{bf52418755328a9f2e313f7f5f79990394185cc1b14abc14b11ffa242e3a3e8a}\end{center}
and the checkpoint record, committed before the first test call,
\begin{center}\scriptsize\texttt{7688850b889728867fff73a8ad89c72a922de8609bd25dfb7d9b756c5fad48d1}\end{center}
The aggregate reports of the analysis are in \texttt{docs/icml\_ext4/ANALYSIS\_20261107/}: the report rerun with deviation D001 applied, which this paper uses, has SHA256
\begin{center}\scriptsize\texttt{f3fed57f0efc7fb368658e907e72db85dd73f434d23bcb3d9ab16e2e4a6374b3}\end{center}
and the original report, whose numeric values are equal to it, \texttt{83d3dbf8ba5d099498333c83e2ca590cdc5defa499aa4a0ba238cf8d830a9e12}.
The licence check of Table~\ref{tab:licences} is \texttt{docs/icml\_ext4/LICENCE\_CHECK\_20261107.md} and the reading of related work \texttt{docs/icml\_ext4/RELATED\_WORK\_READ\_20261107.md}.
The searches, the evaluations and the item files hold item text and model output, are not committed and will not be released; the Manifold conditions forbid releasing question text.

\section{Exposure and disclosure}
\label{app:exposure}

Every contact of ours or of our tools with held-out material is logged in our repository (\texttt{docs/paper\_review/HOLDOUT\_EXPOSURE\_LOG.md}, numbered contacts; \texttt{docs/icml\_ext/ITEM\_EXPOSURE\_LOG.md}, contacts with the extension's items by outside services), and before it happens, with one exception that Table~\ref{tab:contacts} marks.
The case study's holdout contacts are \#1 to \#38; those of this study are \#39 to \#58 and \#60.
Other later contacts, including the preparation of the ancillary files and the posting of this version, are logged in the repository and are not listed here.
Item-exposure contacts 1 and 2 are uploads of the pilot's reading export and runner and of a spot-check and the GEPA code to claude.ai; contact 3 is the AI-assisted reading of the main export (concluded pass; one of us relayed the reading and its conclusion and also checked the conclusion); contact 4 is the copy of the main item files to the project's own GPU instance, which is not an outside service.

\begin{table}[H]
\caption{Holdout contacts of this study (\#39 to \#58 and \#60): what was read or done. Entries \#50 to \#58 and \#60 are after the freeze.}
\label{tab:contacts}
\centering
\small
\begin{tabular}{@{}p{0.07\linewidth}p{0.85\linewidth}@{}}
\toprule
\# & Contact \\
\midrule
\#39 & The ICML formatting draft of the case study copied to a synchronised folder for an upload. \\
\#40 & The LongevityBench split of this study, the earlier-contact flags and the first gold audit built. \\
\#41 & Fourteen files uploaded to claude.ai for a review of the reading export and the runner (item-exposure contact 1). \\
\#42 & The preregistration draft revised and uploaded for reviews in claude.ai and in Codex, after a check for holdout-derived values. \\
\#43 & The main item files built. \\
\#44 & The LB-0042 rows on which B1's fallback gold differs from the derived gold counted. \\
\#45 & The gold audit rerun over the main items with the final scorer. \\
\#46 & The main reading export run. \\
\#47 & The three render-hash lists uploaded to claude.ai by mistake together with the export (see below). \\
\#48 & The main render-hash lists pinned and checked locally. \\
\#49 & Seven replication rows flagged as reading-exposed and the counts redone. \\
\#50 & The main item files, the GEPA arm's rows, the render lists and the frozen code copied to the instance. \\
\#51 & Counts-only scripts over the stopped run of Intern-S1-mini's defect-A pass (addendum A4). \\
\#52 & The checkpoint run on the instance (A6). \\
\#53 & Counts of the test partition's \V{add.test.runs} runs (A7). \\
\#54 & The preregistered analysis (A8). \\
\#55 & The report and its tables committed as aggregates; the never-called rows' composition and the constant-letter baselines computed (A9). \\
\#56 & The constant-label baselines computed (A10). \\
\#57 & The trivial baselines of MMLU-Pro and GSM8K computed (A12). \\
\#58 & The named preprint (its PDF and LaTeX source, nothing else) copied to a folder outside the repository and given to Codex for an independent review at about \V{codex.pt}~Pacific time (\V{codex.utc}~UTC) on 2026-10-03; the review returned \V{codex.n} findings, which this version applies without new computation. \\
\#60 & The PDF of the named preprint, as built at \V{claude.pt}~Pacific time (\V{claude.utc}~UTC) on 2026-10-03, reviewed in claude.ai before the copy of \#58 was made. It was missed at the time and logged afterwards. \\
\bottomrule
\end{tabular}
\end{table}

\paragraph{The mistaken upload (\#47) and its consequence.}
One of us uploaded the three render-hash lists to the claude.ai conversation of the reading, together with the export.
A defect-A entry is the hash of a request that carries the row's reference answer, so for a row whose clean request is known, the answer can be recovered by testing each candidate against the hash.
Seven replication rows are both in the reading's sample and among the defect-A rows, and we flagged them and left them out of every analysis, whether or not anyone tried.
Nothing else in the replication or test partitions was disclosed this way.
The reader's statement, relayed by one of us and not checked independently, is that the reading did not open the lists and did not attempt a reconstruction.

\paragraph{Files derived from held-out rows.}
This paper and the committed analysis report carry aggregates that include the rows flagged \texttt{v21\_holdout} (the sensitivity analysis), so both are listed in \texttt{docs/paper\_review/HOLDOUT\_DERIVED\_FILES.md} of our repository.
They hold no row index or key.
The one place where this paper could repeat row content is the harness texts of the first two studies (Appendices~\ref{app:gepa} and~\ref{app:gepa14}): a small language model wrote them after it had seen search rows, and the copy and key checks that we ran on the original texts do not catch a worked example or a statement about one row.
Version \V{ver.one} of this paper printed the first study's texts in full and version \V{ver.two} printed those of both studies; this version prints them with row-like passages withheld under the rule at the start of each appendix, so that it holds no row content that we know of, and the third study's texts are described, with one shown, in Appendix~\ref{app:third-gepa}.
The run logs, which hold item text, stay off our repository and will not be released.
We drafted the paper with Claude Code and reviewed it in claude.ai and in Codex.
Contact \#58 is the copy given to Codex, which was the two files of the named preprint (then printing the harness texts of the first two studies in full) and not the repository, a run log or any item text, and contact \#60 is the PDF reviewed in claude.ai.
Any further upload of the draft is a new contact, logged first.

\paragraph{The second study.}
Its contacts with item text are logged in the same holdout log and in the item-exposure log: the copy of its draft preregistration that was given to a reviewer (\#62), the copy of the main item file and the frozen code to the instance (\#63) with its notes (the missing list, the checkpoint, the clone, the copy of the outputs, the second copy, each analysis and each scan), and the item-exposure log's contact~5.
No call touched a replication or test item before the second freeze, and we saw only counts and hashes of the main study's outputs until its analysis report existed.

\paragraph{The fourth study.}
Its contacts are logged in \texttt{docs/icml\_ext4/ITEM\_EXPOSURE\_LOG.md}, each by a commit before it happened.
The pilot's calls used development and search-partition rows only, and no call touched a replication or test row before the freeze.
An AI session read the \V{e4.read.n} sample prompts rendered from search rows before the first call on a sealed row (\texttt{docs/icml\_ext4/MAIN\_SAMPLES\_READING.md}, SHA256 beginning \V{e4.read.sha8}); the lead author did not, and no replication or test row was opened, rendered or read before the calls.
Before the preregistration and its supporting files were uploaded to the Open Science Framework, they were scanned against every item text of both tasks, at three levels of normalisation and by a whole-item check that must align to word boundaries, with a positive control, and the final scans found no hit.
Codex reviewed the preregistration drafts, and, after the study, the analysis report (aggregates) and the code that produced it, which led to deviation D001; the report holds no row key or text.
The paper and its committed report carry aggregates that include replication and test rows, and no Manifold question text, no CREAK claim and no user information.

\section{Serving details}
\label{app:serving}

\paragraph{Servers.}
The main study ran each model on a vLLM server started by a pinned script with these flags (the model directory, served name and chat-template file differ by model; \intern\ adds \texttt{--trust-remote-code} and a limit of zero images and videos):
\begin{verbatim}
vllm serve <model directory> --host 127.0.0.1 --port 8000
  --dtype bfloat16 --served-model-name <name>
  --chat-template <pinned file> --generation-config vllm
  --reasoning-parser qwen3
  --reasoning-config '{"reasoning_start_str": "<think>",
                       "reasoning_end_str": "</think>"}'
  --no-enable-prefix-caching --no-enable-chunked-prefill
  --max-model-len 4096
\end{verbatim}
with the environment cleared to a fixed list of variables, among them \texttt{VLLM\_BATCH\_INVARIANT=1} and \texttt{VLLM\_USE\_FLASHINFER\_SAMPLER=0} (exception X4).
Every request sets thinking on or off explicitly, carries a seed (irrelevant at zero temperature) and asks for the token IDs and the rendered prompt; the runner compares both with its own render on every call and stops on a difference, except in the defect-A pass, where the reference answer makes them differ by design.
Each stage runs its start checks before the first call and again after the last (no credential-like variable name, no connection off the machine, loopback-only listening, the model directory and the hash of its configuration, the template, the flags, and the render check against the pinned list), and a watcher samples the server's connections from launch.
The watchers flagged nothing on the vLLM servers; on each FreeToken server they flagged only the single listener of exception X2.

\paragraph{Cost.}
The GEPA search used \V{add.gepa.h} GPU-hours, the replication partition \V{add.rep.h} and the test partition \V{add.test.h}, \V{add.total.h} of the \V{add.budget.h} of the cost bound; the rerun of the determinism check on the new GPU (\V{add.gpu.h} hours) is outside it.

\paragraph{Per-run counts.}
Table~\ref{tab:serving} gives, per run and configuration, the calls, the disagreements between the server's reasoning split and the local one (logged on every call and never a stop; in the pilot they all fell after a forced end of thinking, or where the limit cut a multibyte character, and were equal once the four tags were removed), the calls in which a start-of-turn marker re-opened a turn (role breaks, all in \smol's \cfg{BUDGET-512-1500}), and mean tokens and seconds per call.
The counts include every row called, flagged rows and LB-0038 among them.

\begin{table}[H]
\caption{Serving counts of the zero-temperature grid.}
\label{tab:serving}
\centering
\scriptsize
\setlength{\tabcolsep}{2pt}
\resizebox{\ifdim\width>\linewidth\linewidth\else\width\fi}{!}{%
\begin{tabular}{@{}lllrrrrrr@{}}
\toprule
Partition & Model & Configuration & Calls & Split disagreements & Role breaks & Prompt tokens & Completion tokens & Seconds \\
\midrule
Replication & Qwen3-1.7B & THINK-64 & 3{,}603 & 0 & 0 & 480 & 56 & 1.0 \\
 &  & THINK-512 & 3{,}603 & 3 & 0 & 480 & 257 & 3.7 \\
 &  & THINK-1536 & 3{,}603 & 3 & 0 & 480 & 258 & 3.7 \\
 &  & AUTHORS-1500 & 3{,}603 & 3 & 0 & 484 & 3 & 0.2 \\
 &  & BUDGET-512-1500 & 3{,}603 & 4 & 0 & 480 & 257 & 3.6 \\
\addlinespace
 & SmolLM3 & THINK-64 & 3{,}603 & 0 & 0 & 500 & 64 & 1.8 \\
 &  & THINK-512 & 3{,}603 & 0 & 0 & 500 & 461 & 10.8 \\
\bottomrule
\end{tabular}}
\end{table}

\paragraph{Completion tokens of the second study.}
Table~\ref{tab:app-tokens} in Appendix~\ref{app:z} gives the mean completion tokens per call of Qwen3-14B's configurations and harnesses.

\section{How the related work was searched}
\label{app:related}

We searched arXiv, Semantic Scholar, OpenAlex, Hugging Face papers, GitHub and venue pages from \V{rw.win}, with topic queries only (no item content), starting from the seed papers and their references and citers, in six topic slices: harness and agent-system evaluation; optimisers and agent baselines; reasoning evaluation under token budgets; evaluation-pipeline defects; inference engines and checkpoint identity; preregistration and replication.
The agents behind the slices made \V{rw.calls} web calls.
A shared cap of \V{rw.cap} web searches ran out during the search, the arXiv and Semantic Scholar interfaces were rate-limited, and three of the six slices (reasoning evaluation, inference engines, pipeline defects) were stopped by a rate limit and resumed to write up what they had already read, without new searches.
``We found no study that'' therefore means that none was found, not that none exists, and most recent work on harness evolution is an unreviewed preprint.

Identifiers, titles, authors and dates of the cited papers were checked against arXiv.
The statements in Section~\ref{sec:related} about a paper's design were checked against its text for the papers marked as read in the matrix in our repository (\texttt{docs/icml\_ext/RELATED\_WORK\_MATRIX.md}), and otherwise rest on its abstract.
For the GitHub pull requests cited, the merge date and the release that contains the merge commit were checked through the GitHub interface.

The search was not rerun for this version.
The work added to it, the concurrent studies and the references of the third study, was checked one by one on its arXiv page (identifier, title, authors, dates), and the statements about its design rest on the abstract, except for \citet{lee2026rhsi}, \citet{braun2025acquiescence} and \citet{cho2026optimismbench}, whose text was read in the parts that the statements use.

For this version we did not rerun the search either.
We read six concurrent or earlier arXiv papers and one preprint archived on Zenodo, at their sources, for what bears on the fourth study and on the first (the text of each was searched for thinking budgets and output limits, selection on a reused set, the class balance of a selection set and comparisons of a harness with a setting), and the statements about them in Section~\ref{sec:related} are limited to what the reading notes record (\texttt{docs/icml\_ext4/RELATED\_WORK\_READ\_20261107.md}).
The papers in our reading notes do not report a class-ratio intervention or a comparison of accuracy and balanced-accuracy objectives; the supporting passage of each statement, with its section, is recorded in the notes.
\citet{tomar2026textbook} cites the first two versions of this paper; its reading of our result is broader than ours, as Section~\ref{sec:related} says.

\section{H4 in detail}
\label{app:h4}

This appendix restates, from the frozen preregistration, what Section~\ref{sec:h4design} summarises: the quantity each claim reports, how each defect is built and what it is compared with, the rows of its samples, the bound that decides whether its pair is evaluable, and the predicate and threshold of every check.

\paragraph{The reported quantity and the claim.}
For an instance $I$ (a defect and a model) and its clean counterpart $K$, the reported quantity is the accuracy of a fixed configuration, the pipeline's start $s$, on the letter pool: $a_S=\mathrm{acc}_S(s)$ on the selection sample $S$ and $a_R=\mathrm{acc}_R(s)$ on the check sample $R$.
The start is \cfg{THINK-1536} (for E, the case study's CFG-THINK-1536-ORIG), and \cfg{THINK-64} in the instance of D.
No claim selects anything.
Then $D_S=a_S(I)-a_S(K)$, $D_R=a_R(I)-a_R(K)$ and $\Gamma=D_R-D_S$.
The claim \emph{holds} if the adjusted interval of $D_S$ lies wholly outside $\pm m$, that of $D_R$ lies wholly outside $\pm m$ on the side of $D_S$'s point estimate, and that of $\Gamma$ lies within $\pm m$.
It \emph{fails} if any of the three lies wholly on the other side of its boundary ($D_S$'s wholly within $\pm m$; $D_R$'s wholly within $\pm m$, or wholly outside it on the opposite side; $\Gamma$'s wholly outside $\pm m$), and it is \emph{inconclusive} otherwise.
Here $m=\V{c.minm}$ and the intervals have the adjusted level of the family of \V{c.fam4}; an interval that fails decides the claim even if another is still open, and the error control covers ``holds'' only.

\paragraph{Defect D and the selection.}
For D the reported quantity is the gain $g=\mathrm{acc}(c^*)-\mathrm{acc}(s)$ of the selected configuration $c^*$ over the start, on $S$ and on $R$.
The selection picks $c^*$ as the most accurate member on $S$ of the search space (\cfg{THINK-64}, \cfg{THINK-512}, \cfg{THINK-1536}, \cfg{RESCUE-1500} and \cfg{BUDGET-512-1500}; ties go to the earlier member), and \cfg{AUTHORS-1500} lies outside it, so that it can act as the simple comparator that a search never considered.
Because $c^*$ does not depend on the start, the instance of D and its counterpart select the same $c^*$ in every resample, the selection cancels, and $D_S=\mathrm{acc}_S(\text{\cfg{THINK-1536}})-\mathrm{acc}_S(\text{\cfg{THINK-64}})$, with the same difference on $R$ for $D_R$.
The selection itself, the gains it reports and every interval that follows a selection are descriptive and enter no claim.

\paragraph{Defects, comparators and samples.}
Table~\ref{tab:h4-defects} gives for each defect how it is built, the samples $S$ and $R$, the clean counterpart it is compared with and how the two are resampled, and the bound of its pair.
All samples leave out the flagged rows (Section~\ref{sec:design}).
Where an instance and its counterpart share their rows, one resample of the units serves both, so $D_S$ and $D_R$ are paired; $\Gamma$ pairs the $b$-th resample of $S$ with the $b$-th of $R$, drawn with their own seeds.

Table~\ref{tab:h4} gives the point estimates of every claim.

\begin{table*}[t]
\caption{H4: the \V{c.fam4} (defect, model) pairs of the first study and the \V{z.h4.evaluable} of the second (Qwen3-14B), with the letters of Section~\ref{sec:h4design}; a star marks a claim whose quantity is fixed by an identity of the design; its verdict says little beyond that mechanism, although it still depends on the intervals. The tied output layer (E) exists for \qwen\ only, the one model with two served variants. The bound is the largest distortion the mechanism can cause on $S$, computed without accuracy; a pair is evaluable only if it exceeds \V{c.minm}. $D_S$, $D_R$ and $\Gamma=D_R-D_S$ are point estimates of the distortion of the start's accuracy (for the broken start, of the gain over the start, which equals the accuracy of \cfg{THINK-1536} minus that of \cfg{THINK-64}); intervals are in Figure~\ref{fig:h4} and Appendix~\ref{app:tables}. A dagger marks the claim of the skewed split in \smol, which fails by the general rule of the preregistration (Section~\ref{sec:deviations}).}
\label{tab:h4}
\centering
\footnotesize
\setlength{\tabcolsep}{3pt}
\resizebox{\ifdim\width>\linewidth\linewidth\else\width\fi}{!}{%
\begin{tabular}{@{}llccccl@{}}
\toprule
Defect & Model & Bound & $D_S$ & $D_R$ & $\Gamma$ & Claim \\
\midrule
A reference answer as assistant turn & Qwen3-1.7B& 1.000 & \ensuremath{-0.052} & \ensuremath{-0.030} & \ensuremath{+0.022} & inconclusive \\
 & SmolLM3& 1.000 & \ensuremath{+0.097} & \ensuremath{+0.119} & \ensuremath{+0.022} & inconclusive \\
 & Intern-S1& 1.000 & \ensuremath{+0.142} & \ensuremath{+0.164} & \ensuremath{+0.022} & inconclusive \\
 & Qwen3-14B & 1.000 & \ensuremath{+0.430} & \ensuremath{+0.404} & \ensuremath{-0.026} & inconclusive \\
\addlinespace
B1 gold fallback & Qwen3-1.7B& 0.205 & \ensuremath{-0.145} & \ensuremath{-0.148} & \ensuremath{-0.003} & holds \\
 & SmolLM3& 0.083 & \ensuremath{+0.001} & \ensuremath{+0.007} & \ensuremath{+0.006} & fails \\
 & Intern-S1& 0.057 & \ensuremath{-0.012} & \ensuremath{-0.019} & \ensuremath{-0.007} & fails \\
\bottomrule
\end{tabular}}
\end{table*}

\paragraph{The pairs one by one.}
In the first study, under the broken start the reported gain over the start is inflated by $D_S=\V{h4.d.q.ds}$, $\V{h4.d.s.ds}$ and $\V{h4.d.i.ds}$ (\qwen, \smol, \intern) and by almost the same amount on $R$ ($\Gamma=\V{h4.d.q.g}$, $\V{h4.d.s.g}$, $\V{h4.d.i.g}$), and scoring unparsed answers as~A inflates the start's accuracy by $\V{h4.b2.s.ds}$ and $\V{h4.b2.i.ds}$, exactly the share the start leaves unparsed with gold~A.
The claim fails in four pairs: the gold fallback in \smol\ and \intern, whose starts parse too few of the rows whose gold the fallback changes (bounds \V{h4.b1.s.bound} and \V{h4.b1.i.bound}, distortions $\V{h4.b1.s.ds}$ and $\V{h4.b1.i.ds}$); the skewed split in \smol; and the tied output layer in \qwen, where $D_S=\V{h4.e.q.ds}$ and $D_R=\V{h4.e.q.dr}$ both lie inside the band.
Five pairs are inconclusive, the reference-answer defect in all three models and the skewed split in \qwen\ and \intern; for the reference-answer defect in \intern\ both $D_S=\V{h4.a.i.ds}$ and $D_R=\V{h4.a.i.dr}$ clear the band, but the interval of $\Gamma$, \CI{\V{h4.a.i.g.lo}}{\V{h4.a.i.g.hi}}, is wide because each half has only \V{d.a.comp} components.
The unparsed-as-A scoring in \qwen\ is not evaluable, because its start leaves no row unparsed (bound \V{h4.b2.q.bound}).
In the second study the reference-answer defect has $D_S=\V{z.h4.a.ds}$ and $D_R=\V{z.h4.a.dr}$, both clear of the band, but $\Gamma$, \CI{\V{z.h4.a.g.lo}}{\V{z.h4.a.g.hi}}, is wide; the skewed split has $D_S=\V{z.h4.c.ds}$ and $D_R=\V{z.h4.c.dr}$ with all three components open.

\paragraph{The skewed split is resampled in two pieces.}
The skewed split's instance and its clean counterpart use different rows but the same participant components: the instance's halves are dealt from the unflagged rows of both partitions, and the counterpart's samples are the replication and test rows.
The frozen analysis resamples the two independently, each with its own seed, so the covariance between them is ignored, and the intervals of $D_S$, $D_R$ and $\Gamma$ for this defect may be too wide (if the covariance is positive) or too narrow (if it is negative).
The four claims for C, one failing (\smol) and three inconclusive (\qwen, \intern\ and Qwen3-14B), are the ones that could depend on it; for every other defect one resample serves the instance and its counterpart.

\begin{table}[H]
\caption{The six defects of H4: construction, samples, clean counterpart and evaluability bound. The bound is the largest distortion the defect's mechanism can cause on $S$, computed without accuracy.}
\label{tab:h4-defects}
\centering
\footnotesize
\setlength{\tabcolsep}{3pt}
\begin{tabular}{@{}>{\raggedright\arraybackslash}p{0.035\linewidth}>{\raggedright\arraybackslash}p{0.25\linewidth}>{\raggedright\arraybackslash}p{0.26\linewidth}>{\raggedright\arraybackslash}p{0.195\linewidth}>{\raggedright\arraybackslash}p{0.195\linewidth}@{}}
\toprule
 & Construction & Samples $S$ and $R$ & Clean counterpart and resampling & Evaluability bound \\
\midrule
A & The row's reference answer is sent as a final assistant turn, as the case study's caller did: new calls, built by a separate module. & \V{d.a.rows} replication rows drawn by whole component (seed \V{h4.a.seed}) in proportion to the task counts (\V{h4.a.t30}, \V{h4.a.t34}, \V{h4.a.t38} and \V{h4.a.t42} rows of LB-0030, LB-0034, LB-0038 and LB-0042), halved by component into $S$ and $R$ (\V{h4.a.hcomp} components and \V{h4.a.hrows} rows each; after the flags \V{h4.a.rows.s} and \V{h4.a.rows.r} rows in \V{d.a.comp} components each). Both halves are called in one pass, so $R$ is a disjoint half, not a later sample. & The grid's own outputs on the same rows, in the same halves; one resample serves both. & One: the leaked reference can change any row. \\
\addlinespace
B1 & Offline: the clean outputs are rescored with LB-0042's gold taken from the case study's historical fallback, which with this release's labels differs from the derived gold. Unparsed answers stay wrong; other tasks keep the release's gold. & $S$: the replication partition's letter rows; $R$: the test partition's, called after the checkpoint. The fallback changes the gold of \V{h4.b1.s.n} of \V{h4.b1.s.of} LB-0042 rows of $S$ and \V{h4.b1.r.n} of \V{h4.b1.r.of} of $R$. & The main clean instance on the same rows; one resample serves both. & The share of $S$'s letter rows that the start parses and whose gold B1 changes (exact; B1 changes the gold of \V{h4.b1.share} of them). \\
\addlinespace
B2 & Offline: the clean outputs are rescored by a scorer that never fails closed: an unparsed or conflicting letter answer is scored as the letter A, on every letter task, with the release's gold. & As B1. & As B1. & The share of $S$'s letter rows that the start leaves unparsed or conflicting and whose gold is A (exact; it equals B2's distortion of the start's accuracy). \\
\addlinespace
C & The case study's skewed split, re-implemented. The pooled LongevityBench rows of both partitions that are not flagged (\V{h4.c.pool} of \V{h4.c.all}) are reassigned: components (from all six NHANES tasks, chaining into one giant component that goes to the evolution part) are sorted by size, largest first, and each goes to the part furthest below its target share (evolution \V{h4.c.t1}, validation \V{h4.c.t2}, holdout \V{h4.c.t3}). & $S$ and $R$ are the two halves of the validation part: its components are sorted by identifier, shuffled (seed \V{h4.c.seed}) and dealt alternately. Their rows come from both partitions, so they are not the replication and test rows of the other pairs, and C, computed after the test calls, is not temporally prospective. & The main clean instance, whose $S$ and $R$ are the study's replication and test rows. The rows differ, so the two are resampled independently, each with its own seed. & None: C is evaluable by rule, because its mechanism changes the sample and has no row-level bound. \\
\addlinespace
D & Offline: the start is \cfg{THINK-64} in place of \cfg{THINK-1536}; the reported quantity is the gain over the start. & $S$: the replication partition's letter rows; $R$: the test partition's. & The main clean instance; one resample serves both. & The share of $S$'s letter rows on which the parsed answers of the two starts differ, an unparsed or conflicting answer counting as one value. \\
\addlinespace
E & The tied output layer is served in place of the untied one (\qwen\ only): the case study's six configurations on FreeToken, for both variants, with CFG-THINK-1536-ORIG as the start. The confirmatory instance serves the tied variant on $S$ and on $R$. & $S$: the replication rows; $R$: the test rows. & The untied variant on both samples; one resample serves both. A second instance (tied on $S$, untied on $R$) feeds only the descriptive detection matrix. & The share of $S$'s letter rows on which the tied and untied variants' parsed answers of the start differ. \\
\bottomrule
\end{tabular}
\end{table}

\paragraph{Evaluability.}
A pair is evaluable if its bound exceeds $m$ (a bound of exactly $m$ does not).
Otherwise it is reported as not evaluable, with its bound, makes no claim and is not counted as failing, and the family stays at \V{c.fam4}.
The bounds were computed from parse, gold-change and answer-difference indicators on the replication data at the checkpoint, before the first test call.
For B1 the start must parse the same rows in the instance and its counterpart, and for B2 the gold must be the release's in both; the analysis stops if not.

\paragraph{The checks and their rules.}
Each check is evaluated on the records of an instance, and ``fires'' means that one of its rules holds.
The rules and thresholds were fixed before the study; their outcomes are descriptive, and what is empirical is how often they fire on clean instances (Section~\ref{sec:results}).
The rules of PR and every interval that follows a selection on $S$ are descriptive labels, not tests: the selection makes $\mathrm{acc}_S(c^*)$ and the gain optimistic.
Tables~\ref{tab:h4-checks-a} and~\ref{tab:h4-checks-b} give each rule's predicate and threshold.

\begin{table}[H]
\caption{The rules of PR, TR and TB. $c^*$ is the selected configuration, $s$ the start, $g_R$ the gain on $R$.}
\label{tab:h4-checks-a}
\centering
\footnotesize
\setlength{\tabcolsep}{3pt}
\begin{tabular}{@{}>{\raggedright\arraybackslash}p{0.17\linewidth}>{\raggedright\arraybackslash}p{0.79\linewidth}@{}}
\toprule
Rule & Fires when \\
\midrule
PR, gain rule & $c^*\neq s$ and the lower bound of the \V{c.ci} percentile interval of $\mathrm{acc}_R(c^*)-\mathrm{acc}_R(s)$ is at or below zero: the reported gain does not reproduce on $R$. \\
PR, level rule & The \V{c.ci} percentile interval of $\mathrm{acc}_R(c^*)-\mathrm{acc}_S(c^*)$ excludes zero, whether or not $c^*=s$; the $b$-th resample of $S$ is paired with the $b$-th of $R$, each drawn with its own seed. \\
TR & On any call of the instance the server's prompt-token count differs from the local count of the clean render (the corrected caller's render of the row's messages before the final assistant turn), or, on vLLM, its prompt token IDs differ from the clean render's. For FreeToken only the count is compared. \\
TB1 & A strong trivial baseline (the oracle constant, the frequency of the most frequent gold letter in the sample itself) exceeds \V{h4.thr.tb1} on a task in $S$ or $R$. \\
TB2 & For some task, the trivial constant chosen on $S$ (the most frequent gold letter there, ties to the smallest letter) scores on $R$ more than \V{h4.thr.tb2} away from its score on $S$. \\
TB3 & On LB-0038, some configuration's exact-match rate exceeds three times the best constant age's, in $S$ or $R$. \\
TB4 & On a letter task in $S$ or $R$, a configuration has at least \V{h4.thr.tb4} unparsed or conflicting calls, and every one of them is scored correct exactly when the gold is one and the same constant letter: the scorer treats them as a constant predictor. \\
\bottomrule
\end{tabular}
\end{table}

\begin{table}[H]
\caption{The rules of SCF, IR and SMI. $g_S$ is the reported gain on $S$ and $\Delta_V$ its visibility part (Section~\ref{sec:design}, with the start as P and $c^*$ as R, point estimate on $S$).}
\label{tab:h4-checks-b}
\centering
\footnotesize
\setlength{\tabcolsep}{3pt}
\begin{tabular}{@{}>{\raggedright\arraybackslash}p{0.17\linewidth}>{\raggedright\arraybackslash}p{0.79\linewidth}@{}}
\toprule
Rule & Fires when \\
\midrule
SCF1 & On $S$ the start's parse rate is below \V{h4.thr.scf1.p} or its truncation rate above \V{h4.thr.scf1.t}. \\
SCF2 & $c^*\neq s$ and $\mathrm{acc}_S(c^*)$ is less than \V{h4.thr.scf2} above the best of three panel members on $S$: chance, the strong trivial baseline and \cfg{AUTHORS-1500} (in the instances of E, CFG-THINKING-FALSE-64-ORIG). For the letter pool, chance and the baseline are the row-weighted means of the tasks' values. \\
SCF3 & $c^*\neq s$, $g_S>0$ and $\Delta_V$ is at least \V{h4.thr.scf3} of $g_S$. \\
IR1 & Any request of the instance has an assistant turn, or a last message that is not a user turn. \\
IR2 & The gold screen hits on any request: the row's follow-up text, the LB-0038 age as a standalone integer, or another reference answer of more than one character appears in it. \\
IR3 & On any row the scorer's gold is not among the golds derived from the question and follow-up (apart from the LB-0030 band-endpoint exception). \\
IR4 & For $S$ or $R$ against the release: a task's row share differs by more than \V{h4.thr.ir4.pts} points from its release share; the LB-0042 deceased share or an LB-0030 band share differs by more than \V{h4.thr.ir4.pts} points; or the standardised mean difference of the actual age exceeds \V{h4.thr.ir4.smd}. \\
IR5 & A binary task has a class below \V{h4.thr.ir5} of its rows in $S$ or $R$, or an LB-0030 band is absent from $S$ or $R$. \\
SMI (E only) & A served directory's configuration declares tied word embeddings while its checkpoint's output matrix differs from the embedding matrix; or, on the same rows and request bytes, the two variants' parsed answers differ on more than \V{h4.thr.smi} of rows in some configuration. The second rule applies only where two variants of one checkpoint are served, which is \qwen. \\
\bottomrule
\end{tabular}
\end{table}

\paragraph{The intervals of $D_S$, $D_R$ and $\Gamma$.}
\begin{figure*}[t]
\centering
\includegraphics[width=\textwidth]{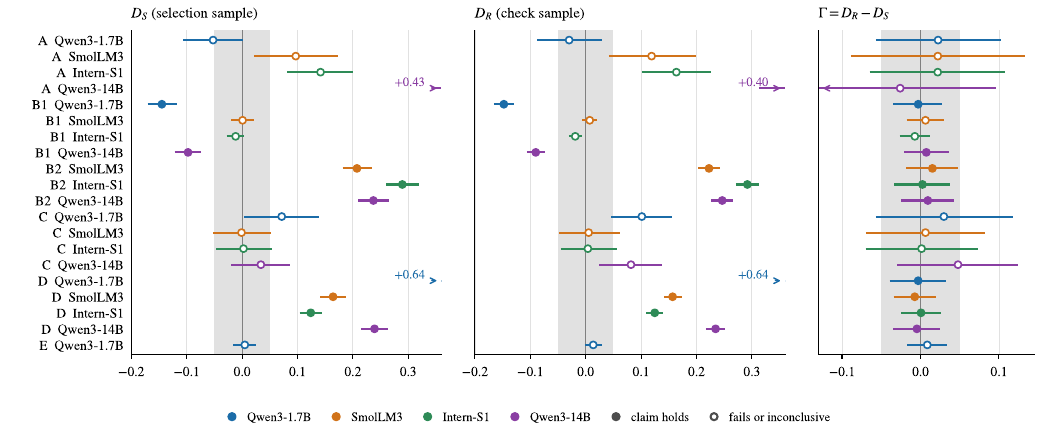}
\caption{H4: the defect's distortion on the selection sample ($D_S$), on the check sample ($D_R$) and their difference ($\Gamma$), with adjusted intervals (family of \V{c.fam4}) and the band $\pm m$ (shaded). Filled markers are claims that hold; arrows mark intervals cut at the axis limit. The unparsed-as-A scoring (B2) in \qwen\ is not evaluable and is not drawn.}
\label{fig:h4}
\end{figure*}

\section{The GEPA arm in detail}
\label{app:gepa}

\paragraph{Texts printed in part.}
Versions \V{ver.one} and \V{ver.two} of this paper printed these texts in full; this version withholds row-like passages under the rule below.
A small language model wrote each text after it had seen search rows and the feedback on its candidates, and such a text can repeat what a row shows; the copy and key checks reported in this appendix do not catch a worked example, a statement about one row or a short measurement value.
The rule withholds every worked example; every participant profile and every measurement value, which here means every number with its comparison sign, range or unit, including each threshold and cut-off, and every length of time; every statement about a specific example, option or row, such as a reference to a numbered example, to a participant by letter, or to an option as invalid; and every named medication, dose or duration.
A passage was withheld when we were in doubt.
Each withheld passage is replaced by \verb|[withheld, N characters]|, where N is the passage's length in characters of the original text, and everything else is as selected.
The SHA256 printed with each text identifies the original, which is not released; the checks in this appendix that name a text were run on the originals, before any passage was withheld.

\paragraph{What GEPA searched.}
GEPA version \V{gp.ver} \citep{agrawal2025gepa} edited two text components and nothing else: the system prompt, which replaces the row's system message (seed: the rows' own system message; limit \V{gp.limit.sys} characters), and an instruction appended to the user turn after a blank line (seed: empty; limit \V{gp.limit.usr} characters).
The request parameters (thinking on or off, the token limit, the temperature) belong to the start and were not searched.
The clean start is the request of \cfg{AUTHORS-1500} (thinking off, \V{k.ans} tokens, zero temperature) and the broken start that of \cfg{THINK-64}; the seed harnesses are the grid's own runs of those two configurations.
The arm therefore searched prompt text only, and its negative result (no selected harness more accurate than the thinking-off baseline) is about that space: it says nothing about searching the thinking switch, the token limits or retry rules.
The task model is \qwen\ (untied) on the batch-invariant server.
The proposer is \intern\ on the standard server at concurrency one, with thinking on, its card's sampling values (temperature \V{smp.i.temp}, top-$p$ \V{smp.i.p}, top-$k$ \V{smp.i.k}), at most \V{gp.prop.tok} tokens and a seed per request.
GEPA's settings are its defaults except: a reflection minibatch of \V{gp.mb} rows; Pareto selection of candidates over instances; a round-robin choice of the component to edit; acceptance only on strict improvement; perfect-score minibatches skipped; no merging; no evaluation cache.

\paragraph{Rows and allocation.}
The arm uses \V{k.gepa.rows} rows of the search partition's three letter tasks (\V{gp.rows.lb30} of LB-0030, \V{gp.rows.lb34} of LB-0034 and \V{gp.rows.lb42} of LB-0042), drawn with seed \V{gp.seed} from search rows that are neither pilot rows nor rows the case study had held out; LB-0038 is left out because its score is an age error.
They form a training set of \V{gp.split} rows (\V{gp.sp.a}, \V{gp.sp.b} and \V{gp.sp.a} by task) and a validation set of \V{gp.split} rows (the same split); the lists are the same in the four runs.
A candidate scores one on a row when the primary scorer's parsed answer equals the gold and zero otherwise, an unparsed answer included.
The proposer sees a row's user message, the task model's output, reasoning and answer, and feedback (correct, incorrect, unparsed, or cut at the token limit); it never sees the gold or a replication or test row.
The four runs are the clean and the broken start, each with GEPA seeds zero and one.

\paragraph{Checks on a candidate, and the copy rule.}
Before any call a candidate is checked, and one that fails is recorded and scored zero without a call: each component is text within its limit; neither holds a chat-template control string; neither holds a run of \V{gp.copy.run} or more consecutive characters of a row's user message that occurs in fewer than \V{gp.copy.rows} of the arm's rows (the copy rule: wording shared by many rows may be copied, a row's own text may not); every render plus the token limit fits the \V{gp.ctx}-token context.
The soft switches \verb|/think| and \verb|/no_think| were allowed and each candidate records whether it holds one; none of the selected harnesses does.
Run on the original texts, a separate script re-checked the four selected texts against the arm's rows and found no violation of the copy rule and no identifier or number sequence of an item.

\paragraph{Budget, selection and ties.}
The budget is \V{k.metric} metric calls per run (\V{gp.n.cand} candidates on \V{k.gepa.rows} rows).
It is soft: GEPA checks it between iterations, so a run may end slightly above it, and the four runs used \V{gepa.gc1.calls} to \V{gepa.gb0.calls} calls.
GEPA spends \V{gp.six} metric calls per proposal, \V{gp.split} on the seed's validation evaluation and \V{gp.split} on each accepted candidate.
The selected harness of a run is its candidate with the highest mean validation score, ties going to the lowest candidate index; if no proposal was accepted it is the seed.
The selected texts were recorded, with their SHA256, before the harness was run on any replication row.
Each was then run on every letter row of the replication and test partitions, on the main server with its start's request parameters.

\paragraph{The four selected harnesses.}
The texts below are as selected, except for the passages withheld under the rule at the start of this appendix, and wrapped for display; a character outside ASCII is written \verb|[U+XXXX]| (\verb|[U+2265]| is the sign for greater than or equal, \verb|[U+03BC]| the Greek letter mu, \verb|[U+00B2]| a superscript two).
Each candidate's SHA256 is the one recorded before any replication call in our repository and identifies the original; the wrapping and the withheld passages mean that it cannot be recomputed from this page.
A small language model generated the texts; we did not check their medical statements, and they must not be used as clinical guidance.
Validation scores and accuracies are in Appendix~\ref{app:tables}.

\FloatBarrier
\begin{minipage}{\linewidth}
\subsection*{GEPA-CLEAN-s0 (clean start, GEPA seed zero)}
Mean validation score \V{gepa.gc0.val}; the candidate has index \V{gepa.gc0.sel} (counted from zero) among the run's \V{gepa.gc0.of} candidates and SHA256
\begin{center}\scriptsize\texttt{ab48481902a519b5c045cd37f61382899ea0e1bce05b7a8345d2873f77b1bb16}\end{center}
The run kept \V{gepa.gc0.kept} candidates, the seed and the others that were accepted on their minibatches, but none of them scored above the seed on the validation rows, so the seed, the thinking-off baseline itself, was selected (ties go to the lowest index; Section~\ref{sec:h2}).
\paragraph{System prompt.}
\begin{small}
\begin{verbatim}
You are a biomedical AI specialized in aging biology, trained on genomic, proteomic, and
clinical data.
\end{verbatim}
\end{small}
\paragraph{Appended instruction.} Empty.
\end{minipage}
\medskip

\FloatBarrier
\harnessneed

\subsection*{GEPA-CLEAN-s1 (clean start, GEPA seed one)}
Mean validation score \V{gepa.gc1.val}; the candidate has index \V{gepa.gc1.sel} (counted from zero) among the run's \V{gepa.gc1.of} candidates and SHA256
\begin{center}\scriptsize\texttt{33819b05a9cd5711ed2b502c48f84a626fd26d8d4286dcadc420cde065e6b7f9}\end{center}
\paragraph{System prompt.}
\begin{small}
\begin{verbatim}
You are a biomedical AI specialized in aging biology, trained on genomic, proteomic, and
clinical data. Your task involves assessing biological age or age-related phenotypes
using clinical biomarkers. Focus exclusively on the biomarkers listed below, which are
validated predictors of aging in human populations.

**Key Biomarkers for Aging Assessment (prioritize these in reasoning):**
1. **Kidney Function:**
   - Cystatin C (mg/L): Increase [withheld, 9 characters] correlates with
[withheld, 9 characters].
   - Creatinine ([U+03BC]mol/L): [withheld, 10 characters] in females suggests age
[withheld, 3 characters].
2. **Liver Metabolism:**
   - Albumin (g/dL): [withheld, 9 characters] indicates malnutrition or chronic disease
linked to advanced age.
   - Bilirubin (mg/dL): Direct increase ([withheld, 4 characters]) reflects impaired
hepatic clearance with age.
3. **Metabolic Health:**
   - LDL Cholesterol (mg/dL): [withheld, 10 characters] in females
[withheld, 3 characters] increases cardiovascular risk.
   - Triglycerides (mg/dL): [withheld, 10 characters] in females
[withheld, 3 characters] indicates insulin resistance.
4. **Inflammation & Stress:**
   - C-reactive protein (mg/dL): [withheld, 10 characters] reflects chronic inflammation
linked to aging.
   - Cadmium (nmol/L): Accumulation ([withheld, 9 characters]) correlates with tissue
damage in older adults.
5. **Cellular & Tissue Health:**
   - Mean Cell Volume (fL): Decline ([withheld, 6 characters] in females
[withheld, 3 characters] indicates deficiency; [withheld, 6 characters] in older adults
suggests anemia of aging).
   - Red Cell Distribution Width (%): [withheld, 6 characters] in females
[withheld, 3 characters] reflects oxidative stress.

**Exceptions & Caveats:**
- Ignore BMI unless explicitly asked to correlate with metabolic syndrome.
- Do not assume chronological age unless explicitly provided [withheld, 79 characters].
- Flag conflicting biomarkers (e.g., high LDL but low triglycerides may require deeper
analysis).

**Generalizable Strategy:**
1. Compare biomarker trajectories across cohorts (e.g., Cystatin C
[withheld, 4 characters] indicates accelerated aging).
2. Prioritize systemic markers over individual parameters (e.g., integrate LDL, CRP, and
Cadmium for metabolic aging profiles).
3. Use thresholds validated in longitudinal cohort studies (e.g., Bone Alkaline
Phosphatase [withheld, 8 characters] in females [withheld, 3 characters]).

For comparative tasks (e.g., "Who is older?"), rank participants by cumulative biomarker
deviation from age-predictive ranges. For age-group classification, use [U+2265]2
biomarkers exceeding thresholds for specific ranges (e.g., [withheld, 49 characters]).
\end{verbatim}
\end{small}
\paragraph{Appended instruction.}
\begin{small}
\begin{verbatim}
Analyze medical data to answer yes/no survival questions or identify older participant
based on blood tests. Input includes demographics, biometrics, blood panel results, and
clinical history. Prioritize evidence-based medical knowledge: For survival prediction,
consider BMI (optimal range [withheld, 9 characters]), albumin ([withheld, 9 characters]
indicates better prognosis), creatinine ([withheld, 10 characters] for women,
[withheld, 4 characters] for men suggests preserved kidney function), LDL
([withheld, 10 characters] ideal), and absence of diabetes/hypertension risk factors.
For age comparison, note cadmium levels (higher in older adults), red cell distribution
width (increases with age), and segmented neutrophils (typically decrease after age
[withheld, 2 characters]). Explicitly flag when data is insufficient to determine
answer. Use structured reasoning: first state key markers supporting conclusion, then
cross-reference contradictions. Never infer survival based solely on absence of provided
data.
\end{verbatim}
\end{small}

\FloatBarrier
\begin{minipage}{\linewidth}
\subsection*{GEPA-BROKEN-s0 (broken start, GEPA seed zero)}
Mean validation score \V{gepa.gb0.val}; the candidate has index \V{gepa.gb0.sel} (counted from zero) among the run's \V{gepa.gb0.of} candidates and SHA256
\begin{center}\scriptsize\texttt{9f38d54e2cc346c8b0c20fcaabfd981906af2ce409a2d04250195c88658482ab}\end{center}
\paragraph{System prompt.}
\begin{small}
\begin{verbatim}
You are a biomedical AI specialized in aging biology, trained on genomic, proteomic, and
clinical data.
\end{verbatim}
\end{small}
\paragraph{Appended instruction.}
\begin{small}
\begin{verbatim}
Analyze clinical data to predict survival probability or age comparisons. Focus on key
biomarkers linked to mortality or aging (e.g., elevated HbA1c [withheld, 5 characters]
for diabetes, RDW [withheld, 6 characters] as mortality predictor, elevated creatinine
[withheld, 11 characters] for chronic kidney disease). Apply validated thresholds for
diabetes (HbA1c [withheld, 5 characters]), cardiovascular risk (LDL
[withheld, 9 characters], HDL [withheld, 9 characters] for males), and organ dysfunction
(eGFR [withheld, 17 characters]). Prioritize metrics like glycohemoglobin, red cell
distribution width, and bone alkaline phosphatase. For age comparisons, use biomarkers
like cadmium accumulation (higher in older adults, [withheld, 11 characters]) and
vitamin B12 levels ([withheld, 11 characters] for deficiency risk). Limit reasoning to
critical data; format answers as A/B per input options. Include domain-specific
knowledge: high RDW correlates with mortality ([withheld, 22 characters]), elevated uric
acid [withheld, 8 characters] predicts cardiovascular events, and low HDL
[withheld, 9 characters] increases dementia risk. Avoid speculative connections beyond
well-established clinical correlations. Respond fully within token limits by
prioritizing high-impact metrics and omitting tangential details.
\end{verbatim}
\end{small}
\end{minipage}
\medskip

\FloatBarrier
\harnessneed

\subsection*{GEPA-BROKEN-s1 (broken start, GEPA seed one)}
Mean validation score \V{gepa.gb1.val}; the candidate has index \V{gepa.gb1.sel} (counted from zero) among the run's \V{gepa.gb1.of} candidates and SHA256
\begin{center}\scriptsize\texttt{ddf8886b2b3edced069fcfa5c3d9baad95643056d772d2ebeb3580770ca017ed}\end{center}
\paragraph{System prompt.}
\begin{small}
\begin{verbatim}
You are a biomedical AI specialized in aging biology, trained on genomic, proteomic, and
clinical data.
\end{verbatim}
\end{small}
\paragraph{Appended instruction.}
\begin{small}
\begin{verbatim}
### Instruction

**Task Overview:**
Analyze provided medical data to answer specific clinical questions. Prioritize
identifying key biomarkers and demographic factors relevant to the question. Answer
strictly using the format "A." or "B." Only, matching the provided options.

---

**Input Format:**
All inputs will include structured medical data:
1. **Demographics:** Age, ethnicity, education level, sex.
2. **Biometrics:** BMI, height, weight, waist circumference.
3. **Blood Panel:** Standard lab results (e.g., glucose, cholesterol, proteins,
electrolytes).
4. **Questionnaire:** Medical history, family history, surgical history.

Data units are standardized (e.g., BMI in kg/m[U+00B2], glucose in mmol/L). Missing
values will not be present.

---

**Domain-Specific Knowledge to Include:**
1. **Survival Prediction:**
   - BMI [withheld, 3 characters] indicates obesity; [withheld, 3 characters] indicates
overweight.
   - LDL [withheld, 11 characters] or triglycerides [withheld, 11 characters] suggest
hyperlipidemia.
   - Glycohemoglobin [withheld, 5 characters] indicates prediabetes/diabetes.
   - Chronic inflammation (C-reactive protein [withheld, 7 characters]) correlates with
mortality risk.
   - Family history of cardiovascular diseases multiplies risk factors.

2. **Age Estimation via Blood Tests:**
   - Red blood cell count and hemoglobin decline with age; compare to age-stratified
norms.
   - Elevated Cystatin C ([withheld, 9 characters]) correlates with aging-associated
renal dysfunction.
   - Decreased Vitamin D ([withheld, 10 characters]) is common in older adults.
   - Glycohemoglobin trends reflect long-term metabolic stability.

---

**Generalizable Strategies:**
1. **Survival Questions:**
   - Flag obesity (BMI [withheld, 3 characters]), hyperlipidemia, diabetes, or chronic
inflammation as high-risk factors.
   - Compare triglycerides/HDL ratio ([withheld, 4 characters]) for cardiovascular risk.
   - Note female reproductive history (e.g., hysterectomy impacts hormone status).

2. **Age Questions:**
   - Prioritize red blood cell indices (mean corpuscular volume) and hemoglobin trends.
   - Analyze biomarkers like Cystatin C and Vitamin D for age-related decline.
   - Contrast metabolic profiles (e.g., glucose, cholesterol) against age-specific
norms.

---

**Output Requirements:**
- Always answer with **A.** or **B.** only.
- No explanations or partial answers.
- Ensure answers align with calculated risk scores and biomarker thresholds.

**Example Alignment:**
[withheld, 245 characters]

**Critical Note:**
Never infer external knowledge (e.g., specific study results). Use only the data and
thresholds provided in the instructions.
\end{verbatim}
\end{small}

\section{Planned power}
\label{app:power}

This paper quotes five planned-power figures (Table~\ref{tab:power}), all from the preregistration and none recomputed after the outputs.
They are rough: made before the study from the pilot's parse shares and from assumed effects, with a normal approximation.
``Expected to be inconclusive'' in Appendix~\ref{app:expected} is an informal expectation read off these figures, not a prediction with a decision rule, and the observed verdicts do not test it.

\paragraph{H2, \intern\ on GSM8K against \cfg{RESCUE-1500}.}
The assumed truth is a $\theta$ of zero: the thinking-off configuration exactly as accurate as the rescue.
The paired difference of correctness takes the values minus one, zero and one, so its variance is at most $d$, the share of rows on which the two configurations' parsed answers differ (an unparsed answer counts as one value); on the pilot's GSM8K items $d=\V{pw.h2.d}$.
The unit is the item and the test partition has $n=\V{pw.h2.n}$ GSM8K items, so the standard error of the estimated difference is at most $\mathrm{SE}=\sqrt{d/n}$.
The lower bound of the interval for $\theta$ clears $-\delta$, with $\delta=\V{c.margin}$, when the estimate exceeds $-\delta+z\,\mathrm{SE}$, where $z$ is the normal quantile of the interval's level: \V{pw.h2.z} for the Bonferroni family of \V{c.fam2} (level \V{c.lvl2}\%) and \V{pw.z.unadj} for an unadjusted interval.
The power at a true $\theta$ of zero is then $\Phi(\delta/\mathrm{SE}-z)$, with $\Phi$ the standard normal distribution function: \V{x.power} adjusted and \V{pw.h2.unadj} unadjusted.

\paragraph{H1, \smol\ and \intern\ on GSM8K, the statistic $T$.}
The inputs are the pilot's joint parse shares of \cfg{THINK-512} (P) and \cfg{AUTHORS-1500} (R) on GSM8K, $v$ (only R parses), $b$ (both parse) and $l$ (only P parses), and $d_B$, the share of the rows both parse on which the parsed answers differ.
No gold or accuracy of the pilot enters.
Two assumed scenarios supply the accuracies, and they are assumptions, not pilot results.
In S1, R is right on half of the rows only it parses and P on half of those only it parses, and neither is better on the rows both parse.
In S2 both are right on \V{pw.s2.acc} of the rows both parse and R is better than P by \V{pw.s2.gap} on them, capped at $d_B$.
The power of $T>0$ for the lower bound of the adjusted interval is a normal approximation on the same $n=\V{pw.h2.n}$ test items (one unit per item) at the level of H1's family of \V{c.fam1} ($z=\V{pw.z9}$); the script, \texttt{prereg\_power\_h1\_v3.py}, is pinned in the preregistration.

\begin{table}[H]
\caption{The planned-power figures quoted in this paper. All are for the test partition, with a normal approximation. S1 and S2 are the two assumed scenarios of H1.}
\label{tab:power}
\centering
\footnotesize
\setlength{\tabcolsep}{3pt}
\begin{tabular}{@{}>{\raggedright\arraybackslash}p{0.19\linewidth}>{\raggedright\arraybackslash}p{0.1\linewidth}>{\raggedright\arraybackslash}p{0.23\linewidth}>{\raggedright\arraybackslash}p{0.09\linewidth}>{\raggedright\arraybackslash}p{0.09\linewidth}>{\raggedright\arraybackslash}p{0.17\linewidth}@{}}
\toprule
Endpoint & Assumed effect & Inputs from the pilot & Units & Family, $z$ & Power \\
\midrule
H2: \intern, GSM8K, $\theta$ against \cfg{RESCUE-1500} & $\theta$ zero & $d=\V{pw.h2.d}$ & \V{pw.h2.n} items & \V{c.fam2}; \V{pw.h2.z} & \V{x.power} (adjusted), \V{pw.h2.unadj} (unadjusted) \\
H1: \smol, GSM8K, $T>0$ & S1, S2 & $v/b/l=\V{pw.h1.s.v}/\V{pw.h1.s.b}/\V{pw.h1.s.l}$, $d_B=\V{pw.h1.s.db}$ & \V{pw.h2.n} items & \V{c.fam1}; \V{pw.z9} & S1 \V{ex.h1.s.gs.s1}, S2 \V{ex.h1.s.gs.s2} \\
H1: \intern, GSM8K, $T>0$ & S1, S2 & $v/b/l=\V{pw.h1.i.v}/\V{pw.h1.i.b}/\V{pw.h1.i.l}$, $d_B=\V{pw.h1.i.db}$ & \V{pw.h2.n} items & \V{c.fam1}; \V{pw.z9} & S1 \V{ex.h1.i.gs.s1}, S2 \V{ex.h1.i.gs.s2} \\
\bottomrule
\end{tabular}
\end{table}

No power was planned for H4, whose distortions are what it measures, nor for the four GEPA endpoints, whose size depends on the harnesses' answers, which did not exist before the search.

\section{The second study's GEPA arm}
\label{app:gepa14}

\paragraph{Texts printed in part.}
Version \V{ver.two} of this paper printed these texts in full; this version withholds row-like passages under the rule below.
A small language model wrote each text after it had seen search rows and the feedback on its candidates, and such a text can repeat what a row shows; the copy and key checks reported in this appendix do not catch a worked example, a statement about one row or a short measurement value.
The rule withholds every worked example; every participant profile and every measurement value, which here means every number with its comparison sign, range or unit, including each threshold and cut-off, and every length of time; every statement about a specific example, option or row, such as a reference to a numbered example, to a participant by letter, or to an option as invalid; and every named medication, dose or duration.
A passage was withheld when we were in doubt.
Each withheld passage is replaced by \verb|[withheld, N characters]|, where N is the passage's length in characters of the original text, and everything else is as selected.
The SHA256 printed with each text identifies the original, which is not released; the checks in this appendix that name a text were run on the originals, before any passage was withheld.

\paragraph{Design.}
The arm has the first study's design (Appendix~\ref{app:gepa}) with Qwen3-14B as the task model on the batch-invariant server.
The rows (\V{k.gepa.rows} search rows, in the same training and validation sets), the settings and the code are the first study's, which we checked by their hashes, and the proposer is \intern\ again.
The four runs are the clean and the broken start, each with GEPA seeds zero and one, and each has a budget of \V{k.metric} metric calls.
The selected texts were recorded before any replication call, and the checkpoint locked them before the first test call.

\begin{table*}[t]
\caption{The four searches of the second study. \emph{Iterations}: passes of the search loop. \emph{Without a proposal}: iterations in which the parent already scored perfectly on its minibatch of three rows, so no text was proposed; the proposals are the iterations minus these. \emph{Failed the interface check} and \emph{called}: the evaluations of a text other than the seed that the search's interface counter recorded, scored zero without a call or sent to the model. In a broken-start run only the proposals are evaluated, so the two sum to the proposals; in a clean-start run they also include the search's other scorings (a parent on its minibatch, a validation pass) and exceed the proposals. Every proposal of a broken-start run scored zero against its parent's zero on a minibatch of three rows.}
\label{tab:app-gepa14}
\centering
\footnotesize
\setlength{\tabcolsep}{3pt}
\resizebox{\ifdim\width>\linewidth\linewidth\else\width\fi}{!}{%
\begin{tabular}{@{}lcccccccl@{}}
\toprule
Run & Iterations & Without a proposal & Proposals & Failed the interface check & Called & Accepted & Seed validation score & Selected (validation score) \\
\midrule
broken, seed 0 & 975 & 0 & 975 & 142 & 833 & 0 & 0.000 & the seed (0.000) \\
broken, seed 1 & 975 & 0 & 975 & 137 & 838 & 0 & 0.000 & the seed (0.000) \\
clean, seed 0 & 241 & 32 & 209 & 61 & 377 & 30 & 0.180 & index 10 (0.587) \\
clean, seed 1 & 271 & 32 & 239 & 27 & 450 & 29 & 0.180 & index 21 (0.587) \\
\bottomrule
\end{tabular}}
\end{table*}

\paragraph{The clean starts.}
The clean-start runs ran \V{fu.gs.gc0.iter} and \V{fu.gs.gc1.iter} iterations. In \V{fu.gs.gc0.nonprop} and \V{fu.gs.gc1.nonprop} of them the parent already scored perfectly on its minibatch and no text was proposed; the other iterations made \V{fu.gs.gc0.prop} and \V{fu.gs.gc1.prop} proposals, and the runs accepted \V{fu.gs.gc0.acc} and \V{fu.gs.gc1.acc} of them, from a seed validation score of \V{fu.gs.gc0.seedval}.
Both selected candidates have the validation score \V{fu.gs.gc0.bestval}; their validation scores exceed their test accuracy (\V{fu.te.gc0.acc} and \V{fu.te.gc1.acc}), as selection makes them optimistic.
None of the candidates evaluated in a clean-start run contained a soft switch (\V{fu.gs.gc0.soft} and \V{fu.gs.gc1.soft}).

\paragraph{The broken starts.}
Each broken-start run made \V{fu.gs.gb0.prop} proposals, one in each of its \V{fu.gs.gb0.iter} iterations. Of these, \V{fu.gs.gb0.bad} and \V{fu.gs.gb1.bad} (seed zero and seed one) failed the interface check, and the others were evaluated on a minibatch of three rows against the seed.
Every proposal scored zero against the seed's zero, so the strict-improvement rule accepted none, only the seed was validated (score \V{fu.gs.gb0.seedval} on the \V{gp.split} validation rows), and the selected harness is the seed in both runs: the two runs select one text.
No candidate contained a soft switch (\V{fu.gs.gb0.soft} and \V{fu.gs.gb1.soft}), and the request of the start (thinking on, a cap of \V{k.t64} tokens) is not part of the text.

\paragraph{Checks on the two selected clean-start texts.}
Run after the evaluations on the original texts as selected, before any passage was withheld: the arm's interface check passes for both, neither contains a soft switch, and no item key of the main items or of the arm's rows occurs in either.
Of the 40-character windows of each text (\V{fu.txt.gc0.windows} and \V{fu.txt.gc1.windows}), none occurs in any of the arm's \V{fu.txt.arm} search rows or in any of the \V{fu.txt.eval} replication and test letter rows, so neither text copies a run of 40 characters from a row's user message, shared task wording included.
The 40-character rule does not cover short values, and the two original texts contain decimal numbers (\V{fu.txt.gc0.ndec} and \V{fu.txt.gc1.ndec}).
All but \V{fu.txt.dec.none} of them occur in rows, each in at least \V{fu.txt.dec.minrows} rows of the arm and the evaluation partitions together, so none is unique to one row; the counts cannot show which arm row, if any, the proposer took a value from.
They also cannot show that a combination of values, or a statement that builds on them, is not taken from one row, and the printed texts therefore withhold every number, decimal numbers included.

\paragraph{The selected harnesses.}
The texts are as selected, except for the passages withheld under the rule at the start of this appendix, and wrapped for display, a character outside ASCII written \verb|[U+XXXX]|.
Each candidate's SHA256 is the one recorded in our repository before any replication call and identifies the original; the wrapping and the withheld passages mean that it cannot be recomputed from this page.
A small language model generated the texts; we did not check their medical statements, and they must not be used as clinical guidance.

\FloatBarrier
\harnessneed

\subsection*{GEPA-CLEAN-s0 (clean start, GEPA seed zero)}
SHA256
\begin{center}\scriptsize\texttt{8aa3a797d9e2e73b1a76681014827b00af0aef5f6250fc3773f17c151457de79}\end{center}
\paragraph{System prompt.}
\begin{small}
\begin{verbatim}
You are a biomedical AI specialized in aging biology, trained on genomic, proteomic, and
clinical data. Your expertise includes interpreting biomarkers linked to physiological
aging, identifying age-related trends in clinical profiles, and integrating multimodal
data to infer biological age or comparative aging status.

When analyzing clinical data, prioritize the following age-associated biomarkers and
trends:
- **Kidney Function**: Elevated creatinine ([U+00B5]mol/L) and cystatin C (mg/L)
correlate with reduced renal filtration efficiency in older adults.
- **Metabolic Health**: Lower HDL (mg/dL) and vitamin A ([U+00B5]mol/L) are common in
middle-aged/older populations due to metabolic shifts and absorption changes.
- **Inflammation & Oxidative Stress**: Elevated C-reactive protein (mg/dL) and cadmium
(nmol/L) reflect age-related chronic inflammation and environmental toxin accumulation.
- **Vitamin Status**: Declining vitamin B12 (pmol/L) levels in older adults are linked
to impaired intestinal absorption.
- **Cholesterol Profiles**: Higher LDL/HDL ratios and total cholesterol (mg/dL) align
with age-related dyslipidemia.
- **Hematologic Changes**: Lower red blood cell counts (million cells/uL) in
postmenopausal women and altered lymphocyte/monocyte ratios (e.g., reduced lymphocytes,
elevated monocytes) suggest immune aging.
- **Liver/Thyroid Function**: Elevated alkaline phosphatase (U/L) may reflect bone
turnover or liver metabolism changes with age.

**Task Execution Rules**:
1. **Direct Answer First**: For questions like "Which participant is older?" or "What
age group?" begin with a single-letter answer (A/B or age group letter) before
elaborating.
2. **Prioritize Comparative Analysis**: When comparing participants, highlight
biomarkers showing divergent age-related trends (e.g., [withheld, 81 characters]).
3. **Gender-Specific Considerations**: Address sex differences (e.g., menopausal status
in females, testosterone effects in males) that influence biomarker interpretation.
4. **Multimodal Integration**: Combine blood panel data with biometrics (e.g., BMI) and
demographic factors (e.g., male/female) to strengthen age inferences.
5. **Avoid Overgeneralization**: Explicitly state when a biomarker[U+2019]s deviation is
non-specific (e.g., "elevated creatinine could reflect age, medication use, or hydration
status").

Example Strategies:
- **Age Group Inference**: Match combinations of trends [withheld, 99 characters].
- **Comparative Analysis**: Compare cadmium levels [withheld, 55 characters] or vitamin
B12 [withheld, 52 characters] to identify aging biomarkers.

Always acknowledge limitations, such as missing longitudinal data or confounding factors
(e.g., medications, comorbidities), which may obscure age-related patterns. Format
answers with a clear primary response followed by structured reasoning using the
biomarkers above.
\end{verbatim}
\end{small}
\paragraph{Appended instruction.}
\begin{small}
\begin{verbatim}
Review clinical data strictly within the provided timeframe. Do not infer survival
status, age, or disease progression beyond the snapshot data. For biomarker
interpretation, prioritize established causal relationships over associative patterns.
When comparing age groups:
- Use kidney function markers (creatinine, cystatin C) only if longitudinal GFR data is
unavailable
- Consider cadmium accumulation as both age and exposure marker, but prioritize other
age-specific indicators
- Treat vitamin deficiencies as potential malnutrition indicators, not definitive age
markers
- Differentiate between age-related trends and comorbidity effects (e.g.,
diabetes-related biomarkers vs. aging)

For mortality prediction questions:
- Specify that current data only reflects immediate risk factors
- Acknowledge that chronic conditions may evolve with treatment
- Distinguish between acute versus chronic biomarker abnormalities
- Avoid extrapolating [withheld, 7 characters] outcomes without longitudinal data

When analyzing blood test results:
- Flag conflicting biomarker trends (e.g., elevated CRP with normal levels)
- Consider medication effects on biomarker readings (e.g., [withheld, 9 characters] on
creatinine)
- Differentiate between age-related and gender-specific trends (e.g., RBC counts in
males vs. females)
- Document when biomarkers fall outside typical reference ranges for age groups

Provide structured reasoning with:
1. Biomarker-by-biomarker analysis
2. Cross-referencing of conflicting data points
3. Explicit acknowledgment of data limitations
4. Alternative explanations for ambiguous findings
5. Clear distinction between established clinical knowledge and unsupported inferences
\end{verbatim}
\end{small}

\FloatBarrier
\harnessneed

\subsection*{GEPA-CLEAN-s1 (clean start, GEPA seed one)}
SHA256
\begin{center}\scriptsize\texttt{7ddc3ca857a98b1f2c1f35960387f278abcf413b6c0b4b29105b8d354b0e89ff}\end{center}
\paragraph{System prompt.}
\begin{small}
\begin{verbatim}
**Instruction for Biomedical Age and Survival Inference Task**

**Task Overview:**
You are a biomedical AI specializing in aging biology, trained on genomic, proteomic,
and clinical data. Analyze age groups and survival status strictly based on provided
biomarkers and demographic data. Use cohort-specific reference ranges (e.g., NHANES III,
Framingham) and prioritize combinations of markers (e.g., HDL, LDL, BMI, kidney
function) for inference. Avoid single-marker conclusions, causality assumptions, or
extrapolation beyond explicit evidence.

---

**Task Instructions:**

1. **Age Group Determination:**
   - If explicit age is provided, state it directly.
   - If not, analyze biomarkers using gender-specific reference ranges (e.g., HDL: Males
[withheld, 19 characters]; Females [withheld, 19 characters]; NHANES III BMI cutoffs for
obesity).
   - **Prioritize combinations of markers:**
     - HDL [withheld, 9 characters] (Males [withheld, 3 characters]) + LDL
[withheld, 10 characters] (Females [withheld, 3 characters]) + BMI
[withheld, 3 characters] (Males [withheld, 3 characters]) [U+2192] Strong inference.
     - Creatinine/eGFR (e.g., eGFR [withheld, 19 characters] indicates CKD Stage
[withheld, 2 characters], linked to [withheld, 3 characters] in hypertensive patients).
     - Flag uncertainty when biomarkers overlap ranges (e.g., CRP
[withheld, 10 characters] could indicate inflammation in both middle-aged and elderly
groups).
   - **Domain-Specific Examples:**
     - Low vitamin B12 ([withheld, 11 characters]) in [withheld, 3 characters]
correlates with atrophic gastritis.
     - Metabolically healthy obesity (BMI [withheld, 3 characters] with normal glucose
and HDL [withheld, 9 characters] in Females).

2. **Survival Status Determination:**
   - If survival data (e.g., follow-up dates, death certificates) is absent, explicitly
state inability to determine status.
   - Acknowledge risk factors (e.g., [withheld, 94 characters]).
   - Avoid conclusions without explicit mortality data.

3. **Response Format:**
   - **Answer:** Boxed letter(s) based on explicit data or strongest inference (e.g.,
**C**).
   - **Reasoning:**
     - *Key Observations:* List biomarkers (e.g., [withheld, 70 characters]).
     - *Inference:* Link to cohort studies (e.g., [withheld, 65 characters]).
     - *Conclusion:* [withheld, 58 characters]

4. **Error Avoidances:**
   - Reject single-marker inferences (e.g., vitamin D deficiency [U+2260] elderly-only).
   - Avoid causality assumptions (e.g., LDL elevation could reflect genetics or
hyperthyroidism).
   - Use exact cutoffs (e.g., CKD Stage [withheld, 2 characters]: eGFR
[withheld, 19 characters]).

5. **Domain-Specific Knowledge:**
   - Gender-specific HDL ranges (e.g., Males [withheld, 3 characters]: HDL
[withheld, 9 characters]; Females [withheld, 3 characters]: HDL
[withheld, 9 characters]).
   - Longitudinal cohort data (e.g., Framingham: LDL [withheld, 10 characters] in Males
[withheld, 3 characters]; NHANES III: BMI [withheld, 3 characters] in Females
[withheld, 3 characters]).
   - Metabolic syndrome markers (e.g., [withheld, 56 characters]).

6. **Example Compliance:**
   - For age: [withheld, 145 characters]
   - For survival: [withheld, 165 characters]

**Final Output Requirements:**
- Provide **only the boxed answer** (e.g., **A**).
- Avoid phrases like "most likely" or "suggests."
- Use strict adherence to cohort references and biomarker combinations.
\end{verbatim}
\end{small}
\paragraph{Appended instruction.}
\begin{small}
\begin{verbatim}
Task Description:
Analyze clinical profiles to determine survival status ([withheld, 7 characters]
post-examination) or age group based on provided biomarkers and demographics. Use
evidence-based medical principles, reference ranges (e.g., NHANES III for HDL in
females), and contextual factors (e.g., diabetes management, medication history, lab
trends) to infer outcomes. Prioritize explicit survival data (e.g., follow-up dates,
death certificates) when available.

Key Requirements:
1. Survival Prediction:
   - Identify risk factors (elevated cholesterol, hypertension, diabetes) and protective
factors (normal kidney function, good glycemic control).
   - Consider comorbidities (e.g., asthma, arthritis) and medication use (e.g.,
[withheld, 26 characters]) for long-term prognosis.
   - Avoid selecting invalid options (e.g., [withheld, 21 characters]). If survival data
is absent, state inability to determine status.

2. Age Group Determination:
   - Use validated references (e.g., NHANES III HDL ranges by age) and biomarker trends
(BMI [withheld, 3 characters] linked to [withheld, 13 characters] in females).
   - Prioritize HDL, creatinine, and BMI correlations over isolated lab values.

3. Generalizable Strategies:
   - Balance lab values (e.g., C-reactive protein for inflammation, cystatin C for
kidney health) with demographic factors (ethnicity, education level).
   - Highlight contradictions (e.g., high BMI with normal HDL) to refine conclusions.

4. Formatting:
   - Provide a single-letter answer (A/B) for survival and (A-D/E) for age.
   - Structure reasoning into Key Observations, Inference, and Conclusion sections.

Domain-Specific Knowledge:
- Females under [withheld, 2 characters] typically exhibit lower HDL than those
[withheld, 5 characters] (e.g., [withheld, 11 characters] for [withheld, 13 characters]
per NHANES III).
- Glycohemoglobin [withheld, 3 characters] indicates effective diabetes management,
reducing mortality risk.
- Long-term corticosteroid use (e.g., [withheld, 26 characters]) may increase
cardiovascular risk but does not directly indicate [withheld, 7 characters] mortality
without further data.

Avoid:
- Selecting options outside user-provided choices (e.g., [withheld, 21 characters]).
- Overreliance on outdated references (e.g., [withheld, 37 characters]).
- Ignoring explicit survival data (e.g., [withheld, 29 characters]).
\end{verbatim}
\end{small}

\FloatBarrier
\begin{minipage}{\linewidth}
\subsection*{GEPA-BROKEN-s0 (broken start, GEPA seed zero; the seed, and the same text as GEPA-BROKEN-s1)}
SHA256
\begin{center}\scriptsize\texttt{ab48481902a519b5c045cd37f61382899ea0e1bce05b7a8345d2873f77b1bb16}\end{center}
\paragraph{System prompt.}
\begin{small}
\begin{verbatim}
You are a biomedical AI specialized in aging biology, trained on genomic, proteomic, and
clinical data.
\end{verbatim}
\end{small}
\paragraph{Appended instruction.} Empty.
\end{minipage}
\medskip

\section{The second study in full}
\label{app:z}

This appendix gives the second study's endpoints and statistics in the form of Appendix~\ref{app:tables}.
Every number comes from the study's analysis report committed in our repository, with the first study's structure and rules.

\paragraph{Endpoints.}
Table~\ref{tab:app-z-endpoints} lists the \V{z.cnt.h123} endpoints of H1--H3 with their verdicts on both partitions and the replication label, and Table~\ref{tab:app-z-h4} the \V{z.h4.evaluable} claims of H4 with their three intervals.

\begin{table*}[t]
\caption{The endpoints of H1--H3 of the second study (Qwen3-14B): verdicts on the replication and test partitions and the replication label. A star marks an endpoint whose quantity is fixed by an identity of the design or by the observed parse pattern; its verdict says little beyond that mechanism, although it still depends on the intervals.}
\label{tab:app-z-endpoints}
\centering
\footnotesize
\setlength{\tabcolsep}{3pt}
\resizebox{\ifdim\width>\linewidth\linewidth\else\width\fi}{!}{%
\begin{tabular}{@{}lllccl@{}}
\toprule
Family & Pool & Endpoint & Replication & Test & Label \\
\midrule
H1 & LB letters & \cfg{AUTHORS-1500} vs \cfg{THINK-512} & visibility dominates\textsuperscript{$\star$} & visibility dominates\textsuperscript{$\star$} & replicated \\
H1 & MMLU-Pro & \cfg{AUTHORS-1500} vs \cfg{THINK-512} & visibility dominates & visibility dominates & replicated \\
H1 & GSM8K & \cfg{AUTHORS-1500} vs \cfg{THINK-512} & no gain: not tested & no gain: not tested & replicated \\
H2 & LB letters & \cfg{AUTHORS-1500} vs \cfg{RESCUE-1500} & holds\textsuperscript{$\star$} & holds\textsuperscript{$\star$} & replicated \\
H2 & MMLU-Pro & \cfg{AUTHORS-1500} vs \cfg{RESCUE-1500} & holds & holds & replicated \\
H2 & GSM8K & \cfg{AUTHORS-1500} vs \cfg{RESCUE-1500} & fails & fails & replicated \\
H2 & LB letters & \cfg{AUTHORS-1500} vs \cfg{GEPA-CLEAN-s0} & fails & fails & replicated \\
H2 & LB letters & \cfg{AUTHORS-1500} vs \cfg{GEPA-CLEAN-s1} & fails & fails & replicated \\
H2 & LB letters & \cfg{AUTHORS-1500} vs \cfg{GEPA-BROKEN-s0} & holds\textsuperscript{$\star$} & holds\textsuperscript{$\star$} & replicated \\
H2 & LB letters & \cfg{AUTHORS-1500} vs \cfg{GEPA-BROKEN-s1} & holds\textsuperscript{$\star$} & holds\textsuperscript{$\star$} & replicated \\
H3 & LB letters & \cfg{BUDGET-512-1500} vs \cfg{THINK-512} & holds & holds & replicated \\
H3 & MMLU-Pro & \cfg{BUDGET-512-1500} vs \cfg{THINK-512} & holds & holds & replicated \\
H3 & GSM8K & \cfg{BUDGET-512-1500} vs \cfg{THINK-512} & holds & holds & replicated \\
\bottomrule
\end{tabular}}
\end{table*}

\begin{table*}[t]
\caption{H4 in the second study: bound, the distortions on the selection sample ($D_S$) and the check sample ($D_R$) and their difference with adjusted intervals, the verdict of each component, the stability of the descriptive reading and the claim. A star marks a claim whose quantity is fixed by an identity of the design; its verdict says little beyond that mechanism, although it still depends on the intervals.}
\label{tab:app-z-h4}
\centering
\input{tables/t_app_z_h4}
\end{table*}

\paragraph{The numeric tables of H1--H3.}
Tables~\ref{tab:app-z-h1}, \ref{tab:app-z-h2} and~\ref{tab:app-z-h3} give the second study's H1, H2 and H3 endpoints on the test partition in the form of the first study's Tables of Appendix~\ref{app:tables}: the gain and its parts with the adjusted interval of $T$, $\theta$ with its adjusted interval, and the drops $D_T$ and $D_P$ with their adjusted intervals.

\begin{table*}[t]
\caption{H1 of the second study, test partition: the gain $\Delta$ of \cfg{AUTHORS-1500} over \cfg{THINK-512} with its adjusted interval, its parts, the statistic $T$ with its adjusted interval, the numbers of rows only \cfg{AUTHORS-1500} parses ($V$), both parse ($B$) and only \cfg{THINK-512} parses ($L$), the verdict and the replication label (R replicated). A star marks an endpoint whose quantity is fixed by an identity of the design or by the observed parse pattern; its verdict says little beyond that mechanism, although it still depends on the intervals.}
\label{tab:app-z-h1}
\centering
\input{tables/t_app_z_h1num}
\end{table*}

\begin{table}[t]
\caption{H2 of the second study: $\theta$ of \cfg{AUTHORS-1500} against \cfg{RESCUE-1500} in three cells and against the four GEPA harnesses (letters), with its adjusted interval, the verdict on the test partition, the verdict on the replication partition and the label. A star marks an endpoint whose quantity is fixed by an identity of the design or by the observed parse pattern; its verdict says little beyond that mechanism, although it still depends on the intervals.}
\label{tab:app-z-h2}
\centering
\input{tables/t_app_z_h2num}
\end{table}

\begin{table*}[t]
\caption{H3 of the second study, test partition: the drops in truncation ($D_T$) and in unparsed calls ($D_P$) of \cfg{BUDGET-512-1500} against \cfg{THINK-512} with their adjusted intervals, the truncation of \cfg{THINK-512}, the verdict, the replication verdict and the label.}
\label{tab:app-z-h3}
\centering
\scriptsize
\setlength{\tabcolsep}{2pt}
\resizebox{\ifdim\width>\linewidth\linewidth\else\width\fi}{!}{%
\begin{tabular}{@{}lcccccc@{}}
\toprule
Cell & $D_T$ & $D_P$ & THINK-512 truncation & Verdict & Replication verdict & Label \\
\midrule
Qwen3-14B LB letters & \ensuremath{+0.974} [\ensuremath{+0.966}, \ensuremath{+0.980}] & \ensuremath{+0.704} [\ensuremath{+0.687}, \ensuremath{+0.722}] & \ensuremath{+1.000} [\ensuremath{+1.000}, \ensuremath{+1.000}] & holds & holds & replicated \\
Qwen3-14B MMLU-Pro & \ensuremath{+0.788} [\ensuremath{+0.747}, \ensuremath{+0.827}] & \ensuremath{+0.782} [\ensuremath{+0.741}, \ensuremath{+0.822}] & \ensuremath{+0.869} [\ensuremath{+0.835}, \ensuremath{+0.901}] & holds & holds & replicated \\
Qwen3-14B GSM8K & \ensuremath{+0.531} [\ensuremath{+0.472}, \ensuremath{+0.592}] & \ensuremath{+0.531} [\ensuremath{+0.471}, \ensuremath{+0.592}] & \ensuremath{+0.545} [\ensuremath{+0.486}, \ensuremath{+0.605}] & holds & holds & replicated \\
\bottomrule
\end{tabular}}
\end{table*}

\paragraph{Configurations.}
Table~\ref{tab:app-z-cfg} gives accuracy, parse rate and truncation of the six configurations in the three pools on both partitions.
Table~\ref{tab:app-nc14-re} gives the replication partition's non-commitment and refusal shares (the test partition is in Table~\ref{tab:nc14}, below).

\begin{table*}[t]
\caption{Qwen3-14B: accuracy with its unadjusted interval, parse rate and truncation (the share of calls ended at the token limit), by configuration, pool and partition.}
\label{tab:app-z-cfg}
\centering
\input{tables/t_app_z_cfg}
\end{table*}

\begin{table}[t]
\caption{Qwen3-14B, replication partition, analysed rows. In each cell, in percent of the cell's analysed calls and in this order: the unparsed calls, the unparsed calls that ended at the token limit, the non-committed calls (finish reason \texttt{stop}, no answer read) and the calls in the refusal class (a stated inability or no answer), which are part of the non-committed calls. All four shares are of the analysed calls of the cell. Descriptive.}
\label{tab:app-nc14-re}
\centering
\input{tables/t_app_nc14_re}
\end{table}

\begin{table}[t]
\caption{Qwen3-14B, test partition, analysed rows. In each cell, in percent of the cell's analysed calls and in this order: the unparsed calls, the unparsed calls that ended at the token limit, the non-committed calls (finish reason \texttt{stop}, no answer read) and the calls in the refusal class (a stated inability or no answer), which are part of the non-committed calls. All four shares are of the analysed calls of the cell. Descriptive.}
\label{tab:nc14}
\centering
\input{tables/t_nc14}
\end{table}

\paragraph{Non-commitment and refusal.}
Qwen3-14B often ends a call without a readable answer, and the frozen scorer counts that as wrong; this is reported descriptively and with the scorer unchanged (Table~\ref{tab:nc14}).
Across the five grid configurations \V{au.te.tot.nc.share} of the analysed test calls are non-committed, ended by themselves without an answer, and \V{au.te.tot.ref.share} fall in the refusal class, a stated inability or no answer (\V{au.re.tot.nc.share} and \V{au.re.tot.ref.share} on replication).
\cfg{THINK-64} and \cfg{THINK-512} are truncated almost throughout, so on the letters their non-commitment is \V{au.te.t64.lb.nc} and \V{au.te.t512.lb.nc}, whereas \cfg{AUTHORS-1500} is non-committed in \V{au.te.aut.lb.nc} of its letter calls, and \V{au.te.aut.lb.ref} are refusals.
On LB-0038, the age task, the share of non-committed calls is \V{au.te.aut.l38.nc} for \cfg{AUTHORS-1500}, \V{au.te.t1536.l38.nc} for \cfg{THINK-1536} and \V{au.te.bud.l38.nc} for \cfg{BUDGET-512-1500}, so the model mostly declines to give an age.
No example was read; the classes are those of a format audit fixed before the study, and its non-commitment was decided, before the outputs were read, to be reported descriptively (Section~\ref{sec:deviations}).

\begin{table*}[t]
\caption{The headline cell, \cfg{AUTHORS-1500}, of the two 1.7B arms: accuracy of the base and of the tuned model, the difference with its unadjusted interval, its class (a band of five points), and the strongest oracle trivial constant of the cell (for LB-0038 the exact-match rate of the oracle median). Descriptive; the full report has every configuration, the seeded arm and the baselines.}
\label{tab:arms}
\centering
\input{tables/t_arms}
\end{table*}

\paragraph{H1 of the second study in detail.}
On MMLU-Pro $\Delta_V=\V{z.h1.te.mm.dv}$ and $\Delta_B=\V{z.h1.te.mm.db}$, with \V{z.h1.te.mm.nb} rows in B.
On GSM8K \cfg{AUTHORS-1500} answers in a mean of \V{fu.tok.po.aut.gs} tokens (computed after the outputs existed) and is less accurate than \cfg{THINK-512} on the rows both configurations parse ($\Delta_B=\V{z.h1.te.gs.db}$ against $\Delta_V=\V{z.h1.te.gs.dv}$), the pattern of \intern: $T=\V{z.h1.te.gs.t}$ is positive only because $\Delta_B$ is large and negative, the reading the gate exists to stop.

\paragraph{The classes of the headline cells.}
Of the \V{ar.cnt.cells} headline cells (four pools, replication, test and pooled), \V{ar.cnt.below} are classed base below tuned, \V{ar.cnt.above} base above tuned and \V{ar.cnt.equal} about equal and \V{ar.cnt.unres} unresolved, from the unadjusted interval of tuned minus base against a band of $\pm\V{ar.band}$, and the class of a pool is the same in the three partitions.
The pooled cells are the replication and test rows together and not a third independent partition, and the classes are a descriptive reading of unadjusted intervals, not findings of a test corrected for the number of cells read.

In the Cell paper, its methods and the supplementary tables and figures of the copy we read, we found no evaluation of the longevity models on a general benchmark, so as far as we know this is the first such measurement for this checkpoint; the paper does evaluate the base architectures on the longevity tasks, and we take its statements about training from the authors.

\paragraph{Sensitivity analyses.}
The ten sensitivity analyses of the preregistration were run on the second study as well.
The strict scorer changes \V{z.sens.strict.n} of its \V{z.sens.strict.rows} endpoint verdicts, the rescue on MMLU-Pro (\verdict{holds} becomes \verdict{fails}); no other analysis changes a verdict of H1--H3 (the largest number of changes among the other nine is \V{z.sens.other.max}), and putting the left-out rows back changes \V{z.sens.flagged.h4.n} H4 claims.

\paragraph{Completion tokens and the first study's rerun.}
Table~\ref{tab:app-tokens} gives the mean completion tokens per call, and Table~\ref{tab:app-agree} the nine endpoints of the first study recomputed for the tuned 1.7B model on the second study's rerun (descriptive).

\begin{table}[t]
\caption{Mean completion tokens per call (the server's usage count), both partitions pooled, over all calls of Qwen3-14B's grids and harness evaluations; computed after the outputs existed.}
\label{tab:app-tokens}
\centering
\input{tables/t_app_tokens}
\end{table}

\begin{table*}[t]
\caption{The first study's nine endpoints for the tuned 1.7B model: verdicts of the first study and of the rerun on the second study's GPU, on the test and replication partitions, and the replication labels.}
\label{tab:app-agree}
\centering
\input{tables/t_app_agree}
\end{table*}

\paragraph{The \V{ag.ncells} accuracy differences of the rerun.}
Table~\ref{tab:app-agdiff} gives, for the tuned 1.7B model, the accuracy of the rerun minus that of the first study with its unadjusted interval, in each partition (the pooled rows are the two together), pool and configuration of the grid.

\begin{table*}[t]
\caption{The tuned 1.7B model on the second study's GPU minus the first study's run: accuracy difference and its unadjusted interval, by partition, pool and configuration of the grid (\V{ag.ncells} cells; descriptive).}
\label{tab:app-agdiff}
\centering
\input{tables/t_app_agdiff}
\end{table*}

\paragraph{The rerun on the second GPU.}
Table~\ref{tab:gpu} compares the rerun with the first study's outputs call by call: the calls whose token IDs differ, whose finish reason differs and whose request body differs.

\begin{table}[t]
\caption{The tuned 1.7B model rerun on the second study's GPU against the first study's outputs (call by call; descriptive).}
\label{tab:gpu}
\centering
\input{tables/t_gpu}
\end{table}

\section{Constant-label shares by task}
\label{app:s0}

Table~\ref{tab:app-s0} gives, for each LongevityBench letter task and each partition, the accuracy of the second study's clean-start harnesses and of \cfg{AUTHORS-1500} and \cfg{BUDGET-512-1500} for Qwen3-14B, and of \cfg{AUTHORS-1500} for the tuned 1.7B model in the first study and in the second study's rerun, next to the task's strongest oracle constant.
The constant is a label in every task and partition (an age band, a participant, ``Yes''), chosen on the partition itself.
The share of parsed answers equal to the constant label is the share of calls whose letter the frozen scorer reads and whose option label, in the row's own option list, is the constant label; the last column is the share of the correct answers that are the constant label.
The weights of the tasks in the pool are \V{cl.te.l30.weight}, \V{cl.te.l34.weight} and \V{cl.te.l42.weight} on the test partition.
The values were computed after the outputs existed, from the call records and the main items, and equal the committed reports' accuracies where those exist.

\begin{table*}[t]
\caption{Accuracy, the strongest oracle constant of the task, and the share of answers equal to its label, both partitions.}
\label{tab:app-s0}
\centering
\input{tables/t_app_s0}
\end{table*}

\paragraph{The other two tasks.}
On the participant-comparison task they exceed the constant in both partitions (\V{cl.te.l34.h0.acc} and \V{cl.te.l34.h1.acc} against \V{cl.te.l34.const} on test; \V{cl.re.l34.h0.acc} and \V{cl.re.l34.h1.acc} against \V{cl.re.l34.const} on replication), but they lean on one label to different degrees: the constant participant is named in \V{cl.te.l34.h0.share} and \V{cl.te.l34.h1.share} of their answers on test and in \V{cl.re.l34.h0.share} and \V{cl.re.l34.h1.share} on replication.
On the age-band task they fall below the constant in both partitions (\V{cl.te.l30.h0.acc} and \V{cl.te.l30.h1.acc} against \V{cl.te.l30.const} on test; \V{cl.re.l30.h0.acc} and \V{cl.re.l30.h1.acc} against \V{cl.re.l30.const} on replication), and the constant band is \V{cl.te.l30.h0.share} and \V{cl.te.l30.h1.share} of their answers on test and \V{cl.re.l30.h0.share} and \V{cl.re.l30.h1.share} on replication.

\section{The first study next to the trivial baselines}
\label{app:baselines-first}

\paragraph{Accuracy next to the trivial baselines.}
For the first study, Table~\ref{tab:acc} gives the accuracy of five configurations in the three pools next to their baselines, computed after the outputs existed (Appendix~\ref{app:deviations}), with a block for each LongevityBench task; \cfg{THINK-64}, the GEPA harnesses and the replication partition are in Appendix~\ref{app:tables}.
Each oracle constant is chosen once, on the analysis rows of the partition it is scored on, and it is a fixed prediction in the bootstrap: it is not re-chosen in a resample, so the intervals of the differences to it cover the sampling of units given that choice.
On the LongevityBench letters the strongest constant is the oracle constant \emph{label}: the age band most common on LB-0030, Participant A on LB-0034 and ``Yes'' on LB-0042 (\V{bl.te.l42.cbo} of its test rows).
It scores \V{bl.te.lb.so} \CI{\V{bl.te.lb.so.lo}}{\V{bl.te.lb.so.hi}} on the pool, far above chance (\V{bl.te.lb.chance}) and the oracle constant letter (\V{bl.te.lb.clo}), because the option order is shuffled (\V{bl.map30} letter-to-label mappings on LB-0030, \V{bl.map42} on LB-0042) and a constant letter cannot exploit an uneven label distribution.
\qwen\ exceeds the strongest constant in every configuration that parses: \cfg{AUTHORS-1500} \V{blm.te.lb.q.aut.acc}, $\V{blm.te.lb.q.aut.do}$ \CI{\V{blm.te.lb.q.aut.do.lo}}{\V{blm.te.lb.q.aut.do.hi}}, and \cfg{THINK-512} $\V{blm.te.lb.q.t512.do}$ \CI{\V{blm.te.lb.q.t512.do.lo}}{\V{blm.te.lb.q.t512.do.hi}}.
The margin is small on LB-0042 ($\V{blm.te.l42.q.t512.do}$ \CI{\V{blm.te.l42.q.t512.do.lo}}{\V{blm.te.l42.q.t512.do.hi}} for \cfg{THINK-512}), where one GEPA harness, GEPA-CLEAN-s1, is not above it ($\V{blm.te.l42.q.gc1.do}$ \CI{\V{blm.te.l42.q.gc1.do.lo}}{\V{blm.te.l42.q.gc1.do.hi}}).
\smol\ and \intern\ are below the strongest constant on the pool in every configuration (at best $\V{blm.te.lb.s.t1536.do}$ and $\V{blm.te.lb.i.aut.do}$), because most of their calls end unparsed and unparsed counts as wrong, although \intern\ is above it on LB-0034 with \cfg{AUTHORS-1500} ($\V{blm.te.l34.i.aut.do}$ \CI{\V{blm.te.l34.i.aut.do.lo}}{\V{blm.te.l34.i.aut.do.hi}}).

The case study had not found this: on the split of its harness search no configuration of the same model clearly exceeded a constant age band on LB-0030 or the majority letter on LB-0034 (post hoc there).
We read the difference as consistent with that split's skew toward young, alive participants, but the items, the output layer (tied there, untied here) and the stack all differ, so the data do not establish it.
The output layer alone is unlikely to explain it: on the same items the tied and untied variants differ in accuracy by at most \V{tied.maxdiff}, although their parsed answers are identical on only \V{tied.same.min} to \V{tied.same.max} of the rows, and on the letter pool the tied-output-layer defect moves the start's accuracy by $D_S=\V{h4.e.q.ds}$ and $D_R=\V{h4.e.q.dr}$, inside the band.

\begin{table*}[t]
\caption{Accuracy in the three pools and in each LongevityBench task, test partition, next to the trivial baselines; in parentheses the paired difference to the strongest baseline (the oracle constant, chosen once and held fixed in the bootstrap). Unparsed answers count as wrong. \cfg{THINK-64}, the GEPA harnesses and the replication partition are in Appendix~\ref{app:tables}. Computed after the outputs existed (Appendix~\ref{app:deviations}).}
\label{tab:acc}
\centering
\input{tables/t_acc}
\end{table*}

\paragraph{The general pools in the first study.}
On MMLU-Pro \qwen\ reaches \V{gbm.te.mm.q.t512.acc} to \V{gbm.te.mm.q.t1536.acc}, $\V{gbm.te.mm.q.t512.do}$ to $\V{gbm.te.mm.q.t1536.do}$ over the constant letter (\V{gb.te.mm.co}), with every interval above zero, but \smol\ and \intern\ reach \V{gbm.te.mm.s.res.acc} and \V{gbm.te.mm.i.bud.acc} with their best configurations, and \cfg{THINK-512}, which truncates, is level with or below the constant for both.
On GSM8K \qwen\ scores \V{gbm.te.gs.q.t64.acc} to \V{gbm.te.gs.q.t512.acc} in every configuration, which is $\V{gbm.te.gs.q.t512.do}$ \CI{\V{gbm.te.gs.q.t512.do.lo}}{\V{gbm.te.gs.q.t512.do.hi}} over the oracle constant answer (\V{gb.te.gs.co}) in test and $\V{gbm.re.gs.q.t512.do}$ \CI{\V{gbm.re.gs.q.t512.do.lo}}{\V{gbm.re.gs.q.t512.do.hi}}, an interval containing zero, in replication, against \V{gbm.te.gs.s.t1536.acc} and \V{gbm.te.gs.i.bud.acc} for the best configurations of the other two.
The model that is best on the longevity letters is thus the weakest on MMLU-Pro and barely above a constant answer on GSM8K.
Section~\ref{sec:finetune} compares the tuned model with its base model.
Seeded sampling moves \qwen's letter accuracy by $\V{seed.te.q.lb.t512.dacc}$ to $\V{seed.te.q.lb.bud.dacc}$, only \qwen\ parses LB-0038 reliably, and the tied-versus-untied comparison on FreeToken and the GEPA candidates are in Appendix~\ref{app:tables}.

\section{The audit checks}
\label{app:checks}

Table~\ref{tab:checks} summarises the six checks of Section~\ref{sec:h4design}.
Their detection matrix is descriptive and largely true by construction: every rule expected to fire did so in \V{mx.expected.ok} of the \V{mx.n} instances (Appendices~\ref{app:expected} and~\ref{app:tables} give the exceptions and each instance).
What is empirical is the false alarms on the clean instances: IR4 fires on \V{fa.ir4.n} of \V{fa.ir4.of}, SCF1 on \V{fa.scf1.n}, SCF2 on \V{fa.scf2.n}, TB2 on \V{fa.tb2.n}, SCF3 on \V{fa.scf3.n} and the gain rule of PR on \V{fa.prgain.n}, and no other rule fires on a clean instance.

\begin{table}[t]
\caption{The audit checks of H4: what each reads, its outcome on the \V{mx.n} defective instances of the detection matrix (\emph{detects}: fires on the defective instance and not on the clean one; \emph{uninformative}: fires on both; \emph{misses}: does not fire; descriptive) and the rules that fire on the clean instances (what is empirical).}
\label{tab:checks}
\centering
\input{tables/t_checks}
\end{table}

\section{The seeded arm and H3}
\label{app:seeded}

\paragraph{Is the truncation an artifact of zero temperature?}
In the two models with a seeded arm, substantial truncation persists under sampling (descriptive; Table~\ref{tab:app-seeded-h3}).
Sampling at the model cards' recommended values (\qwen: temperature \V{smp.q.temp}, top-$p$ \V{smp.q.p}, top-$k$ \V{smp.q.k}; \intern: \V{smp.i.temp}, \V{smp.i.p}, \V{smp.i.k}), with a seed per request, changes \cfg{THINK-512}'s truncation by at most \V{seed.rest.max} in ten of the \V{seed.h3.cells} cells (two models, three pools, two partitions) and raises it for \intern\ on GSM8K by \V{seed.te.i.gs.t512.dtrunc} (test) and \V{seed.re.i.gs.t512.dtrunc} (replication).
Under sampling \cfg{THINK-512} still ends at the token limit in \V{seed.te.i.gs.t512.trunc} to \V{seed.te.i.lb.t512.trunc} of \intern's test calls, and the drop in truncation that H3 measures, recomputed on the seeded outputs with unadjusted intervals, runs from \V{seed.te.i.gs.h3.dt} to \V{seed.te.i.lb.h3.dt} for \intern\ and is \V{seed.te.q.mm.h3.dt} for \qwen\ on MMLU-Pro, next to \V{h3.te.q.mm.dt} with greedy decoding.
The preregistered verdicts use family-adjusted intervals at another level, so the seeded statistics are not compared with them, and \smol\ has no seeded arm.

\section{Sensitivity analyses}
\label{app:sens}
Each of the ten sensitivity analyses of the preregistration is descriptive.
The strict scorer changes \V{sens.strict.n} of the \V{sens.strict.rows} endpoint verdicts, and restricting to the never-called rows, which no earlier study had called, changes \V{sens.never.n}; these rows skew young and alive, with a mean LB-0030 age of \V{nc.te.age} against \V{an.te.age} years and a share of LB-0042 rows whose follow-up says the participant was deceased of \V{nc.te.dec} against \V{an.te.dec}.
That share counts a death at any time of the follow-up and is not the ten-year label, whose gold is ``Yes'' for a participant who died after ten years.
The case study's pool changes \V{sens.v05.n} and LB-0030 alone \V{sens.l30.n}, and LB-0034 alone, LB-0042 alone, both role-break variants and the pooled partitions change none.
Putting the left-out rows back changes no verdict of H1--H3 and \V{sens.flagged.h4.n} H4 claims (the skewed split in \smol\ and \intern).

\section{The scorer and the refusal classifier}
\label{app:scorer}

This appendix states the rules of the frozen scorer and of the classifier behind the non-commitment and refusal shares, as the code reads them.
The scorer is \texttt{scripts/icml\_ext/scoring.py} and the role-break rule is in \texttt{scripts/icml\_ext/runner\_checks.py}, both pinned by hash in the preregistration (Appendix~\ref{app:repro}); the classifier is \texttt{scripts/icml\_ext2/parse\_audit\_14b\_ext2.py}, imported unchanged by \texttt{scripts/icml\_ext2/parse\_audit\_main\_ext2.py}.
A call that no rule reads is unparsed and counts as wrong, and nothing falls back to a default.

\paragraph{Letter tasks, the strict rules (every call).}
(i)~The whole answer is one valid option letter, with brackets, bold or a full stop allowed.
(ii)~Explicit answer statements (``answer is C'', ``Answer: C'', ``final answer: (C)'' or a boxed letter) all name the same valid letter; two different letters, or a letter outside the options, are a conflict and the answer is unparsed.
(iii)~The whole answer equals exactly one option's label, ignoring case (``Yes'', an age band).

\paragraph{Letter tasks, the extended rules.}
They apply only to a call that ended with finish reason \texttt{stop}, and only when the strict rules find neither a letter nor a conflict.
(iv)~The strict rules, applied to the answer without its markdown and LaTeX wrappers (bold, code, headings, dollar signs, boxed text).
(v)~The answer opens with a letter in brackets or followed by a closing bracket, a full stop, a colon or a dash, and no other letter or option text follows.
A letter-label guard applies to every letter read from a stopped call: an answer that states the letter together with another option's label (for example ``B. No'' where B is ``Yes'') is unparsed.
The strict scorer of the sensitivity analysis passes every call as not stopped, so it applies the strict rules only.

\paragraph{Number tasks (LB-0038 and GSM8K).}
The strict rules: (i)~the whole answer is one number, with an optional unit word; (ii)~explicit statements (``answer is'', ``Answer:'', the marker \verb|####|, a boxed number) agree on one number, and a number followed by an arithmetic operator is not taken; (iii)~an age phrase that holds the only number of the answer.
The extended rules, for a stopped call only: (iv)~the strict rules without markdown and LaTeX; and, for GSM8K only, (v)~the answer holds exactly one distinct number; (vi)~the last non-empty line ends with an equals sign and a number or consists of one number with an optional currency sign or unit; (vii)~the last sentence holds exactly one distinct number and no arithmetic operator.
LB-0038 is read by the first four rules only, because the fifth had read laboratory values from answers that decline to give an age (decided after the pilot's parse audit).
The last number of prose is never taken for being last.

\paragraph{The role-break rule.}
A model may emit the start-of-turn marker of its chat template after its answer.
Each marker opens a turn whose role is the rest of its line.
A re-opened assistant turn is the model going on with its own answer: the marker and the role line are dropped and the text that follows is kept.
Any other turn (a user turn, a system turn or a line that names no role) ends the answer at its marker.
Two sensitivity analyses vary the rule: one removes the markers and keeps everything, the other cuts the content at the first marker.

\paragraph{The refusal classifier.}
It applies to the calls that ended with finish reason \texttt{stop} and were unparsed, that is, the non-committed calls, and it puts each in the first of five classes that fits.
(a)~\emph{No answer or refusal}: the final text is empty, or it matches a pattern of a keyword list, namely an apology or an inability (``I am sorry'', ``I cannot'', ``I do not know''), a statement that the answer cannot be determined, derived or given (``cannot be determined'', ``not derivable'', ``not provided'', ``does not contain'', ``insufficient information'', ``not enough data''), ``none of the options'', ``unknown'', ``unanswerable'', ``indeterminate'', ``without additional context'', ``decline'' or ``apologize''.
For a letter task the next classes are (b)~\emph{several candidate letters} (two or more different letters or option texts), (c)~\emph{a form the scorer does not read} (one candidate, as an option text instead of its letter, inside markup or inside prose) and (d)~\emph{other} (only a letter outside the options, or no candidate).
For a number task they are (b)~\emph{a range or several numbers} (a range such as ``a to b'' or ``between a and b'', or two different numbers on the answer line), (c)~\emph{a form the scorer does not read} (one number inside markup, with units or words around it, in plain form, or given only in words) and (d)~\emph{other} (no number on the answer line).
The non-commitment share is the share of the cell's analysed calls that are non-committed, and the refusal share is the share of the cell's analysed calls in the first class.
The keyword list is a heuristic: a refusal in other words falls into another class, and a harmless use of ``unknown'' falls into the first.
The classifier reads counts only, and no output text was read for this paper.

\paragraph{What the classes do not tell.}
The harness texts of the second study's clean-start searches tell the model not to infer survival beyond the snapshot data, or to conclude that survival cannot be determined when follow-up data are missing (Appendix~\ref{app:gepa14}), and the harnesses still answer ``Yes'' in almost all of their parsed mortality answers.
Because the outputs were not read, why they do is not examined here.

\section{The third study: preregistration and freeze}
\label{app:third-prereg}

This appendix gives the third study's preregistration in summary and records its freeze; the study's text is in Sections~\ref{sec:third} to~\ref{sec:dev3}.
As for the earlier studies, every path, commit and SHA256 value is that of a file in our repository, which is not public at present.

\paragraph{The frozen record.}
In our repository the preregistration (\texttt{docs/icml\_ext3/PREREG\_ICML\_EXT3.md}) was frozen at \V{e3.s3.time}~UTC on \V{e3.s3.date}, in commit \texttt{\V{e3.s3.commit}}.
Its SHA256 is
\begin{center}\small\texttt{\V{e3.s3.hash.full}}\end{center}
The file \texttt{docs/icml\_ext3/PREREG\_ICML\_EXT3.sha256} holds the hash and the time, and the addenda are in \texttt{docs/icml\_ext3/PREREG\_ICML\_EXT3\_ADDENDA.md}.
As for the earlier studies, the freeze time is recorded only in our private repository and is not independently timestamped; the hash lets a reader check a copy of the preregistration.
The freeze also pinned every script of the study by SHA256 in a manifest (\texttt{docs/icml\_ext3/EXT3\_FILES.sha256}), and the analysis stops if a pinned file differs from it; the later amendments of the manifest are named in the addenda (Appendix~\ref{app:third-deviations}).

\paragraph{What the preregistration fixes.}
It fixes the two models and the serving stack of the earlier studies; the list of candidate benchmarks and the rule for admitting each; the grouping of rows and the seeded, hash-based partitions, with the replication and test rows sealed and opened only after the freeze; the constant label of a benchmark (the label with the largest count in its search partition, a tie going to the first label in a fixed order); the eligibility rules and the screening rule of Section~\ref{sec:third}; the clean-start GEPA arm with two seeds and its selection rule; hypotheses H1, H2 and H5 with their estimands, margins, families (\V{e3.fam.h1}, \V{e3.fam.h2} and \V{e3.fam.h5}), interval rules and the statuses of every cell; the descriptive readings; the bootstrap (\V{e3.ana.b} resamples, with the first study's seed rule and percentile rule); the run plan with its stopping rules; and the rule that a repair which changes request bytes, parsing, scoring or call order needs a new namespace.
Every H1, H2 or H5 quantity of Autocast is exploratory by the same document (Appendix~\ref{app:third-autocast}).

\paragraph{What was stated in advance.}
Three descriptive expectations were written before the freeze: that MMLU-Pro would be \verdict{above the constant} for every harness; that the 14B pattern would read as H2 \verdict{fails} together with H5 \verdict{at the constant} for the same harness, and that where a harness did not show it the paper may not say that the harness reached the constant; and that the premise might fail, leaving no qualifying pair, in which case ``premise not met'' is a result.
The planned power of H5 was computed for assumed effects, assumed per-row spread and the measured sizes; its values are in the preregistration, the low power for a verdict of \verdict{at the constant} was accepted, and no size was changed.
The study was written after the first two studies' results were known and is not blind to the direction of the pattern it tests.

\paragraph{The outputs.}
The report of the preregistered analysis is committed, as aggregates, in \texttt{docs/icml\_ext3/ANALYSIS\_20261028/}; its SHA256 is
\begin{center}\scriptsize\texttt{\V{e3.ana.sha}}\end{center}
and it records the bootstrap size \V{e3.ana.b} and the pin of the main item files, which begins \texttt{\V{e3.items.sha8}}.
The agreement with the earlier studies' outputs on MMLU-Pro was computed after the outputs existed and is committed in the same folder (its SHA256 begins \texttt{\V{e3.agr.sha8}}); the report's own field for it is empty, because the analysis ran without the earlier outputs.
The account of the results (\texttt{docs/icml\_ext3/PART2\_REPORT.md}) and the screening account (\texttt{docs/icml\_ext3/STAGE1\_REPORT.md}, \texttt{docs/icml\_ext3/STAGE0\_REPORT.md}) are committed too.
The run logs hold benchmark text and model output; they are not committed and will not be released.

\paragraph{The order of the run.}
Stage 0 made no model call: it downloaded the candidates, grouped and partitioned them and counted rows and search-partition labels.
Stage 1 ran on search-partition rows only: the screening, a pilot of the kept candidates, the start checks and a rehearsal of the chain on search rows and mock records.
The freeze followed.
Part 1 then ran the searches, the replication grid, the harness evaluations on the replication partition and the verification of the rescue, and wrote the checkpoint record, which locks the selected harnesses and was committed alone before the first test call.
Part 2 ran the test grid, the harness evaluations on the test partition, the determinism recheck and the analysis.

\section{The third study: deviations and addenda}
\label{app:third-deviations}

The preregistration of the third study allows a change only by a written addendum, and a repair that changes request bytes, parsing, scoring or call order needs a new namespace; none was needed.
This appendix lists the decisions before the freeze and every addendum after it.

\subsection{Before the freeze (D1--D6)}
Six deviations from the screening specification were recorded before the freeze, each before the step it concerns and none after a screening result existed.
\begin{description}
\item[D1] The first run of the grouping script on the Autocast archive read a macOS resource-fork file in place of the data file; the script now takes the data file by its exact name, and its test holds the decoy files.
\item[D2] Autocast fell short of the size rule by \V{e3.s0.au.short} rows, and we decided, from the counts alone and before any model call, to run it as a declared exploratory arm outside every family (Section~\ref{sec:third} and Appendix~\ref{app:third-autocast}); the confirmatory rule stands and the minimum was not lowered.
\item[D3] The specification fixed no seeds for MMLU-Pro, the control, which has no screening and no partition of its own; its lists are content-free seeded draws of question ids from the first study's sets, written before any call.
\item[D4] An Autocast prompt is the question text and the instruction to answer yes or no, and nothing else, after a reading of search-partition rows found that the background field of one question stated its outcome.
\item[D5] The script that computes the screening results in one file was added after stage 1 had made its calls and before any screening result had been computed; it adds checks and re-implements no rule.
\item[D6] The freeze manifest was extended to pin the scripts that run after the freeze, and the guard that reads the manifest was corrected: it had split each line at its first white space, so every pinned file counted as absent in a check that was not strict (addendum A3 records that no earlier check had passed through the defect).
\end{description}

\subsection{After the freeze (A1--A\V{e3.add.last})}
Table~\ref{tab:third-addenda} lists every addendum with the effect it states.
A8 and A9 are not entries; they are the numbers of two amendment manifests that list every pinned file as it stood after a supplement to A7.
The four that bear on a result (A5, A7, A10 and A13) are described in Section~\ref{sec:dev3}.
Two further facts belong here.
A12 gave standing permission to repeat a run once, with identical requests, after a failure that the request bytes did not cause, and a second failure stops the run.
The note to A10 records that the repeat of the run which stopped at gate~(b) reproduced the stopped attempt's whole candidate sequence, although A10 had said a restart need not; the repeat then went on to \V{e3.run.bs.fp.s1.evald} distinct candidate texts with a task call and completed, and the candidate that tripped the gate was not selected.
Every failed attempt is kept aside and excluded from every analysis.

\begin{table}[H]
\caption{The third study's addenda after the freeze, with each entry's own statement of its effect on requests, scoring and verdict rules (a dash where the entry records an event and states no effect).}
\label{tab:third-addenda}
\centering
\input{tables/t3_app_addenda}
\end{table}

\section{The third study: stage 0 and the screening}
\label{app:third-screen}

\paragraph{Stage 0.}
Stage 0 made no model call and read no label of a replication or test row.
A pinned script grouped the rows by a hash of their shared text (the BoolQ passage, the FPB sentence, the Autocast question), assigned each group to a partition by a seeded hash so that no group spans two partitions, and counted rows, groups and token lengths.
For BoolQ the official train rows are the search partition and the dev rows go to replication and test, groups first: \V{e3.s0.bq.moved} dev rows whose group also holds a train row went to search.
The context rule dropped no row of any candidate (the row limit is \V{e3.s0.bq.limit} tokens).
Table~\ref{tab:third-stage0} gives the sizes, groups, design effects, constants and terms.
The licences and terms are those the sources state: BoolQ \V{e3.s0.terms.bq}, FPB \V{e3.s0.terms.fp}, Autocast \V{e3.s0.terms.au} and MMLU-Pro MIT; \smol\ is licensed \V{e3.lic.sm} and \basem\ \V{e3.lic.bs}.
The Autocast release holds \V{e3.s0.au.records} records, not the number its paper gives: \V{e3.s0.au.q.tf} true/false, \V{e3.s0.au.q.mc} multiple-choice and \V{e3.s0.au.q.num} numerical questions, of which \V{e3.s0.au.st.resolved} are resolved; the true/false questions with a resolution number \V{e3.s0.au.considered}, and the hash gave them a test partition of \V{e3.s0.au.T.rows} rows.
MRPC, the fourth candidate, was not downloaded, because no terms of use were stated for it; it was not screened.

\begin{table}[H]
\caption{Stage 0: rows and groups of each partition (search, replication, test), the largest group and the design-effect bound of the test partition, the constant label with its share of the search partition, the terms, and the test rows against the minimum. MMLU-Pro has the first study's partitions and its unit is the item.}
\label{tab:third-stage0}
\centering
\input{tables/t3_app_stage0}
\end{table}

\paragraph{Screening.}
The screening ran \cfg{AUTHORS-1500} on \V{e3.sc.bq.sm.n} search rows of each eligible candidate with each model, and applied the preregistered rule once to the first complete attempt (Table~\ref{tab:third-screen}).
The eligible candidates were BoolQ and FPB and both were kept.
The ranking width, a deterministic tie-break that carries no claim about precision, was \V{e3.sc.bq.width} for BoolQ, \V{e3.sc.fp.width} for FPB and \V{e3.sc.au.width} for Autocast, and it decided nothing, because exactly two candidates were eligible.
Autocast's two pairs also qualified, and its result affects no choice.
The start checks compared outputs at concurrency one and at concurrency \V{d.conc} with descending order, on the first rows of each candidate, in both configurations and for both models: \V{e3.sc.sm.g2b.c1.th.same} of \V{e3.sc.sm.g2b.c1.th.n} calls were identical in each comparison, and so were \V{e3.sc.sm.shared.same} of \V{e3.sc.sm.shared.n} calls of the repeat at the GPU share used with the proposer.
Gate~(b) found no trip in the screening of either model.

\begin{table}[H]
\caption{The screening: the constant label and its share of the search partition, and for each model the number of correct answers among the \V{e3.sc.bq.sm.n} screening rows, the accuracy, the number of rows that carry the constant label (a pair qualifies if the correct count is at most that number), the parse and truncation counts of the rows, the non-committed calls and the calls with conflicting answers. Autocast is outside the ranking.}
\label{tab:third-screen}
\centering
\input{tables/t3_app_screen}
\end{table}

\section{The third study: all endpoints}
\label{app:third-endpoints}

Every number in this appendix is read from the analysis report committed in our repository (Appendix~\ref{app:third-prereg}).
Confirmatory endpoints carry the adjusted interval of their family (levels \V{e3.level.h1}, \V{e3.level.h2} and \V{e3.level.h5}), descriptive endpoints and every accuracy, parse rate and truncation rate the unadjusted \V{c.ci} interval; exploratory Autocast rows carry unadjusted intervals and no verdict.
A star marks an H2 endpoint whose selected harness is the seed, so that $\theta$ is exactly zero and \verdict{holds} is trivial.
Points are fractions of the rows: $\Delta$, $\theta$ and $\Delta_c$ are differences of accuracies.

\subsection{Verdicts on both partitions}
\begin{table}[H]
\caption{All \V{e3.cnt.n.all} endpoints of the third study: family, model, benchmark, status, the verdict on the replication and the test partitions, and the replication label. Autocast has no verdict and no label (a dash).}
\label{tab:third-endpoints}
\centering
\input{tables/t3_app_endpoints}
\end{table}

\subsection{H1}
\begin{table}[H]
\caption{H1, test partition: the gain $\Delta$ of \cfg{AUTHORS-1500} over \cfg{THINK-512} with its interval, its parts $\Delta_V$, $\Delta_B$ and $\Delta_L$, the statistic $T$ with its interval, the numbers of rows only \cfg{AUTHORS-1500} parses ($V$), both parse ($B$) and only \cfg{THINK-512} parses ($L$), the verdict and the visibility share $\Delta_V/\Delta$. The interval is the adjusted one for confirmatory cells and the unadjusted one for MMLU-Pro (descriptive) and Autocast (exploratory).}
\label{tab:third-h1te}
\centering
\input{tables/t3_app_h1num_te}
\end{table}

\begin{table}[H]
\caption{H1, replication partition, in the form of Table~\ref{tab:third-h1te}.}
\label{tab:third-h1re}
\centering
\input{tables/t3_app_h1num_re}
\end{table}

\subsection{H2}
\begin{table}[H]
\caption{H2: $\theta$, the accuracy of \cfg{AUTHORS-1500} minus that of the selected harness, on the test and replication partitions with its interval (adjusted for the twelve confirmatory endpoints, unadjusted for Autocast), the verdicts and the label. A negative $\theta$ means the harness is more accurate.}
\label{tab:third-h2}
\centering
\input{tables/t3_app_h2num}
\end{table}

\subsection{H5}
\begin{table}[H]
\caption{H5: $\Delta_c$, the accuracy of the selected harness minus that of the search partition's constant label, with the interval the report used (adjusted, family of \V{e3.fam.h5}, for confirmatory endpoints; unadjusted for descriptive and exploratory ones), the verdicts and the label.}
\label{tab:third-h5}
\centering
\input{tables/t3_app_h5num}
\end{table}

\subsection{Configurations}
Tables~\ref{tab:third-cfg-te-sm} to~\ref{tab:third-cfg-re-bs} give accuracy with its unadjusted interval, the parse rate, the truncation rate (finish reason \texttt{length}) and the non-committed rate for every configuration, harness and constant label of each model and partition.
The constant label is the one chosen on the search partition.

\begin{table}[H]
\caption{\smol, test partition.}
\label{tab:third-cfg-te-sm}
\centering
\input{tables/t3_app_cfg_te_sm}
\end{table}

\begin{table}[H]
\caption{\basem, test partition.}
\label{tab:third-cfg-te-bs}
\centering
\input{tables/t3_app_cfg_te_bs}
\end{table}

\begin{table}[H]
\caption{\smol, replication partition.}
\label{tab:third-cfg-re-sm}
\centering
\input{tables/t3_app_cfg_re_sm}
\end{table}

\begin{table}[H]
\caption{\basem, replication partition.}
\label{tab:third-cfg-re-bs}
\centering
\input{tables/t3_app_cfg_re_bs}
\end{table}

\section{The third study: the GEPA runs and the selected harnesses}
\label{app:third-runs}

\paragraph{The runs.}
There are \V{e3.cnt.runs} runs: two models, four benchmarks (Autocast included) and two search seeds.
Table~\ref{tab:third-runs} gives, for each, the pool of candidates GEPA accepted (the seed and the proposals accepted on their minibatches; \V{e3.cnt.pool.min} to \V{e3.cnt.pool.max} in the \V{e3.cnt.runs} runs), the number of candidate texts that received at least one task call, on a minibatch or on the validation split, the proposals, the metric calls, whether the selected harness is the seed, outcome flag (ii) as the report computes it and as the preregistration defines it, and the gate~(b) count.
The task server was preempted from \V{e3.run.bs.fp.s1.pre} to \V{e3.run.sm.mm.s1.pre} times in a run, never as an error; the preemptions of a run that was repeated are those of the attempt that completed.
The proposer's server had none.
None of the selected harnesses holds a soft switch (\texttt{/think} or \texttt{/no\_think}).
Three runs completed on a repeat: \smol\ on BoolQ with seed zero (after the stop of A4 and the change of A5), \smol\ on FPB with seed zero (the one retry that A7 allows, after an out-of-memory failure) and \basem\ on FPB with seed one (after the stop at gate~(b) that led to A10).

\begin{table}[H]
\caption{The \V{e3.cnt.runs} GEPA runs (search rows only). Pool: the seed and the proposals accepted. Candidate texts with a task call: the distinct candidate texts that have at least one thinking-off task call in the run's response records, counted after the outputs existed (not the number of full validation passes). Flag (ii) is ``no accepted proposal'': the first column is the value in the report, which the code sets when the selected harness is the seed (a star), and the second the value under the preregistration's definition, that the run accepted no candidate. Gate (b): trips among the run's thinking-off calls.}
\label{tab:third-runs}
\centering
\input{tables/t3_app_runs}
\end{table}

\paragraph{What the column of candidate texts counts.}
The column counts every distinct candidate text with at least one thinking-off task call in a run's response records, whether the call was on a minibatch of \V{gp.mb} rows (which a proposal receives before it is accepted or rejected) or on the validation split.
The counts come from the call keys, which carry the first twelve characters of the candidate's hash, and they are the ``candidates in the folder'' of Table~\ref{tab:third-gateb}; they were computed after the outputs existed.
It is not the number of candidates that received a full validation pass.
Such a pass is one call for each row of the validation half of the \V{k.gepa.rows}-row sample, and this column times the validation rows exceeds the run's metric calls in every run.
The call budget rules out reading this column as the number of candidates that received a full validation pass: it includes candidates evaluated on a minibatch only, and the reports do not record full-validation counts for every run.
In every run more candidate texts received a call than entered the pool.
The pool is the nearest recorded count, because the selection compares the validation scores of the pool's members, and the one run for which a report gives both agrees with it: addendum A5 records \V{e3.a5.full} full validation evaluations for the \V{e3.run.sm.bq.s0.pool} candidates of the first run, which is its pool.
For the other runs we therefore give the pool and no separate count.

\paragraph{The four runs whose harness is the seed.}
Table~\ref{tab:third-a13} gives the four runs of addendum A13 with both facts: the pool held \V{e3.cnt.pool.seed.min} to \V{e3.cnt.pool.seed.max} candidates, and the seed's validation score was the highest, strictly, so the selection rule returned the seed because no other candidate beat it and not because no proposal was accepted.
In these runs the selected harness equals \cfg{AUTHORS-1500}, so $\theta$ is zero on both partitions and the verdict \verdict{holds} is trivial.
No verdict, interval, label, family or exclusion changes with the flag, which only describes the run.

\begin{table}[H]
\caption{The four runs in which the selected harness is the seed: pool, proposals, validation scores in points (the seed and the best other candidate), the flag as the report computes it and under the frozen definition, $\theta$ and the H2 verdicts on test and replication.}
\label{tab:third-a13}
\centering
\input{tables/t3_app_a13}
\end{table}

\paragraph{The harnesses.}
Each model has one seed harness for each of the four benchmarks (an empty system prompt and no instruction), eight in all; for every one, the rendered requests equal those of \cfg{AUTHORS-1500} on every row of the benchmark's \V{k.gepa.rows}-row sample, text and token ids, before each run.
Table~\ref{tab:third-harness} describes the \V{e3.cnt.runs} selected harnesses by what the committed report records: the parse rate on the test partition, the share of parsed answers equal to the majority label on both partitions, whether that reaches the collapse threshold, whether the harness is the seed or equals \cfg{AUTHORS-1500}, the flag as the report computes it and the first twelve characters of the candidate's hash.
The two \smol\ harnesses on MMLU-Pro are the same text.
The twelve selected texts that are not the seed are described, and one is reproduced, in Appendix~\ref{app:third-gepa}; why a harness answers as it does is not examined.

\begin{table}[H]
\caption{The selected harnesses: parse rate on test, majority share on test and replication, label collapse (test, replication; the threshold is \V{e3.collapse.thr}\% of parsed answers), whether the harness is the seed, whether it is identical to \cfg{AUTHORS-1500}, flag (ii) as the report computes it, and the candidate's hash.}
\label{tab:third-harness}
\centering
\input{tables/t3_app_harness}
\end{table}

\paragraph{Gate (b) by run.}
Gate~(b) recorded \V{e3.gb.said.trip} trips in \V{e3.gb.said.calls} thinking-off task calls of the completed searches (\V{e3.gb.share}), in two runs of the base model (Table~\ref{tab:third-gateb}).
In \basem\ on FPB with seed one, \V{e3.gb.rep3} of the \V{e3.gb.rep153} thinking-off calls of candidate \texttt{\V{e3.gb.cand1}} began with the model's think-end token; in \basem\ on MMLU-Pro with seed one, the trips are in candidate \texttt{\V{e3.gb.cand2}}.
Neither candidate was selected (the selected candidates are \texttt{\V{e3.gb.cand1.sel}} and \texttt{\V{e3.gb.cand2.sel}}), so no evaluated harness carries a trip.
The \V{e3.gb.sel.re.calls} thinking-off calls of the selected harnesses on replication and the \V{e3.gb.sel.te.calls} on test had none, and neither had the \V{e3.gb.grid.calls} calls of the grid, rescue and determinism runs.
The failed attempts, kept aside and excluded, hold \V{e3.gb.fail.trip} trips in \V{e3.gb.fail.calls} calls, those of the candidate of addendum A10.
The counts were computed after the outputs existed from the calls' flags and keys only.

\begin{table}[H]
\caption{Gate (b) in the searches, the evaluations of the selected harnesses, and the other runs: thinking-off calls, trips and their share, and the candidates of each search folder; computed after the outputs existed.}
\label{tab:third-gateb}
\centering
\input{tables/t3_app_gateb}
\end{table}

\section{The third study: the Autocast arm in full}
\label{app:third-autocast}

Autocast \citep{zou2022autocast} is an exploratory arm.
Its test partition has \V{e3.sc.au.ntest} rows against the preregistered minimum of \V{e3.s0.min.test}, a shortfall of \V{e3.s0.au.short}, so it was not eligible; its other checks passed (a search partition of at least \V{e3.s0.min.search} rows, a replication partition of at least \V{e3.s0.min.rep}, no row lost to the context rule, a majority share of \V{e3.s0.au.share}\% in the search partition and \V{e3.s0.au.k} of the \V{e3.sc.au.sm.n} screening rows carrying the constant label, inside the allowed range).
The confirmatory rule stands and the minimum was not lowered.
Under deviation D2 the arm ran by the same procedure as a kept benchmark, with every H1, H2 and H5 quantity exploratory: an estimate with an unadjusted \V{c.ci} interval, no verdict, no replication label, no place in any family, and calls that follow every confirmatory call of a step.
Its results change no family size or level and no choice of benchmark.

\paragraph{The data.}
The prompt is the question text and the instruction to answer yes or no (deviation D4).
The search partition has \V{e3.s0.au.S.rows} rows, the replication partition \V{e3.s0.au.R.rows} and the test partition \V{e3.s0.au.T.rows}, in groups of at most \V{e3.s0.au.T.largest} questions.
The constant label is ``no'': \V{e3.lab.te.au.lc.no} of the test rows and \V{e3.lab.re.au.lc.no} of the replication rows carry it, against \V{e3.lab.te.au.lc.yes} and \V{e3.lab.re.au.lc.yes} that carry ``yes''.
The questions resolved before the models were trained, so an answer may come from memory of the outcome and not from forecasting; the arm measures answering from the question text alone, and a gain or a collapse there says nothing about forecasting skill.

\paragraph{The result.}
Table~\ref{tab:third-autocast} gives accuracy, $\Delta_c$ with its unadjusted interval and the share of parsed answers equal to ``no'', on both partitions.
\cfg{THINK-512} is truncated on \V{e3.cfg.te.sm.au.t512.trunc.pp} (\smol) and \V{e3.cfg.te.bs.au.t512.trunc.pp} (\basem) of the test calls, and \cfg{AUTHORS-1500} reaches \V{e3.cfg.te.sm.au.aut.acc.pp} and \V{e3.cfg.te.bs.au.aut.acc.pp}, against \V{e3.cell.te.sm.au.oracle.pp} for the constant.
The four selected harnesses answer ``no'' on \V{e3.rng.au.mj.te.min}\% to \V{e3.rng.au.mj.te.max}\% of their parsed test answers and \V{e3.rng.au.mj.re.min}\% to \V{e3.rng.au.mj.re.max}\% of their parsed replication answers, so all \V{e3.cnt.coll.au.of} cells reach the collapse threshold, and their accuracy is within \V{e3.rng.au.dc.te.max} points of the constant on test.
On replication the largest $\Delta_c$ is \V{e3.rng.au.dc.re.max} points (\basem, seed zero).
H1 and H2 values for the arm are in Appendix~\ref{app:third-endpoints}, where the Autocast rows carry no verdict.
\smol\ with seed zero parsed \V{e3.hn.te.sm.au.s0.parse.pp} of the test rows and the other three harnesses at least \V{e3.hn.te.sm.au.s1.parse.pp}.

\begin{table}[H]
\caption{Autocast, exploratory: for each partition and model, the accuracy of the constant label ``no'', of \cfg{THINK-512}, of \cfg{AUTHORS-1500} and of the two selected harnesses, $\Delta_c$ of each harness with its unadjusted interval, and the share of the harness's parsed answers that are ``no''. No verdict or label is attached.}
\label{tab:third-autocast}
\centering
\input{tables/t3_app_autocast}
\end{table}

\section{The third study: determinism and agreement with the earlier studies}
\label{app:third-checks}

\paragraph{Determinism recheck.}
After the test partition, \cfg{THINK-512} and \cfg{AUTHORS-1500} were rerun on a fixed list of \V{e3.det.rows} test rows of each model on a fresh server, and compared with the test grid token for token, finish reasons included (Table~\ref{tab:third-det}).
All \V{e3.det.calls} calls of each model were identical, with none differing or missing; the result is per model, the two models being rerun separately, and neither rerun had a preemption.
The two comparison files are byte for byte equal (SHA256 beginning \texttt{\V{e3.det.sha8}}) because each holds only counts, and both models gave the same counts; the equality does not show that either comparison ran, and the evidence that each did is the log lines of the chain, the \V{e3.det.calls} calls in each rerun's own log and the report's flag.
The preflight at the start of the test partition showed the same GPU identifier as part 1's, so the preregistered recheck for a change of GPU was not triggered.

\begin{table}[H]
\caption{The determinism recheck: calls compared with the test grid, calls identical, and the report's flag.}
\label{tab:third-det}
\centering
\input{tables/t3_app_det}
\end{table}

\paragraph{Agreement with the earlier studies on MMLU-Pro.}
MMLU-Pro was called again for both models and both configurations, so that each harness and its seed are paired on one stack, and the earlier outputs were kept for a descriptive comparison.
It was computed after the outputs existed: the preregistered analysis code has the comparison for the test partition, but the report's field for it is empty because the analysis ran without the earlier runs, and we computed it afterwards from the earlier and the new first calls with the pinned scorer, on the rows both called, adding the replication partition to the test-only pinned reading.
``Agree'' is the correctness agreement, the share of rows where the two correctness readings are equal, and ``same label'' the label agreement, the share where the parsed answer letter, or its absence, is equal; neither compares tokens or response bytes.
Table~\ref{tab:third-agree} gives the comparison.
The earlier side is the first study's run for \smol\ and the second study's base-model arm for \basem.

\smol's parsed labels and correctness readings are the same on all rows, on both partitions and both configurations (\V{e3.ag.te.sm.au.nag} of \V{e3.ag.te.sm.au.rows} test rows and \V{e3.ag.re.sm.au.nag} of \V{e3.ag.re.sm.au.rows} replication rows for \cfg{AUTHORS-1500}, and the same for \cfg{THINK-512}), and the accuracies and truncation counts are equal; this does not show equal tokens.
The first study and this one ran on the same kind of GPU, an \V{e3.gpu.old}.
\basem's are not: for \cfg{AUTHORS-1500} the correctness agreement is \V{e3.ag.te.bs.au.agree.pp} on test rows and \V{e3.ag.re.bs.au.agree.pp} on replication rows, with \V{e3.ag.te.bs.au.nc} rows correct only in the new run and \V{e3.ag.te.bs.au.oc} only in the old on test, and the label agreement is \V{e3.ag.te.bs.au.same.pp} and \V{e3.ag.re.bs.au.same.pp}; \cfg{THINK-512}, truncated on almost every row in both runs, agrees on \V{e3.ag.te.bs.th.agree.pp} and \V{e3.ag.re.bs.th.agree.pp}.
The accuracy of \cfg{AUTHORS-1500} differs by \V{e3.ag.te.bs.au.oacc.pp} against \V{e3.ag.te.bs.au.nacc.pp} (old and new) on test and by \V{e3.ag.re.bs.au.oacc.pp} against \V{e3.ag.re.bs.au.nacc.pp} on replication.
The second study ran on \V{e3.gpu.new} GPUs, which is consistent with the cross-GPU disagreement that the preregistration warned of; we did not test it, and the equality of \smol's rows shows that the scoring itself reproduces.

\begin{table}[H]
\caption{MMLU-Pro, the earlier run against the new run (computed after the outputs existed): rows, agreement of the correctness readings, the four correctness cells, agreement of the parsed label, accuracy and truncated calls (old and new).}
\label{tab:third-agree}
\centering
\input{tables/t3_app_agree}
\end{table}

\addtocounter{section}{-\value{section}}
\renewcommand{\thesection}{A\Alph{section}}
\section{The third study: the selected harness texts}
\label{app:third-gepa}

\paragraph{What the texts are.}
\intern, a small language model, wrote every harness text of the third study's GEPA arm.
It read search rows only, and the feedback it received for each answer was whether the answer was correct.
We did not check the statements in the texts, some of which look wrong, and the texts are not guidance for anyone.
Of the \V{e3.cnt.runs} runs, twelve selected a text other than the seed; the other four (addendum A13, Table~\ref{tab:third-a13}) selected the seed, an empty system prompt and no instruction, so they show no text.
The order was this.
Two of the Autocast texts were read before the lexical specification was written.
The specification was fixed before any count of the lexical check was computed.
We read all twelve texts, those two included, after the counts, and the readings are logged in the third study's contact log.
To keep item content out of this paper we describe eleven of them by their length, the kinds of statements they make and their counts, and reproduce the one text for which the lexical check below flags no row of any partition.

\paragraph{Checks run before any text was read.}
For each of the twelve texts, none contains an item key or an item pattern, and none contains a stretch of \V{gp.copy.run} characters of any row's user message in the search, replication or test partition of the four benchmarks, exactly or after lower-casing and collapsing white space; the candidate's hash equals the record's.
The copy rule does not cover short values: the twelve texts hold \V{e3.ov.dec.total} decimal numbers in all (the FPB texts and one MMLU-Pro text), and all but one of them also occur in at least one row of the evaluation partitions.

\paragraph{The lexical check (computed after the outputs existed; descriptive; not preregistered).}
The copy rule does not catch a text that states what a row says in other words.
GEPA's feedback reveals the correct label of a binary task, so a search can write row-derived statements into a text.
We therefore ran a lexical check that flags, for each text and each partition of its own benchmark, the rows that share a string with the text.
It compares sets of strings and reads no meaning, and it is not named-entity recognition: a flag shows a shared string, not shared content or a copy, and a row that is not flagged is not shown to be free of shared content.
A row is \emph{flagged} if it shares at least one item with the text, of four kinds: a capitalised word or an acronym, a date, a number, or a run of five words.
The flag of a row depends on the harness's own text.

\paragraph{What is compared, and what is left out.}
The text is the system prompt and the appended instruction joined by a line break.
A row's content is its prompt, the user message the model saw without any harness, with the benchmark's fixed template removed: the passage and the question for BoolQ, the sentence for FPB, the question for Autocast, and the question and the option lines without their letter labels for MMLU-Pro.
Every row of every partition had to match its template, a mismatch would have stopped the run, and none occurred.
Because the template is removed, the task's own instruction wording cannot create a match.
From the capitalised words and acronyms of the text and of the rows the check also drops the words of the benchmark's template, the label words (yes, no, true, false, positive, negative and neutral), the option letters A to J, and a fixed list of English function words (articles, prepositions, conjunctions, pronouns, auxiliaries, determiners and question words), the last only for capitalised words, so that an article that opens a title is not an entity.
There is no list of common words beyond these, so a capitalised common word inside a sentence counts as an entity.

\paragraph{Normalisation, numbers and dates.}
A word is a run of letters with internal hyphens and apostrophes kept; an initialism written with periods, such as U.S., is the acronym US.
A capitalised word (a capital, then a lower-case letter, at least three characters) is lower-cased, and an acronym (two or more capitals, once hyphens and apostrophes are removed) keeps its capitals, so the acronym US and the word us do not meet.
A capitalised word at the start of a sentence is kept only if the next word is also a capitalised candidate, so a single capitalised word that opens a sentence is missed; each word of a capitalised sequence is a separate entity.
Dates are read first, in this order, each matched span being removed before the next pattern and before numbers are read: an ISO date; a month name with a day and a year, in either order; a month name with a year; a month name with a day; and a four-digit number from \V{e3.ovs.y0} to \V{e3.ovs.y1}, read as a year unless it is part of a longer number or of a decimal.
A date that carries a year also counts as that year, so a shared year alone flags a row.
Dates written with slashes, relative dates, fiscal periods and two-digit years are not read, and ``May'' counts only inside a pattern with a day or a year.
Numbers are read from what is left: an integer of two digits or more and any number with a decimal part count; one-digit integers do not, because they would match nearly every row; a comma-grouped integer loses its commas and is a number, not a year; a trailing zero is kept, so two decimals that differ only in a trailing zero do not match; a percent sign or a unit is ignored; numbers written as words are not read.
For runs of five words, the text and the content are lower-cased, apostrophes are removed, and both are cut into runs of letters and digits; a run may cross a line break.
The check is broad: its entity rule over-counts shared content, and a run of five words is shared with at most \V{e3.ovr.all.g5max.n} rows of any partition.
It misses paraphrase and statements in plain lowercase words, in particular claims about events.

\paragraph{Empty and small subsets, intervals and the second comparison.}
A side with no row has no accuracy, no difference and no interval, and is shown with dashes; a side with fewer than thirty rows is shown with its numbers and interval and a dagger.
We compute no interval, and run no test, for the difference between the flagged and the not flagged side.
The intervals are the unadjusted \V{c.ci} intervals of the study's group bootstrap, conditional on the split (the flag is a function of the rows) and not adjusted for the twelve harnesses, the two partitions or the two sides.
The difference of a harness from \cfg{AUTHORS-1500} is computed as H2's $\theta$ is, on the same resampled groups, with the opposite sign, so that a positive value means that the harness is more accurate.
Table~\ref{tab:gtexts} describes the texts, Table~\ref{tab:goverlap} gives the counts, and Tables~\ref{tab:gb-test} and~\ref{tab:gb-rep} give, for the flagged and the not flagged rows of each harness, its accuracy, the constant's and \cfg{AUTHORS-1500}'s, and its difference from each with the unadjusted interval.

\paragraph{When the rule was fixed.}
The study's outputs and its analysis report existed before the check was written.
We had also read two of the Autocast texts in full, after the copy rule had passed on all \V{e3.cnt.runs} selection records and before any other text was read, and that reading is how we learned that the texts carry statements that look derived from rows.
The specification and the checker, with synthetic tests, were committed before the checker ran on the real files, and nothing in them changed after a count had been seen: the rule was fixed before the counts and after the outputs, and it was not preregistered, so every result of this appendix is descriptive and was computed after the outputs existed.
The comparison with \cfg{AUTHORS-1500} was added later, after the first counts had been seen, with its own specification committed before its numbers and the checker unchanged; its stop rule was that an interval of an FPB harness on the not flagged test rows that included zero would stop the work, and none did.
The checker's SHA256 begins \texttt{\V{e3.ov.chk}}, that of the first specification \texttt{\V{e3.ov.spec}}, that of the second \texttt{\V{e3.ova.spec}} and that of the second script \texttt{\V{e3.ova.script}}.
The first specification, that of the lexical check itself (\texttt{OVERLAP\_CHECK\_SPEC.md}), is an ancillary file of this arXiv version, byte for byte, and its SHA256 is the one that begins as given above; the second specification, the scripts and their tests are in the authors' repository, which is not public at present (Appendix~\ref{app:repro}).

\begin{table*}[t]
\caption{The twelve selected texts that are not the seed. \emph{Candidate}: index of the selected candidate in the search's list of candidates, and the length of that list. \emph{Characters} and \emph{Words}: of the text (system prompt and appended instruction). \emph{Decimals}: decimal numbers in the text. \emph{Contains}: the kinds of statements, from our reading, with no quotation.}
\label{tab:gtexts}
\centering
\input{tables/t3_app_gtexts}
\end{table*}

\begin{table*}[t]
\caption{Rows that the lexical check flags and does not flag for each text, by partition of the text's own benchmark (computed after the outputs existed; descriptive; not preregistered), and, on test, the rows flagged for a capitalised word or acronym (entity), a date, a number or a run of five words (a row can be flagged for several kinds).}
\label{tab:goverlap}
\centering
\input{tables/t3_app_goverlap}
\end{table*}

\begin{table*}[t]
\caption{Test partition: each harness's accuracy, the constant's and \cfg{AUTHORS-1500}'s on the rows the lexical check flags and does not flag for its own text, the harness's difference from the constant and from \cfg{AUTHORS-1500} (columns headed Minus) in points with its unadjusted \V{c.ci} interval, and the share of parsed answers that equal the constant label (computed after the outputs existed; descriptive; not preregistered). No interval is given for the difference between the two sides. A dagger marks fewer than thirty rows and a dash a side with no row.}
\label{tab:gb-test}
\centering
\input{tables/t3_app_gb_te}
\end{table*}

\begin{table*}[t]
\caption{Replication partition: as Table~\ref{tab:gb-test}.}
\label{tab:gb-rep}
\centering
\input{tables/t3_app_gb_re}
\end{table*}

\paragraph{The one text shown.}
The text of base Qwen3-1.7B on BoolQ with GEPA seed one has no row in the search, replication or test partition that the lexical check flags, and no copied stretch.
It is a system prompt with no appended instruction.
It is reproduced as selected, wrapped for display, a character outside ASCII written \verb|[U+XXXX]|.
The candidate's SHA256, which equals the record's and the committed report's, cannot be recomputed from this page because of the wrapping.
The lexical check does not see lowercase claims, so we read the text as well and found no statement of fact about a person, an organisation, an event, a date or a figure; it holds two short quoted phrases of legal wording as examples of a condition.
Whether these two phrases came from a search row, the lexical check cannot tell.
Intern-S1-mini wrote it, and it is not guidance.

\begin{minipage}{\linewidth}
\subsection*{\basem, BoolQ, GEPA seed one}
SHA256
\begin{center}\scriptsize\texttt{5171d6a6dfb263e7f95241ca83f4aac01decc76c3cb98d188fca3e1bc1bbd26d}\end{center}
\paragraph{System prompt.}
\begin{small}
\begin{verbatim}
When responding to yes/no questions based on provided text, you must strictly answer
"Yes" or "No" without additional explanation, reasoning, or elaboration. Your response
should directly reflect factual assertions from the input text alone, even if nuanced
conditions (e.g., "reasonable suspicion" requirements) are mentioned. If the input text
explicitly states a condition that negates a general affirmative (e.g., "not required
unless..."), the answer must be "No" for general queries. Avoid inferring beyond the
text. For example, if the text clarifies that a legal requirement applies only under
specific circumstances not implied by the question, your answer must reflect that
limitation. Always adhere strictly to the input's explicit claims, ignoring external
knowledge or assumptions.
\end{verbatim}
\end{small}
\paragraph{Appended instruction.} Empty.
\end{minipage}
\medskip

\section{The third study: prompts and answer readers}
\label{app:third-scoring}

The prompts and the answer readers of the third study were fixed before the screening, in Section~\V{e3.prereg.sec.prompts} of its preregistration, which is an ancillary file of this arXiv version (\texttt{PREREG\_ICML\_EXT3.md}; Appendix~\ref{app:third-prereg} gives its hash).
That section states the label sets, the two reading layers, the treatment of conflicts, refusals and labels that occur only in reasoning, the mapping of ``true'' and ``false'', and the wording of each instruction.
It does not state the layout of the templates (the blank lines and the label before a BoolQ question), which a pinned script and a specification fix and which are not ancillary files, so this appendix prints the templates with placeholders; it shows the rules with synthetic answers only, and no benchmark text appears in it.

\paragraph{The prompts.}
Each item is one user message, and there is no system message; a seed harness's empty system prompt also produces none, so its request equals \cfg{AUTHORS-1500}'s.
The templates, in which a blank line separates the parts, are as follows.
BoolQ: \verb|{passage}|, then \verb|Question: {question}|, then \verb|Answer with Yes or No.|
FPB: \verb|{sentence}|, then, on one line, \verb|Is the sentiment of this sentence for an investor negative, neutral or positive?| and \verb|Answer with one word.|
Autocast: \verb|{question}|, then \verb|Answer with Yes or No.|, with the question field alone and no background field (deviation D4 of Appendix~\ref{app:third-deviations}).
MMLU-Pro uses the first study's prompt.
A searched harness adds a system prompt and an instruction appended to the user message (Appendix~\ref{app:third-gepa}).

\paragraph{What \cfg{AUTHORS-1500} means here.}
In the third study \cfg{AUTHORS-1500} is the first study's configuration applied unchanged: thinking off, an answer allowance of \V{k.ans} tokens and temperature zero, which the first study took from its benchmark authors' serving protocol.
The authors of BoolQ, FPB, Autocast and MMLU-Pro did not supply it, and the prompts above are the ones we fixed in the preregistration; the name is historical.

\paragraph{The readers.}
The label set is yes and no for BoolQ and Autocast, and negative, neutral and positive for FPB; for the two yes-or-no tasks ``true'' becomes yes and ``false'' becomes no before anything else is checked.
Reading is case-insensitive and applies to the visible answer, the text after any reasoning, so a label that occurs only in the reasoning is not read.
The strict layer is applied to every call.
(a)~If the whole visible answer is one label, with punctuation, quotes, bold markers or a full stop around it, that label is the answer.
(b)~Otherwise the explicit answer statements (``the answer is X'', ``Answer: X'', a boxed label) are collected: if they all name one label, that is the answer, and if two of them name different labels the call is a conflict and unparsed.
The extended layer applies only to a call that stopped of its own accord (finish reason stop, not length) and only when the strict layer found neither a label nor a conflict.
(c)~The strict layer is applied again after markdown and LaTeX wrappers are removed.
(d)~If that finds nothing, an answer that opens with a label word followed by punctuation, a space or its end is read as that label.
Nothing else is read.
A label word elsewhere in prose, an empty answer, a refusal and a label that occurs only in the reasoning are unparsed, and an unparsed call is wrong.
The rule (d) has a known limit: an answer that opens with ``No'' in the sense of ``no one'' is read as no.
The preregistration gives the reason for not reading the last label word of an answer: that rule would count an answer such as ``Yes. There is no evidence that\ldots'' as a conflict and so lower the accuracy of wordy answers by the reader alone.
The strict layer by itself is the strict scorer, the preregistration's descriptive sensitivity analysis.

\paragraph{The pinned reader and synthetic examples.}
The reader is \texttt{scripts/icml\_ext3/scoring3.py}.
Its SHA256 begins \V{e3.scorer.sha}, as the screening specification pinned it at commit \V{e3.scorer.commit}, and it is the file in the repository at the date of this paper; \V{e3.scorer.tests} synthetic cases, none of which uses benchmark text, tested it before the screening.
Table~\ref{tab:third-scoring} gives examples, each a synthetic string run through the reader, with the layer that decides and the reading of the strict scorer alone.

\begin{table}[H]
\caption{Synthetic visible answers and how the reader reads them. Call ended: stop, or length (the call hit its token limit). Layer: the rule above that decides. Strict only: the reading of the strict layer alone.}
\label{tab:third-scoring}
\centering
\scriptsize
\setlength{\tabcolsep}{3pt}
\begin{tabular}{@{}llp{6.2cm}lllp{1.7cm}@{}}
\toprule
Task & Ended & Visible answer & Read as & Layer & Strict only \\
\midrule
yes or no & stop & \verb|Yes| & yes & (a) & yes \\
yes or no & stop & \verb|**No.**| & no & (a) & no \\
yes or no & stop & \verb|True| & yes & (a) & yes \\
yes or no & stop & \verb|False.| & no & (a) & no \\
yes or no & stop & \verb|The answer is: no| & no & (b) & no \\
yes or no & stop & \verb|Answer: True. Answer: Yes.| & yes & (b) & yes \\
yes or no & stop & \verb|Answer: Yes. On reflection, Answer: No.| & conflict, unparsed & (b) & conflict \\
yes or no & stop & \verb|Yes. There is no evidence that it changed.| & yes & (d) & unparsed \\
yes or no & length & \verb|Yes. There is no evidence that it changed.| & unparsed & none & unparsed \\
yes or no & stop & \verb|No one can say.| & no & (d) & unparsed \\
yes or no & stop & \verb|I think it is probably yes, but I am not sure.| & unparsed & none & unparsed \\
yes or no & stop & \verb|I cannot determine this.| & unparsed & none & unparsed \\
yes or no & stop & \verb|Yesterday it was cold.| & unparsed & none & unparsed \\
yes or no & stop & (empty; the label is only in the reasoning) & unparsed & none & unparsed \\
sentiment & stop & \verb|Positive| & positive & (a) & positive \\
sentiment & stop & \verb|The answer is neutral| & neutral & (b) & neutral \\
sentiment & stop & \verb|Neutral. The outlook is positive.| & neutral & (d) & unparsed \\
sentiment & stop & \verb|Negative, I think.| & negative & (d) & unparsed \\
sentiment & length & \verb|Neutral. The outlook is positive.| & unparsed & none & unparsed \\
\bottomrule
\end{tabular}
\end{table}

\section{The fourth study: preregistration, pilot and run}
\label{app:fourth-prereg}

This appendix summarises the fourth study's preregistration, what was known before the freeze, the stages of the run and the checks around it.
The preregistration (SHA256 beginning \V{e4.pre.sha8}, \V{e4.pre.bytes} bytes) was registered at \url{https://osf.io/g5apw} at \V{e4.osf.reg.time}~UTC on \V{e4.osf.reg.date} and frozen in commit \V{e4.osf.commit}; it was registered with an embargo until \V{e4.osf.embargo}, which we end when this version is posted, so the registration is public from then.
Its time is the time as reported to us, to the minute; the seconds are not known.
The paths, commits and SHA256 values named here are those of files in our repository, as for the earlier studies (Appendix~\ref{app:repro}).

\subsection*{What the preregistration fixes}
\paragraph{Prompts, scoring and denominators.}
The task prompt is the claim or the question followed by ``Answer with True or False.'' (CREAK) or ``Answer with Yes or No.'' (Manifold), with the third study's harness interface and copy rule, and the seed is the empty harness.
The primary scoring is the third study's two-layer reader (the strict layer for every call and the extended layer for a call that stopped), with ``true'' and ``false'' mapped to ``yes'' and ``no'' before any conflict check; a strict-only reading is reported beside it as a sensitivity.
Every row of a class is in that class's denominator whatever the answer: an unparsed answer and a conflict are wrong and never dropped, and a truncated call is read by the strict layer only.
A missing response or any other technical failure is not a row: the run stops and re-sends the identical request.

\paragraph{Serving, gates and hardware.}
The serving command is the third study's, with a context of \V{e4.d.ctx.task} tokens for the task model and \V{e4.d.ctx.prop} for the proposer, one batch-invariant server at a time (two while GEPA runs) and client concurrency \V{e4.d.conc}.
The thinking-switch check records a thinking-off call that returns reasoning content in the searches and in the evaluations of harnesses and baselines, and stops the chain only in the seed checks and the determinism rechecks.
The preregistration names the instance, card and driver, allows one change of card and driver before the first replication call (used once, Section~\ref{sec:fourth-run}) and stops the chain on any later change.
A shutdown guard shuts the instance down after a success or a failure, and a budget gate pauses the chain at a ceiling of \V{e4.run.ceiling}~CNY at a run boundary.

\paragraph{Statistics, labels and deviations.}
Section~\ref{sec:fourth} gives the interval.
The preregistration's labels for confirmatory claims are words of a verdict; its rule for a recorded deviation is that every claim becomes ``descriptive only'' with the deviation's name, and that rule, applied to D001, is why this paper uses no verdict word for the fourth study.
The preregistration also fixes the descriptive items that are reported without a family and without a label: both constant labels and the oracle constant, the parse shares of both scorings, the completion-token summary, a low-parse flag, the flagged-versus-not-flagged overlap sensitivity (a test row is flagged if it shares a mid-frequency word five-gram with a search row), and the comparison of the re-selected harness with the balanced-objective harness.

\subsection*{What was known before the freeze}
The pilot, its numbers and what it changed are in Section~\ref{sec:fourth}.
Four changes were made after the pilot and before the preregistration, on \V{e4.pil.dev.date}: the addition of hypothesis (d) and its family, the change of (d)'s balanced-accuracy requirement from a rise to a guard of \V{e4.d.margin}, the diagnostic work on the pilot's logs, and the pilot-effect power calculations.
They were not allowed by the pilot's rule that no pilot output changes the hypotheses; each was authorised by us and dated.
The pilot also evaluated three of the six re-selected harnesses that re-selection would have chosen, as another role's results, and a complete pilot estimate of (d) was not computed.
Neither the pilot's outputs nor its harnesses enter the main study.
The change of source came earlier: the proposal named a forecasting benchmark as the source of Arm~B and Manifold as the fallback under a rule fixed in advance; the benchmark did not reach the prespecified minima of eligible questions and test rows, so the fallback was used before any model output was seen.

\subsection*{The reading before the calls}
Before the first call on a replication or test row we rendered \V{e4.read.n} sample prompts from search rows, for each model and task under the seed harness, \cfg{THINK-512} and a placeholder harness, and read them under the preregistration's rule (the count and the files are in \texttt{docs/icml\_ext4/MAIN\_SAMPLES\_READING.md}, SHA256 beginning \V{e4.read.sha8}, and entry \V{e4.read.entry} of the contact log).
An AI session (Claude Code) read them alone; the study's lead author did not read them, and no replication or test row was opened, rendered or read.
The written record of the reading states who read what and quotes no item text.

\subsection*{The run}
The stages ran in the preregistered order on one card.
All \V{e4.d.runs} searches finished first, \V{e4.run.gepa.h} hours after the chain's first start on \V{e4.run.sta.date}; the harness list (\V{e4.run.harn} entries: the \V{e4.d.runs} selected and the \V{e4.a11.runs} re-selected harnesses; SHA256 beginning \V{e4.run.list.sha8}) was committed before the first replication call, and one re-selected harness equalled its selection, so the study has \V{e4.run.rep} replication and \V{e4.run.test} test evaluations.
The replication stage ran \V{e4.run.rep.h} hours; the checkpoint record (SHA256 beginning \V{e4.run.ckpt.sha8}) was committed before the first test call; the test stage, the determinism rechecks and the analysis followed, and the chain finished on \V{e4.run.end.date} at \V{e4.run.end.time}~UTC.
The first start of the chain stopped before any server, call or read of an item file, at \V{e4.run.a5.stop}~UTC on \V{e4.run.card.date} (addendum A5).
No search was retried and the chain logged no stop, no out-of-memory event and no budget pause.
The logged estimate of the spend was \V{e4.run.spend}~CNY (\V{e4.run.hours} billed hours) after the last determinism recheck, against the ceiling of \V{e4.run.ceiling}~CNY, and the budget gate was called after every run, the rechecks included (addendum A11 corrects an earlier statement that they were not gated).

\subsection*{Checks}
The determinism rechecks (\V{e4.rc.n}, each of \V{e4.rc.rows} rows) were identical to the first calls in every comparison.
No call with reasoning content occurred in any search or evaluation, and no selected harness was flagged for a low parse rate.
The diagnostic of whether the weighted objective changes the minibatch decision is in Section~\ref{sec:fourth-results}.

\section{The fourth study: addenda and the deviation}
\label{app:fourth-addenda}

The preregistration's addenda file is append-only; an entry is never edited, and each gives its date, its reason, the rule it applies and whether outputs on the items it concerns had been seen.
Table~\ref{tab:fourth-addenda} lists the thirteen entries, by the paper's wording and not by the file's, and says which of them bear on a result.
The one recorded deviation, D001, is described in Section~\ref{sec:fourth-results}.

\begin{table}[t]
\caption{The fourth study's addenda. ``Bears on a result'' means that the entry concerns a request, a score, a verdict rule, a stopping rule or the interpretation of a result; none changed a request or a score.}
\label{tab:fourth-addenda}
\centering
\footnotesize
\setlength{\tabcolsep}{3pt}
\begin{tabular}{@{}p{0.9cm}p{2cm}p{8.4cm}p{2.2cm}@{}}
\toprule
Entry & Date & What it records & Bears on a result \\
\midrule
A1 & 2026-11-02 & The freeze and the OSF registration; statements of the frozen file that the last steps made out of date. & no \\
A2 & 2026-11-04 & Our decision to use the one change of card and driver that the preregistration allows before the first replication call: a clone of the instance on a new card. The original host never freed a card. & no \\
A3 & 2026-11-04 & The clone's identity (container, card, driver) and a read-only check of its disks. & no \\
A4 & 2026-11-04 & The clone cleared of the pilot's state and its pins rewritten; the console ID; the render-hash lists are produced per evaluation, not at the sealed stage. & yes: D001 \\
A5 & 2026-11-04 & The first start of the chain stopped before any call, because the clone has no \texttt{python3} on the path of a non-login shell; the start of the chain. & no \\
A6 & 2026-11-04 & Two checks after A5; the clone's region and host; the copy-back rule. & no \\
A7 & 2026-11-07 & The searches are complete; the harness list and its SHA256, logged before any replication call. & no \\
A8 & 2026-11-07 & The replication stage is complete; the checkpoint record and its SHA256, logged before the first test call. & no \\
A9 & 2026-11-07 & The study is complete; the analysis report and what surrounds it. A later entry corrects one statement (the rechecks were gated). & yes: before D001 it said that no deviation exists \\
A10 & 2026-11-07 & A date correction to A9 and to three entries of the contact log. & no \\
A11 & 2026-11-07 & After the independent review: deviation D001 recorded; A9's statement on the recheck gate corrected; the main-study diagnostic of the minibatch decisions; a stale sentence in the re-selection module's description. & yes: D001 \\
A12 & 2026-11-07 & A transcription error in A9 and in one contact-log entry (a carriage-return byte in an archive path). & no \\
A13 & 2026-11-08 & The analysis rerun with D001 applied: the new report, equal to the original in every numeric value, with the labels changed. & yes: the report used here \\
\bottomrule
\end{tabular}
\end{table}

\paragraph{D001 in full.}
The preregistration's section on what is generated after the freeze lists the render-hash lists as computed at the sealed stage and logged before the first replication call.
The pinned code has no separate rendering pass at that stage; its item builder writes the item files, the groups file and the item hashes and renders nothing.
Each evaluation of the chain first renders its own rows and configuration, writes a render record (the SHA256 of every row's rendered text and one hash over them), puts that hash in the chain log, and then runs the evaluation with a check that refuses to start if its own render differs.
The full request message list and the SHA256 of the rendered text of every request were saved before each call was sent.
The searched harness texts were unavailable at the sealed stage. They were fixed after the searches and before replication, but we generated the render-hash lists per evaluation and did not commit them before the first replication call.
So the hashes of an evaluation were recorded just before its calls, on the instance and in its logs, and not in a commit before the first replication call.
We wrote this up in A4, before the sealed stage, and treated it, on our own instruction, as a change of no rule; that was wrong under the preregistration's rules, as the independent review found, because a departure from a registered procedure is a deviation, and the journal records that it cannot recreate the promised advance commitment.
The deviation was recorded with the frozen deviation tool on \V{e4.d001.date} at \V{e4.d001.time}~UTC, after every result existed; the instance had been shut down, so the journal was written on a scratch folder and copied back to the instance before the analysis was rerun.
The rerun used the pinned analysis environment (NumPy~\V{e4.sw.numpy}, SciPy~\V{e4.sw.scipy}); it compared equal to the original in \V{e4.cmp.n} numeric values and differs in the labels, the statuses and the deviation fields only (the original report's SHA256 begins \V{e4.orig.sha8}, the rerun's \V{e4.rep.sha8}).
No request, row, selection, rule, card or driver changed, and the only thing that D001 changed is the status of the claims.

\section{The fourth study: all comparisons, both partitions}
\label{app:fourth-endpoints}

Every number here is read from the analysis report rerun with D001 applied; every cell is descriptive.
Estimates are differences of means over \V{e4.d.reps} searches per condition, in points, and the interval after each is the preregistered one (\V{e4.d.lvl1}\% for (a), \V{e4.d.lvl2}\% for (b) and (d)).
The percentile intervals are two-way bootstrap intervals at \V{e4.ci95}\% over searches and groups and are descriptive; they are not the intervals of the preregistered comparisons.
The test-partition tables of the main text are repeated here with the replication partition, and the means and the baselines are given for both.

\begin{table}[t]
\caption{Comparisons (a) and (b), test partition (as Table~\ref{tab:fourth-ab}).}
\label{tab:fourth-ab-te}
\centering
\input{tables/t4_ab_te}
\end{table}

\begin{table}[t]
\caption{Comparisons (a) and (b), replication partition. The replication rows are \V{e4.re.cr.sm.rows} (CREAK) and \V{e4.re.mf.sm.rows} (Manifold); intervals as for the test partition.}
\label{tab:fourth-ab-re}
\centering
\input{tables/t4_ab_re}
\end{table}

\begin{table}[t]
\caption{Comparison (d), test partition (as Table~\ref{tab:fourth-d}).}
\label{tab:fourth-d-te}
\centering
\input{tables/t4_d_te}
\end{table}

\begin{table}[t]
\caption{Comparison (d), replication partition.}
\label{tab:fourth-d-re}
\centering
\input{tables/t4_d_re}
\end{table}

\begin{table}[t]
\caption{Percentile two-way bootstrap intervals (\V{e4.ci95}\%, descriptive) for the estimates of the preregistered comparisons, test partition, in points. Columns are the estimates of (a), of (b) in balanced accuracy and minority recall, and of (d) in the same two.}
\label{tab:fourth-pct-te}
\centering
\input{tables/t4_pct_te}
\end{table}

\begin{table}[t]
\caption{As Table~\ref{tab:fourth-pct-te}, replication partition.}
\label{tab:fourth-pct-re}
\centering
\input{tables/t4_pct_re}
\end{table}

\begin{table}[t]
\caption{Mean replication-partition recall of the minority and majority labels, in percent, for the four conditions (as Table~\ref{tab:fourth-means} for the test partition).}
\label{tab:fourth-means-re}
\centering
\input{tables/t4_means_re}
\end{table}

\begin{table}[t]
\caption{Mean balanced accuracy and accuracy in percent for the four conditions, test partition.}
\label{tab:fourth-meansba-te}
\centering
\input{tables/t4_meansba_te}
\end{table}

\begin{table}[t]
\caption{Mean balanced accuracy and accuracy in percent for the four conditions, replication partition.}
\label{tab:fourth-meansba-re}
\centering
\input{tables/t4_meansba_re}
\end{table}

\begin{table}[t]
\caption{Mean accuracy of each condition minus the accuracy of the constant label with the higher accuracy on the partition (``Yes'' on CREAK's test partition, ``No'' on Manifold's), test partition, in points. The constant is the oracle constant of the report (the label with the higher accuracy on the partition's rows, a tie going to the majority label), so on a task near chance the difference is at most about zero.}
\label{tab:fourth-hc-te}
\centering
\input{tables/t4_hc_te}
\end{table}

\begin{table}[t]
\caption{As Table~\ref{tab:fourth-hc-te}, replication partition.}
\label{tab:fourth-hc-re}
\centering
\input{tables/t4_hc_re}
\end{table}

\begin{table}[t]
\caption{Baselines, test partition, in percent: balanced accuracy and recall of the minority and the majority label. \cfg{AUTHORS-1500} is the empty harness with the longevity model's authors' inference settings; the last row of each block gives the constant labels.}
\label{tab:fourth-base-te}
\centering
\input{tables/t4_base_te}
\end{table}

\begin{table}[t]
\caption{Baselines, replication partition, in percent, with \cfg{THINK-LONG-2048} under the model card's sampling (temperature \V{e4.d.card.t}, top-p \V{e4.d.card.p}).}
\label{tab:fourth-base-re}
\centering
\input{tables/t4_base_re}
\end{table}

\end{document}